\documentclass[11pt,a4paper]{article}

\usepackage[utf8]{inputenc}
\usepackage[T1]{fontenc}
\usepackage{microtype}                       % optical margin alignment
\usepackage[margin=1in, headheight=22pt]{geometry}

\usepackage{mathpazo}                        % Palatino + Pazo math
\usepackage[scaled=0.85]{helvet}             % Helvetica for sans-serif
\usepackage{courier}                         % Courier for monospace

\usepackage{amsmath,amssymb,amsthm}
\usepackage{bm}
\usepackage{mathtools}

\usepackage[dvipsnames]{xcolor}

\definecolor{DPink1}{HTML}{F3CDF0}
\definecolor{DPink2}{HTML}{E993D0}
\definecolor{DPink3}{HTML}{CF50A0}
\definecolor{DPink4}{HTML}{94133F}

\definecolor{DBlue1}{HTML}{C7DDDE}
\definecolor{DBlue2}{HTML}{7DBDC2}
\definecolor{DBlue3}{HTML}{007A87}
\definecolor{DBlue4}{HTML}{004751}

\definecolor{DYellow1}{HTML}{FFF8D4}
\definecolor{DYellow2}{HTML}{F7C886}
\definecolor{DYellow3}{HTML}{F8B65A}
\definecolor{DYellow4}{HTML}{DE9B35}

\definecolor{DPurple1}{HTML}{D6D4F5}
\definecolor{DPurple2}{HTML}{B4B4ED}
\definecolor{DPurple3}{HTML}{504FB3}
\definecolor{DPurple4}{HTML}{2B2B7F}

\definecolor{DGray1}{HTML}{FBFAF4}
\definecolor{DGray2}{HTML}{EDEBE2}
\definecolor{DGray3}{HTML}{CFCDC3}
\definecolor{DGray4}{HTML}{9D9D9C}

\usepackage[
  colorlinks  = true,
  linkcolor   = DBlue4,
  citecolor   = DPink4,
  urlcolor    = DBlue3,
  bookmarks   = true,
]{hyperref}

\usepackage[nameinlink,capitalise]{cleveref}
\crefrangelabelformat{figure}{#3#1#4--#5#2#6}

\crefrangelabelformat{subfigure}{#3#1#4--#5\crefstripprefix{#1}{#2}#6}

\usepackage{titlesec}
\titleformat{\section}
  {\Large\bfseries\color{DBlue4}}            % format
  {\thesection}                               % label
  {0.75em}                                    % sep
  {}                                          % before-code
  [\vspace{2pt}{\color{DBlue2}\titlerule[0.6pt]}] % after-code

\titleformat{\subsection}
  {\large\bfseries\color{DBlue3}}
  {\thesubsection}{0.6em}{}

\titleformat{\subsubsection}
  {\normalsize\bfseries\color{DPink4}}
  {\thesubsubsection}{0.5em}{}

\titleformat{\paragraph}[runin]
  {\normalsize\bfseries\color{DPurple4}}
  {}
  {0em}
  {}[.]
\titlespacing*{\paragraph}{0pt}{0.8ex plus .2ex minus .1ex}{0.6em}

\usepackage{fancyhdr}
\renewcommand{\headrulewidth}{0.4pt}
\renewcommand{\headrule}{\hbox to\headwidth{%
  \color{DBlue2}\leaders\hrule height \headrulewidth\hfill}}
\fancypagestyle{firstpage}{%
  \fancyhf{}
  \renewcommand{\headrulewidth}{0pt}
  \fancyfoot[C]{\small\color{DGray3}\thepage}
}

\fancypagestyle{toc}{%
\fancyhf{} % Clear all header/footer fields
\renewcommand{\headrulewidth}{0.4pt}
\renewcommand{\headrule}{\hbox to\headwidth{%
  \color{DBlue2}\leaders\hrule height \headrulewidth\hfill}}
\fancyhead[L]{\small\color{DBlue4}\textsc{DISCO}}
\fancyhead[R]{\small\color{DGray3}\thepage}
\fancyfoot[C]{}
\fancyfoot[R]{\hyperlink{toc}{\footnotesize{(Back to Table of Contents)}}} % Link in top right
}
\usepackage[
  font={small},
  labelfont={bf,color=DBlue4},
  textfont={color=black},
  labelsep=period,
  margin=1em,
]{caption}
\usepackage{subcaption}

\usepackage{tcolorbox}
\tcbuselibrary{skins,breakable}

\newtcolorbox{discoabstract}{%
  enhanced,
  breakable,
  colback    = DGray1,
  colframe   = DBlue2,
  boxrule    = 0.8pt,
  arc        = 3pt,
  left       = 12pt,
  right      = 12pt,
  top        = 10pt,
  bottom     = 10pt,
  fontupper  = \small,
  before upper = {\noindent\textbf{\color{DBlue4}Abstract}\quad},
}

\newtcolorbox{theorembox}[1][]{%
  enhanced,
  colback     = DPurple1!30!white,
  colframe    = DPurple3,
  boxrule     = 0.6pt,
  arc         = 2pt,
  left        = 8pt,
  right       = 8pt,
  top         = 6pt,
  bottom      = 6pt,
  fonttitle   = \bfseries\color{DPurple4},
  title       = {#1},
  attach boxed title to top left = {yshift=-2mm, xshift=4mm},
  boxed title style = {colback=DPurple1!30!white, colframe=DPurple1!30!white},
}

\newtcolorbox{proofbox}{%
  enhanced,
  breakable,
  colback     = DYellow1!50!white,
  colframe    = DYellow3,
  boxrule     = 0.4pt,
  arc         = 2pt,
  left        = 8pt,
  right       = 8pt,
  top         = 6pt,
  bottom      = 6pt,
}

\newtcolorbox{corollarybox}[1][]{%
  enhanced,
  colback     = DPink1!30!white,
  colframe    = DPink3,
  boxrule     = 0.6pt,
  arc         = 2pt,
  left        = 8pt,
  right       = 8pt,
  top         = 6pt,
  bottom      = 6pt,
  fonttitle   = \bfseries\color{DPink4},
  title       = {#1},
  attach boxed title to top left = {yshift=-2mm, xshift=4mm},
  boxed title style = {colback=DPink1!30!white, colframe=DPink1!30!white},
}

\usepackage{listings}
\usepackage[
  numbers,
  sort&compress,
]{natbib}
\usepackage{algorithm}
\usepackage{algpseudocode}

\makeatletter
\renewcommand{\ALG@name}{\color{DBlue4}\textbf{Algorithm}}
\makeatother

\usepackage{booktabs}
\usepackage{longtable}
\usepackage{multirow}
\usepackage{colortbl}

\usepackage{graphicx}
\usepackage{float}
\usepackage{enumitem}
\usepackage{xspace}
\usepackage{mhchem}
\usepackage{seqsplit}
\usepackage{url}
\usepackage{mdframed}
\usepackage{makecell}
\usepackage{adjustbox}

\setlist{nosep, leftmargin=1.5em}

\newcommand{\name}{\textsc{Disco}\xspace}

\definecolor{citeblue}{HTML}{007A87} 
\definecolor{codeblue}{HTML}{007A87} 
\definecolor{codepink}{HTML}{CF50A0}

\hypersetup{
	colorlinks=true,       % false: boxed links; true: colored links
	linkcolor=citeblue,        % color of internal links
	citecolor=citeblue,        % color of links to bibliography
	filecolor=magenta,     % color of file links
	urlcolor=citeblue         
}

\mdfdefinestyle{MyFrame2}{%
    linecolor=white,
    outerlinewidth=1pt,
    roundcorner=5pt,
    innertopmargin=5pt,
    innerbottommargin=5pt,
    innerrightmargin=10pt,
    innerleftmargin=10pt,
    backgroundcolor=black!3!white}

\newenvironment{breakablealgorithm}[1]
{\vspace{15pt}
   \par\noindent
   \hrule height 0.8pt
   \vspace{2pt}
   \refstepcounter{algorithm}
   \noindent\mbox{\textbf{Algorithm \thealgorithm}\ #1}
   \par\vspace{2pt}
   \hrule height 0.4pt
}
{
   \vspace{2pt}
   \hrule height 0.8pt
   \par
}

\usepackage{setspace}
\newcommand{\nameshort}{\textsc{Disco}\xspace}
\newcommand{\namelong}{Diffusion for Sequence-Structure Co-design\xspace}
\newcommand{\benchmark}{\textsc{Studio-179}\xspace}

\newcommand{\xhdr}[1]{{\noindent\bfseries\color{DBlue4} #1.}\enspace}
\newcommand{\prot}{\textrm{prot}}
\newcommand{\nuc}{\textrm{nuc}}
\newcommand{\ligand}{\textrm{ligand}}

\newcommand{\clust}{\textrm{clust}}
\newcommand{\seq}{\textrm{seq}}
\newcommand{\struct}{\textrm{struct}}
\newcommand{\txtskip}{\textrm{skip}}

\newcommand{\angstrom}{\mathring{A}}

\newcommand{\R}{\mathbb{R}}
\newcommand{\gV}{\mathcal{V}}
\newcommand{\gD}{\mathcal{D}}
\newcommand{\mb}[1]{\mathbf{#1}}
\newcommand{\deriv}[2]{\frac{\partial #1}{\partial #2}}
\newcommand{\inner}[2]{\langle #1, #2 \rangle}
\newcommand{\mean}{\mathbb{E}}
\newcommand{\cond}{\,|\,}
\newcommand{\norm}[1]{\left\lVert #1 \right\rVert}

\mdfdefinestyle{MyFrame2}{%
  linecolor  = DPink3,
  linewidth  = 0.8pt,
  roundcorner= 3pt,
  backgroundcolor = DPink1!20!white,
  innertopmargin  = 8pt,
  innerbottommargin = 8pt,
  innerleftmargin = 10pt,
  innerrightmargin= 10pt,
}

\usepackage[symbol]{footmisc}

\title{%
  \vspace{-1.5em}%
  {\color{DBlue4}\rule{\textwidth}{1.2pt}}\\[0.8em]%
  {\huge\bfseries\color{DBlue4}%LARGE
    General Multimodal Protein Design\\[0.15em]
    Enables DNA-Encoding of Chemistry}%
  \\[0.6em]%
  {\color{DPink3}\rule{0.4\textwidth}{0.6pt}}%
  \vspace{0.3em}%
}

\author{%
  Jarrid Rector-Brooks\textsuperscript{1,2,3\,\dag\,*},\;
  Théophile Lambert\textsuperscript{1,4\,\dag},\;
  Marta Skreta\textsuperscript{2,3\,\dag},\;
  Daniel Roth\textsuperscript{1\,\dag}\\[3pt]
  Yueming Long\textsuperscript{1},\;
  Zi-Qi Li\textsuperscript{1},\;
  Xi Zhang\textsuperscript{2,5},\;
  Miruna Cretu\textsuperscript{6},\;
  Francesca-Zhoufan Li\textsuperscript{1}\\[3pt]
  Tanvi Ganapathy\textsuperscript{1},\;
  Emily Jin\textsuperscript{7},\;
  Avishek Joey Bose\textsuperscript{2,8},\;
  Jason Yang\textsuperscript{1}\\[3pt]
  Kirill Neklyudov\textsuperscript{2,3,9},\;
  Yoshua Bengio\textsuperscript{2,3,10},\;
  Alexander Tong\textsuperscript{11}\\[3pt]
  Frances H. Arnold\textsuperscript{1\,*},\;
  Cheng-Hao Liu\textsuperscript{1,12\,\dag\,*}\\[10pt]
  {\small\color{DBlue4}%
    \textsuperscript{1}California Institute of Technology\enspace
    \textsuperscript{2}Mila -- Québec AI Institute\enspace
    \textsuperscript{3}Université de Montréal}\\
  {\small\color{DBlue4}%
    \textsuperscript{4}Université Paris-Saclay\enspace
    \textsuperscript{5}McGill University\enspace
    \textsuperscript{6}University of Cambridge\enspace
    \textsuperscript{7}University of Oxford}\\
  {\small\color{DBlue4}%
    \textsuperscript{8}Imperial College London\enspace
    \textsuperscript{9}Institut Courtois\enspace
    \textsuperscript{10}LawZero\enspace
    \textsuperscript{11}AITHYRA\enspace
    \textsuperscript{12}FutureHouse}\\[6pt]
  {\small\color{DGray4}%
    \textsuperscript{*}Correspondence:
    \texttt{\color{DBlue3}chl@caltech.edu},}\\
  {\small\color{DGray3}%
    \texttt{\color{DBlue3}jarrid.rector-brooks@mila.quebec},
    \texttt{\color{DBlue3}frances@cheme.caltech.edu}}\\
  {\small\color{DGray4}\textsuperscript{\dag}Equal contribution}
}

\date{}

\begin{document}
\maketitle
\thispagestyle{firstpage}

%% ── Abstract ──
\begin{discoabstract}
Evolution is an extraordinary engine for enzymatic diversity, yet the chemistry it has explored remains a narrow slice of what DNA can encode. Deep generative models can design new proteins that bind ligands, but none have created enzymes without pre-specifying catalytic residues. We introduce \nameshort, a multimodal model that co-designs protein sequence and 3D structure around arbitrary biomolecules, as well as inference-time scaling methods that optimize objectives across both modalities. Conditioned solely on reactive intermediates, \nameshort designs diverse heme enzymes with novel active-site geometries. These enzymes catalyze new-to-nature carbene-transfer reactions, including alkene cyclopropanation, spirocyclopropanation, B–H, and C(sp$^3$)–H insertions, with high activities exceeding those of engineered enzymes.  Random mutagenesis of a selected design further confirmed that enzyme activity can be improved through directed evolution. By providing a scalable route to evolvable enzymes, \nameshort broadens the potential scope of genetically encodable transformations.
\end{discoabstract}

\vspace{2.0em}

\begin{center}
\begin{tcolorbox}[
  enhanced,
  hyperref,
  colback=DBlue1!30!white,
  colframe=DBlue3,
  boxrule=0.5pt,
  arc=4pt,
  width=0.75\textwidth,
  left=8pt, right=8pt, top=6pt, bottom=6pt,
]
\centering\small
\textbf{\color{DBlue4}Code \& Models}\enspace\textbar\enspace
\url{https://github.com/DISCO-design/DISCO}
\end{tcolorbox}
\end{center}

\vspace{0.5em}

%% ── Main sections ──

\section*{Introduction} 
\looseness=-1
Enzymes catalyze chemical reactions under mild conditions with high efficiency and specificity, underpinning applications in chemical manufacturing, drug synthesis, and emerging therapeutic modalities. \cite{buller_nature_2023,reisenbauer_catalyzing_2024} Enzyme engineering has typically relied on iterative rounds of mutation and screening to accumulate beneficial sequence changes toward desired functions – an evolutionary process that can enable enzymes to perform functions not previously known in nature. \cite{chen_engineering_2020} However, every such campaign requires an initial protein with measurable activity. For well-studied transformations, suitable starting points can be found among natural proteins or their close variants; but for many desirable new-to-nature chemistries, nature offers no obvious candidates. Identifying one remains largely a matter of chemical intuition – a laborious process fundamentally limited by what evolution has already sampled. How can we systematically design functional enzymes for chemical transformations that have no precedent in known biology? 

Generative deep learning models have transformed protein design, successfully enabling the creation of novel binders and the scaffolding of pre-defined structural motifs. \cite{watson_novo_2023,wang_scaffolding_2022,pacesa_one-shot_2025} However, \textit{de novo} design of enzymes for new-to-nature reactions remains largely unrealized. Current computational pipelines face two fundamental limitations. First, they rely heavily on pre-specified, fixed geometric arrangements of active-site residues or theozymes. \cite{ahern_atom-level_2026,lauko_computational_2025,yeh_novo_2023} 
This precludes reactions for which no motif or precise mechanism is known and prevents the discovery of novel active-site geometries. Second, experimentally validated generative pipelines predominantly operate sequentially: they first generate a backbone and then devise a sequence to match the backbone design (inverse folding). \cite{ahern_atom-level_2026,butcher2025novo,stark_boltzgen_2025} Because sequence and structure jointly determine function, this decoupled paradigm cannot leverage sequence-based objectives during critical stages of backbone formation, and vice versa.  

\looseness=-1
We introduce \nameshort (DIffusion for Sequence-structure CO-design), a general multimodal framework that simultaneously designs protein sequences and 3D structures \textit{de novo}. \nameshort designs can be conditioned on and co-folded with arbitrary biomolecules, without pre-defined residue motifs. A novel multimodal Feynman-Kac corrector method allows \nameshort to optimize sequence and structure objectives together. \nameshort achieves state-of-the-art \textit{in silico} results in generating binders for diverse biomolecular targets with fine-grained property control. Applied to new-to-nature catalysis experimentally, \nameshort generates diverse, evolvable carbene transferases with novel active sites. No inverse-folding is required. These enzymes have activities that surpass typical starting points for directed evolution and, in selected cases, extensively evolved variants. Ultimately, \nameshort enables the discovery of biocatalysts and protein motifs previously unknown to biology, expanding the searchable space of DNA-encoded chemical reactivity.  

\begin{figure}
	\centering
	\includegraphics[width=1.0\textwidth]{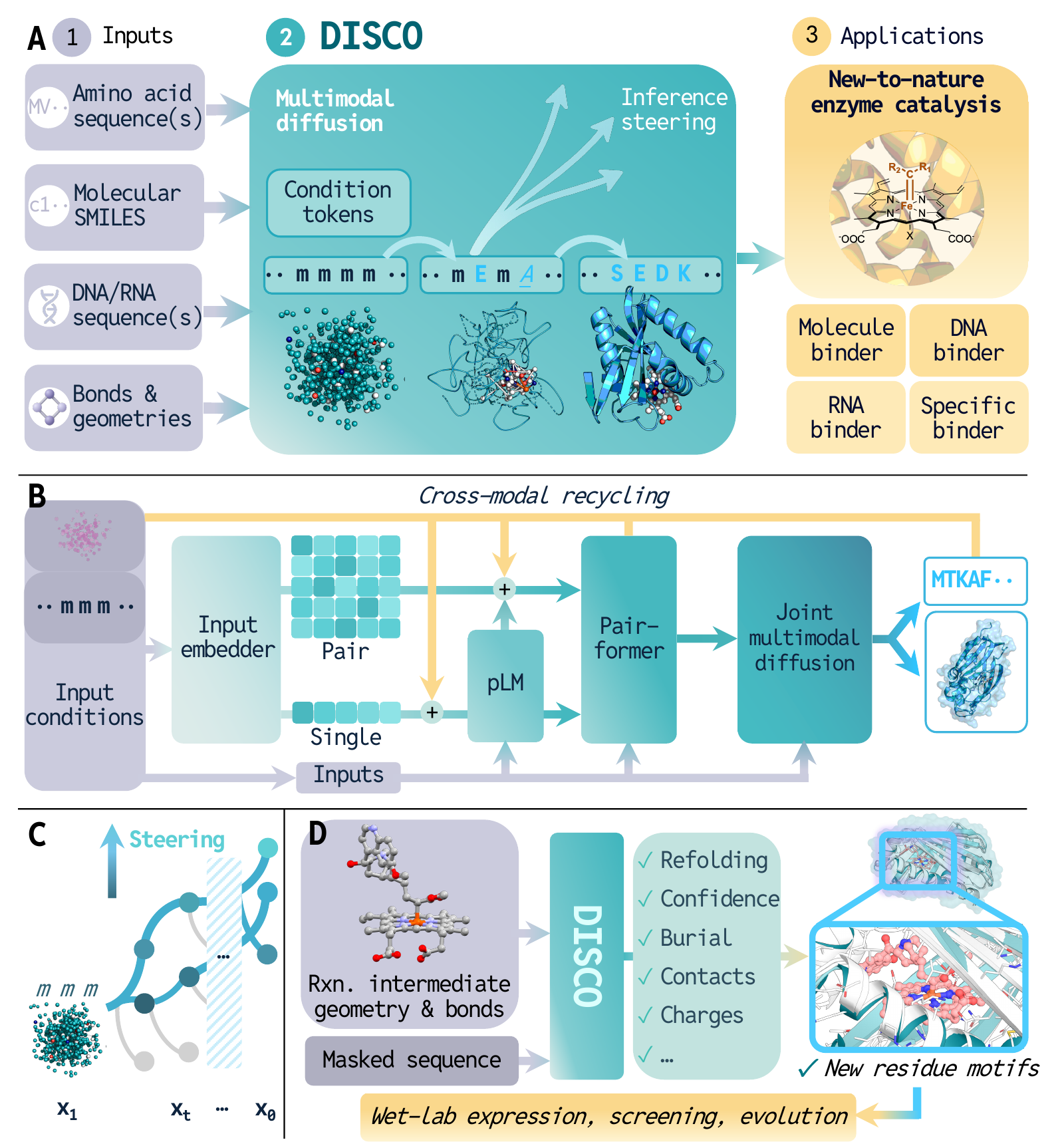}
    {\phantomsubcaption\label{fig:1A}}
    {\phantomsubcaption\label{fig:1B}}
    {\phantomsubcaption\label{fig:1C}}
    {\phantomsubcaption\label{fig:1D}}
	\caption{\textbf{Multimodal protein design workflow with \nameshort.} (\textbf{A}) Inference overview, highlighting the arbitrary molecular context, sequence-structure co-generation with sequence correction, and representative applications. (\textbf{B}) Architecture overview. (\textbf{C}) Multimodal inference-time steering schematics with Feynman-Kac corrector. (\textbf{D}) Design-to-test workflow for generating enzymes with new active sites from reactive intermediates.}
	\label{fig:1}
\end{figure}

\section*{Multimodal protein generation with \nameshort}

\nameshort models discrete sequences and 3D structures as a joint distribution that is denoised through a unified generative process (\cref{fig:1}). We achieve this by training a single deep neural network that employs a masked discrete diffusion process for sequences and, in parallel, leverages continuous diffusion for 3D atomic coordinates. By independently sampling noise per modality during training, one can provably learn the joint reverse process using only unimodal diffusion losses \cite{rojas2025diffuseeverythingmultimodaldiffusion}: the standard denoising score matching loss for coordinates and the masked diffusion language model loss for sequences. We trained \nameshort using only data from the Protein Data Bank without special filtering, avoiding any selection biases introduced when models are trained exclusively on ``designable'' structures that folding algorithms can successfully predict.  

\nameshort's architecture comprises an input embedder that produces single and pair representations of all input modalities, a frozen protein language model (pLM) that encodes partial sequences, \cite{wang2024diffusion} a Pairformer stack that contextualizes these representations, and a joint diffusion module built on atom-level attention that predicts both sequence logits and denoised coordinates of the backbone (\cref{fig:1B}). \cite{abramson_accurate_2024} We inject SE(3) symmetries softly through data augmentation rather than architectural constraints.  

Central to our approach is a cross-modal recycling mechanism that conditions each generation step on four distinct encodings: the model's current predicted clean sequence $\hat{x}_0^\seq$ and structure $\hat{x}_0^\struct$, as well as the current noised sequence $x_\tau^\seq$ and structure $x_\tau^\struct$. We train a structure encoder \cite{dauparas2025atomic} and employ the pLM to inject sequence information. This bidirectional conditioning ensures that sequence predictions are informed by emerging structural features while structural predictions adapt to evolving sequence identity. 

\begin{figure}
	\centering
	\includegraphics[width=1.0\textwidth]{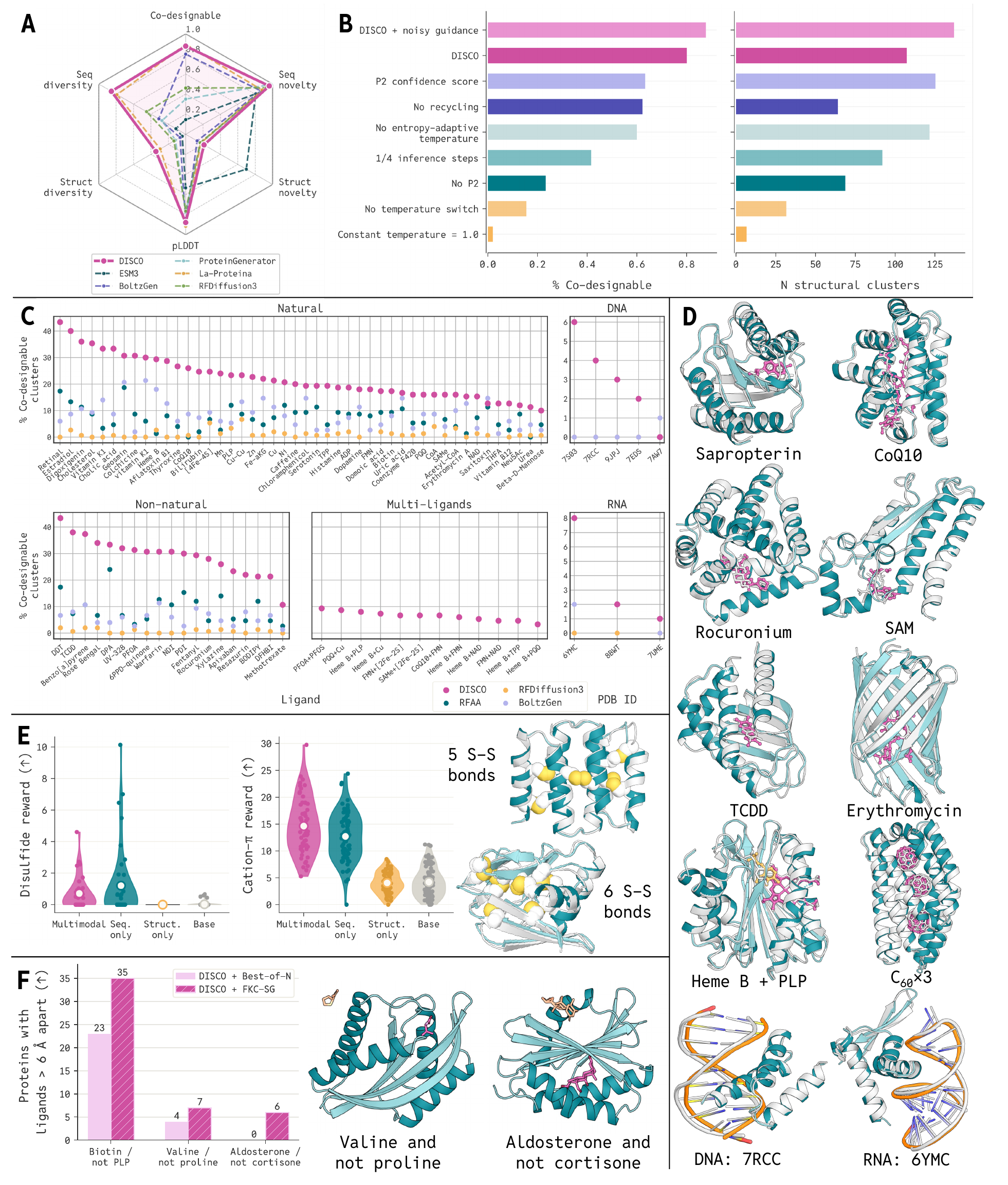}
    {\phantomsubcaption\label{fig:2A}}
    {\phantomsubcaption\label{fig:2B}}
    {\phantomsubcaption\label{fig:2C}}
    {\phantomsubcaption\label{fig:2D}}
    {\phantomsubcaption\label{fig:2E}}
    {\phantomsubcaption\label{fig:2F}}
	\caption{\textbf{\nameshort generates diverse, designable protein sequences and structures across a wide variety of functions evaluated \textit{in silico.}} (\textbf{A}) In unconditional monomer generation, \nameshort outperforms existing methods in terms of co-designability, novelty, and diversity (400 samples).  (\textbf{B}) Ablation of unconditional monomer generation with \nameshort to identify best inference-time practices.  (\textbf{C}) \nameshort outperforms existing methods for conditional generation across diverse targets, including natural and non-natural molecules, multi-ligands, DNA, and RNA. (\textbf{D}) Examples of designed (blue) and refolded (white) protein complexes show high agreement. (\textbf{E}) FKC-MM enables the joint steering of sequence and structure, as demonstrated by two multimodal rewards: disulfide bridges and cation-$\pi$ interactions. Example  generated proteins with high disulfide bridge counts are shown at right. (\textbf{F}) FKC-SG generates more proteins with predicted binding \textit{specificity} (a large separation between binding sites of on- and off-targets) compared to existing filtering methods. Example refolded structures are shown at right (on-target in pink, off-target in yellow).}
	\label{fig:2}
\end{figure}

\nameshort's inference strategy proved critical to generation quality (\cref{fig:2B,fig:unconditional_n_steps_cycles}). Standard masked diffusion—iteratively unmasking tokens without revision \cite{mdlm}—yields sequences that fail to fold into their generated geometries. By enabling self-correction during sequence inference \cite{peng2025pathplanningmaskeddiffusion,wang2025remaskingdiscretediffusionmodels} and introducing a novel sequence temperature mechanism that smooths the amino acid distribution for overconfident tokens early in the trajectory, we drastically improve co-designability. Combining this with noisy guidance \cite{rojas2025diffuseeverythingmultimodaldiffusion} allows \nameshort to achieve state-of-the-art performance in sequence-structure co-design. On unconditional monomer design, approximately 90\% of generated sequences refold to within 2 Å RMSD of their designed backbone structures using ESMFold (\cref{fig:2A,fig:unconditional_metrics_by_length,tab:main-results}). \cite{lin2023evolutionary} Compared to baselines, \nameshort achieves the highest sequence and structure diversity and novelty without sacrificing co-designability (\cref{fig:2A}).

\subsection*{Conditioning on arbitrary biomolecular contexts}

\nameshort flexibly conditions on arbitrary biomolecular contexts, including small molecules, metallocofactors, reactive intermediates, and nucleic acids. Non-protein molecules are represented identically to proteins – with discrete tokens for chemical identity and corresponding atomic coordinates. The model simultaneously unmasks protein sequences while denoising the coordinates of targets ranging from small molecules and metal cofactors to reactive intermediates and nucleic acids. These contexts co-fold with the designed protein throughout the generative trajectory: \nameshort adjusts the coordinates of conditioning biomolecules while designing the protein, capturing conformational changes induced by molecular interactions and extending generation beyond fixed, pre-defined atomic scaffolds. 

To benchmark this capability, we designed a new benchmark, \benchmark, a library of 179 natural and non-natural ligands spanning applications in catalysis, pharmaceuticals, luminescence, and sensing. The library encompasses rigid molecules (e.g., the pollutant tetrachlorodibenzodioxin), large or flexible molecules (e.g., the cofactor CoQ$^{10}$), and metals/metalloclusters (e.g., [4Fe-4S]). For each ligand, we quantify the fraction of generated designs that are both structurally diverse and co-designable, defining co-designable as when the protein backbone and all ligand centroids have an RMSD < 2 Å upon refolding with Chai-1. \cite{chai2024chai} Compared to baselines, \nameshort generates the highest proportion of diverse, co-designable complexes for 178 of 179 evaluated ligands (\cref{fig:2C,fig:2D,fig:p2_p3_ligands,fig:conditional_metrics_by_length,fig:cond_prot_examples}). Our formulation naturally supports multi-context conditioning—such as designing proteins to interface with multiple distinct ligands simultaneously—again yielding high-quality complexes (\cref{fig:2D}). 

Extending beyond small molecules, we evaluated \nameshort's performance on macromolecular interfaces. \nameshort successfully generated co-designable proteins predicted to bind to both sequence-specific DNA and RNA sequences, outperforming existing models (\cref{fig:2C}). Note that because the frozen pLM in \nameshort was not trained on multimeric complexes, we constrained our current evaluation to non-protein-binding tasks.  

\subsection*{Multimodal inference steering using Feynman-Kac Correctors}

The multimodality of \nameshort enables inference-time scaling with both sequence and structure signals (\cref{fig:1C}). Rather than relying on inefficient generate-and-filter (best-of-$N$) approaches, which often fail when desired traits are rare, we utilize the Feynman-Kac Corrector (FKC) framework to incorporate target properties into the generation process directly and tilt sample distributions toward having desirable traits without necessarily invoking expensive reward oracles. \cite{skreta2025fkc,hasan2026discrete} We derive two novel FKC approaches.  

First, FKC - Multimodal (FKC-MM) allows for reward tilting with functions defined jointly over both discrete sequences and continuous structures. When using multimodal reward functions for increasing disulfide bond content or cation-$\pi$ interactions, FKC-MM generates samples with enriched target properties compared to structure-only FKC or unconditional generations (\cref{fig:2E}). Notably, while unconditional generation underrepresents cysteines (\cref{fig:aa_distribution}), the top 2\% of designable, 100-amino acid proteins produced by FKC-MM contain six disulfide bonds – a density matched by only the top 0.2\% of comparable training proteins (\cref{fig:disulfide_distribution}). These results underscore the benefit of being able to generate sequences alongside protein structures.  

We further introduce FKC - Specificity Guidance (FKC-SG) to steer \nameshort towards binding exclusively to a target molecule while avoiding a structurally similar decoy. This is achieved by sampling from a tilted distribution that encourages samples that are likely under the on-target model while penalizing samples that are likely under the off-target model (\cref{sec:inference_steering}). We evaluated FKC-SG on discriminating two dissimilar molecules (e.g., biotin from pyridoxal 5'-phosphate (PLP)), as well as very similar molecules, such as aldosterone from cortisone (two constitutional isomeric steroids) and valine from proline (two hydrophobic, aliphatic amino acids). We find that FKC-SG generates proteins that have high separation between on- and off-target complexes (\cref{fig:2F}), improving upon filters like best-of-$N$, which in some cases do not produce any hits passing \textit{in silico} filters. 

\subsection*{\nameshort designs exhibit realistic protein features with novel, complementary motifs}

\begin{figure}
	\centering
	\includegraphics[width=1.0\textwidth]{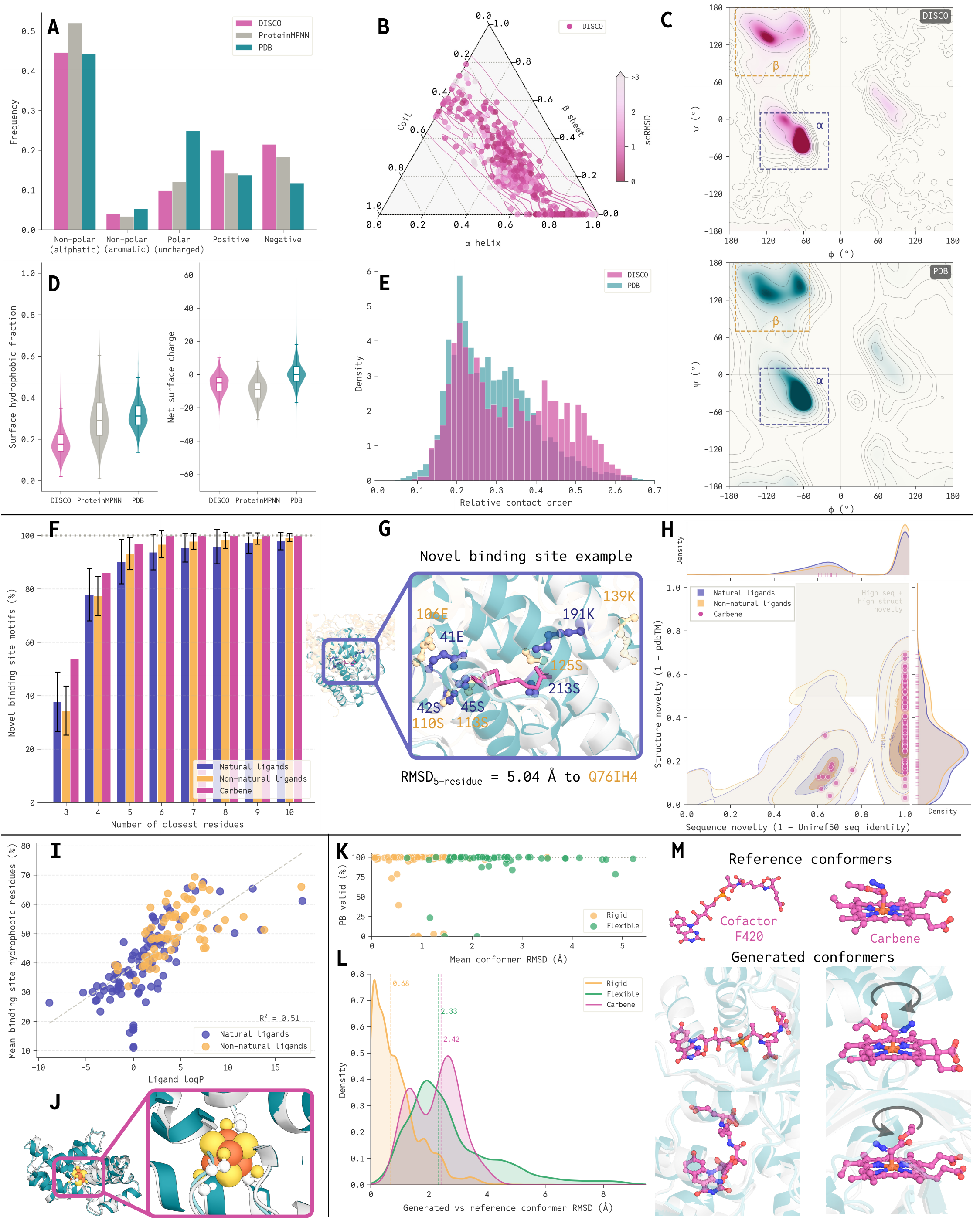}
    {\phantomsubcaption\label{fig:3A}}
    {\phantomsubcaption\label{fig:3B}}
    {\phantomsubcaption\label{fig:3C}}
    {\phantomsubcaption\label{fig:3D}}
    {\phantomsubcaption\label{fig:3E}}
    {\phantomsubcaption\label{fig:3F}}
    {\phantomsubcaption\label{fig:3G}}
    {\phantomsubcaption\label{fig:3H}}
    {\phantomsubcaption\label{fig:3I}}
    {\phantomsubcaption\label{fig:3J}}
    {\phantomsubcaption\label{fig:3K}}
    {\phantomsubcaption\label{fig:3L}}
    {\phantomsubcaption\label{fig:3M}}
    \vspace{-7mm}
	\caption{\textbf{\nameshort designs biophysically realistic proteins with novel motifs, flexible ligands, and target-informed sequences.} (\textbf{A}-\textbf{E}) Unconditional generation results: (\textbf{A}) Amino acid type distribution as compared to PDB and inverse-folded samples. (\textbf{B}) Secondary structure diversity. (\textbf{C}) Ramachandran plot. (\textbf{D}) Surface hydrophobicity and net charge. (\textbf{E}) Distribution of long-range contact order. (\textbf{F}-\textbf{M}) Conditional generation results: (\textbf{F}) Folddisco motif similarity searches reveal no close matches to designed residue motifs in AlphaFoldDB. (\textbf{G}) Example active-site (purple) compared to the closest known motif (yellow). (\textbf{H}) Distribution of global sequence/structure novelty. (\textbf{I}) \nameshort designs binding sites with consideration of ligand properties, as illustrated by matching lipophilicity. (\textbf{J}) Example design satisfying ligand constraints, building four disulfide bonds to a [4Fe-4S] cluster. (\textbf{K}) \nameshort-generated conformers are valid as evaluated by PoseBusters, even if they deviate from the reference. (\textbf{L}) \nameshort samples different conformers. (\textbf{M}) Examples of sampled conformers that are rare or absent in the training set.}
	\label{fig:3}
\end{figure}

\nameshort's designs capture the complex statistical properties of proteins. In unconditional generation, samples display natural amino-acid compositions (\cref{fig:3A,fig:aa_distribution}), diverse secondary structures (\cref{fig:3B,fig:contact_types}), favorable local Ramachandran geometries (\cref{fig:3C}), reasonable radii of gyration (\cref{fig:rog}), and native-like surface properties including appropriate hydrophobicity and net charge (\cref{fig:3D}). The designs further exhibit high long-range contact order (\cref{fig:3E}), indicating complex, well-connected topologies rather than trivial folds often seen in \textit{de novo} designs. \cite{listov_opportunities_2024} These properties extend to conditional generation (\cref{fig:cond_aa_distribution,fig:cond_ss_distribution,fig:cond_contact_distribution,fig:cond_surface_distribution,fig:cond_rama_distribution,fig:cond_rog_distribution}).

We hypothesized that jointly generating sequence and structure allows the designed binding site to adapt its chemistry and geometry as the target adjusts its conformation, yielding a mutually compatible interface. This is a capability unique to co-design, allowing \nameshort to construct chemically appropriate microenvironments without inverse folding or any pre-specified residue motif. Indeed, ligand-conditioned generation yields pockets that are quantitatively responsive to ligand identity: binding-site lipophilicity correlates with ligand hydrophobicity (\cref{fig:3I,fig:num_rot_vs_rmsd,fig:binding_site_residue_composition}), appropriate coordinating residues emerge for specific cofactors (\cref{fig:3J,fig:cond_chem_motif_examples}), and sufficiently large cavities form to avoid steric clashes (\cref{fig:steric_clashes}). Prior co-design models capture these physiochemical trends less consistently (\cref{fig:binding_site_method_comparison}), which aligns with their lower co-designability and diversity. Furthermore, \nameshort samples valid ligand conformers beyond the reference input (\cref{fig:3K,fig:3L,fig:3M,fig:cond_rog_distribution,fig:cond_conf_examples}), by selectively exploring rotatable bonds while preserving rigid geometries. 

These complementary pockets generated by \nameshort exhibit striking novelty and diversity. For coordination spheres up to ten residues, motif similarity searches using Folddisco \cite{kim2025structural} show that most binding motifs in co-designable generations lack natural homologs in AlphaFoldDB (novel motif defined as no match or RMSD > 3 Å; \cref{fig:3F,fig:3G,fig:cond_novel_motif_examples}). Clustering the generated motifs via pairwise distance shows over 90\% cluster diversity (\cref{fig:binding_site_diversity_cluster,fig:binding_site_diversity_k5}). \nameshort also generates globally novel sequences, and many folds have no close homologs in the Protein Data Bank (\cref{fig:3H}). Such novelty and diversity are only relevant when the motifs are foldable and physicochemically realistic – standards \nameshort meet while expanding well beyond the space of known global scaffolds and local motifs.

\section*{Designing enzymes for new-to-nature biocatalysis}

Enzyme design is among the most challenging applications for conditional protein generation, especially for new-to-nature catalytic activities that lie outside the model’s training distribution. Carbene transfer reactions, unexplored in known natural enzymes, have proven valuable for expanding the catalytic repertoire of biology. \cite{yang_navigating_2021} While directed evolution of cytochrome P450, cytochrome \textit{c}, and various globin variants has produced numerous carbene transferases, identifying initial activity for a desired reaction remains challenging and laborious, usually limited to screening a limited set of known scaffolds. Previous computational approaches have designed heme enzymes for only a handful of carbene reactions using physics-based theozymes and pre-specified active-site residue geometries \cite{kalvet_design_2023,hou_novo_2025,huang_novo_2025}—a strategy that is expensive, biased by the choice of motif, and difficult to generalize, especially when the catalytic mechanism is not known. We hypothesized that \nameshort's ability to co-generate sequence and structure around arbitrary molecules could bypass residue motif or theozyme specification. 

Carbene-transfer reactions proceed via formation of an iron–carbenoid intermediate, most often generated by nitrogen extrusion from a diazo precursor. This intermediate subsequently transfers the carbene fragment to a substrate through diverse pathways, including alkene cyclopropanation, or insertion into C–H, B–H, Si–H, or N–H bonds. \cite{yang_navigating_2021,damiano_iron_2020} Formation of the heme–carbene intermediate is generally the rate-determining step (>20 kcal/mol, Fig.~S3). \cite{garcia-borras_origin_2021} Therefore, we conditioned \nameshort on DFT-derived geometries and bonding patterns of the heme-carbene precursor complex rather than on a complete transition-state model. This choice reflects a deliberate simplification: rather than fixing an exact transition state calculated \textit{in vacuo}, we let \nameshort's co-folding mechanism sample conformations of the reactive intermediate that would be compatible with the protein it is constructing. This allows us to bypass the theozyme-based design entirely, where the limited mechanistic understanding for new-to-nature reactions may preclude the design of an accurate theozyme. \nameshort is free to explore catalytic solutions without being constrained by human assumptions about the protein residues required for activity or a fixed transition state geometry.  

From $\sim10^4$ \textit{in silico} generated sequence-structure pairs – one to two orders of magnitude fewer than recent structure-based pipelines \cite{ahern_atom-level_2026,butcher2025novo}  – we applied a computational filter (\cref{fig:1D}) combining predicted structures (AlphaFold 3 \cite{abramson_accurate_2024} and Chai-1 \cite{chai2024chai}), confidence measures such as chain pAE and ipTM, active-site contact counts, solvent exposure, net charge, and exposed hydrophobicity, yielding 90 designs for experimental testing (\cref{sec:filtering}). Neither the sequences nor the structures were redesigned after generation. Computational analysis confirms that the designed active sites represent new residue geometries (\cref{fig:3F}, >80\% with >3 Å RMSD in the five nearest residues). Moreover, while some designs closely recapitulate the reference geometry (RMSD < 1 Å), a substantial fraction adopt more divergent conformations (RMSD > 2 Å), primarily through rotations about the iron–carbenoid bond (\cref{fig:3L,fig:3M}). This indicates that \nameshort actively explores conformational space even for reaction intermediates. The designed enzymes are highly diverse: no pair shares more than 50\% sequence identity, and the 90 designs form 75 distinct structural clusters (TM-score < 0.5). 

To assess the catalytic generality, we selected four substrates spanning distinct carbene transfer reactions: \textit{p}-methoxystyrene (\textbf{1a}) with ethyl diazoacetate (EDA, \textbf{5a}) for cyclopropanation, \cite{coelho_olefin_2013} 1,3-dimethylimidazol-2-ylidene borane (\textbf{2a}) with ethyl 2-diazopropanoate (EDP, \textbf{5b}) for B–H insertion, \cite{kan_genetically_2017} 1-phenylpyrrolidine (\textbf{3a}) with \textbf{5a} for C(sp$^3$)–H insertion, \cite{zhang2019enzymatic} and tert-butyl-3-methyleneazetidine-1-carboxylate (\textbf{4a}) with \textbf{5a} for spirocyclopropanation of a pharmaceutically relevant scaffold. \cite{kennemur2025enzymatic} 

\begin{figure}
	\centering
	\includegraphics[width=1.0\textwidth]{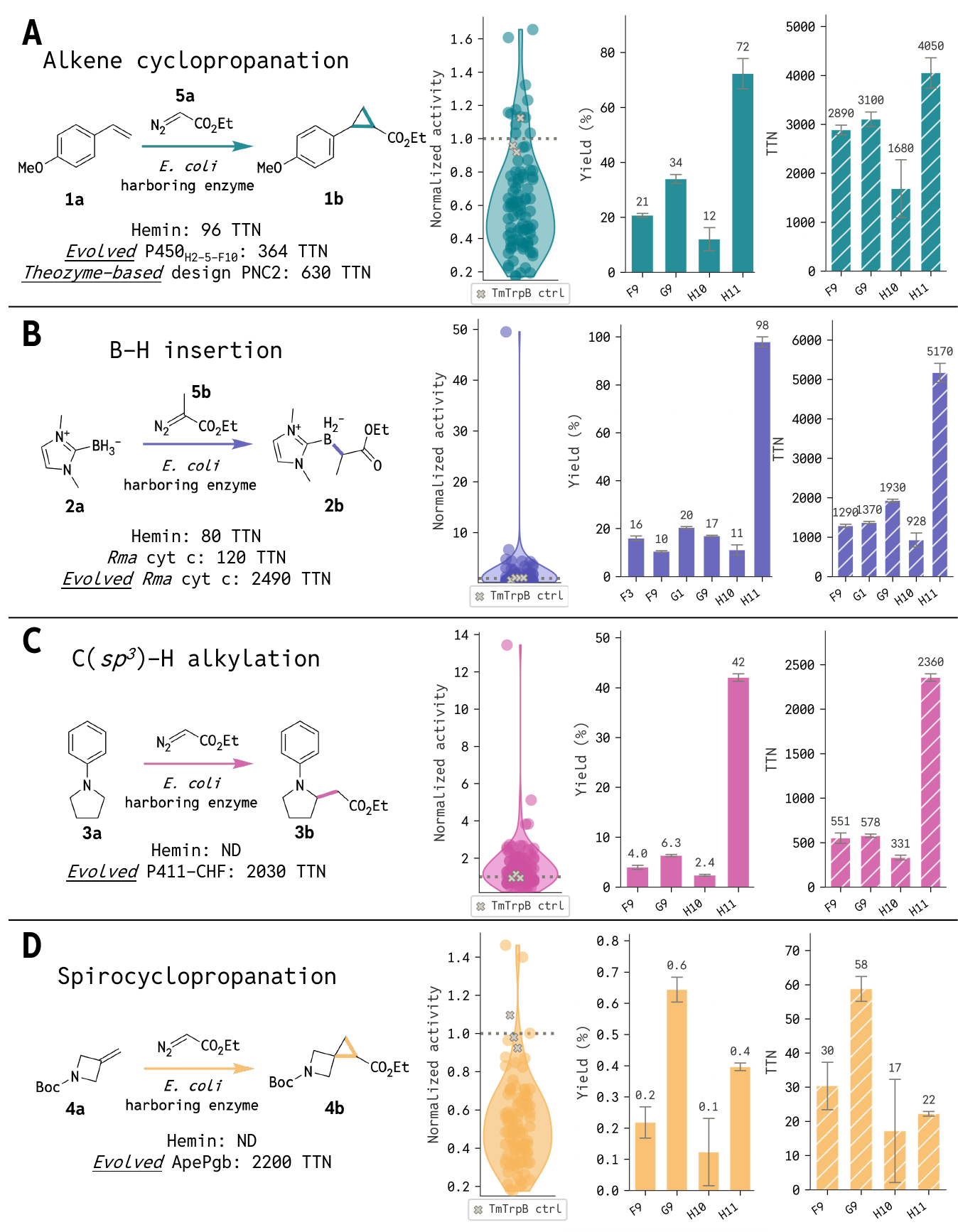}
    {\phantomsubcaption\label{fig:4A}}
    {\phantomsubcaption\label{fig:4B}}
    {\phantomsubcaption\label{fig:4C}}
    {\phantomsubcaption\label{fig:4D}}
	\caption{\textbf{\nameshort-designed carbene transferases (dCTs) catalyze multiple carbene transfer reactions with high activity at room temperature,} including (\textbf{A}) \textit{p}-methoxystyrene cyclopropanation. (\textbf{B}) B–H bond insertion of carbenes. (\textbf{C}) C(sp$^3$)–H alkylation. (\textbf{D}) Spirocyclopropanation of an unsaturated exocyclic N-heterocycle. Each panel shows a reaction scheme, results for hemin/previous evolved variants, plate-based screening data, and yield and TTN results from selected validations. ND: Not Detectable.}
	\label{fig:4}
\end{figure}

All 90 designs were screened in whole-cell format in \textit{Escherichia coli} across all four reactions. To control for background reactivity from endogenous hemoproteins, we overexpressed an unrelated protein (TrpB from \textit{Thermotoga maritima}) under identical conditions. Top hits were selected for validation (\cref{fig:4}). Several reactions exhibited high background activity from endogenous \textit{E.\ coli} hemoproteins, which complicated identification of lower-activity designs, especially since overexpressing these designs may disrupt heme homeostasis and further reduce their activity. We note that hemin alone has no detectable activity for the spirocyclopropanation or the C(sp$^3$)–H insertion reaction. 

Remarkably, screening across our selected carbene reactions confirmed that the designed variants were not only functional (2.2 – 66\% exceeded the TmTrpB control) but, in several cases, exhibited high catalytic activity. For the cyclopropanation of \textit{p}-methoxystyrene, the best design achieved 72\% yield and a total turnover number (TTN) of 4,050 with a 99:1 diastereomeric ratio (\cref{fig:dr_styrene}), surpassing both early evolved P411 enzymes and the recently reported designed enzyme PNC2 which scaffolded a helix bundle around a porphyrin-based theozyme (\cref{fig:4A}). \cite{coelho_olefin_2013,hou_novo_2025} For B–H insertion, the top variant achieved 98\% yield and 5,170 TTN (\cref{fig:4B}), far exceeding both the prior starting point (120 TTN) and the laboratory-evolved variant (2490 TTN). \cite{kan_genetically_2017} Most notably, for C(sp$^3$)–H insertion – a highly challenging transformation whose previous engineering campaign required 14 rounds of directed evolution and for which mechanistic uncertainty precludes theozyme construction – our best design reached 42\% yield and 2,360 TTN for the alkylation of 1-phenylpyrrolidine, rivaling the performance of previously evolved P411-CHF catalysts on this substrate (2,030 TTN) (\cref{fig:4C}). \cite{zhang2019enzymatic} The sterically and electronically demanding spirocyclopropanation proved more challenging, \cite{kennemur2025enzymatic} yielding fewer active variants with modest activity. Although enantioselectivity was not a design criterion, modest enantioselectivities were observed across all reactions, with enantiomeric excesses up to 35\% for the spirocyclopropanation (\cref{fig:ee_styrene,fig:ee_bh,fig:ee_ch}). Notably, enzymes with a preference for either enantiomer were identified for three reactions.   

\begin{figure}[t!]
	\centering
	\includegraphics[width=1.0\textwidth]{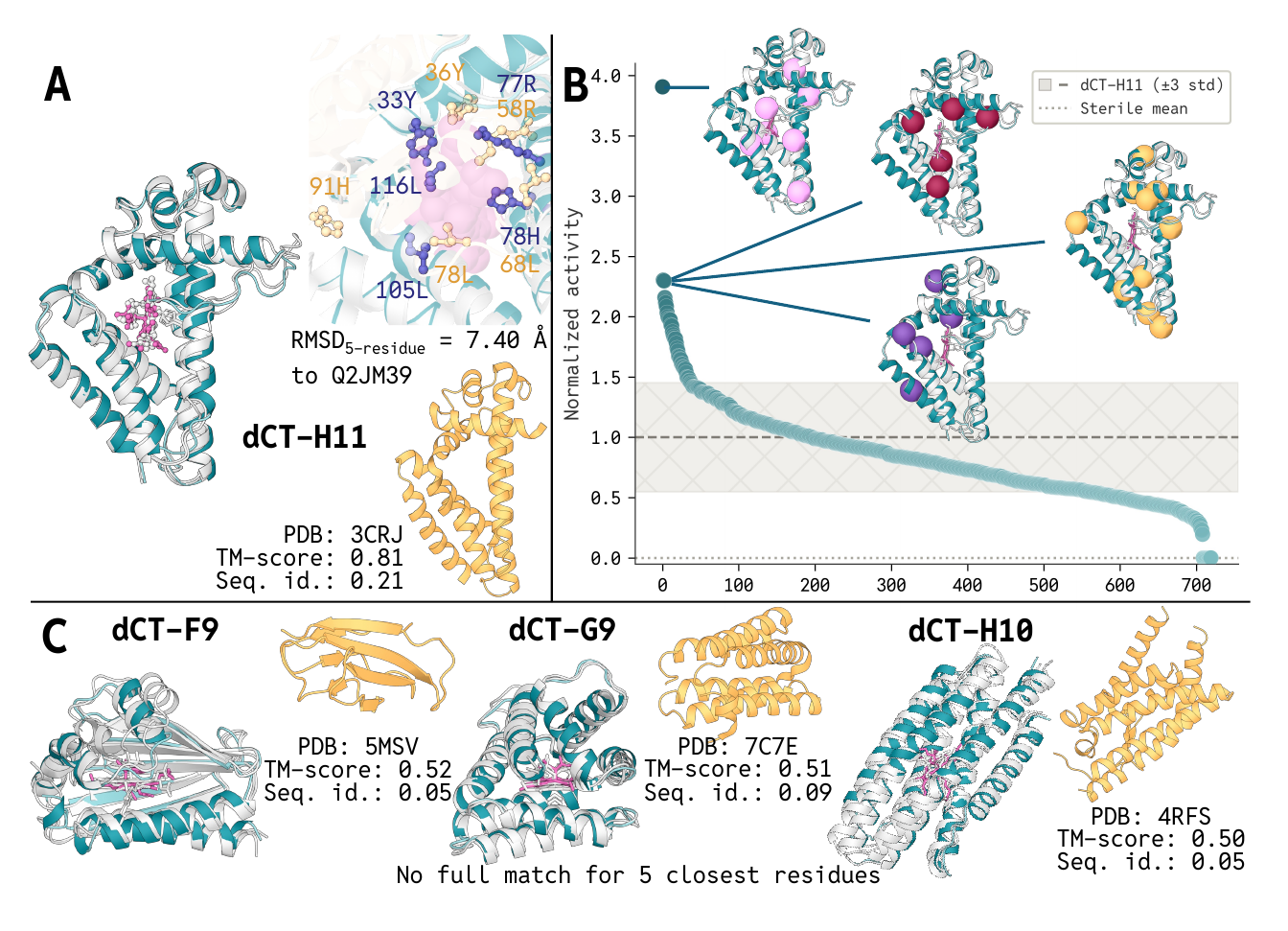}
    {\phantomsubcaption\label{fig:5A}}
    {\phantomsubcaption\label{fig:5B}}
    {\phantomsubcaption\label{fig:5C}}
	\caption{\textbf{ Novelty of selected, designed enzymes and evaluation of evolvability.} (\textbf{A}) dCT-H11 design (green/pink) structure and the AlphaFold 3/Chai predicted structures (grey), alongside the closest PDB match by TM-score. The inset shows the five-residue active-site arrangement predicted by AlphaFold 3 (purple), compared to the closest motif in AlphaFoldDB (yellow). (\textbf{B}) Mutational landscape of error-prone PCR variants in plate-based screening of dCT-H11 for the spirocyclopropanation reaction. (\textbf{C}) Designed and predicted structures of dCT-F9, dCT-G9, dCT-H10, alongside the closest PDB matches by TM-score. The five closest active-site residues yielded no full match in AlphaFoldDB. }
	\label{fig:5}
\end{figure}

To probe the evolvability of \nameshort's designs, we selected the spirocyclopropanation reaction, where design alone had yielded only modest activity. We subjected the designed carbene transferase dCT-H11 to a single round of error-prone PCR and screened approximately 700 mutants (\cref{fig:5B}). Around 35 variants displayed noticeably improved activity relative to the parent enzyme with divergent enantioselectivity, either enhancing the parent's preference (from +35\% to +49\% \textit{ee}) or inverting it entirely (+35\% to -35\% \textit{ee}, \cref{fig:ee_azetidine}). The top variants harbor substitutions scattered across the protein, consistent with the long-range epistatic effects characteristic of natural enzyme evolution. These results suggest that \nameshort's designs can occupy regions of sequence space with accessible uphill paths for further optimization.  

The top hits span diverse folds with novel motifs (\cref{fig:5}). The closest structural match to dCT-H11 is a TetR-family transcription factor from the extremophile \textit{Haloarcula marismortui} (PDB ID: 3CRJ, TM-score: 0.81) that lacks known catalytic activity. dCT-H11 shares only 21\% sequence identity with 3CRJ, suggesting that \nameshort repurposed a topology not pre-specified for catalysis. The most similar residue motif to the five closest active-site residues, drawn from AlphaFoldDB, shows a large mismatch (RMSD > 7 Å), confirming that the geometry of the catalytic residues is itself novel. Other designs are even more divergent: dCT-F9 (closest match: 5MSV, TM-score 0.52, 5\% sequence identity) and dCT-G9 (closest match: 7C7E, TM-score 0.51, 9\% sequence identity) adopt folds that are substantially more remote from known structures, with no corresponding motif identified in AlphaFoldDB based on their five nearest active-site residues (\cref{fig:5C}). Notably, none of the closest structural matches correspond to naturally heme-binding proteins, highlighting the ability of the model to capture underlying biochemical principles and repurpose unrelated protein folds for heme binding and carbene transfer reactions. Together, these results demonstrate that the model can create novel proteins with emergent functions, extending beyond the boundaries of natural protein space.  

\section*{General biomolecular design unlocks new-to-nature reactivities} % diverse

These results establish a general platform for functional protein design in which sequence and 3D structure are generated as a coupled object and flexibly conditioned on any biomolecules. Computationally, careful inference alignment across modalities via cross-modal recycling, self-correction, and entropy regularization enables \nameshort to generate diverse, high-quality proteins that outperform existing co-generation methods, while inference-time steering on both sequence and structure turns generative sampling into a controllable search process. Experimentally, conditioning only on reaction intermediates was sufficient to produce promiscuous carbene transferases across four new-to-nature reactions, bypassing transition-state calculations, theozyme scaffolding, and extensive wet-lab screening. From just 90 genes, top designs achieved 98\% yield for B–H insertion and >2,300 TTN for C(sp$^3$)–H insertion, exceeding previously evolved biocatalysts. Importantly, the designed active-site residues are novel, exhibit stereoselectivity, and occupy accessible fitness landscapes that support subsequent evolution. 

The ability to genetically encode arbitrary chemical transformations requires global discovery with local optimization. \nameshort provides a global search for diverse, evolvable enzymes. Yet the reactions explored here represent only a small subset of synthetically valuable transformations not found in nature. Mastering increasingly complex mechanisms likely requires closing the loop between generative discovery and directed evolution, informed by richer biophysical constraints. More broadly, co-designing proteins with arbitrary biomolecular partners charts a path toward programmable biological systems whose complexity rivals that of natural evolution, and whose function serves human intent. 

%%%%%%%%%%%%%%%% REFERENCES %%%%%%%%%%%%%%%

\clearpage % Clear all remaining figures and tables then start a new page

% The list of references goes after the main text and before the acknowledgements
% When preparing an initial submission, we recommend you use BibTeX, like this:
%
\bibliography{zotero,main} % for a file named science_template.bib

\clearpage
\section*{Acknowledgments}
The authors thank Sabine Brinkmann-Chen, Olexa Bilaniuk, Almer van der Sloot, Luca Ambrogioni, Jessica Wu, Dejan Stančević, Danyal Rehman, Guillaume Huguet, Ella Miray Rajaonson, Fred Zhangzhi Peng, Andrei Rekesh, Ariane Mora, Julia Reisenbauer for very fruitful discussions and Ziyan Zhang for help with compound synthesis. 
\paragraph*{Funding}
The authors acknowledge funding from Amgen, AMD, Wendy and Eric Schmidt Foundation, Amazon Web Services, the Natural Sciences and Engineering Research Council of Canada, National Research Council Canada, Microsoft, IVADO, and Canada First Research Excellence Fund. We thank Twist Bioscience for support with gene fragment synthesis. The work is enabled by high-performance computing resources at Mila, AMD, and the Digital Research Alliance of Canada. J.R.B. is supported by the St-Pierre-Larochelle Scholarship in AI for the Environment. D.R. thanks the Alexander von Humboldt Foundation for a Feodor Lynen Postdoctoral Fellowship. M.S. is supported by the IVADO 2025 Postdoctoral Research Funding Program.
\paragraph*{Author contributions}
All authors participated in the review and editing of the paper.  
\begin{itemize}
\item Conceptualization: CHL, JRB, AT, FHA 
\item Model development: JRB, CHL, AT, MS, MC, NZ, AJB  
\item Inference steering: MS, JRB, KN, CHL 
\item Benchmarks, ablations, and baselines: JRB, CHL, MC, JY, EJ  
\item Design filtering: CHL, FZL 
\item Experimental characterization: TL, DR, YL, ZL, TG 
\item Funding acquisition: FHA, CHL, YB 
\item Supervision: CHL, FHA, AT, YB, KN 
\item Writing – original draft: CHL, JRB, TL, MS 
\end{itemize}

\paragraph*{Competing interests}
There are no competing interests to declare.
\paragraph*{Data and materials availability}

Code and model checkpoints associated with this paper are available at \url{https://github.com/DISCO-design/DISCO}. Analyzed results, including protein sequence, structure, and functional outcomes, will be available in the supplementary information or in an online database. Lyophilized cell powders are available upon request under a materials transfer agreement.
%% ═══════════════════════════════════════════════════
%%  SUPPLEMENTARY INFORMATION
%% ═══════════════════════════════════════════════════
\clearpage
\appendix

% Reset counters for supplement
\renewcommand{\thefigure}{S\arabic{figure}}
\renewcommand{\thetable}{S\arabic{table}}
\renewcommand{\theequation}{S\arabic{equation}}
\setcounter{figure}{0}
\setcounter{table}{0}
\setcounter{equation}{0}

% ── Supplement title banner ──
\begin{center}
  %{\color{DBlue2}\rule{\textwidth}{0.6pt}}\\[0.8em]
  {\Large\bfseries\color{DBlue4} Supplementary Information}\\[0.3em]
  {\large General Multimodal Protein Design Enables DNA-Encoding of Chemistry}\\[0.3em]
  {\color{DPink3}\rule{0.3\textwidth}{0.4pt}}
\end{center}
\vspace{1em}
\phantomsection\hypertarget{toc}{}
\tableofcontents

\clearpage
\pagestyle{toc}

\section{\textit{In silico} methods and results}

\subsection{Notation}

Denote the space of real numbers in dimension $d$ as $\mathbb{R}^d$. Let $\gV = \{1, \dots, V\}$ be a finite vocabulary set. We designate the final element of this set as a specialized mask token $V=\mb{m}$, whereas the remaining $V-1$ elements in $\gV$ form the set of natural amino acids, including one unknown amino acid UNK.  We use the diffusion time convention where time $\tau=0$ is associated with the data distribution $p_{\mathrm{data}}$ and $\tau=1$ is associated with $p_{noise}$. We describe proteins with $L$ residues with backbone coordinates $x^\struct$ and sequence $x^\seq$ as $x = (x^\struct, x^\seq) \in (\mathcal{X}^\struct, \mathcal{X}^\seq) := (\mathbb{R}^{L \times 4 \times 3}, \mathcal{V}^L) = \mathcal{X}$ where $\mathcal{X}$ denotes the space of protein backbone coordinates with associated amino acid types. However, for simplicity, whenever it is clear from context, we also omit superscripts ${}^\struct$ and ${}^\seq$ to avoid notational clutter. 

We use $\tau$ to denote inference time, with $t(\tau)$ to denote structure time and $r(\tau)$ to denote sequence time. We also use the shorthand $t$ to denote $t(\tau)$ and $r$ to denote $r(\tau)$.

Let $\Delta^d\coloneqq \lbrace v\in \mathbb{R}^d:v^i\geq 0,i=1,\dots, d,\sum_{i=1}^d v^i=1\rbrace$ represent the $d$-dimensional probability simplex. Each point on $u \in \Delta^{d}$ corresponds to a categorical distribution $\text{Cat}(j; u)=u^j$ for $j\in \gV$. Let $\delta(x) \in \Delta^d$ be the one-hot distribution that places all its mass on $x$.
The data distribution $p_{\mathrm{data}}$ is provided as an empirical distribution on $n$ sequences in the form of a training set $\gD = \{ x\}^n\subset \mathcal{X}$. We denote for $x\in \gV$, $\delta(x)\in\Delta^{d}$ given by $\text{Cat}(j; \delta(x))=1$ if $j=x$ and $0$ otherwise. Finally, we reserve superscripts for set indexing purposes, e.g.\ $x^i, i \in[d]$, while subscripts are used to represent positions in time of a discrete sample $x_r^\seq, r \in[0,1]$.

\subsection{Theoretical foundations}

For a multi-modal space where elements contain multiple components, we describe the theoretical framework for modeling such systems, first individually, then in combination.

\subsubsection{Continuous-space diffusion modeling}

Generative modeling in continuous space systems can be formulated as the simulation of a Stochastic Differential Equation (SDE) forward in time to destroy information, then reverse in time to generate new samples from the target distribution. 

In particular, during training, information is gradually destroyed from the training dataset by simulation with samples $x \sim p_{\mathrm{data}}(x)$ using the following SDE:
\begin{equation}\label{eq:forward_continuous_diffusion}
    d x_t = f_t(x_t) dt + g_t d W_t, \quad x_{t=0} \sim p_{\mathrm{data}}(x)
\end{equation}
where $f_t(x_t): [0,1] \times \mathcal{X}^\struct \to \mathcal{X}^\struct$ is generally some pre-defined linear \textit{drift}, $W_t$ is the standard Wiener process, $g_t \in \mathbb{R}_{\ge 0}$ is the diffusion coefficient, a scaling factor for the noise. The drift $f_t$ and the diffusion $g_t$ are chosen such that at $p_{t=1} \approx \mathcal{N}(0, I)$.

Given this forward process, a reverse-time denoising SDE can be defined in the opposite direction,
\begin{equation}
    d x_t = \left (- f_t(x_t) + g_t^2 \nabla \log p_t(x_t) \right ) dt + g_t d W_t, \quad x_{t=1} \sim \mathcal{N}(0, I)
\end{equation}
where $p_t$ is the marginal distribution induced by the stochastic process in \eqref{eq:forward_continuous_diffusion}. By training a model of the score function $\nabla \log p_t(x_t)$, the SDE can be reversed to generate points approximately from the data distribution. This leads to a natural parameterization using a denoising network $D_\theta(x_t, t)$ which can be used to generate new samples using the following reverse-time denoising SDE:
\begin{equation}
    d x_t = \left (- f_t(x_t) + g_t^2 \frac{D_\theta(x_t, t) - x_t}{\sigma_t^2} \right ) dt + g_t d W_t, \quad x_{t=1} \sim \mathcal{N}(0, I)
\end{equation}
where $\sigma_t^2 = \int_0^t g_r^2 dr$ represents the total noise of the system at time $t$. It is possible to learn $D_\theta(x_t)$ using the standard denoising score matching loss \cite{ho2020denoisingdiffusionprobabilisticmodels}:
\begin{equation}\label{eq:continuous_loss}
    \mathbb{E}_{t, x_0 \sim p_{\mathrm{data}}, x_t \sim p_t(x_t | x_0)} \left [\frac{1}{\sigma_t^2} \left \| D_\theta(x_t, t) - x_0 \right \|_2^2 \right ]
\end{equation}

\subsubsection{Discrete-Space diffusion modeling}

In the discrete-space setting, we require a slightly different treatment of discrete variables. We follow the discrete diffusion literature, which defines a similar noising and denoising process based on a masking process. In this case, continuous-time Markov chains (CTMC) or jump processes describe transitions on the discrete state spaces. In this section, we focus on the discrete time setting where time is divided up into $T$ sub-intervals such that $\tilde r \in [0, T)$ represents normalized time $r := \tilde r /T$. This process can also be described in the fully continuous-time setting, which requires more rigorous treatment and can be found in related works such as \cite{peng2025pathplanningmaskeddiffusion,rojas2025diffuseeverythingmultimodaldiffusion}.

We focus on ``masked'' discrete diffusion modeling where information is destroyed by transitioning to a special mask token $\bf{m}$ independently for each $x_r^i \neq \bf{m}$:
\begin{equation}
    \label{eq:forward_discrete_diffusion}
    p_r (x_r | x_0) = \prod_{i=1}^L p_t(x_r^{i}|x_0^{i}) = \prod_{i=1}^L\text{Cat}(x^i_r; \alpha_r \delta(x^i_0) + (1 - \alpha_r) \delta(\mathbf m)).
\end{equation}

where $\alpha_r$ describes the noise schedule similarly to $g_t$ in the continuous case, and is specified such that $\alpha_0 = 1$ and $\alpha_1 = 0$. This describes a process where at time $r$ a fraction $\alpha_r$ of the tokens are in their original state and $1 - \alpha_r$ of the tokens are in the masked state. The simplest conditional reverse transition kernel for a token $x_r^i$ is described by:
\begin{equation}
\label{eqn:posterior_mdlm}
q_t(x^i_{r-1} | x^i_r, x^i_0) =
    \begin{cases}   
        \text{Cat}(x^i_{r-1}; \delta (x^i_r)) &  x^i_r \neq \mathbf m \\
        \text{Cat}\left (x^i_{r-1}; \frac{(1 - \alpha_{r-1}) \delta(\mathbf{m}) + (\alpha_{r-1} - \alpha_r) \delta(x^i_0)}{1-\alpha_r} \right) & x^i_r = \mathbf{m}. 
    \end{cases}    
\end{equation}

This leads to a natural parameterization for the transitions of a discrete time generative model. We first define a (possibly) time-dependent denoiser network $D_{r, \theta}: \gV^L \times [0,1] \to \Delta^L$ that predicts the probabilities of a clean sample $x_0 \sim D_{r,\theta}(x_r)$:
\begin{equation}
\label{eqn:parametrized_posterior_mdlm}
p_{r, \theta}(x^i_{r-1} | x^i_r, D^i_{r, \theta}(x_r)) =  
    \begin{cases}   
        \text{Cat}(x^i_{r-1}; \delta (x^i_r)) &  x^i_r \neq \mathbf m \\
        \text{Cat}\left (x^i_{r-1}; \frac{(1 - \alpha_{r-1}) \delta(\mathbf{m}) + (\alpha_{r-1} - \alpha_t) D^i_{r, \theta}(x_r)}{1-\alpha_r} \right) & x^i_r = \mathbf{m}. 
    \end{cases}    
\end{equation}
where $D_\theta^i$ represents the predicted distribution over tokens for the $i$-th index of the sequence. Using this reverse parameterization, as $T \to \infty$, there exists an evidence lower bound (ELBO) to the log likelihood of the data distribution:
\begin{equation}
\label{eq:MDMELBO}
    \log p_{\theta}(x_0) \geq -\int^1_0 \frac{d\alpha_r}{d r}\cdot \frac{1}{1 - \alpha_r}\mathbb{E}_{x_r \sim \mb{p}_{r}(\cdot | x_0)} \left[ \sum_{i=1,x_r^i=\mb{m}}^L \delta(x_0^i)^T\log D^i_{r, \theta}(x_r) \right] dr
\end{equation}
This can be used to train $D_\theta$ using standard stochastic gradient-descent type algorithms. It can be thought of as training a masked discrete diffusion model by optimizing a weighted cross-entropy loss over masking levels $\alpha_r$ weighted by $\frac{d\alpha_r}{d r}\cdot \frac{1}{1 - \alpha_r}$.

However, there exist several major limitations of the denoising process in \eqref{eqn:posterior_mdlm}, which is that as $T \to \infty$,
\begin{enumerate}
    \item \textbf{Lack of Order Control}. An analytic Gillespie-style sampler \cite{GILLESPIE1976403} shows that the denoising proceeds by uniformly randomly selecting a masked position, offering no control over the generation order of tokens.
    \item \textbf{Token Commitment}. \eqref{eqn:parametrized_posterior_mdlm} reveals that once unmasked, a specific token is committed to and never revisited. This means that if ``mistakes'' are made in the generation process, they can never be corrected.
\end{enumerate}

These two issues have led to a variety of solutions involving alternative reverse-time processes for discrete generation. We use a \textit{path planning} (P2) approach \cite{peng2025pathplanningmaskeddiffusion} which allows for ``planned'' unmasking and remasking of tokens during the diffusion process. 

P2 departs from the standard MDM inference procedure in \eqref{eqn:parametrized_posterior_mdlm} where the reverse-time transition $p_{r, \theta}(x_{r-1}^i | x_r^i, D_{r, \theta}^i(x_r))$ is independently sampled for each index by instead assigning the likelihood of denoising index $i$ as a function of a separate planner $G_U$ which represents the probability of unmasking each index. This addresses limitation (1) of the standard denoising process. To address limitation (2), P2 also includes a remasking planner $G_M$ which represents the probability of an already unmasked token being first remasked then denoised to (potentially) a new value. This can be described with the following transition kernel, where $z \sim D_\theta(x_tr)$ is a sampled instantiation from the denoiser, and $\bar{x}_r$ represents $x_r$ after remasking according to $G_M$.

\begin{equation}
\label{eqn:p2_reverse}
p_{r, \theta}(x^i_{r-1}| x_r, z ) =  
    \begin{cases}   
        \text{Cat}\left (x^i_{r-1}; \frac{ \alpha_{r-1} - \alpha_r }{1-\alpha_r}G^i_{U}(z,x_r)\delta(z^i) \right) & x^i_r = \mathbf{m} \\
        \text{Cat}\left(x^i_{r-1};  \frac{(\alpha_{r-1} - \alpha_r)G^i_{M}(z, x_r)}{(1-\alpha_r)(1-\text{Cat}(x^i_r,D^{i}_{\theta}(\bar{x}_r)))}D^i_{\theta}(\bar{x}_r)  \right)&  x^i_r \neq \mathbf m, \\
    \end{cases}
\end{equation}
in practice, we use a uniform $G_U$ and $G_M$, as we find that a $G_M$ which depends on the confidence of the 
$D_\theta$ reduces structural diversity.

\subsubsection{Multi-modal diffusion modeling}
Having described separate processes for structure and sequence, it now remains to combine them to perform cogeneration in \namelong (\nameshort).

We describe a protein $x$ with $L$ residues with sequence component $x^\seq \in \mathcal{V}^L$ and backbone structure component $x^\struct \in (\R^{4})^L$ as $x = (x^\seq, x^\struct)$. A noisy protein with sequence noise level $r$ and structure noise level $t$ is then denoted as $x_{r, t} = (x_r^\seq,x_t^\struct)$. We define the conditional forward process $p_{r, t|0}(x_{r,t} | x_0)$ to factorize over modality
\begin{equation}
    p_{r, t | 0} ( x_{r,t} | x_0) := p_{r |0}(x^{\seq}_r | x^{\seq}_0)  p_{t |0}(x^{\struct}_t | x^{\struct}_0)  
\end{equation}
where $p_{r |0}(x^{\seq}_r | x^{i, \seq}_0)$ and $p_{t |0}(x^{\struct}_t | x^{\struct}_0)  $ are defined similarly to the individual cases i.e.
\begin{equation}
    p_{r |0}(x^{\seq}_r | x^{\seq}_0) := \prod_{i=1}^L \text{Cat}(x^{i, \seq}_r; \alpha_r \delta(x^{\seq}_0) + (1 - \alpha_r) \delta(\mathbf m))
\end{equation}
and $p_{t |0}(x^{\struct}_t | x^{\struct}_0)$ is defined implicitly through simulation of the forward SDE
\begin{equation}
    d x^\struct_t = f_t(x^\struct_t) dt + g_t d W_t, \quad x^\struct_{t=0} = x^\struct_0.
\end{equation}

To reverse this process, we need to define a reverse over both spaces that depends on increments $\Delta_s$ and $\Delta_t$.
\begin{equation}
    p(x_{r-\Delta_r, t - \Delta_t} | x_{r, t}) = p_{r - \Delta_r} (x^\seq_{r - \Delta_r} | x_{r, t}) p_{t - \Delta_t} (x^\struct_{t - \Delta_t} | x_{r, t})
\end{equation}
where $p_{t - \Delta_t} (x^\struct_{t - \Delta_t} | x_{r, t})$ is governed by the reverse-time SDE
\begin{equation}
    d x_t^\struct = \left (- f_t(x^\struct_t) + g_t^2 \nabla_{x^\struct_t} \log p_t(x_{r,t}) \right ) dt + g_t d W_t
\end{equation}
and 
$p(x^\seq_{r - \Delta_r} | x_{r,t}) = \prod_{i=1}^L p(x^{i, \seq}_{r-\Delta_r}| x_{r,t})$
with

These equations govern the dynamics of the reverse process, but we still need to parameterize both the structure score relative to the partially noised $x_{r,t}$ as well as the sequence denoiser relative to the same state. To do this, we use a single model $D_\theta(x_{r,t})$. This allows us to perform inference along any trajectory of $r, t$ pairs from $(1,1)$ to $(0,0)$ for valid generation up to numerical and learning errors. We use the following generative process given our denoising model $D_\theta$ with structure output component $D^\struct_\theta$ and sequence logit output component $D^\seq_\theta$: 
\begin{align}
    p_{s - \Delta_s, \theta}(x^i_{r-\Delta_r}| x_{r,t}, z ) =  
    \begin{cases}   
        \text{Cat}\left (x^i_{r-1}; \frac{ \alpha_{r-1} - \alpha_r }{1-\alpha_r}(z,x_{r,t})\delta(x_r^i) \right) & x^{i,\seq}_{r,t} = \mathbf{m} \\
        \text{Cat}\left(x^i_{r-1};  \frac{(\alpha_{r-1} - \alpha_r)G^i_{M}(\mathbf{z},x_{r,t})}{(1-\alpha_r)(1-\text{Cat}(x^i_r, D^{i, \seq}_{\theta}(x_{r,t}, r, t))}D^{i, \seq}_{\theta}(x_{r,t}, r, t)  \right)&  x^i_r \neq \mathbf m, \\
    \end{cases}
    \label{eq:seq_update}
\end{align}

\begin{equation}
    d x_t^\struct = \left (- f_t(x^\struct_t) + g_t^2 \frac{D^\struct_\theta(x_{r,t}, r, t) - x^\struct}{\sigma_t^2} \right ) dt + g_t d W_t
    \label{eq:struct_update}
\end{equation}
Finally, it remains to define the training scheme for $D_\theta$. Here, we combine independent losses and noising for each component individually. Specifically, we define independent distributions and losses over separate domains as in \eqref{eq:continuous_loss} and \eqref{eq:MDMELBO}, but replace the modality-specific denoisers with our joint denoiser $D_\theta$, which operates on a noisy input containing both modalities:
\begin{equation} \mathbb{E}_{r, t, x \sim p_{\mathrm{data}} x_{r,t} \sim p(x_{r,t} | x)} \!\left [ \frac{1}{\sigma^2_t}\| D^\struct_\theta(x_{r,t})\!-\!x^\struct \|^2 \!-\! \left ( \frac{d \alpha_r}{dr} \cdot \frac{1}{1 \!-\! \alpha_r} \!\sum_{\substack{i=1 \\ x_t^{i, \seq} \!= \mathbf{m}}}^L \delta(x^\seq)^T \log D^{i,\seq}_\theta(x_{r,t})\right )\!\right]
\end{equation}

These tools provide the necessary theory to perform multimodal training and inference. While theoretically any schedule over $r, t$ can be used, we lock together the two schedules during inference only by defining the schedules in terms of a unifying time variable $\tau$.

% In this section, consider a distribution $p_{\mathrm{data}}(x, y)$ where $x \in \R^d, y \in \mathcal{V}^L$. Combining the ideas from the above continuous-space and discrete space noising processes leads us to the following forward process,
% \begin{equation}
% \begin{cases}
%     x_{t+1} = x_t + f_t(x_t) \Delta t + g_t \Delta W_n \\
%     \by_{t+1} = \text{Cat}(\by_{t+1}; \frac{\alpha_{t+1}}{\alpha_t} \by_t + (1 - \frac{\alpha_{t+1}}{\alpha_t}) \bf{m})
% \end{cases}
% \end{equation}

\subsection{Data pipeline}
We trained on the weighted PDB dataset described in AlphaFold 3 with a cutoff date of 2021-09-30. Data processing, including clustering and filtering, follows the procedures described in Sections 2.5.3 and 2.5.4 of the AlphaFold 3 Supplementary Material. The dataset was obtained by downloading with the Protenix \texttt{scripts/database/download\_training\_data.sh} script from commit \texttt{a4059a6894c826b7bb81ba753cbdb89029313244}. 

We sample either single chains or chain-pair interfaces, with interfaces defined as pairs of chains with minimum heavy atom separation less than $5 \angstrom$. For $r \in \{\textrm{chain, interface}\}$, the dataset is weighted as

\begin{equation*}
    w \propto \frac{\beta_r}{N_{\clust}}(\alpha_{\prot} * 
    n_{\prot} + \alpha_{\nuc}  * n_{\nuc} + \alpha_{\ligand}  * n_{\ligand})
\end{equation*}

with $\beta_{\textrm{chain}}=0.5$, $\beta_{\textrm{interface}}=1$, $\alpha_\prot=3$, $\alpha_\nuc=3$, $\alpha_\ligand=1$, and $N_\clust$ defined by the Protenix dataset clusters.

Cropping is performed per sample, where, as in AlphaFold 3 we use a contiguous crop, a spatial crop, and a spatial interface crop. For detailed algorithms, please refer to Section 2.7 of the AlphaFold 3 Supplementary Material \cite{abramson_accurate_2024}. We crop with the following weights: $20\%$ contiguous cropping, $40\%$ spatial cropping, and $40\%$ spatial interface cropping.

\subsection{Model architecture}
While our architecture is broadly based on AlphaFold 3, we make a number of changes to enable protein sequence and structure co-design. Unless otherwise specified, we make use of the AlphaFold 3 architectural components and the interested reader may refer to the AlphaFold 3 Supplementary Information for detailed module-wise algorithms. As in AlphaFold 3, our model takes the form of a conditional diffusion model. However, unlike AlphaFold 3, we remove a number of the conditioning modules and add other domain-specific modules. We begin by removing the Template Module from the architecture as well as the MSA Module as computing MSAs for an ever evolving protein sequence trajectory during inference is computationally intractable. To maintain a source of evolutionary information, we replace the MSA Module with a protein language model, in our case DPLM 650M \cite{wang2024diffusion}.We present the main model architecture in \cref{alg:denoiser}. 

Our tokenization scheme sees each amino acid residue and nucleotide correspond to a single token, while for other molecules, we encode each heavy atom as a separate token. Input features correspond to those used in AlphaFold 3, with the exception of all MSA and template features (see Table 5 in the AlphaFold 3 S.I. \cite{abramson_accurate_2024}). To obtain the initial trunk features, we feed the input features to the InputFeatureEmbedder (Algorithm 2 in AlphaFold 3 S.I.)

\subsubsection{Denoiser architecture}
The core of our model is the Denoiser (\cref{alg:denoiser}), which iteratively refines the single ($s_i$) and pair ($z_{ij}$) representations over a series of cycles ($N_{\text{cycle}} = 4$). We first derive the initial features, $\{s_i^{\text{inputs}}\}$, using the \texttt{InputFeatureEmbedder} based on the input features $\{f^*\}$. The initial single and pair representations, $s_i^{\text{init}} \in \mathbb{R}^{c_s}$ and $z_{ij}^{\text{init}} \in \mathbb{R}^{c_z}$, are generated via linear projections of the embedded inputs. We enrich the initial pair representations with relative position encodings and token bond features.

During each cycle $c \in [1, \dots, N_{\text{cycle}}]$, the pair representation is first updated using a layer-normalized linear projection. For cycles beyond the first ($c > 1$), the architecture employs a \texttt{CrossModalEncode} module to update representations based on the current sequence and structure states, $x_l^{\text{seq}}$ and $x_l^{\text{struct}}$, as well as skip connections. The model then integrates evolutionary information via the \texttt{PlmModule}, our drop-in replacement for the MSA Module of AlphaFold 3 and set to be DPLM 650M \cite{wang2024diffusion}, updating the pair representations using features from the underlying protein language model. To encode chain breaks as input for the pLM, we insert a length 25 glycine linker between all protein chains. To obtain the single representations, we average the hidden states for each transformer block, while for the pair representations, we concatenate all attention weights for each attention block. Following an update to the single representation $s_i$, both representations are processed through the \texttt{PairformerStack}. As we believe that for \textit{de novo} design we require a lighter pairformer than for typical folding tasks, we reduce the number of blocks in the Pairformer from 48 in AlphaFold 3 to 8. The outputs of the first cycle are cached as skip connections ($\{s_i^{\text{skip}}\}$ and $\{z_{ij}^{\text{skip}}\}$) for use in subsequent modules. 

After completing all cycles, a final \texttt{CrossModalEncode} is applied before passing the representations to the \texttt{DiffusionModule} to produce the denoised sequence and structure estimates, $\hat{x}_0^{\text{seq}}$ and $\hat{x}_0^{\text{struct}}$.

\subsubsection{Cross-modal encoding}
To inform the trunk representations with the current trajectory of the diffusion process and to better encourage self-consistency between both sequence and structure modalities, the \texttt{CrossModalEncode} module (Algorithm 2) integrates intermediate sequence and structure states. It first calls the \texttt{DiffusionModule} to obtain a preliminary estimate of the fully denoised states, $\hat{x}_0^{\text{seq}}$ and $\hat{x}_0^{\text{struct}}$. 

The noisy and denoised sequences are encoded using the frozen DPLM 650M, while the noised and denoised structures are encoded via a separate structure encoder, in our case a Message-Passing Neural Network following the LigandMPNN architecture \cite{dauparas2025atomic}, yielding embeddings $u^{\text{seq}}_0$, $u^{\text{seq}}_l$, $v^{\text{seq}}_0$, and $v^{\text{seq}}_l$, respectively. The structure encoder weights are initialized from LigandMPNN weights and subsequently trained end-to-end along with the rest of the weights in the overall model, with the structure encoder embeddings being passed through a final layer norm and linear layer before being added to the single representation. These are subsequently added to the single representation $s_i$. Inspired by CarbonNovo  \cite{ren2024carbonnovo}, a distance map $D_{ij}$ is computed from the step 0 structure estimate $\hat{x}_0^{\text{struct}}$ using a distance one-hot encoder. This spatial information is injected directly into the pair representation $z_{ij}$. 

\subsubsection{Diffusion module}
The \texttt{DiffusionModule} (Algorithm 3) dictates the generation of both sequence and structural updates. It begins by conditioning the trunk representations on the current structure noise ($\hat{t}$) and sequence noise ($\hat{r}$) via \texttt{DiffusionConditioning}. The noisy structure coordinates, $x_l^{\text{noisy}}$, are then scaled to dimensionless vectors with approximately unit variance ($r_l^{\text{noisy}}$).

To process atom-level geometries, the model utilizes an \texttt{AtomAttentionEncoder} which aggregates atomic features into coarse-grained, sequence-local token representations $\{a_i\}$. These token representations are then passed through a 24-block \texttt{DiffusionTransformer} featuring full self-attention.

Following the transformer, the architecture decodes the token activations back to the atomic level using the \texttt{AtomAttentionDecoder}. This decoder is run twice with different configurations for the sequence and structure outputs, with each configuration being a separately trained \texttt{AtomAttentionDecoder} head:
\begin{enumerate}
    \item \textbf{Structure Update:} Run with \texttt{seq\_mode = False} to compute coordinate updates ($r_l^{\text{update}}$).
    \item \textbf{Sequence Logits:} Run with \texttt{seq\_mode = True} to predict sequence logits ($x_l^{\text{logits}}$), where in sequence mode we scatter add the atom-level embeddings back to the token level before finally predicting the sequence logits.
\end{enumerate}

Finally, the predicted positional update $r_l^{\text{update}}$ is rescaled and combined with the input noisy positions using the following schedule equation to yield the output structure $x_l^{\text{out}}$:

\begin{equation}
x_l^{\text{out}} = \frac{\sigma_{\text{data}}^2}{\sigma_{\text{data}}^2 + \hat{t}^2} x_l^{\text{noisy}} + \frac{\sigma_{\text{data}} \cdot \hat{t}}{\sqrt{\sigma_{\text{data}}^2 + \hat{t}^2}} r_l^{\text{update}}
\end{equation}

\subsubsection{Diffusion conditioning}
The \texttt{DiffusionConditioning} block (Algorithm 4) modulates the input representations based on the current noise levels. The pair representations $z_{ij}$ are concatenated with relative position encodings and processed through a 2-layer \texttt{Transition} block. 

For the single representations $s_i$, trunk outputs and input features are concatenated and projected. As the sequence and structure times may be different, we opt to embed them separately to help the model distinguish between sequence and structure noise levels. We embed the structure and sequence noise levels using a \texttt{FourierEmbedding} layer. The structure noise $\hat{t}$ is embedded logarithmically as $\frac{1}{4} \log(\hat{t}/\sigma_{\text{data}})$, while the sequence noise $\hat{r}$ is embedded directly. Both noise embeddings are added to $s_i$ before it undergoes a 2-layer \texttt{Transition} block.

\subsubsection{Atom attention decoder}
The \texttt{AtomAttentionDecoder} (Algorithm 5) maps the processed token-level features back to the required output space. Token activations $a_i$ are broadcast to their constituent atoms and summed with the corresponding atom-level skip connections $q_l^{\text{skip}}$. This representation is passed through a 3-block, 4-head \texttt{AtomTransformer} using cross-attention. 

If the decoder is operating in sequence mode (\texttt{seq\_mode = True}), the updated atomic features are aggregated back into token-level representations using a scatter-add operation. The final output $\hat{y}_l$ is then generated via layer normalization and a linear projection without biases.

\begin{algorithm}
\caption{Denoiser}
\label{alg:denoiser}
\begin{algorithmic}[1]
\Function{Denoiser}{$\{f^*\}$, $x_{l}^\seq$, $x_{l}^\struct$, $N_{\text{cycle}} = 4$, $c_s = 384$, $c_z = 128$}
    \State $\{s^{\text{inputs}}_i\} = \textsc{InputFeatureEmbedder}(\{f^*\})$
    \State $s^{\text{init}}_i = \textsc{LinearNoBias}(s^{\text{inputs}}_i)$
        \hfill $s^{\text{init}}_i \in \mathbb{R}^{c_s}$
    \State $z^{\text{init}}_{ij} = \textsc{LinearNoBias}(s^{\text{inputs}}_i)
        + \textsc{LinearNoBias}(s^{\text{inputs}}_j)$
        \hfill $z^{\text{init}}_{ij} \in \mathbb{R}^{c_z}$
    \State $z^{\text{init}}_{ij}\ \mathrel{+}= \textsc{RelativePositionEncoding}(\{f^*\})$
    \State $z^{\text{init}}_{ij}\ \mathrel{+}= \textsc{LinearNoBias}(f^{\text{token\_bonds}}_{ij})$
    \State $\{\hat{z}_{ij}\},\ \{\hat{s}_i\}, \{s_i^\txtskip\}, \{z_{ij}^\txtskip\} = \mathbf{0},\ \mathbf{0},\  \mathbf{0},\ \mathbf{0}$
    \For{all $c \in [1, \ldots, N_{\text{cycle}}]$}
        \State $z_{ij} = z^{\text{init}}_{ij}
            + \textsc{LinearNoBias}(\textsc{LayerNorm}(\hat{z}_{ij}))$
            \hfill $z_{ij} \in \mathbb{R}^{c_z}$
        \If{$c > 1$}
            \State $\{s_i\}, \{z_{ij}\} = \textsc{CrossModalEncode}(\{s_i\},\{z_{ij}\},x_l^\seq,x_l^\struct, \{s_i^\txtskip\}, \{z_{ij}^\txtskip\})$
        \EndIf
        \State $\{z_{ij}\}\ \mathrel{+}= \textsc{PlmModule}(\{f_{S_i}\},\ \{z_{ij}\},\ \{s^{\text{inputs}}_i\})$
        \State $s_i = s^{\text{init}}_i
            + \textsc{LinearNoBias}(\textsc{LayerNorm}(\hat{s}_i))$
            \hfill $s_i \in \mathbb{R}^{c_s}$
        \State $\{s_i\},\ \{z_{ij}\} = \textsc{PairformerStack}(\{s_i\},\ \{z_{ij}\})$
        \State $\{\hat{s}_i\},\ \{\hat{z}_{ij}\} \leftarrow \{s_i\},\ \{z_{ij}\}$
        \If{$c == 1$}
            \State $\{s_i^\txtskip\}, \{z_{ij}^\txtskip\} = \{\hat{s}_i\}, \{\hat{z}_{ij}\}$
        \EndIf
    \EndFor
    \State $\{s_i\}, \{z_{ij}\} = \textsc{CrossModalEncode}(\{s_i\},\{z_{ij}\},x_l^\seq,x_l^\struct, \{s_i^\txtskip\}, \{z_{ij}^\txtskip\})$
    \State $\{\hat{x}_0^\seq\}, \{\hat{x}_0^\struct\} = \textsc{DiffusionModule}(\{f^*\}, \{s_i^{\textrm{inputs}}\}, \{s_i\},\{z_{ij}\},\{s_i^\txtskip \}, \{z_{ij}^\txtskip \}, x_l^\seq, x_l^\struct)$
\EndFunction
\end{algorithmic}
\end{algorithm}

\begin{algorithm}
\caption{Cross-Modal Encoding}
\begin{algorithmic}[1]
\Function{CrossModalEncode}{$\{s_i\},\{z_{ij}\},x_l^\seq,x_l^\struct, \{s_i^\txtskip\}, \{z_{ij}^\txtskip\}, x_l^\seq, x_l^\struct$}
    \State $\{\hat{x}_0^\seq\}, \{\hat{x}_0^\struct\} = \textsc{DiffusionModule}(\{f^*\}, \{s_i^{\textrm{inputs}}\}, \{s_i\},\{z_{ij}\},\{s_i^\txtskip \}, \{z_{ij}^\txtskip \}, x_l^\seq, x_l^\struct)$

    \Statex

    \State $u_0^\seq = \textsc{LinearNoBias}(\textsc{LayerNorm}(\textsc{PlmEncode}(\hat{x}_0^\seq)))$
    
    \State $u_l^\seq = \textsc{LinearNoBias}(\textsc{LayerNorm}(\textsc{PlmEncode}(\hat{x}_l^\seq)))$

    \State $v_0^\seq = \textsc{LinearNoBias}(\textsc{LayerNorm}(\textsc{StructureEncode}(\hat{x}_0^\struct)))$
    
    \State $v_l^\seq = \textsc{LinearNoBias}(\textsc{LayerNorm}(\textsc{StructureEncode}(\hat{x}_l^\struct)))$

    \State $D_{ij} = \textsc{LinearNoBias}(\textsc{DistanceOneHotEncoder}(\hat{x}_0^\struct))$

    \Statex

    \State $s_i = s_i + u_{0i}^\seq + u_{li}^\seq + v_{0i}^\struct + v_{li}^\struct$

    \State $z_{ij} = z_{ij} + D_{ij}$
    
    \State \Return $\{s_i\}, \{z_{ij}\}$
\EndFunction
\end{algorithmic}
\end{algorithm}

\begin{algorithm}
\caption{Diffusion Module}
\begin{algorithmic}[1]
\State \textbf{function} \textsc{DiffusionModule}(
    $\{x^{\text{noisy struct}}_l\}$, $\{x^{\text{seq}}_l\}$, $\hat{t}$, $\hat{r}$, $\{f^*\}$, $\{s^{\text{inputs}}_i\}$,
    $\{s^{\text{trunk}}_i\}$, $\{z^{\text{trunk}}_{ij}\}$, $\{s_i^\txtskip \},$
    $\{z_{ij}^\txtskip \}$, $\sigma_{\text{data}} = 16$, $c_{\text{atom}} = 128$,
    $c_{\text{atompair}} = 16$, $c_{\text{token}} = 768$)

    \Statex \textit{\# Conditioning}
    \State $\{s_i\},\ \{z_{ij}\} = \textsc{DiffusionConditioning}(\hat{t}, \hat{r}, \ \{f^*\},\ \{s^{\text{inputs}}_i\},\ \{s^{\text{trunk}}_i\},\ \{z^{\text{trunk}}_{ij}\},\ \sigma_{\text{data}})$

    \Statex \textit{\# Scale positions to dimensionless vectors with approximately unit variance.}
    \State $r^{\text{noisy}}_l = x^{\text{noisy}}_l \Big/ \sqrt{\hat{t}^2 + \sigma^2_{\text{data}}}$
        \hfill $r^{\text{noisy}}_l \in \mathbb{R}^3$

    \State $\{s_i^{\text{trunk}}\} += \textsc{LinearNoBias}(\textsc{LayerNorm}(\{s_i^\txtskip \}))$
    \State $\{z_{ij}^{\text{trunk}}\} += \textsc{LinearNoBias}(\textsc{LayerNorm}(\{z_{ij}^\txtskip \}))$

    \Statex \textit{\# Sequence-local Atom Attention and aggregation to coarse-grained tokens}
\State $\{a_i\},\ \{q^{\text{skip}}_l\},\ \{c^{\text{skip}}_l\},\ \{p^{\text{skip}}_{lm}\} =
    \textsc{AtomAttentionEncoder}($
\Statex \hspace{5em}
    $\{f^*\},\ \{r^{\text{noisy}}_l\},\ \{s^{\text{trunk}}_i\},\ \{z_{ij}\},c_{\text{atom}},\ c_{\text{atompair}},\ c_{\text{token}}$
\Statex )  $a_i \in \mathbb{R}^{c_{\text{token}}}$
    \Statex \textit{\# Full self-attention on token level.}
    \State $a_i\ \mathrel{+}= \textsc{LinearNoBias}(\textsc{LayerNorm}(s_i))$
    \State $\{a_i\} \leftarrow \textsc{DiffusionTransformer}(\{a_i\},\ \{s_i\},\ \{z_{ij}\},\ \beta_{ij} = 0,\ N_{\text{block}} = 24,\ N_{\text{head}} = 16)$
    \State $a_i \leftarrow \textsc{LayerNorm}(a_i)$

    \Statex \textit{\# Broadcast token activations to atoms and run Sequence-local Atom Attention}
    \State $\{r^{\text{update}}_l\} = \textsc{AtomAttentionDecoder}(\{a_i\},\ \{q^{\text{skip}}_l\},\ \{c^{\text{skip}}_l\},\ \{p^{\text{skip}}_{lm}\}, \text{seq\_mode} = \text{False})$

    \State $\{x^{\text{logits}}_l\} = \textsc{AtomAttentionDecoder}(\{a_i\},\ \{q^{\text{skip}}_l\},\ \{c^{\text{skip}}_l\},\ \{p^{\text{skip}}_{lm}\}, \text{seq\_mode} = \text{True})$

    \Statex \textit{\# Rescale updates to positions and combine with input positions}
    \State $x^{\text{out}}_l = \dfrac{\sigma^2_{\text{data}}}{\sigma^2_{\text{data}} + \hat{t}^2} \cdot x^{\text{noisy}}_l
        + \dfrac{\sigma_{\text{data}} \cdot \hat{t}}{\sqrt{\sigma^2_{\text{data}} + \hat{t}^2}} \cdot r^{\text{update}}_l$

    \State \Return $\{x^{\text{out}}_l\}, \{x^{\text{logits}}_l\}$
\State \textbf{end function}
\end{algorithmic}
\end{algorithm}

\begin{algorithm}
\caption{Diffusion Conditioning}
\begin{algorithmic}[1]
\Function{DiffusionConditioning}{$\hat{t}$, $\hat{r}$, $\{f^*\}$, $\{s^{\text{inputs}}_i\}$, $\{s^{\text{trunk}}_i\}$, $\{z^{\text{trunk}}_{ij}\}$, $\sigma_{\text{data}}$, $c_z = 128$, $c_s = 384$}

    \Statex \hspace{\algorithmicindent} \textit{\# Pair conditioning}
    \State $z_{ij} = \text{concat}([z^{\text{trunk}}_{ij},\ \textsc{RelativePositionEncoding}(\{f^*\})])$
    \State $z_{ij} \leftarrow \textsc{LinearNoBias}(\textsc{LayerNorm}(z_{ij}))$
        \hfill $z_{ij} \in \mathbb{R}^{c_z}$
    \For{all $b \in [1, 2]$}
        \State $z_{ij}\ \mathrel{+}= \textsc{Transition}(z_{ij},\ n = 2)$
    \EndFor

    \Statex \hspace{\algorithmicindent} \textit{\# Single conditioning, $t$ is structure noise, $r$ is sequence noise}
    \State $s_i = \text{concat}([s^{\text{trunk}}_i,\ s^{\text{inputs}}_i])$
    \State $s_i \leftarrow \textsc{LinearNoBias}(\textsc{LayerNorm}(s_i))$
        \hfill $s_i \in \mathbb{R}^{c_s}$
    \State $n^\struct = \textsc{FourierEmbedding}\!\left(\tfrac{1}{4}\log(\hat{t}/\sigma_{\text{data}}),\ 256\right)$
    \State $n^\seq = \textsc{FourierEmbedding}\!\left(\hat{r},\ 256\right)$

    \State $s_i\ \mathrel{+}= \textsc{LinearNoBias}(\textsc{LayerNorm}(n^\struct)) + \textsc{LinearNoBias}(\textsc{LayerNorm}(n^\seq))$
    \For{all $b \in [1, 2]$}
        \State $s_i\ \mathrel{+}= \textsc{Transition}(s_i,\ n = 2)$
    \EndFor

    \State \Return $\{s_i\},\ \{z_{ij}\}$
\EndFunction
\end{algorithmic}
\end{algorithm}

\begin{algorithm}
\caption{Atom Attention Decoder}
\begin{algorithmic}[1]
\Function{AtomAttentionDecoder}{$\{a_i\}$, $\{q^{\text{skip}}_l\}$, $\{c^{\text{skip}}_l\}$, $\{p^{\text{skip}}_{lm}\}, \text{seq\_mode}$}
    \State \textit{\# Broadcast per-token activations to per-atom activations and add the skip connection}
    \State $q_l = \textsc{LinearNoBias}(a_{\text{tok\_idx}(l)}) + q^{\text{skip}}_l$
    \State \textit{\# Cross attention transformer}
    \State $\{q_l\} = \textsc{AtomTransformer}(\{q_l\},\ \{c^{\text{skip}}_l\},\ \{p^{\text{skip}}_{lm}\},\ N_{\text{block}} = 3,\ N_{\text{head}} = 4)$
    \State \textit{\# Scatter add atom embeddings back to token embeddings if sequence mode}
    \If{\text{seq\_mode}}
        \State $q_l = \textsc{ScatterAdd}(q_l, \text{tok\_idx}(l))$
    \EndIf
    \State \textit{\# Map to positions update or logits}
    \State $\hat{y}_l = \textsc{LinearNoBias}(\textsc{LayerNorm}(q_l))$
    \State \Return $\{\hat{y}_l\}$
\EndFunction
\end{algorithmic}
\end{algorithm}

\subsection{Training}
All training hyperparameters are detailed in Table \ref{tab:hyperparameters}. The model was trained with distributed data parallel across 32 L40S GPUs for 11 days for 160,000 training steps. We treated any atomic position with probability of being resolved less than $0.9$ as unresolved and mask out these atoms (or residues) when evaluating the losses. Dropout is only applied in the Pairformer module, and we fully freeze the pre-trained DPLM 650M weights during training. We use a series of losses to train \nameshort as described below. \nameshort is comprised of 888 million parameters in total, with 235 million trainable parameters.

\subsubsection{Structure diffusion losses}
The losses used for the structure diffusion module are largely the same as those used in AlphaFold 3. We recapitulate them here for convenience. The final structure loss is a weighted combination of the following loss.

\xhdr{MSE Loss}
The main loss used for the structure portion of our diffusion model takes the form of a weighted aligned MSE loss to the predicted denoised structure. We first perform a rigid alignment of the ground truth structure $\vec{x}_l^{\text{GT}}$

\begin{equation}
    \{\vec{x}_l^{\text{GT-aligned}}\} = \text{weighted\_rigid\_align}(\{\vec{x}_l^{\text{GT}}\}, \{\vec{x}_l^\struct \}, \{w_l\})
\end{equation}

where the weights $w_l$ are defined as

\begin{equation}
    w_l = 1 + f_l^\text{is\_dna}\alpha^\text{dna} + f_l^\text{is\_rna}\alpha^\text{rna} + f_l^\text{is\_ligand}\alpha^\text{ligand}
\end{equation}

and $\alpha^\text{dna}=\alpha^\text{rna}=5$ and $\alpha^\text{ligand}=10$.

The MSE loss is then defined as 
\begin{equation}
    \mathcal{L}_\text{MSE} = \frac{1}{3} \underset{l}{\text{mean}}\left(w_l\left\|\vec{x}_l^\struct - \vec{x}_l^\text{GT-aligned}\right\|^2\right)
\end{equation}

Notably, we do not take the loss on any atomic positions which are unresolved. 

\xhdr{Smooth LDDT Loss}
The Smooth LDDT Loss measures the accuracy of predicted atomic coordinates relative to ground truth structures. For each pair of atoms, we compute the difference between predicted and ground truth pairwise distances, then score each pair using a differentiable approximation of the discrete LDDT thresholds via a sum of sigmoid $\varepsilon_{lm} \in [0, 1]$. Pairs are masked by an inclusion radius that is modality-aware: nucleotide atoms (DNA/RNA) are evaluated within $30$ \AA, while all other atoms are evaluated within $15$ \AA. The final loss is one minus the masked mean per-pair score, such that a loss of $0$ corresponds to perfect coordinate recovery. The Smooth LDDT loss is detailed in Algorithm \ref{alg:smooth_lddt}.

\xhdr{Full Structure Diffusion Loss}
The full structure diffusion loss is a weighted combination of the MSE and smooth LDDT losses with 

\begin{equation}
    \mathcal{L}_\struct = \frac{\hat{t}^2 + \sigma_\text{data}^2}{(\hat{t} + \sigma_\text{data})^2}\mathcal{L}_\text{MSE} + \mathcal{L}_\text{smooth\_lddt}
\end{equation}

where $\hat{t}$ is the sampled noise level and $\sigma_\text{data}$ is a constant based on the variance of the data (we use $\sigma_\text{data} = 16$). During training the structure noise level $\hat{t}$ is sampled from $\sigma_\text{data} \cdot \exp(-1.2 + 1.5 \cdot \mathcal{N}(0,1))$. Before computing the structure diffusion losses, we apply an optimal ground truth chain assignment as described in Subsection 4.2 of the AlphaFold 3 S.I.

\begin{algorithm}
\caption{Smooth LDDT Loss}
\label{alg:smooth_lddt}
\begin{algorithmic}[1]
\Function{SmoothLDDTLoss}{$\{x_l\}$, $\{x^{\text{GT}}_l\}$, $\{f^{\text{is\_dna}}_l\}$, $\{f^{\text{is\_rna}}_l\}$}
    \State \textit{\# Compute distances between all pairs of atoms}
    \State $\delta x_{lm} \leftarrow \|x_l - x_m\|$
    \State $\delta x^{\text{GT}}_{lm} \leftarrow \|x^{\text{GT}}_l - x^{\text{GT}}_m\|$
    \State \textit{\# Compute distance difference for all pairs of atoms}
    \State $\delta_{lm} \leftarrow \left| \delta x^{\text{GT}}_{lm} - \delta x_{lm} \right|$
    \State $\varepsilon_{lm} \leftarrow \frac{1}{4} \Big[
        \text{sigmoid}\!\left(\tfrac{1}{2} - \delta_{lm}\right)
        + \text{sigmoid}\!\left(1 - \delta_{lm}\right)
        + \text{sigmoid}\!\left(2 - \delta_{lm}\right)
        + \text{sigmoid}\!\left(4 - \delta_{lm}\right)
    \Big]$
    \State \textit{\# Restrict to bespoke inclusion radius}
    \State $f^{\text{is\_nucleotide}}_l \leftarrow f^{\text{is\_dna}}_l + f^{\text{is\_rna}}_l$
    \State $c_{lm} \leftarrow \left(\delta x^{\text{GT}}_{lm} < 30\,\text{\AA}\right) f^{\text{is\_nucleotide}}_l
        + \left(\delta x^{\text{GT}}_{lm} < 15\,\text{\AA}\right)\!\left(1 - f^{\text{is\_nucleotide}}_l\right)$
    \State \textit{\# Compute mean, avoiding self term}
    \State $\text{lddt} = \underset{l \neq m}{\text{mean}}\,(c_{lm}\,\varepsilon_{lm})\ /\ \underset{l \neq m}{\text{mean}}\,(c_{lm})$
    \State \Return $1 - \text{lddt}$
\EndFunction
\end{algorithmic}
\end{algorithm}

\subsubsection{Sequence diffusion losses}
While the structure diffusion module is a weighted combination of multiple losses, for the sequence diffusion head, we instead make use exclusively of the masked diffusion loss from \eqref{eq:MDMELBO}, introduced in \cite{mdlm,shi2024simplified}. In our paper we use a linear schedule $\alpha_r = 1 - r$ so that the loss weight $\frac{d\alpha_r}{dr}\cdot\frac{1}{1-\alpha_r}=-\frac{1}{r}$. When computing the loss, we mask out unreliable residues. In particular, we mask out any residues for which any of its constituent backbone atoms have occupancy less than $0.8$, any residue for which any backbone atom is unresolved, and any residue which is part of a histag. We define a residue to be part of a histag if it is part of a consecutive run of 5 or more histidine residues.

In parameterizing our log denoiser $D^{\seq}_\theta$, we utilize the SUBS parameterization from \cite{mdlm}, wherein as the reverse process never predicts a $\mathbf{m}$ token, we substitute $-\infty$ in for the logit corresponding to the $\mathbf{m}$ token. During training, we sample the sequence time $r$ uniformly from $[0, 1]$. When masking the sequence, we ensure that all amino acid residue information is not leaked to the model by masking all reference information features in addition to any sequence identity features. Specifically, we replace the reference position, mask and charge features with masked versions of themselves. We set the masked reference position to be the average reference conformer position for each backbone atom, averaged over all 20 standard amino acid residues, while for the masked reference charge and mask, we set them to be identical to those of the glycine backbone atoms.

\subsubsection{Full training loss}
The full training loss is a weighted combination of the sequence and structure losses, as well as a distogram loss wherein our distogram head and loss are identical to AlphaFold 2 \cite{jumper2021highly}. 

\begin{equation}
    \mathcal{L}_\text{total} = \alpha_\seq \mathcal{L}_\text{seq} + \alpha_\text{MSE} \mathcal{L}_\text{MSE} + \alpha_\text{smooth\_lddt}\mathcal{L}_\text{smooth\_lddt} + \alpha_\text{distogram}\mathcal{L}_\text{distogram}
\end{equation}

where we set $\alpha_\seq = 1, \alpha_\text{MSE}=4,\alpha_\text{smooth\_lddt}=4,\alpha_\text{distogram}=0.03$.

\subsubsection{Training hyperparameters}
We present all training hyperparameters in Table \ref{tab:hyperparameters}.

\subsubsection{Model selection}

\begin{figure}[t]
    \centering
    \includegraphics[width=\linewidth]{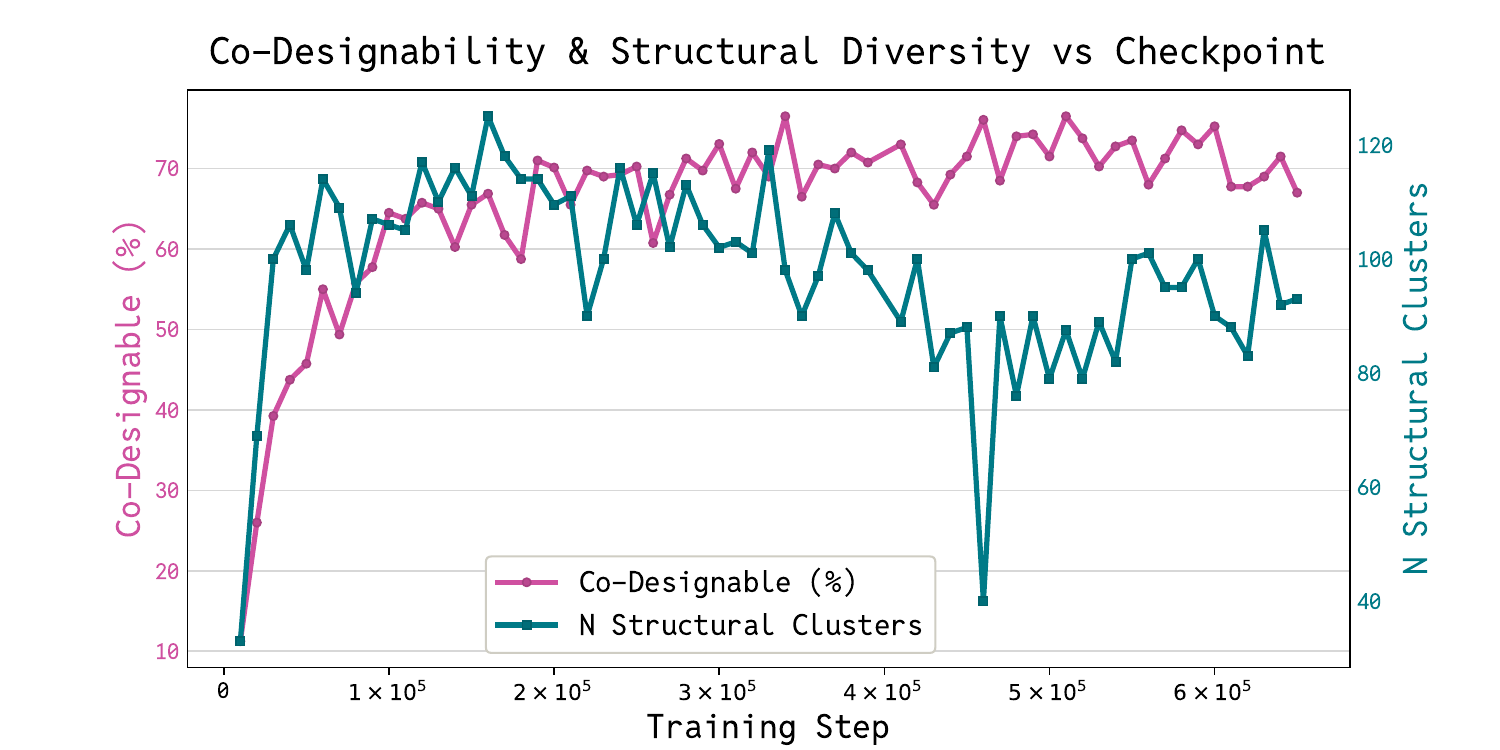}
    \caption{Co-designability increases while structural diversity decreases as training progresses.}
    \label{fig:ckpt_designability_vs_struc_div}
\end{figure}

We choose our final checkpoint by evaluating EMA checkpoints on the unconditional generation task. We select the checkpoint with the best tradeoff of \% co-designability and structural cluster designability. Interestingly, in Figure \ref{fig:ckpt_designability_vs_struc_div} we observe that structural diversity peaks around 200k training steps and decreases throughout the rest of training, indicating overfitting. We select checkpoint 160k as our final model, given its superior tradeoffs. We note that the inference settings used to generate this plot were those of Table \ref{tab:cond_inf_hyperparameters}, which yields improved structural clusters with slightly degraded co-designability.

\begin{longtable}{llc}
\caption{Model and Training Hyperparameters}\label{tab:hyperparameters}\\
\toprule
\textbf{Category} & \textbf{Hyperparameter} & \textbf{Value} \\
\midrule
\endfirsthead
% Header for the second and subsequent pages
\multicolumn{3}{c}%
{{\bfseries \tablename\ \thetable{} -- continued from previous page}} \\
\toprule
\textbf{Category} & \textbf{Hyperparameter} & \textbf{Value} \\
\midrule
\endhead
% Footer for all pages except the last one
\midrule
\multicolumn{3}{r}{{Continued on next page}} \\
\bottomrule
\endfoot
% Footer for the final page
\bottomrule
\endlastfoot
% --- Table Content Begins Here ---
\multicolumn{3}{c}{\textit{Global Dimensions \& Training}} \\
\midrule
Global & Single Representation Dimension ($c_s$) & 384 \\
Global & Pair Representation Dimension ($c_z$) & 128 \\
Global & Gradient Clipping Norm & 10.0 \\
Global & EMA Decay Rate & 0.999 \\
Global & Sequence Batch Size & 32 \\
Global & Structure Batch Size & 96 \\
Global & Crop Size & 384 \\
\midrule
\multicolumn{3}{c}{\textit{Optimizer (Adam)}} \\
\midrule
Adam & Learning Rate & 0.00018 \\
Adam & $\beta_1$ & 0.9 \\
Adam & $\beta_2$ & 0.95 \\
Adam & Weight Decay & $1 \times 10^{-8}$ \\
\midrule
\multicolumn{3}{c}{\textit{Learning Rate Scheduler}} \\
\midrule
Scheduler & Warmup Steps & 1000 \\
Scheduler & Warmup Type & Linear \\
Scheduler & Decay Factor & 0.95 \\
Scheduler & Decay Every $N$ Steps & 50000 \\
\midrule
\multicolumn{3}{c}{\textit{Core Model Architecture}} \\
\midrule
Model & Max Number of Cycles ($N_{\text{cycle}}$) & 4 \\
PLM & PLM Base Model & DPLM 650M \\
Input Embedder & Token Dimension ($c_{\text{token}}$) & 384 \\
Input Embedder & Atom Dimension ($c_{\text{atom}}$) & 128 \\
Input Embedder & Atom Pair Dimension ($c_{\text{atompair}}$) & 16 \\
Relative Position & $r_{\text{max}}$ & 32 \\
Relative Position & $s_{\text{max}}$ & 2 \\
\midrule
\multicolumn{3}{c}{\textit{Pairformer Stack}} \\
\midrule
Pairformer & Number of Blocks & 8 \\
Pairformer & Number of Attention Heads & 16 \\
Pairformer & Dropout & 0.25 \\
\midrule
\multicolumn{3}{c}{\textit{Diffusion Module}} \\
\midrule
Diffusion & Token Dimension ($c_{\text{token}}$) & 768 \\
Diffusion & $\sigma_{\text{data}}$ & 16 \\
Diffusion Transformer & Number of Blocks & 24 \\
Diffusion Transformer & Number of Attention Heads & 16 \\
Atom Encoder & Number of Blocks & 3 \\
Atom Encoder & Number of Attention Heads & 4 \\
Structure Atom Decoder & Number of Blocks & 3 \\
Structure Atom Decoder & Number of Attention Heads & 4 \\
Sequence Atom Decoder & Number of Blocks & 3 \\
Sequence Atom Decoder & Number of Attention Heads & 4 
\end{longtable}
\subsection{Inference}
We found that inference techniques are critical for model performance, observing in our experiments that the co-designability metrics can scale from 16\% to 88\% with the same checkpoint, depending on inference settings. Below, we detail the inference pipeline and progressively introduce each technique responsible for this improvement.

\subsubsection{Inference sampling}

Our inference-time sampling scheme is shown in \cref{alg:sampling}, where the structure inference procedure is largely based off of that of AF3.

\begin{algorithm}
\caption{SampleDiffusion}
\begin{algorithmic}[1]
\Require $\{f^*\}$, Structure Noise Schedule $[c_0, c_1, \ldots, c_T]$, Sequence Time Schedule $[r_0, r_1, ..., r_T]$, $\gamma_0 = 0.8$, $\gamma_{\min} = 1.0$, noise scale $\lambda = 0.1$, step scale $\eta = 1.5$, $\text{do\_noisy\_guidance} = \text{false}$

\State $x_l^\struct \sim c_0 \cdot \mathcal{N}(0, \mathbf{I}_3)$ \hfill $x_l \in \mathbb{R}^3$
\State $x_l^\seq = [\bf{m}, \bf{m}, \ldots, \bf{m}]$
\For{each $(c_\tau, r_\tau) \in \text{zip}([c_1, \ldots, c_T], [r_1, \ldots, r_T])$}
    \State $\{x_l^\struct\} \leftarrow \textsc{CentreRandomAugmentation}(\{x_l^\struct\})$
    \State $\gamma = \gamma_0 \text{ if } c_\tau > \gamma_{\min} \text{ else } 0$
    \State $\hat{t} = c_{\tau-1}(\gamma + 1)$
    \State $\xi_l = \lambda\sqrt{\hat{t}^2 - c_{\tau-1}^2} \cdot \mathcal{N}(0, \mathbf{I}_3)$ \hfill $\xi_l \in \mathbb{R}^3$
    \State $x^{\struct,\text{noisy}}_l = x_l^\struct + \xi_l$
    %\State $\{x^{\struct,\text{denoised}}_l\}, \{x^{\seq,\text{logits}}_l\} = \textsc{Denoiser}( $
    %\Statex $\hspace{4em}\{x^{\struct,\text{noisy}}_l\},\{x_l^\seq\},\, \hat{t},\, r_\tau, \{f^*\}$
    %\Statex $\hspace{1.8em})$
    \State $\{x^{\struct,\text{denoised}}_l\}, \{x^{\seq,\text{logits}}_l\} = \textsc{Denoiser}( \{x^{\struct,\text{noisy}}_l\},\{x_l^\seq\},\, \hat{t},\, r_\tau, \{f^*\})$
    \If{$\text{do\_noisy\_guidance}$}
       % \State $\{x^{\struct,\text{denoised}}_l\}, \{x^{\seq,\text{logits}}_l\} = \textsc{NoisyGuidance}($
       % \Statex $\hspace{4em}\{x^{\struct,\text{noisy}}_l\},\{x_l^\seq\},\, \hat{t},\, r_\tau, \{f^*\}$
       % \Statex $\hspace{3em})$
        \State $\{x^{\struct,\text{denoised}}_l\}, \{x^{\seq,\text{logits}}_l\} = \textsc{NoisyGuidance}(\{x^{\struct,\text{noisy}}_l\},\{x_l^\seq\},\, \hat{t},\, r_\tau, \{f^*\})$
    \EndIf
    
    \State $\delta_l = (x_l - x^{\text{denoised}}_l) / \hat{t}$
    \State $dt = c_\tau - \hat{t}$
    \State $x_l^\struct \leftarrow x^{\struct,\text{noisy}}_l + \eta \cdot dt \cdot \delta_l$
    \State $x_l^\seq = \textsc{PathPlanningStep}(x_l^{\seq,\text{logits}}, b_\tau)$
\EndFor
\State \Return $\{x_l\}$
\end{algorithmic}
\label{alg:sampling}
\end{algorithm}

\subsubsection{Sequence self-correction}
Standard masked diffusion inference is limited due to the structure of its reverse process. Specifically, the masked diffusion reverse process requires that once a particular token is unmasked, it retains its value throughout the rest of the inference trajectory. This, however, prevents the model from correcting its mistakes and, for optimal performance, requires the model to have learned perfectly how to generate a particular sequence in every $n!$ possible unmasking orderings. Given limited compute budgets learning, such a model is impractical, and significant performance gains have been observed by allowing a masked discrete diffusion model to self-correct its errors \cite{peng2025pathplanningmaskeddiffusion,nie2025largelanguagediffusionmodels,wang2025remaskingdiscretediffusionmodels}. 

We similarly find that using standard masked discrete diffusion inference performs poorly. To remedy, this we adopt a random remasking strategy \cite{peng2025pathplanningmaskeddiffusion} wherein at each sequence time $r$ we uniformly choose $r$ percent of tokens to be masked and $1-r$ percent to be unmasked with the full procedure summarized in \cref{alg:p2}. This allows the model to correct mistakes it made during the inference trajectory, especially at critical structural residues such as those of loopy regions of a structure and at regions where the structure shifts from disordered to structured. %\todo{FIGURE OUT IF WE WANT TO SAY THIS CAUSE I NEED TO RUN MORE EXPERIMENTS IF SO, THE NEXT SENTENCE I MEAN} We find that random remasking outperforms other schemes such as confidence based remasking or annealed confidence masking, with random remasking offering a higher percent of co-designable sequence-structure pairs while slightly trading off for number of structure clusters.

\begin{algorithm}
\caption{Path Planning Sampling}
\label{alg:p2}
\begin{algorithmic}[1]
\Require $x_1 \leftarrow [\mathbf{m}, \ldots, \mathbf{m}],\ \text{Seq length } N,\ \text{Denoiser } D_\theta^\text{seq},\ \text{Schedule } \kappa,\ \text{Num sampling steps } T$
\For{$t \in \left[1, \frac{T-1}{T}, ..., 0\right]$}
    \State \textbf{Plan:}
    \State \hspace*{\algorithmicindent} $z \sim D_\theta^\text{seq}(x_t)$
    \State \hspace*{\algorithmicindent} $\texttt{UnmaskPos} \leftarrow \text{RandSampleUniform}\!\left(\lfloor \kappa(t) \cdot N \rfloor,\ \{1, \ldots, N\}\right)$
    \State \hspace*{\algorithmicindent} $\texttt{MaskPos} \leftarrow \{1,\ldots,N\} \setminus \texttt{UnmaskPos}$
    \State \textbf{Denoise:}
    \State \hspace*{\algorithmicindent}$x_t^j \leftarrow
        \begin{cases}
            z^j & \text{if } j \in \texttt{UnmaskPos} \land x_t^j = \mathbf{m} \\
            \mathbf{m}   & \text{if } j \in \texttt{MaskPos}
        \end{cases}$
\EndFor
\State \Return $x_L$
\end{algorithmic}
\end{algorithm}

\subsubsection{Entropy adaptive sequence temperature}
We observed that the model tends to select particular overconfident residues early on during an inference trajectory, which harms overall co-designability. To remedy this, we applied a novel adaptive temperature scheme for the sequence logits, whereby the distribution of overly confident tokens is flattened early in the inference trajectory.

\begin{equation}
    \tau^{\text{entropy}, (n)}_r = 1 + r^\gamma\beta \frac{H_\text{max} - H^{(n)}}{H_\text{max}} 
\end{equation}

where $H^{(n)}$ is the entropy of the $n$th token's distribution induced by the denoiser $\mu_\theta$, $H_\text{max} = -\log(20)$ is the maximum entropy of a distribution over all amino acids, $\beta$ is a constant controlling the strength of the entropy scaling, and $\gamma$ controls how quickly we turn off entropy adaptive tempering. We also maintain a global logit temperature $\tau_r^\text{global}$ for which we use a step function.

\begin{equation}
    \tau_r^\text{global} = \begin{cases}
        0.8 & \text{if } r \geq 0.2 \\
        0.1 & \text{if }r < 0.2
    \end{cases}
\end{equation}

The final logit temperature at time $r$ for token position $n$ is then

\begin{align}
    \tau_r^{(n)} &= \tau_r^\text{global} \cdot  \tau_r^{\text{entropy},(n)} \\
    &= \tau_r^\text{global} + \tau_r^\text{global} r^\gamma \beta \frac{H_\text{max} - H^{(n)}}{H_\text{max}}
\end{align}

where the entropy term approaches $0$ as the sequence time $r \rightarrow 0$. Writing the logits for the $n$th token induced by the denoiser as $\{x^{\text{seq,logits,}(n)}\}$ we finally obtain the tempered distribution we draw sequence samples from $p_r^\tau(x_i^{(n)}) \propto \exp(x^{\text{seq,logits,}(n)} / \tau_r^{(n)})$. We find that the application of both these sequence temperature methods in conjunction is critical for performance.

\subsubsection{Inference schedule}
\begin{figure}[t]
    \centering
    \includegraphics[width=\linewidth]{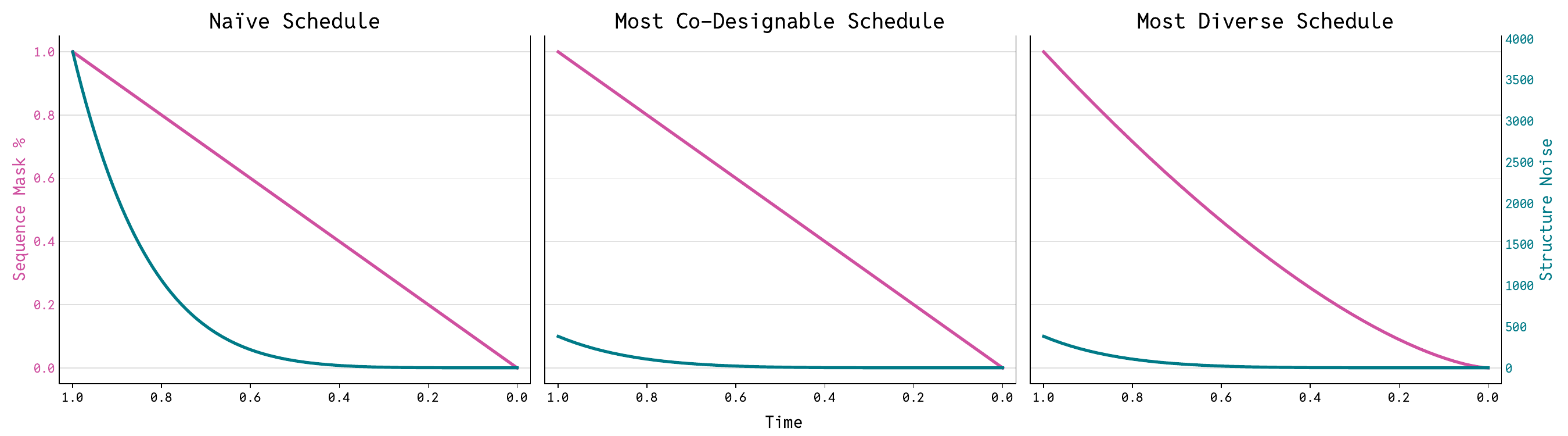}
    \caption{Various noise schedules used as inference settings. The leftmost plot displays a naïve schedule representing the default AlphaFold 3 settings paired with a linear sequence schedule. The middle plot is the schedule used in our most co-designable settings, while the right is our most diverse settings (using $\text{mask\_proportion}=t^{1.5}$ for sequence).}
    \label{fig:inf_schedules}
\end{figure}

Figure~\ref{fig:inf_schedules} illustrates the three noise schedules evaluated as inference settings for joint sequence-structure co-design. The naïve schedule (left) pairs the default AlphaFold 3 structure noise schedule with a linear sequence masking schedule, serving as our baseline. The most co-designable schedule (middle) and most diverse schedule (right) modify the sequence masking trajectory, with the latter employing a $\texttt{mask\_rate} = t^{1.5}$ to concentrate unmasking toward the end of the denoising process, encouraging greater sequence diversity. Consistent with prior works on diffusion-based protein structure design~\cite{watson_novo_2023}, we found that the noise scale $\lambda$ applied to the structure diffusion process is critical for sample quality---excessively large noise scales destabilize generation, while overly small values collapse diversity. In practice, we set $\lambda = 0.1$ for all reported experiments, which we found to strike the best balance between structural plausibility and sequence diversity across our benchmark tasks.

\subsubsection{Conditional generation}
Conditional generation of proteins in complex with other biomolecules extends naturally from our framework. During inference, the conditioning molecule (e.g.\ a small-molecule ligand or nucleic acid chain) is provided as an input feature: its reference conformer geometry and bond information are supplied through the standard AlphaFold~3 conditioning features, and its atomic coordinates are allowed to be denoised throughout the reverse diffusion trajectory. The protein sequence and structure are generated jointly in the same setting as above, while the conditioning molecule's features participate in attention and pairformer computations, allowing the model to shape the protein around the target binding partner (and vice versa). No additional training or fine-tuning is required.

\subsubsection{Noisy guidance}

\begin{algorithm}
\caption{Multimodal Noisy Guidance}
\begin{algorithmic}[1]
\Require $\{x^{\struct,\text{noisy}}_l\}$, $\{x_l^\seq\}$, $\hat{t}$, $r_\tau$, $\{f^*\}$,
    guidance weight $\omega$,
    noisy times $\psi^\seq > r_\tau$, $\psi^\struct > \hat{t}$,
    rescaling weight $\phi$,
    guidance intervals $(\xi^\struct_{\min}, \xi^\struct_{\max})$, $(\xi^\seq_{\min}, \xi^\seq_{\max})$

\State \textit{\# Conditional denoiser pass at current noise levels}
\State $\{x^{\struct,\text{denoised}}_{\text{cond}}\}, \{\ell^\seq_{\text{cond}}\} = \textsc{Denoiser}(\{x^{\struct,\text{noisy}}_l\}, \{x_l^\seq\}, \hat{t}, r_\tau, \{f^*\})$

\State \textit{\# Structure noisy guidance: corrupt sequence}
\If{$\hat{t} \in (\xi^\struct_{\min},\, \xi^\struct_{\max})$} 
    \State $\{x^{\seq,\psi}_l\} \leftarrow \textsc{Corrupt}(\{x_l^\seq\},\, \psi^\seq)$
    \State $\{x^{\struct,\text{denoised}}_{\text{ng}}\}, \_ = \textsc{Denoiser}(\{x^{\struct,\text{noisy}}_l\}, \{x^{\seq,\psi}_l\}, \hat{t}, \psi^\seq, \{f^*\})$
    \State $\sigma_{\text{cond}} \leftarrow \text{std}\!\left(\{x^{\struct,\text{denoised}}_{\text{cond}}\}\right),\quad
           \sigma_{\text{ng}} \leftarrow \text{std}\!\left(\{x^{\struct,\text{denoised}}_{\text{ng}}\}\right)$
    \State $x^{\struct,\text{rescaled}}_l \leftarrow x^{\struct,\text{denoised}}_{\text{ng}} \cdot \dfrac{\sigma_{\text{cond}}}{\sigma_{\text{ng}}}$
    \State $\{x^{\struct,\text{denoised}}_l\} \leftarrow \phi \cdot x^{\struct,\text{rescaled}}_l + (1-\phi) \cdot x^{\struct,\text{denoised}}_{\text{cond}}$
\Else
    \State $\{x^{\struct,\text{denoised}}_l\} \leftarrow \{x^{\struct,\text{denoised}}_{\text{cond}}\}$
\EndIf

\State \textit{\# Sequence noisy guidance: corrupt structure}
\If{$r_\tau \in (\xi^\seq_{\min},\, \xi^\seq_{\max})$} 
    \State $\xi^\struct_l = \sqrt{(\psi^\struct)^2 - \hat{t}^2} \cdot \mathcal{N}(0, \mathbf{I}_3)$
    \State $\{x^{\struct,\psi}_l\} \leftarrow \{x^{\struct,\text{noisy}}_l\} + \xi^\struct_l$
    \State $\_, \{\ell^\seq_{\text{ng}}\} = \textsc{Denoiser}(\{x^{\struct,\psi}_l\}, \{x_l^\seq\}, \psi^\struct, r_\tau, \{f^*\})$
    \State $\{\ell^{\seq,\text{denoised}}_l\}  =\omega \cdot \{\ell^\seq_{\text{cond}}\} + (1-\omega) \cdot \{\ell^\seq_{\text{ng}}\}$
\Else
    \State $\{\ell^{\seq,\text{denoised}}_l\} \leftarrow \{\ell^\seq_{\text{cond}}\}$
\EndIf

\State \Return $\{x^{\struct,\text{denoised}}_l\}$,\; $\{\ell^{\seq,\text{denoised}}_l\}$
\end{algorithmic}
\label{alg:noisy_guidance}
\end{algorithm}

 Noisy guidance~\cite{rojas2025diffuseeverythingmultimodaldiffusion} is a guidance mechanism that arises naturally from a multimodal diffusion framework with decoupled noise schedules. We detail the entire noisy guidance algorithm in Algorithm \ref{alg:noisy_guidance}. Standard Classifier-Free Guidance (CFG) interpolates between a conditional and unconditional score with guidance strength $\omega$:
\begin{equation}
    \omega s_\theta(x_t^\struct, t, c) + (1 - \omega) s_\theta(x_t^\struct, t, \varnothing).
\end{equation}
Noisy guidance exploits the fact that we train with a decoupled noise schedule for sequence and structure to replace the unconditional score with a
conditional score evaluated at a \emph{partially} noised condition $x_\rho^\seq$ at noise level $\rho > r$:
\begin{equation}
    \omega s_\theta^\struct(x_t^\struct, x_r^\seq, t, r) + (1 - \omega) s_\theta^\struct(x_t^\struct, x_\rho^\seq, t, \rho), \quad \rho > r.
\end{equation}
This recovers vanilla CFG when $r = 0$ and $\rho = 1$, but more generally allows the guiding model to
be a partially conditioned score rather than a fully unconditional one. This is conceptually related to
Autoguidance \cite{karras2024guiding}, which uses a degraded model as the corrector to reduce errors in the conditional score ---
here the same effect is achieved by degrading the \emph{condition} rather than the model. Noisy guidance
can also be applied to unconditional joint generation by adaptively choosing $\rho > r$ as both
modalities are being generated simultaneously.

For the discrete modality, the guidance interpolation is performed in \emph{logit space} rather than
directly averaging the discrete scores. Concretely, given conditional and unconditional logits
$\ell_{\mathrm{cond}}$ and $\ell_{\mathrm{uncond}}$, the guided discrete score is
\begin{equation}
    s_\theta^\seq \leftarrow \mathrm{softmax}\!\left(\omega \cdot \ell_{\mathrm{cond}} + (1 - \omega) \cdot \ell_{\mathrm{uncond}}\right),
\end{equation}
rather than a geometric average of the softmax outputs as proposed in some prior discrete guidance
work. This arithmetic interpolation in logit space has been shown to carry theoretical advantages
over direct probability-space averaging \cite{chang2022maskgit}.

In practice, we find that in continuous space, the magnitude of the conditional score does not match the magnitude of the noised score, or equivalently, the conditional and noised structure denoiser outputs. To remedy this, we take the approach of \cite{lin2024common} and rescale the guided structure denoiser output based on the ratio of standard deviations of conditional and noised structure denoiser outputs $D_{\theta,\text{cond}}^\struct$, $D_{\theta,\text{ng}}^\struct$ as
\begin{equation}
    \sigma_\text{cond} = \text{std}(D_{\theta,\text{cond}}^\struct), \;\;\sigma_\text{ng} = \text{std}(D_{\theta,\text{ng}}^\struct)
\end{equation}
\begin{equation}
    D_{\theta,\text{rescaled}}^\struct = D_{\theta,\text{cond}}^\struct \cdot \frac{\sigma_\text{cond}}{\sigma_\text{ng}}
\end{equation}
\begin{equation}
    D_{\theta,\text{final}}^\struct = \phi \cdot D_{\theta,\text{rescaled}}^\struct + (1 - \phi) \cdot D_{\theta,\text{cond}}^\struct
\end{equation}

Similar to the usage of CFG, we only apply noisy guidance within a guidance interval $t \in (\xi_\text{min}^\struct,\xi_\text{max}^\struct)$ and $r \in (\xi_\text{min}^\seq,\xi_\text{max}^\seq)$. To enable multimodal noisy guidance, for each modality, we fix a constant modality and add noise to the other modality before feeding both through the model to produce denoiser predictions on the current timestep and noisy inputs before adjusting the respective score function based on the modality we noised. Finally, we found that setting the noisy sequence and structure times $\psi^\seq$ and $\psi^\struct$ to constants worked best and performed experiments using this.

\subsubsection{Inference hyperparameters}
We specify the hyperparameters used for both the unconditional benchmark in Table \ref{tab:uncond_inf_hyperparameters} and those for the conditional benchmark in Table \ref{tab:cond_inf_hyperparameters}. We note that we used different parameters for the unconditional and conditional settings as we found that a different set of parameters yielded more co-designable, diverse structures while slightly degrading the \% co-designability, with each setting sitting at different points on the Pareto frontier.
\clearpage
\begin{longtable}{llc}
\caption{Unconditional Inference Hyperparameters}
\label{tab:uncond_inf_hyperparameters} \\
\toprule
\textbf{Category} & \textbf{Hyperparameter} & \textbf{Value} \\
\midrule
\endfirsthead
% Header for the second and subsequent pages
\multicolumn{3}{c}%
{{\bfseries \tablename\ \thetable{} -- continued from previous page}} \\
\toprule
\textbf{Category} & \textbf{Hyperparameter} & \textbf{Value} \\
\midrule
\endhead
% Footer for all pages except the last one
\midrule
\multicolumn{3}{r}{{Continued on next page}} \\
\bottomrule
\endfoot
% Footer for the final page
\bottomrule
\endlastfoot
% --- Table Content Begins Here ---
\multicolumn{3}{c}{\textit{Model}} \\
\midrule
Model & Number of Cycles ($N_{\text{cycle}}$) & 4 \\
\midrule
\multicolumn{3}{c}{\textit{Inference Noise Scheduler}} \\
\midrule
Noise Scheduler & $\sigma_{\max}$ & 160.0 \\
Noise Scheduler & $\sigma_{\min}$ & 0.0004 \\
Noise Scheduler & $\rho$ & 7 \\
Noise Scheduler & $\sigma_{\text{data}}$ & 16.0 \\
\midrule
\multicolumn{3}{c}{\textit{Structure Diffusion Sampling}} \\
\midrule
Sampling & $\gamma_0$ & 0.8 \\
Sampling & $\gamma_{\min}$ & 1.0 \\
Sampling & Noise Scale ($\lambda$) & 0.1 \\
Sampling & Step Scale ($\eta$) & 1.5 \\
Sampling & Number of Steps ($N_{\text{step}}$) & 200 \\
\midrule
\multicolumn{3}{c}{\textit{Noisy Guidance}} \\
\midrule
Noisy Guidance & Enabled & True \\
Noisy Guidance & $\omega_{\text{struct}}$ & 1.5 \\
Noisy Guidance & $\omega_{\text{seq}}$ & 2.0 \\
Noisy Guidance & Unconditional Seq.\ Time & 0.6 \\
Noisy Guidance & Unconditional Struct.\ Time & 0.8 \\
Noisy Guidance & Rescale $\phi$ & 0.7 \\
Noisy Guidance & Guidance End Time & 0.3 \\
Noisy Guidance & Guidance Start Time & 0.8 \\
\midrule
\multicolumn{3}{c}{\textit{Sequence Sampling Strategy}} \\
\midrule
Seq.\ Sampling & Strategy & Path Planning \\
Seq.\ Sampling & Score Type & Random \\
Seq.\ Sampling & Logits Temperature & $\begin{cases}0.8 & \text{if $t \geq 0.2$} \\ 0.1 & \text{if $t < 0.2$}\end{cases}$ \\
Seq.\ Sampling & Entropy Adaptive $\beta$ & 1.0 \\
Seq.\ Sampling & Entropy Adaptive $\gamma$ & 1.0 \\
\midrule
\multicolumn{3}{c}{\textit{Sequence Noise Scheduler}} \\
\midrule
Seq.\ Noise Scheduler & Schedule & $t$
\end{longtable}

\begin{longtable}{llc}
\caption{Conditional Inference Hyperparameters}
\label{tab:cond_inf_hyperparameters} \\
\toprule
\textbf{Category} & \textbf{Hyperparameter} & \textbf{Value} \\
\midrule
\endfirsthead
% Header for the second and subsequent pages
\multicolumn{3}{c}%
{{\bfseries \tablename\ \thetable{} -- continued from previous page}} \\
\toprule
\textbf{Category} & \textbf{Hyperparameter} & \textbf{Value} \\
\midrule
\endhead
% Footer for all pages except the last one
\midrule
\multicolumn{3}{r}{{Continued on next page}} \\
\bottomrule
\endfoot
% Footer for the final page
\bottomrule
\endlastfoot
% --- Table Content Begins Here ---
\multicolumn{3}{c}{\textit{Model}} \\
\midrule
Model & Number of Cycles ($N_{\text{cycle}}$) & 4 \\
\midrule
\multicolumn{3}{c}{\textit{Inference Noise Scheduler}} \\
\midrule
Noise Scheduler & $\sigma_{\max}$ & 160.0 \\
Noise Scheduler & $\sigma_{\min}$ & 0.0004 \\
Noise Scheduler & $\rho$ & 7 \\
Noise Scheduler & $\sigma_{\text{data}}$ & 16.0 \\
\midrule
\multicolumn{3}{c}{\textit{Structure Diffusion Sampling}} \\
\midrule
Sampling & $\gamma_0$ & 1.6 \\
Sampling & $\gamma_{\min}$ & 1.0 \\
Sampling & Noise Scale ($\lambda$) & 0.1 \\
Sampling & Step Scale ($\eta$) & 1.5 \\
Sampling & Number of Steps ($N_{\text{step}}$) & 200 \\
\midrule
\multicolumn{3}{c}{\textit{Noisy Guidance}} \\
\midrule
Noisy Guidance & Enabled & False \\
\midrule
\multicolumn{3}{c}{\textit{Sequence Sampling Strategy}} \\
\midrule
Seq.\ Sampling & Strategy & Path Planning \\
Seq.\ Sampling & Score Type & Random \\
Seq.\ Sampling & Logits Temperature & $\begin{cases}0.8 & \text{if $t \geq 0.2$} \\ 0.1 & \text{if $t < 0.2$}\end{cases}$ \\
Seq.\ Sampling & Entropy Adaptive Temp. & False \\
\midrule
\multicolumn{3}{c}{\textit{Sequence Noise Scheduler}} \\
\midrule
Seq.\ Noise Scheduler & Schedule & $t^{1.5}$
\end{longtable}

\subsection{Inference steering}\label{sec:inference_steering}

Inference-time steering enables the adaptation of a single pretrained model to diverse and evolving design objectives at sampling time, without the need for retraining. Desired properties are incorporated directly into the generation process, tilting the distribution of samples at each step. Common guidance approaches rely on heuristic score modifications in diffusion models, which only approximately target conditional distributions and often compromise control over the resulting marginals. In this work, we instead consider the Feynman-Kac Corrector (FKC) framework~\cite{skreta2025fkc}, which enables principled sampling from desired target distributions. In the continuous case, this is done by updating samples using a weighted stochastic differential equation (SDE). The Feynman-Kac PDE describes the time-evolution of the density of samples as  

\begin{align}
    % \deriv{p_t^{\textsc{fk}}(x)}{t} &= \deriv{p_t^{\textsc{sde}}(x)}{t} + \deriv{p_t^{w}(x)}{t} \nonumber\\
    \deriv{p_t^{\textsc{fk}}(x)}{t} &= -\inner{\nabla}{p_t^{\textsc{fk}}(x)v_t(x)} + \frac{g_t^2}{2}\Delta p_t^{\textsc{fk}}(x) + p_t^{\textsc{fk}}(x)(h_t(x) -  \int  h_t(x)p_t^{\textsc{fk}}(x) dx)\,, \label{eq:fk_pde}
\end{align}

where the first two terms correspond to the Fokker-Planck PDE and the last term is a reweighting PDE, assigning each sample a time-dependent log-weight $w_t$ based on the weighting function $h_t(x)$. In the discrete setting~\cite{hasan2026discrete}, the analogous time-evolution of weighted marginals can be described by weighted Forward Kolmogorov Equations (FKEs) of the form. 

\begin{align}
\begin{split}
    \deriv{p_t^{\textsc{fk}}(i)}{t} = \sum_{j \neq i} \left(A_{t}(j,i)p_t(j) - A_t(i,j)p_t(i)\right) + p_t(i)\left(h_t(i)-\mean_{p_t(i)}h_t(i)\right)\,,
    \label{eq:wFKE}
\end{split}
\end{align}

where $A_t(i,j)$ is the rate matrix.

To sample from $p_t^{\textsc{fk}}(x)$ using FKC, three general steps are required at every timestep $t$. We start with a batch of $K$ particles (or samples); at each timestep, we do the following to each particle $k$:

\begin{enumerate}
    \item Update its state $x_t$ based on the reverse process using \cref{eq:struct_update} (continuous) or  \cref{eq:seq_update} (discrete).
    %\begin{equation}
    %\begin{aligned}
     %   &x_{t+dt}^k = x_t^k + (v_t(x_t^k)dt + \sigma_t dW_t) 
      %  && \text{[continuous]} \\
       % &\quad \text{or} \\
      %  &x_{t+dt}^k \sim \texttt{Cat}(x_{t+dt}^k = j\cond \delta_{ij} + A_t(i,j)dt)\,, \text{ for } x_t^k = i 
      %  && \hspace{1.2em} \text{[discrete]}
    %\end{aligned}
    %\end{equation}

    \item Compute its log-weight  update $\log w^k_{t+dt} = \log w^k_{t} + h_t(x^k)dt$.

 %       \begin{equation}
  %  \begin{aligned}
   %     &\log w^k_{t+dt} = \log w^k_{t} + g_t(x^k)dt 
    %    && \text{[continuous]} \\
     %   &\quad \text{or} \\
      %  &\log w^k_{t+dt} = \log w^k_{t} + g_t(i)dt  
       % && \hspace{1.2em} \text{[discrete]}
    %\end{aligned}
    %\end{equation}

    \item Normalize the weights and use them to resample the particles in the batch.
    
\end{enumerate}

The state and weight updates are derived based on the target distribution we wish to sample from. In \cref{sec:fkcsg} and \cref{sec:fkcmm}, we demonstrate two use cases of FKC steering and write the state and weight updates for each. 

\subsubsection{Feynman-Kac Correctors - Specificity Guidance}
\label{sec:fkcsg}

The goal of specificity guidance is to generate protein samples that preferentially bind to a desired ligand (on-target) while avoiding binding to an undesired one (off-target). We formalize this as sampling from the target density. 

\begin{equation}
p_{t, \beta}^{\text{SG}}(x) \propto q_{t}^{\text{on}}(x)^{(1-\beta)}q_{t}^{\text{off}}(x)^{\beta}
\end{equation}

with $\beta < 0$. By setting $\beta < 0$, the off-target distribution acts as a repulsive term, encouraging samples that are likely under $q^{\text{on}}$ while penalizing those that are likely under $q^{\text{off}}$. In \cref{cor:fkcsg}, we present FKC - Specificity Guidance (FKC-SG) and define the corresponding state and weight updates to sample from $p_{t, \beta}^{\text{SG}}(x)$. This corollary follows from FKC - Geometric Average defined in \cite{skreta2025fkc}.

% \begin{mdframed}[style=MyFrame2]
\begin{corollarybox}[FKC-SG]
\label{cor:fkcsg}
    Consider two diffusion models $q_t^{\text{on}}(x), q_t^{\text{off}}(x)$ and inverse temperature $\beta < 0$.
    The weighted SDE corresponding to the repulsion of marginals $p_{t, \beta}^{\text{SG}}(x) \propto q_{t}^{\text{on}}(x)^{(1-\beta)}q_{t}^{\text{off}}(x)^{\beta}$ is
    \begin{align}
        dx_t =~& g_t^2((1-\beta)\nabla \log q_t^{\text{on}}(x_t)+\beta\nabla \log q_t^{\text{off}}(x_t))dt -f_t(x_t)dt + g_t dW_t\,,\\
        dw_t =~& \frac{g_t^2}{2}\beta(\beta-1)\norm{\nabla \log q_t^{\text{on}}(x_t)-\nabla \log q_t^{\text{off}}(x_t)}^2dt
    \end{align}
\end{corollarybox}
% \end{mdframed}

In \cref{alg:fkcsg}, we show how this is implemented in our sampling framework. Importantly, since our model does not enforce invariance to rotations and translations, one cannot naively combine the scores from the two distributions since they may not necessarily have the same frame of reference. Thus, we first compute the translation and rotation matrices $ q_t^{\text{off}}(x_t)$ to the space of $q_t^{\text{on}}(x_t)$ using Kabsch alignment before doing guidance or computing weights. Furthermore, we only do steering on the protein state and use the ligand state for conditioning. In \cref{alg:fkcsg}, we denote operations restricted to the protein or ligand components using $x^{\struct}_{\text{prot}}$ and $x^{\struct}_{\text{lig}}$, respectively. Similarly, operations corresponding to the on-target and off-target states are denoted by $x_{\text{on}}$ and $x_{\text{off}}$.

\begin{breakablealgorithm}{SampleDiffusion with FKC-SG}
\label{alg:fkcsg}
\begin{algorithmic}[1]
\footnotesize
\Require $\{f^*\}$, Structure Noise Schedule $[c_0, c_1, \ldots, c_T]$, Sequence Time Schedule $[r_0, r_1, ..., r_T]$, $\gamma_0 = 0.8$, $\gamma_{\min} = 1.0$, noise scale $\lambda = 0.1$, step scale $\eta = 1.5$,  $\text{guidance\_on\_sequence} = \text{true}$, inverse temperature $\beta=-0.5$, $\text{resample} = \text{true}$, number of particles   $K = 4$
%\State $\{x^{\struct}_{l,\text{on}}\}_{k=1}^K \sim \mathcal{N}(0, c_0^2 \mathbf{I}_3)$
\State $\{x^{\struct}_{k,l,\text{on}}\}_{k=1}^K,\quad x^{\struct}_{k,l} \sim \mathcal{N}(0, c_0^2 \mathbf{I}_3)\ \forall l \in \{1,\dots,L\}$ \hfill $x^\struct_{k,l} \in \mathbb{R}^3$
\State $x^{\struct}_{\text{off}} = x^{\struct}_{\text{on}}\texttt{.clone()}$
%\State $\{x_{k}^{\seq}\}_{k=1}^{2K} = [\textbf{m}, \textbf{m}, ..., \textbf{m}]$
\State $\{x^{\seq}_{k,l}\}_{k=1}^{2K},\quad x^{\seq}_{k,l} = \mathbf{m}\ \forall l \in \{1,\dots,L\}$ 

\State $\{w\}_{k=1}^K = 1/K$

\State $R = \mathbf{I}_3$
\State $t = 0$
\For{each $(c_\tau, r_\tau) \in \text{zip}([c_1, \ldots, c_T], [r_1, ..., r_T])$}
    \State $\gamma = \gamma_0 \text{ if } c_\tau > \gamma_{\min} \text{ else } 0$
    \State $\hat{t} = c_{\tau-1}(\gamma + 1)$
    \State $\{\xi_{l,k}\}_{k=1}^{K},\quad \xi_l = \lambda\sqrt{\hat{t}^2 - c_{\tau-1}^2} \cdot \mathcal{N}(0, \mathbf{I}_3)\forall l \in \{1,\dots,L\}$ 
    \State $x^{\struct,\text{noisy}}_{\text{on}} = x^\struct_{\text{on}} + \xi$
    %\State $x^{\struct,\text{noisy}}_{\text{off}} = x^\struct_{\text{off}} + \xi$

    % \State $x^{\struct,\text{noisy}}_\text{off} = x_l^\struct + \xi_l$

    \State $x^{\struct,\text{noisy}}_{\text{prot,off}} \gets \textsc{ApplyTransform}(x^{\struct,\text{noisy}}_{\text{prot,on}}, R, t)$  \Comment{\textcolor{codeblue}{\texttt{rotate on-prot struct to off-prot space}}}

    \State $\{x^{\struct,\text{denoised}}\}, \{x^{\seq,\text{logits}}\} = \textsc{Denoiser}( \{x^{\struct,\text{noisy}}\},\{x^\seq\},\, \hat{t},\, r_\tau, \{f^*\})$

    \State $dt = c_\tau - \hat{t}$

    %\If{$\text{guidance\_on\_structure}$}

    \Statex \textcolor{codepink}{\texttt{\# update protein state}}

    \State $(R, t) \gets \textsc{Kabsch}(x^{\struct,\text{noisy}}_{\text{prot,on}}, x^{\struct,\text{noisy}}_{\text{prot,off}})$ \Comment{\textcolor{codeblue}{\texttt{maps on-prot to off-prot space}}}
    \State $(R^{-1}, t^{-1}) \gets \textsc{InvertTransform}(R, t)$ \Comment{\textcolor{codeblue}{\texttt{maps off-prot to on-prot space}}}
    
    \State $\tilde{x}^{\struct,\text{denoised}}_{\text{prot,off}} \gets \textsc{ApplyTransform}(x^{\struct,\text{denoised}}_{\text{prot,off}}, R^{-1}, t^{-1})$  
    
    \State $\delta_{\text{prot, on}} \gets (x^{\struct,\text{noisy}}_{\text{prot, on}} - x^{\struct,\text{denoised}}_{\text{prot, on}}) / \hat{t}$
    \State $\delta_{\text{prot,off}} \gets (x^{\struct,\text{noisy}}_{\text{prot, on}} - \tilde{x}^{\struct,\text{denoised}}_{\text{prot,off}}) / \hat{t}$
    
    %\Statex \textcolor{codeblue}{\# guidance for on-target (repel off-target)}
    \State $\delta_{\text{prot,on}} \gets (1-\beta)\delta_{\text{prot, on}} + \beta\delta_{\text{prot, off}}$ \Comment{\textcolor{codeblue}{\texttt{guidance for on-target (repel off-target)}}}
    \State $x^\struct_{\text{prot,on}} \leftarrow x^{\struct,\text{noisy}}_{\text{prot,on}} + \eta \cdot dt \cdot \delta_{\text{prot,on}}$

    \Statex \textcolor{codepink}{\texttt{\# update ligand state}}

    \State $\delta_{\text{lig}} \gets (x^{\struct,\text{noisy}}_{\text{lig}} - x^{\struct,\text{denoised}}_{\text{lig}}) / \hat{t}$
    \State $x^\struct_{\text{lig}} \leftarrow x^{\struct,\text{noisy}}_{\text{lig}} + \eta \cdot dt \cdot \delta_{\text{lig}}$

    \Statex \textcolor{codepink}{\texttt{\# calculate scores for on- and off-target protein states}}
    \State $\text{score}_{\text{prot,on}} \gets (x^{\struct,\text{noisy}}_{\text{prot, on}} - x^{\struct,\text{denoised}}_{\text{prot, on}}) / \hat{t}^2$
    \State $\text{score}_{\text{prot,off}} \gets (x^{\struct,\text{noisy}}_{\text{prot, on}} - \tilde{x}^{\struct,\text{denoised}}_{\text{prot,off}}) / \hat{t}^2$

    \Statex \textcolor{codepink}{\texttt{\# optionally adjust sequence logits}}
    \If{$\text{guidance\_on\_sequence}$}
        \State $x^{\seq,\,\text{logits}}_{\text{on}} = (1 - \beta)x^{\seq,\,\text{logits}}_{\text{on}} + (1 - \beta)x^{\seq,\,\text{logits}}_{\text{off}}$ %\Comment{\textcolor{codeblue}{optionally guide sequence logits}}

    \EndIf
    
    \State $x^\seq_{\text{on}} = \textsc{PathPlanningStep}(x^{\seq,\text{logits}}_{\text{on}}, b_\tau)$
    
    \Statex \textcolor{codepink}{\texttt{\# compute weights \& resample particles in batch}}
    \State $dw = -\beta(\beta - 1)\cdot\norm{\text{score}_{\text{prot,on}} - \text{score}_{\text{prot,off}}}^2 \cdot \hat{t}$
    \State $w \gets w + dw \cdot dt$
    
    \If{$\text{resample}$}
        \State $w_k = w_k/\left(\sum_i^K w_i\right)$ \Comment{\textcolor{codeblue}{\texttt{normalize weights}}}
        \State $\ell \sim \texttt{Cat}(K\cond w)$ \Comment{\textcolor{codeblue}{\texttt{sample index}}}
        \State $x_{k,\text{prot,on}}^{\struct} \gets x_{\ell,\text{prot,on}}^{\struct} $ 
        \State $x_{k,\text{lig,on}}^{\struct} \gets x_{\ell,\text{lig,on}}^{\struct} $ 
        \State $x_{k,\text{lig,off}}^{\struct} \gets x_{\ell,\text{lig,off}}^{\struct} $ 
        \State $x_{k}^{\seq} \gets x_{\ell}^{\seq} $ 

        \State $w_k \gets 1/K$ \Comment{\textcolor{codeblue}{\texttt{re-initialize weights}}}
    \EndIf
    \State $x^\seq \gets \textsc{Concat}([x^\seq_{\text{on}}, x^\seq_{\text{on}}])$

    % \State $x^\struct \gets \textsc{Concat}([x^\struct_{\text{prot}}, x^\struct_{\text{lig}}], \text{dim}=1)$

\EndFor
\State \Return $x^{\struct}_{\text{on}}, x^{\seq}_{\text{on}}$
\end{algorithmic}
\end{breakablealgorithm}

\subsubsection{FKC-SG experimental details}

We investigate the ability of FKC-SG to generate specific-binding proteins for three sets of on/off-targets. The first pair is amino acids valine (on-target) and proline (off-target); the second is steroids aldosterone (on-target) and cortisone (off-target); the third is biotin (on-target) and pyridoxal 5'-phosphate (PLP, off-target). For each, we sweep over inverse temperatures $\beta \in \{-0.5, -1.0, -2.0\}$. Previous work also found that resampling at every step reduced sample diversity significantly due to variance of the weights\cite{skreta2025fkc}, and so we also searched for the best time point to stop resampling ($\text{resample} = \text{false when } \tau \in \{0.4, 0.6, 0.8, 1.0\}$). Finally, we tried particle batch sizes of $K \in \{2, 4, 8\}$. We found that $\beta = -0.5$ worked best, as well as setting $\text{resample} = \text{false when } \tau = 0.6$. For the valine/proline pair, we used 4 particles and for the other two pairs, we found 8 particles were better.

\subsubsection{FKC-SG evaluation}
 
For each ligand pair, we generated $K \cdot 200$ protein structures and sequences of length 150 using \name with FKC-SG or \name on its own. We refolded all $K \cdot 200$ sequences with the on-target using Chai-1 and aligned them with the structure generated by \name. If the complex was designable (i.e. RMSDs of both the ligand centroids and proteins were less than $2$~\AA), we kept the structure. We then folded each off-target with the sequence and computed the RMSD between the ligand centroids after aligning the complex with the protein folded with the on-target. For each batch of $K$ particles, we kept the sample with the largest ligand centroid RMSD (max reward) for further evaluation (max. $200$ samples for each method if each batch had one designable complex). For each protein sequence, we folded it three times with Chai-1 for each of the on/off-targets.  For each sequence, we measured the pairwise ligand centroid RMSD between all remaining Chai-1 folds (9 comparisons) and kept the minimum RMSD. A sequence was considered to show separation between the on- and off- targets if this minimum RMSD exceeded a fixed threshold ($6$~\AA). We report the number of such specific sequences for each condition.

\subsubsection{Feynman-Kac Correctors - Multimodal}
\label{sec:fkcmm}
We observe that the reward tilting framework of unimodal Feynman-Kac correctors in continuous and discrete spaces can be extended to the multimodal case. To this end, we derive in Theorem \ref{thm:mm_reward_fke} expressions for the augmented drift and rate matrices as well as the weight term $h$. We consider the one-dimensional discrete case as the extension to multi-dimensional discrete objects is trivial.

\begin{theorembox}[FKC-MM Reward-Tilting]
\label{thm:mm_reward_fke}
    \; Identify $p_\tau(x_\tau, i) := p_{t(\tau),s(\tau)}(x_{t(\tau)},i)$. For the reward tilted marginals $q_\tau(x, i) \propto p_\tau(x,i)\exp(\beta_\tau R(x,i))$ the following equations holds

    \begin{align}
        \frac{\partial q_\tau(x_t,i)}{\partial \tau} &= \dot{t}(\tau)\left(\!-\!\left\langle \nabla, q_\tau(x_t,i)(-f_\tau(x) + g_\tau^2\nabla \log p_\tau(x) + \beta_\tau \frac{g_\tau^2}{2} \nabla R(x,i) \right\rangle + \frac{g_\tau^2}{2} \Delta q_\tau(x,i)\right) \nonumber \\
        &+ \dot{s}(\tau) \left(\sum_{j\neq i} A_\tau^R(i,j,x)q_\tau(x,i) - A_\tau^R(y,i,x) q_\tau(x,j)\right) \nonumber \\
        &+ q_\tau(x,i)\left(h_\tau(x,i) - \mathbb{E}_{y, j \sim p_\tau(y,j)}[h_\tau(y,j)]\!\right) 
        \end{align}
        \begin{align}
        h_\tau(x,i) &= \dot{t}(\tau)\left\langle \beta_\tau \nabla_x R(x,i), \frac{g_\tau^2}{2} \nabla_x \log p_\tau(x,i) - f_\tau(x) \right\rangle \nonumber \\
        &+ \dot{s}(\tau) \sum_{j\neq i}\left(A^R_\tau(i,j,x) - A_\tau(i,j,x)\right) + \frac{\partial \beta_\tau}{\partial \tau}R(x,i)
        \end{align}
        \begin{align}
        A_\tau^R(i,j) &= A_\tau(i,j,x)\frac{\exp(\beta_\tau R(x,j))}{\exp(\beta_\tau R(x,i))}
    \end{align}
\end{theorembox}

\begin{proofbox}
    This proof follows closely the steps for unimodal reward-tilted FKC in continuous and discrete space, specifically Proposition D.6 of \cite{skreta2025fkc} and Theorem 3.5 of \cite{hasan2026discrete}. We begin by denoting a time schedule for each modality $t(\tau)$ and $s(\tau)$ with $t$ corresponding to the continuous modality and $s$ to the discrete modality. We then define

    \begin{equation}
        p_\tau(x,i) \coloneqq p_{t(\tau),s(\tau)}(x,i) 
    \end{equation}
    \begin{equation}
        q_\tau(x,i) \coloneqq \frac{1}{Z_\tau} p_\tau(x,i) \exp(\beta_\tau R(x,i)), Z_\tau = \sum_i\int p_\tau(x,i)\exp(\beta_\tau R(x,i)) dx
    \end{equation}
    
    Breaking apart the time derivative

    \begin{equation}
        \frac{\partial}{\partial \tau} p_\tau(x,i) = \dot{t}(\tau) \frac{\partial}{\partial t} p_{t,s}(x,i) + \dot{s}(\tau) \frac{\partial}{\partial s}p_{t,s}(x,i)
    \end{equation}

    From Lemma 3 of \cite{rojas2025diffuseeverythingmultimodaldiffusion} we have
    \begin{equation}
        \frac{\partial}{\partial t}p_{t,s}(x,i) = - \nabla \cdot (v_t(x) p_{t,s}(x,i)) + \frac{g_t^2}{2}\Delta p_{t,s}(x,i)
    \end{equation}
    \begin{equation}
        \frac{\partial}{\partial s}p_{t,s}(x,i) = \sum_{j \neq i} \left[A_s(x,j,i) p_{t,s}(x,j) - A_s(x,i,j)p_{t,s}(x,i)\right]
    \end{equation}

    where in the latter we define $A_s(x,i,i)$ to be the negative sum over the off-diagonal.
    Consider the time derivative of the change in log density

    \begin{align}
        \frac{\partial}{\partial \tau} \log q_\tau(x,i) &= \frac{\partial}{\partial \tau} (\log p_\tau(x,i) + \beta_\tau R(x,i) - \log Z_\tau)\nonumber \\ 
        &= \dot{t}(\tau) \frac{\partial}{\partial t} \log p_{t,s}(x,i) + \dot{s}(\tau) \frac{\partial}{\partial s} \log p_{t,s}(x,i) + \frac{\partial \beta_\tau}{\partial \tau}R(x,i) - \frac{\partial}{\partial \tau} \log Z_\tau
    \end{align}

    We break up the expression by modality, considering the continuous modality first
    % \begin{adjustbox}
    {\small
    \begin{align}
        \frac{\partial}{\partial t} \log p_{t,s}(x,i)  &= - \langle \nabla, v_t(x)\rangle - \langle \nabla \log p_{t,s}(x,i),v_t(x)\rangle + \frac{g_t^2}{2}\Delta \log p_{t,s}(x,i) + \frac{g_t^2}{2}\|\nabla \log p_{t,s}(x,i)\|^2 \nonumber \\
        &= -\langle \nabla, v_t(x)\rangle - \langle \nabla \log q_{t,s}(x,i),v_t(x)\rangle + \frac{g_t^2}{2}\Delta \log q_{t,s}(x,i) + \frac{g_t^2}{2} \| \nabla \log q_{t,s}(x,i)\|^2 \nonumber \\
        &\phantom{{}={}} + \left\langle \beta_\tau \nabla R(x,i), v_t(x) - g_t^2 \nabla \log p_{t,s}(x,i) -\frac{g_t^2}{2}\beta_\tau \nabla R(x,i)\right\rangle - \beta_\tau \frac{g_t^2}{2}\Delta R(x,i) \nonumber
    \end{align}}
    % \end{adjustbox}
    We will eventually evaluate the change in marginal density $\frac{\partial p_\tau(x,i)}{\partial \tau}$ and to do so we will use the term $q_{t,s}(x,i) \frac{\partial}{\partial t} \log p_{t,s}(x,i)$ which we derive now. Making use of the identities $p(x) \langle \nabla, f(x)\rangle = \langle \nabla, f(x) p(x) \rangle -\langle f(x), \nabla p(x)\rangle  $ for vector field $f$ and scalar field $p$, that $\Delta p(x) = p(x)(\Delta \log p(x) + \| \log p(x)\|^2)$, and that $\frac{\partial p_t(x)}{\partial t} = p_t(x) \frac{\partial}{\partial t} \log p_t(x)$ we can evaluate the time derivative of the non-log density
{\small
    \begin{align*}
        q_{t,s}(x,i) \frac{\partial}{\partial t} \log p_{t,s}(x,i) &= -q_{t,s}(x,i)\langle \nabla , v_t(x)\rangle - q_{t,s}(x,i)\langle \nabla \log q_{t,s}(x,i),v_t(x)\rangle \\ 
        &\phantom{{}={}} + q_{t,s}(x,i)\frac{g_t^2}{2}\left(\Delta \log q_{t,s}(x,i) + \|\nabla \log q_{t,s}(x,i)\|^2\right) \\
        &\phantom{{}={}} + q_{t,s}(x,i) \left\langle \beta_\tau \nabla R(x,i), v_t(x) - g_t^2 \nabla \log p_{t,s}(x,i) -\frac{g_t^2}{2}\beta_\tau \nabla R(x,i)\right\rangle \nonumber \\ 
        &\phantom{{}={}}-q_{t,s}(x,i)\beta_\tau \frac{g_t^2}{2}\Delta R(x,i) \\
        &= - \langle \nabla, v_t(x)q_{t,s}(x,i)\rangle + \langle \nabla q_{t,s}(x,i), v_t(x)\rangle - \langle \nabla q_{t,s}(x,i), v_t(x)\rangle \\
        &\phantom{{}={}} +\frac{g_t^2}{2} \Delta q_{t,s}(x,i) + q_{t,s}(x,i) h_{t,s}^\text{cont}(x,i) \\
        &= - \langle \nabla, v_t(x)q_{t,s}(x,i)\rangle +\frac{\sigma_t^2}{2} \Delta q_{t,s}(x,i) + q_{t,s}(x,i) h_{t,s}^\text{cont\_pre}(x,i)
    \end{align*}}

    To help with performance during inference, we can add the reward gradient to the drift. To do so we recall that for a scalar field $p(x)$ and vector fields $f(x)$ and $h(x)$ we have that $\langle \nabla, p(x)f(x)\rangle + \langle \nabla, p(x)g(x)\rangle = \langle \nabla, p(x)(f(x) + h(x))\rangle$
{\small
    \begin{align*}
        q_{t,s}(x,i) \frac{\partial}{\partial t} \log p_{t,s}(x,i) &=- \langle \nabla, v_t(x)q_{t,s}(x,i)\rangle +\frac{g_t^2}{2} \Delta q_{t,s}(x,i) + q_{t,s}(x,i) h_{t,s}^\text{cont\_pre}(x,i) \\
        &\phantom{{}={}}- \langle \nabla, q_{t,s}(x,i) a\nabla R(x,i)\rangle + \langle \nabla, q_{t,s}(x,i) a\nabla R(x,i)\rangle \\
        &=-\langle \nabla, q_{t,s}(x,i)(v_t(x) + a\nabla R(x))\rangle  +\frac{g_t^2}{2} \Delta q_{t,s}(x,i) + q_{t,s}(x,i) h_{t,s}^\text{cont}(x,i)
    \end{align*}
}
    where
    \begin{align*}
        h_{t,s}^\text{cont}(x,i) &= a \Delta R(x,i) + a\langle\nabla \log q_{t,s}(x,i),\nabla R(x,i)\rangle -\beta_\tau \frac{g_t^2}{2}\Delta R(x,i) \\
        &\phantom{{}={}}+ \left\langle \beta_\tau \nabla R(x,i), v_t(x) - g_t^2 \nabla \log p_{t,s}(x,i) -\frac{g_t^2}{2}\beta_\tau \nabla R(x,i)\right\rangle
    \end{align*}

    Plugging in $v_t(x) = -f_t(x) + g_t^2 \nabla \log p_{t,s}(x,i)$ and $a=\beta_\tau g_t^2/2$ we have

    \begin{align*}
        q_{t,s}(x,i) \frac{\partial}{\partial t} \log p_{t,s}(x,i) &= -\left\langle \nabla, q_{t,s}(x,i)(-f_t(x)+g_t^2\nabla \log p_{t,s}(x,i) + \beta_\tau \frac{g_t^2}{2}\nabla R(x,i))\right\rangle \nonumber \\
        &\phantom{{}={}}+ \frac{g_t^2}{2}\Delta q_{t,s}(x,i) + q_{t,s}(x,i) h_{t,s}^\text{cont}(x,i) \\
        h_{t,s}^\text{cont}(x,i)&= \beta_\tau\frac{g_t^2}{2}\Delta R(x,i) - \beta_\tau\frac{g_t^2}{2}\Delta R(x,i) + \left\langle \beta_\tau \nabla R(x,i), \frac{g_t^2}{2}\nabla \log q_{t,s}(x,i)\right\rangle \nonumber \\
        &\phantom{{}={}}+\left\langle \beta_\tau \nabla R(x,i),-f_t(x) -\frac{g_t^2}{2}\beta_\tau \nabla R(x,i) \right\rangle \\
        &= \left\langle \beta_\tau \nabla R(x,i), -f_t(x) + \frac{g_t^2}{2}\nabla \log (q_{t,s}(x,i)/\exp(\beta_\tau R(x,i))\right\rangle \\ 
        &= \left\langle \beta_\tau \nabla R(x,i), -f_t(x) + \frac{g_t^2}{2} \nabla \log p_{t,s}(x,i)\right\rangle
    \end{align*}

    We now consider the discrete modality

    \begin{align*}
        \frac{\partial}{\partial s}\log p_{t,s}(x,i) &= \sum_{j\neq i}\left[ A_s(x,j,i)\frac{p_{t,s}(x,j)}{p_{t,s}(x,i)} - A_s(x,i,j) \right] \\
        &= \sum_{j\neq i}\bigg[ \underbrace{A_s(x,j,i)\frac{\exp(\beta_\tau R(x,j)) }{\exp(\beta_\tau R(x,i))}}_{\coloneqq A_s^R(x,j,i)}\frac{q_{t,s}(x,j)}{q_{t,s}(x,i)} - A_s(x,i,j)\bigg] \\
        &= \sum_{j\neq i} \left[A_s^R(x,j,i)\frac{q_{t,s}(x,j)}{q_{t,s}(x,i)} - A_s^R(x,i,j)\right] + \sum_{j\neq i}\left[A_s^R(x,i,j) - A_s(x,i,j)\right]
    \end{align*}

    Again, we consider the weighted term 
    \begin{align*}
        q_{t,s}(x,i)\frac{\partial}{\partial s}\log p_{t,s}(x,i) &= \sum_{j\neq i} \left[A_s^R(x,j,i)q_{t,s}(x,j) - A_s^R(x,i,j)q_{t,s}(x,i)\right] \\
        &\phantom{{}={}} + q_{t,s}(x,i) \underbrace{\sum_{j\neq i}\left[A_s^R(x,i,j) - A_s(x,i,j)\right]}_{\coloneqq g_{t,s}^\text{discrete}(x,i)}
    \end{align*}

    Finally we have

    \begin{align*}
        \frac{\partial q_\tau(x,i)}{\partial\tau} &= -\dot{t}(\tau)\left\langle \nabla, q_{t,s}(x,i)(-f_t(x)+g_t^2\nabla \log p_{t,s}(x,i) + \beta_\tau \frac{g_t^2}{2}\nabla R(x,i))\right\rangle \\
        &\phantom{{}={}}+ \dot{t}(\tau)\frac{g_t^2}{2}\Delta q_{t,s}(x,i) + \dot{s}(\tau)\sum_{j\neq i}\left[A_s^R(x,j,i)q_{t,s}(x,j)-A_s^R(x,i,j)q_{t,s}(x,i)\right] \\
        &\phantom{{}={}}+q_{t,s}(x,i)(h_{t,s}(x,i) - \mathbb{E}_{x,i \sim q_{t,s}(x,i)}[h_{t,s}(x,i)]) \\
        h_{t,s}(x,i) &= \dot{t}(\tau) h_{t,s}^\text{cont}(x,i) + \dot{s}(\tau) h_{t,s}^\text{discrete}(x,i) + \frac{\partial \beta_\tau}{\partial \tau}R(x,i) \\
        &= \dot{t}(\tau)\left\langle \beta_\tau \nabla R(x,i), -f_t(x) + \frac{g_t^2}{2} \nabla \log p_{t,s}(x,i)\right\rangle \\
        &\phantom{{}={}}+ \dot{s}(\tau) \sum_{j\neq i}\left(A_s^R(x,i,j) - A_s(x,i,j)\right) + \frac{\partial \beta_\tau}{\partial \tau}R(x,i)
    \end{align*}
\end{proofbox}

\subsubsection{FKC-MM reward functions}
A central motivation for joint sequence--structure guidance is that many biophysically meaningful objectives are inherently \emph{multimodal}: they depend simultaneously on residue identity (sequence) and inter-residue geometry (structure), and thus multimodal steering would be favorable compared to sequence-only or structure-only methods that are available to current two-stage design methods.
We designed two such rewards---disulfide bond formation and cation-$\pi$ interactions---that exemplify this coupling.

Both rewards share the same form. Given residue-type masks $\phi_i, \psi_i \in \{0,1\}$ derived from the sequence, the soft contact count between eligible pairs is
\begin{equation}
  c_{ij} = \sigma\!\!\left(\frac{d_0 - \lVert \mathbf{c}_i - \mathbf{c}_j \rVert}{\tau}\right)
  \cdot m_{ij}^{\text{type}}
  \cdot \mathbf{1}[|i-j| \ge \Delta_{\min}]
  \cdot \mathbf{1}[i < j],
\end{equation}
where $\sigma(\cdot)$ is the sigmoid function, $\mathbf{c}_i$ are reconstructed C$\beta$ coordinates, $d_0$ is a distance threshold, and $\tau$ controls sigmoid sharpness.
The per-sample reward aggregates these contacts with a log-sum form to enable better gradient flow:
\begin{equation}
  R \;=\; \sum_{i=1}^{L} \log\!\left(1 + \textstyle\sum_{j} c_{ij}\right).
  \label{eq:reward_agg}
\end{equation}

\textbf{Disulfide bond reward.}\quad
C$\beta$ positions are reconstructed from the backbone N, C$\alpha$, C atoms of each residue using the ideal tetrahedral placement.
Disulfide C$\beta$--C$\beta$ distances peak at ${\sim}3.8$~\AA\ in the PDB.
We set $d_0 = 4.5$~\AA, $\tau = 0.3$~\AA, $\Delta_{\min} = 3$, and
$m_{ij}^{\text{SS}} = \mathbf{1}[s_i {=} \text{Cys}]\cdot\mathbf{1}[s_j {=} \text{Cys}]$.

\textbf{Cation-$\boldsymbol{\pi}$ reward.}\quad
Cation-$\pi$ C$\beta$--C$\beta$ distances span ${\sim}6$--$8$~\AA.
We define $\mathcal{C}^{+} = \{\text{Arg, Lys}\}$ and $\mathcal{A}^{\pi} = \{\text{Phe, Tyr, Trp, His}\}$ (His treated as aromatic only), with
$m_{ij}^{\text{cat-}\pi} = \mathbf{1}[s_i {\in} \mathcal{C}^{+}]\cdot\mathbf{1}[s_j {\in} \mathcal{A}^{\pi}] + \mathbf{1}[s_i {\in} \mathcal{A}^{\pi}]\cdot\mathbf{1}[s_j {\in} \mathcal{C}^{+}]$,
$d_0 = 8.0$~\AA, $\tau = 0.5$~\AA, $\Delta_{\min} = 4$.

\subsection{In-silico evaluation}

\subsubsection{Unconditional protein generation}

We sampled 100 structures at each of four target lengths (70, 100, 200, and 300 residues) using \nameshort with noisy guidance, as discussed above. 
Each generated structure was paired with the sequence produced jointly by the model. 
A design was classified as \textit{co-designable} if the generated sequence from the model can fold into the generated structure, where the threshold is root-mean-square deviation (RMSD) of $<$ 2.0 \AA\ of the backbone, as folded by ESMFold \cite{lin2023evolutionary}. 
The mean predicted local distance difference test (pLDDT) score was extracted from ESMFold and averaged over all residues for each design. We note that novelty/diversity metrics are only valid for co-designable samples (e.g. random, unfoldable sequences will have high novelty/diversity, but they are not useful), but other analyses (e.g. secondary structure, radius of gyration, etc.) are conducted for all samples as they reflect how well the model captures protein statistics.

\begin{figure}[htp]
    \centering
    \includegraphics[width=\linewidth]{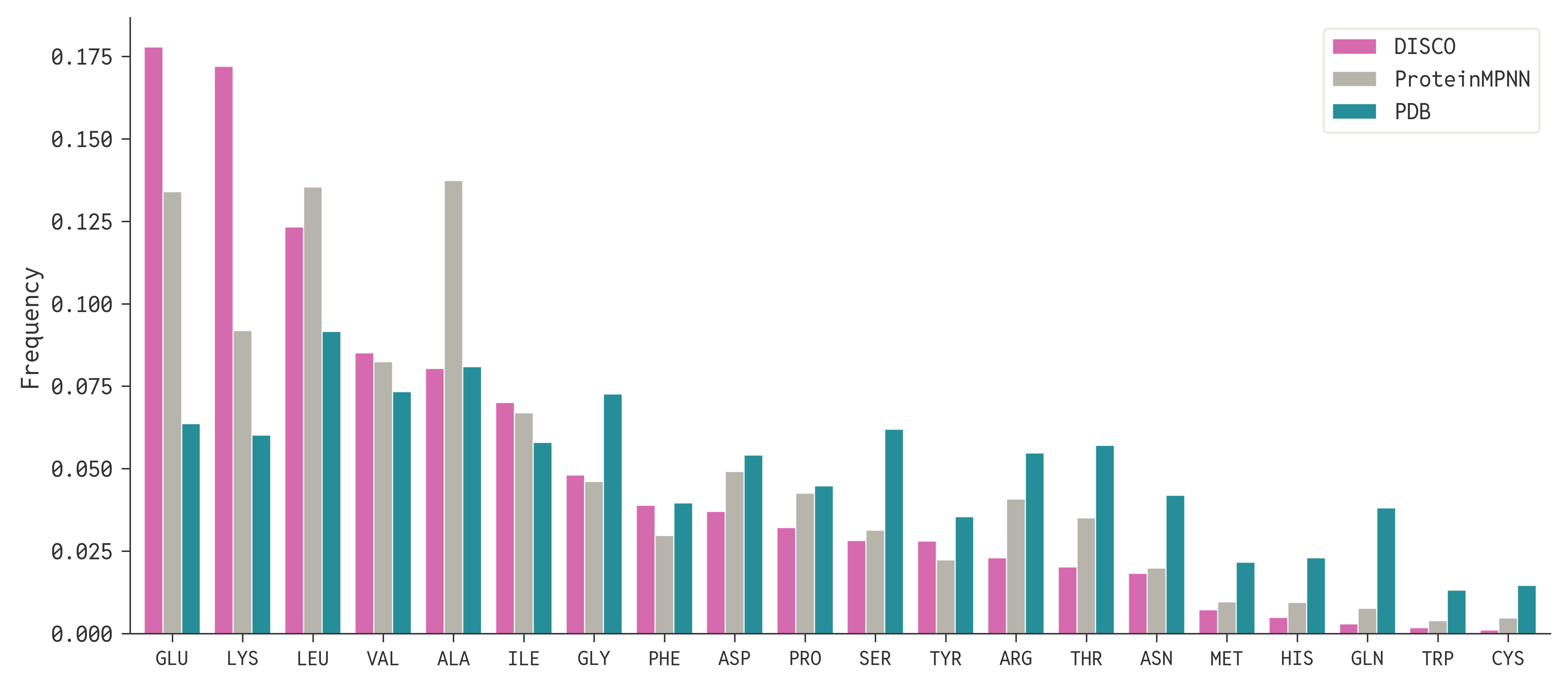}
    \caption{Amino acid distribution for unconditional generation using our sequence–structure model. Shown are sequences designed by our model, sequences obtained by inverse folding our generated structures with ProteinMPNN, and sequences from 50,000 randomly sampled PDB structures from the training data. }
    \label{fig:aa_distribution}
\end{figure}

To compare generated sequence properties against inverse-folding sequences (Figure \ref{fig:3}, Figure \ref{fig:aa_distribution}), ProteinMPNN \cite{dauparas2022robust} was applied to each generated backbone, producing 8 candidate sequences per structure.
Each was refolded with ESMFold, and the sequence yielding the lowest backbone RMSD was retained for comparison.

Secondary structure was assigned using the tool DSSP.
Eight-state assignments were reduced to three states: helix (H, G, I $\to$ H), strand (E, B $\to$ E), and coil (all others $\to$ C).

The relative contact order is defined by
\begin{equation}
  \text{RCO} = \frac{1}{L \cdot N} \sum_{\text{contacts}} |i - j|,
\end{equation}
where $L$ is the chain length and $N$ is the total number of qualifying contacts. A long-range contact between residues $i$ and $j$ was recorded if
\begin{equation}
  d_{\mathrm{C}\alpha}(i,j) < 8.0\;\text{\AA} \quad\text{and}\quad |i - j| \ge 12.
\end{equation}

\subsubsection{Sequence diversity and novelty}
Sequence diversity was quantified by clustering co-designable sequences with MMseqs2 \cite{steinegger2017mmseqs2} \texttt{easy-cluster}:
\begin{lstlisting}
mmseqs easy-cluster input.fasta clust tmp \
  --min-seq-id 0.3 -c 0.8 --cov-mode 0
\end{lstlisting}
The number of clusters divided by the total number of samples (400) served as the diversity metric.
Sequence novelty was assessed by searching against UniRef50 at maximum sensitivity:
\begin{lstlisting}
mmseqs easy-search input.fasta uniref50 result.m8 tmp --gpu 1
\end{lstlisting}
Sequence novelty was computed as the average of $1-\text{seq. identity}$.

\subsubsection{Structural diversity and novelty}

Structural diversity was measured by clustering co-designable backbones with Foldseek \cite{van2024fast} \texttt{easy-cluster} using TM-score--based alignment:
\begin{lstlisting}
foldseek easy-cluster input_dir clust tmp \
  --alignment-type 1 --cov-mode 0 \
  --min-seq-id 0 --tmscore-threshold 0.5
\end{lstlisting}
The number of clusters divided by the total number of samples (400) served as the diversity metric.

Structural novelty was evaluated by searching each backbone against the PDB:
\begin{lstlisting}
foldseek easy-search input_dir pdb result.m8 tmp \
  --alignment-type 1 --tmscore-threshold 0.0
\end{lstlisting}
Structure novelty was computed as the average of $1-\text{TM-score}$.

\subsubsection{Conditional evaluation of small-molecule ligands}

We assembled a curated set of 179 chemically diverse molecules (cofactors, metals, drug-like compounds, fluorophores, saccharides, pollutants, metabolites, etc.), including multiple different molecules. We call this benchmark \benchmark.

For each molecule, we optimized the geometry of RDKit ETKDG conformer with the semi-empirical method GFN2-xTB\cite{bannwarth2019gfn2}. The RDKit-derived bond information and the optimized geometry were used as the input bond/reference conformer information to the model. 50 protein--ligand complexes were generated at each of three target lengths (150, 200, and 250 residues) using  \nameshort with the inference settings specified in Table \ref{tab:cond_inf_hyperparameters}. 

Co-designability was assessed by refolding each complex with Chai-1 \cite{chai2024chai}.
A design was considered co-designable if both:
\begin{align}
  \text{RMSD}_{backbone}(\mathbf{X}_{\text{design}},\, \mathbf{X}_{\text{Chai}}) &< 2.0\;\text{\AA}, \label{eq:prot_cond}\\[4pt]
  \lVert \bar{\mathbf{r}}^{\,\ell}_{\text{design}} - \bar{\mathbf{r}}^{\,\ell}_{\text{Chai}} \rVert &< 2.0\;\text{\AA} \quad \forall\;\text{ligand}\;\ell, \label{eq:lig_cond}
\end{align}
where $\bar{\mathbf{r}}^{\,\ell}$ denotes the centroid of ligand $\ell$.
For multi-ligand systems, \cref{eq:lig_cond} was required to hold for every ligand independently.

A protein--ligand steric clash was flagged if any protein heavy atom in the backbone structure $i$ and
ligand heavy atom $j$ satisfied
$d_{ij} < r_i^{\text{vdW}} + r_j^{\text{vdW}} - 0.5\;\text{\AA}$,
where $r^{\text{vdW}}$ denotes the van der Waals radius and the
0.5~\AA\ tolerance accounts for minor overlap in experimental structures.

For logP and pocket hydrophobicity analysis, the binding site is defined as the set of protein residues whose backbone atoms lie within 10 Å of any heavy atom of the cofactor/ligand, retaining up to the 10 closest residues sorted by minimum protein–ligand distance.

\subsubsection{Conditional evaluation: nucleic acid binding}

We selected DNA/RNA sequences that were released in the PDB after our training cut-off date. Each DNA chain was cropped to a maximum length of 10. Designed protein--nucleic acid complexes were refolded with Chai-1. We followed the experimental design of the RFDiffusion3 DNA benchmark and for all DNA and RNA targets generated 100 samples with lengths sampled uniformly between lengths 50 to 80. Co-designability was assessed by first aligning nucleotide reference atoms between the design and the Chai-1 prediction, using phosphorus (P) atoms as the primary anchor.

A design was considered co-designable if
\begin{equation}
  \text{RMSD}_{backbone}^{\,\text{(global-aligned)}}(\mathbf{X}_{\text{design}},\, \mathbf{X}_{\text{Chai}}) < 2.0\;\text{\AA},
\end{equation}
ensuring that the protein refolds correctly and maintains the designed spatial relationship with its nucleic acid partner.

\subsubsection{Baseline methods}

For \emph{unconditional monomer generation}, we benchmarked against La-Proteina\cite{geffner2025proteina}, ProteinGenerator\cite{lisanza2025multistate}, ESM3\cite{hayes2025simulating}, RFdiffusion3\cite{butcher2025novo}, and BoltzGen\cite{stark_boltzgen_2025}.
For \emph{conditional generation} (small-molecule and nucleic acid binding), only RFdiffusion3 and BoltzGen were included, as the remaining baselines do not natively support ligand- or nucleic acid--conditioned generation. We additionally included RFDiffusion All-Atom \cite{krishna2024generalized} as a baseline for small molecule conditioning. For each designed backbone for RFDiffusion All-Atom, one sequence was generated using LigandMPNN\cite{dauparas2025atomic}, and was selected for downstream evaluation.
All baselines were run with publicly released code and recommended default parameters; the same downstream evaluation pipeline was applied identically. For RFDiffusion3, BoltzGen, and RFDiffusion All-Atom to allow for comparison to \nameshort we require each to co-fold the ligand or nucleic acid positions along with the protein backbone.

\subsubsection{Binding motif novelty}
\label{subsubsec:motif_novelty}
To assess whether \nameshort generates novel binding motifs, we searched designed active sites against known structures using Folddisco\cite{kim2025structural} with default parameters.
For each \emph{co-designable} protein--ligand complex, the binding motif was defined as the $k$ protein residues closest to the ligand (by side chain--ligand heavy-atom distance of refolded structure with Chai-1).
Searches were performed on Chai-1 refolded structures (which provide side-chain coordinates required by Folddisco) against the \texttt{afdb\_swissprot\_v4} Foldcomp database. Folddisco calculations are only carried out for co-designable structures, to avoid searching for ligands whose binding site cannot be agreed upon by Chai-1 and \nameshort .

A binding motif was classified as \emph{novel} if either: (i) Folddisco returned no match, or (ii) Folddisco returned a full match to all $k$ query residues but the matched-residue RMSD exceeded 3.0~\AA. 

\subsubsection{Binding motif diversity}
We now turn to assess the diversity of the generated binding motifs. The binding motif's definition is the same as ~\cref{subsubsec:motif_novelty}. Given two $K$-residue motifs $(P,\tau^P)$ and $(Q,\tau^Q)$ from
\emph{co-designable} proteins sharing the same ligand type, we report the aggregated per-sample metric of \emph{nearest-neighbour diversity} $\text{NN}_i = \min_{j\neq i}\,d(i,j)$ for a metric~$d$ detailed below. Furthermore, we also report 

\textbf{(i)~C$\alpha$-RMSD.}
Both coordinate sets are centred and then iteratively aligned using
the Kabsch algorithm \cite{kabsch1983} with the Hungarian
(linear-sum-assignment) optimal matching.  At each of $n=3$
iterations, the current assignment is solved on the squared-distance
cost matrix, and the optimal rotation is recomputed on the matched
pairs.  The final RMSD is
$d_{\text{RMSD}} = \sqrt{\frac{1}{K}\sum_{k=1}^{K}\|\mathbf{p}_k -
\mathbf{q}_{\sigma(k)}\|^2}$,
where $\sigma$ is the terminal assignment.

\textbf{(ii)~Chem-cost.}
A combined spatial--chemical distance is used.  Each residue is
assigned a chemical class (hydrophobic, polar, positive, negative, or
special).  A chemical distance matrix $D^{\text{chem}}_{ij}$ is
defined as $0$ for identical residue types, $0.5$ for same class, and
$1.0$ otherwise.  The Hungarian assignment is solved on
$C_{ij} = \alpha\,\|\mathbf{p}_i-\mathbf{q}_j\|^2 +
\beta\,D^{\text{chem}}_{ij}$
($\alpha=1$, $\beta=5$), alternated with Kabsch alignment for $n=3$
iterations.  The reported score is the mean per-residue combined
distance under the final matching:
$d_{\text{chem}} = \frac{1}{K}\sum_k \bigl(\alpha\,\|\mathbf{p}_k -
\mathbf{q}_{\sigma(k)}\| + \beta\,D^{\text{chem}}_{k,\sigma(k)}\bigr)$.

\textbf{(iii)~$1-\text{lDDT}$.}
After the spatial (RMSD) assignment $\sigma$ is established, the
internal distance matrices $D^P_{ij}=\|\mathbf{p}_i-\mathbf{p}_j\|$
and $D^Q_{ij}=\|\mathbf{q}_{\sigma(i)}-\mathbf{q}_{\sigma(j)}\|$ are
compared.  For every upper-triangular pair $(i,j)$, the absolute
difference $|D^P_{ij}-D^Q_{ij}|$ is tested against thresholds
$\{0.5,1.0,2.0,4.0\}$\,\AA.  The local distance difference test
score is $\text{lDDT}=(\text{preserved pairs})/
(|\text{pairs}|\times|\text{thresholds}|)$, and we report
$1-\text{lDDT}$ so that larger values indicate greater diversity.

\textbf{(iv)~Frobenius distance.}
This rotation-invariant metric compares the internal C$\alpha$
distance matrices directly.  An optimal residue permutation is found
by iteratively solving a Hungarian assignment on the row-permuted
distance-matrix cost ($n=3$ iterations).  The Frobenius distance is
$d_F = \sqrt{\frac{1}{\binom{K}{2}}
\sum_{i<j}(D^P_{ij}-D^Q_{\pi(i),\pi(j)})^2}$.

\textbf{(v) Motif cluster diversity.} 
With all pairwise C$\alpha$-RMSD values between motifs using iterative Hungarian matching with Kabsch alignment, we perform complete-linkage agglomerative clustering with a distance threshold
of $t = 2.0$\,\AA. The cluster ratio is defined as
$r = n_{\text{clusters}} / n_{\text{designs}}$, where $r \approx 1$ indicates that nearly every design adopts a structurally distinct active-site configuration, while $r \approx 0$ indicates that designs collapse onto a small number of shared motif geometries.

\subsection{Density Functional Theory calculations}
Density functional theory (DFT) calculations were performed with ORCA 6.0.1 using the hybrid B3LYP functional %\cite{becke1993density}
in combination with the def2-TZVP basis set \cite{ORCA}. %\cite{weigend2006accurate}. 
Dispersion effects were included with Grimme’s D3 correction %\cite{grimme2010consistent} 
and Becke–Johnson damping (D3BJ).% \cite{grimme2011effect}. 
Unless otherwise noted, all species were treated as triplet-state open-shell systems and were therefore computed using unrestricted DFT. Solvent effects were modeled with the SMD implicit solvation model %\cite{marenich2009universal}
using ethyl acetate as the solvent, chosen to approximate the dielectric environment of the enzyme active site. 

For carbene insertions, we optimized the pre-reactive intermediate corresponding to the approach of the carbene precursor to the heme iron center. With heme, multiple carbene insertion intermediates were computed. The heme cofactor was modeled with a 5-methylimidazole axial ligand; this ligand was included in the electronic-structure model but removed in the subsequent design stage. All reaction intermediates were geometry optimized and characterized by vibrational frequency analysis. For the calculation of energy barriers, transition state geometries were located using the NEB-TS method and frequency analysis indicates that they possess a single imaginary frequency along the reaction coordinate.
%\cite{henkelman2000climbing}
% \todo{[PLACEHOLDER FIGURE, UPDATE WITH NEW RXN, LATEST NUMBERS, SULFOXONIUM.]}
\begin{figure}[t]
    \centering
    \includegraphics[width=\linewidth]{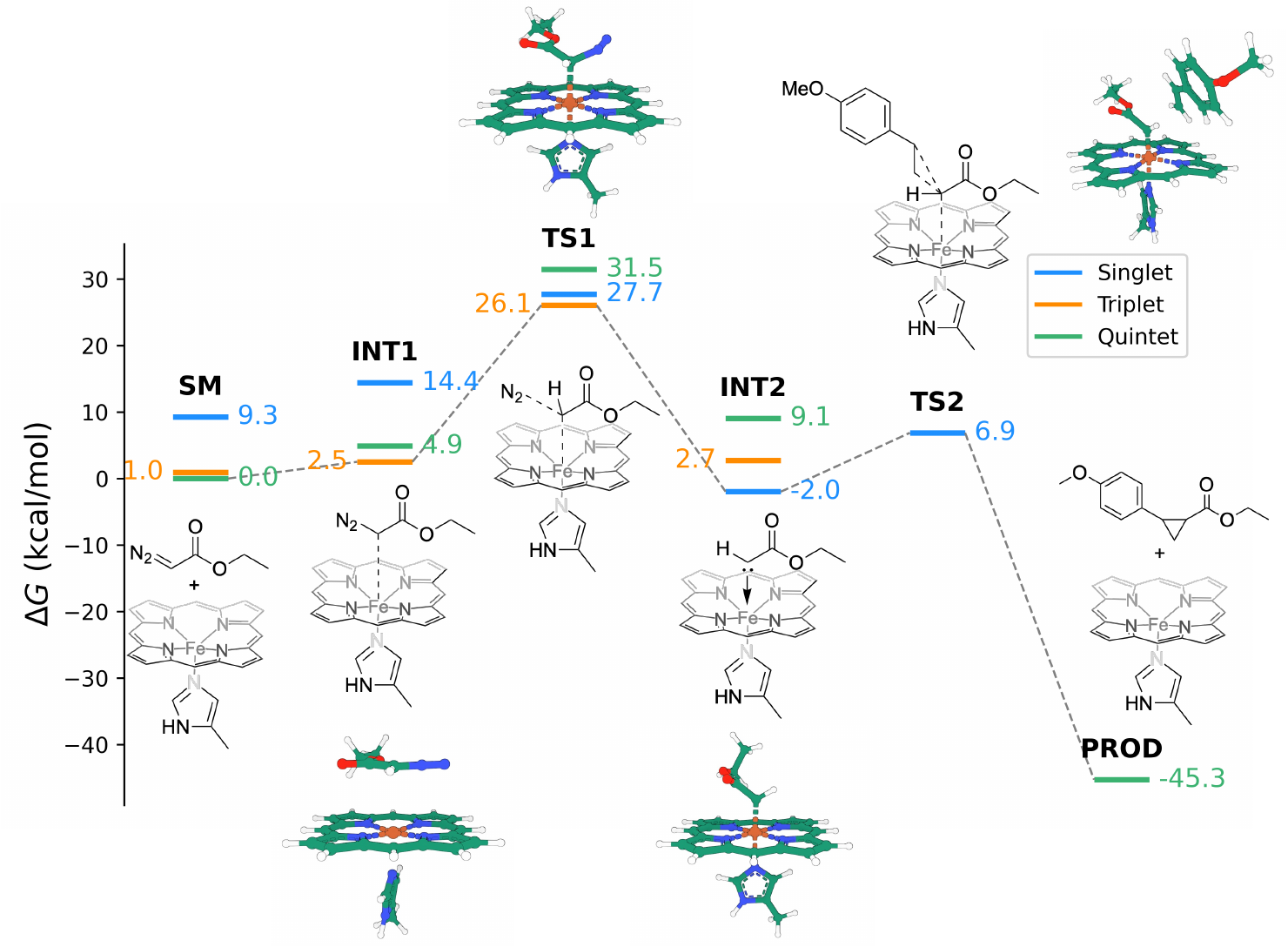}
    \caption{DFT-calculated reaction mechanism for the carbene-insertion reactions using Fe-porphyrin as the cofactor. All design campaigns used intermediate 1 reoptimized with heme b.}
    \label{fig:radius_of_gyration}
\end{figure}

\subsection{Design filtering for enzymes}\label{sec:filtering}
For every enzyme design campaign, \nameshort generated on the order of $10^4$ sequence--structure pairs conditioned on the target intermediate structure (including reference conformer and bonds), where \nameshort generates the structure for the reactive intermediate. 
These candidates were subjected to a multi-stage filtering pipeline combining orthogonal structure-prediction oracles, physicochemical filters, and diversity-aware selection.

Each candidate was first refolded with Chai-1, and structures satisfying the co-designability criterion (\cref{eq:lig_cond}) were advanced to a second round of structure prediction with AlphaFold 3.
In both rounds, the protein was co-folded with the reactive intermediate to assess active-site geometry.
AlphaFold 3 predictions were additionally performed with only the cofactor (omitting the intermediate).
A design was retained only if it was co-designable under both Chai-1 and AlphaFold 3.

Dual-oracle passing designs were subjected to the following filters:
\begin{enumerate}
  \item \textbf{Contacts.} The number of protein heavy atoms within 4.0~\AA\ of any intermediate heavy atom was required to be $\ge 5$.
  \item \textbf{Active-site burial.} 
  Three complementary metrics assessed ligand burial.
  First, the SASA burial fraction,
  \begin{equation}
    f_{\text{burial}} = 1 - \frac{A_{\text{complex}}}{A_{\text{free}}},
  \end{equation}
  was required to exceed 0.5.
  Second, we computed a solid-angle \emph{enclosure} score.
  From each of up to 15 ligand heavy-atom origins (selected by farthest-point sampling), $n = 500$ rays were cast uniformly over $S^2$ via a Fibonacci lattice.
  A ray from origin $\mathbf{o}$ was considered blocked by protein atom $j$ if
  \begin{equation}
    \cos\angle(\hat{\mathbf{r}},\,\hat{\mathbf{d}}_j) \;\ge\; \cos\!\left(\arctan\frac{r_j^{\text{vdW}} + r_{\text{probe}}}{\lVert \mathbf{p}_j - \mathbf{o}\rVert}\right),
  \end{equation}
  where $\hat{\mathbf{r}}$ is the ray direction, $\hat{\mathbf{d}}_j$ the unit vector from $\mathbf{o}$ to protein atom $\mathbf{p}_j$, $r_j^{\text{vdW}}$ its van der Waals radius, and $r_{\text{probe}} = 1.4$~\AA.
  The enclosure from each origin is the fraction of blocked rays; the \emph{worst-case} (minimum) across all origins was required to exceed 0.5.
  Third, the \emph{maximum exposure angle} was estimated from the centroid origin by clustering unblocked rays via greedy single-linkage and reporting the bounding-cone half-angle of the largest cluster; this was required to be $\le 65^{\circ}$.
  \item \textbf{Metal coordination.} For metalloenzyme campaigns, proper coordination geometry of catalytic metal ions was verified (correct number and identity of coordinating residues, bond lengths within $\pm$0.5~\AA\ of ideal values).
  \item \textbf{Surface hydrophobicity.} The fraction of solvent-exposed hydrophobic residues (Ala, Val, Ile, Leu, Met, Phe, Trp, Pro) was required to be $<$50\% to reduce aggregation propensity.
  \item \textbf{Net charge.} The formal net charge at pH~7 was checked to be within $\pm 15$.
\end{enumerate}

Designs were then ranked within each cluster by a simple composite confidence score:
\begin{equation}
  S = 0.5 \cdot \text{ipTM} + 0.25 \cdot \text{pTM} - 0.5 \cdot \text{PAE}_{\text{chain}} + 1.0 \cdot \text{ipTM}_{\text{chain}},
  \label{eq:rank_score}
\end{equation}
where ipTM and pTM are the AlphaFold 3 global interface and predicted TM-scores, $\text{PAE}_{\text{chain}}$ is the pairwise aligned error between the protein and ligand chains, and $\text{ipTM}_{\text{chain}}$ is the pairwise interface pTM between the ligand and protein chains.
In addition to the composite score, hard cutoffs were enforced: ipTM $\ge 0.7$, pTM $\ge 0.7$, and chain-pair ipTM $\ge 0.5$.

Passing designs were clustered at both the sequence level (MMseqs2, 30\% identity) and the structure level (Foldseek, TM-score 0.5). 
Within each sequence cluster, at most 2 sequences were retained. Within each structure cluster, at most 8 structures were retained.

\clearpage
\subsection{Additional in-silico results}
\subsubsection{Additional unconditional generation results}
We note that as co-designability is very high for unconditional generation results, sample analysis of generated proteins (e.g. amino acid distribution, secondary structure distribution, radius of gyration, relative contact order, etc.) show no qualitative changes when filtered for co-designability. 

\begin{table}[htb]
  \centering
  \caption{Unconditional generation metrics. \textbf{Bold} indicates best per column; \underline{underline} indicates second best. Co-designable structural and sequence diversity are the most informative metrics because they reflect both sample quality and the absence of mode collapse, whereas one should not pursue the maximization of co-designability alone, as folding models cannot refold every PDB protein within 2~\AA~RMSD. Note that only \nameshort, BoltzGen and RFDiffusion3 can condition on other biomolecules.}
  \label{tab:main-results}
\begin{tabular}{@{}l cccccc@{}}
    \toprule
    Method
      & Co-designable
      & \makecell{Struct.\\Diversity}
      & \makecell{Seq.\\Diversity}
      & \makecell{Struct.\\Novelty}
      & \makecell{Seq.\\Novelty}
      & pLDDT \\
    \midrule
    ESM3              & 0.145          & 0.040          & 0.113          & \textbf{0.697} & 0.806          & 0.533 \\
    ProteinGenerator  & 0.355          & 0.118          & 0.273          & 0.173          & \textbf{0.998} & 0.779 \\
    La-Prote\'{i}na   & \textbf{0.895} & \underline{0.293} & \underline{0.790} & 0.173       & 0.879          & \textbf{0.917} \\
    BoltzGen          & 0.798          & 0.065          & 0.310          & 0.132          & 0.865          & \underline{0.882} \\
    RFDiffusion3      & 0.460          & 0.135          & 0.450          & 0.166          & 0.926          & 0.766 \\
    \midrule
    \nameshort (ours) & \underline{0.878} & \textbf{0.343} & \textbf{0.855} & \underline{0.211} & \underline{0.961} & 0.879 \\
    \bottomrule
  \end{tabular}
\end{table}

\begin{figure}[h]
    \centering
    \includegraphics[width=\linewidth]{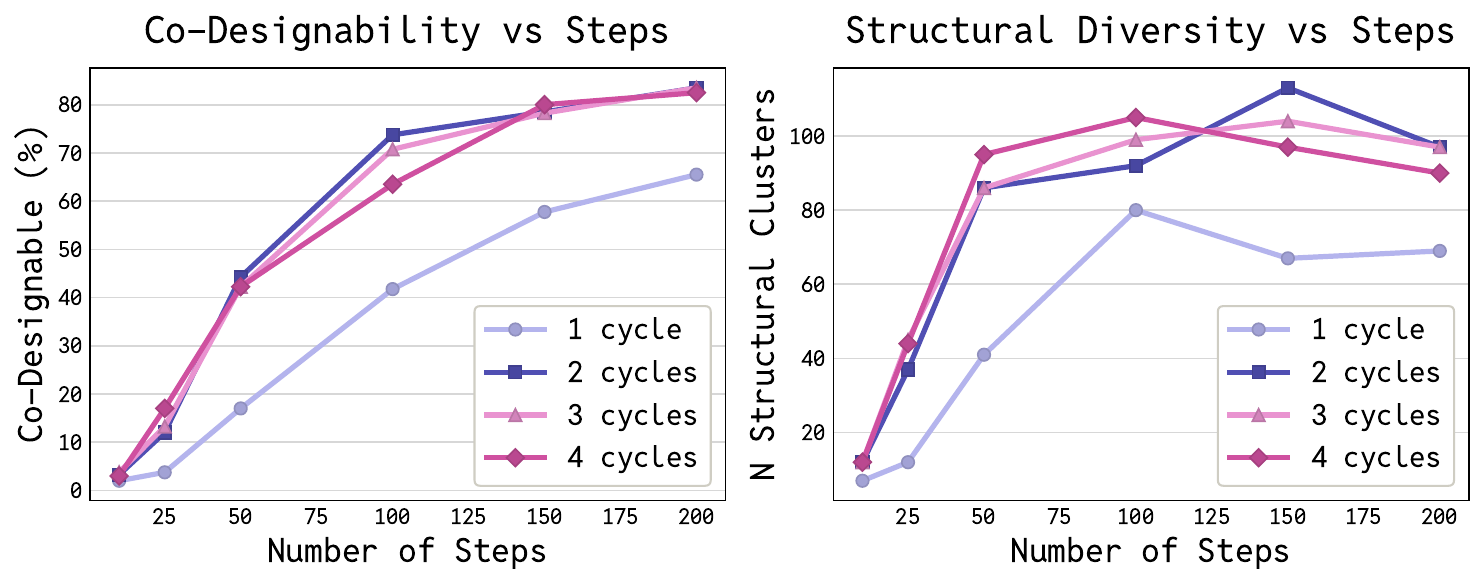}
    \caption{Co-designability and number of structure clusters for different numbers of steps and recycles. Performed on the unconditional task with the settings of Table \ref{tab:uncond_inf_hyperparameters} and noisy guidance disabled.}
    \label{fig:unconditional_n_steps_cycles}
\end{figure}
 
\begin{figure}[h]
    \centering
    \begin{subfigure}[b]{0.48\textwidth}
        \centering
        \includegraphics[width=\textwidth]{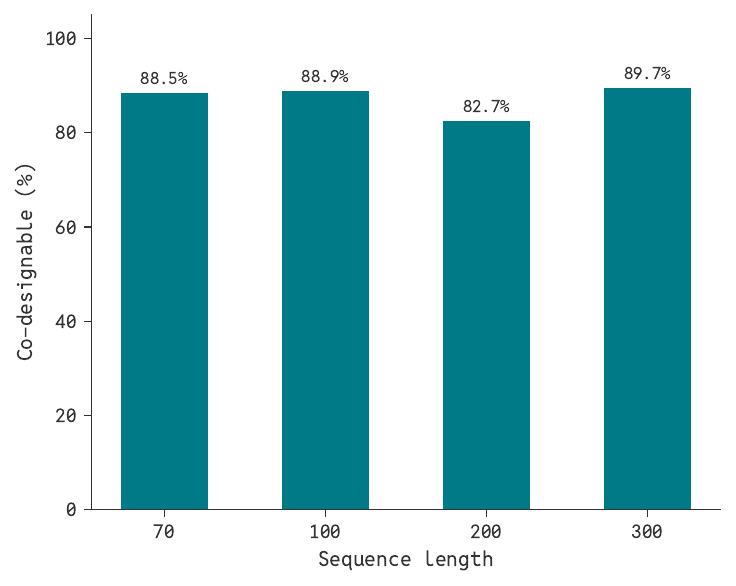}
    \end{subfigure}
    \hfill
    \begin{subfigure}[b]{0.48\textwidth}
        \centering
        \includegraphics[width=\textwidth]{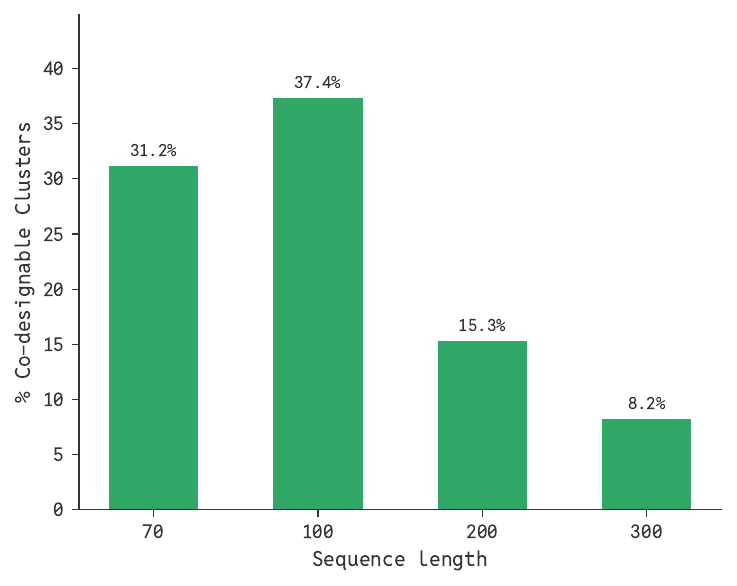}
    \end{subfigure}
    \caption{Distribution of unconditional generation co-designability and the proportion of co-designable clusters across protein lengths.}
    \label{fig:unconditional_metrics_by_length}
\end{figure}

\begin{figure}[t]
    \centering
    \includegraphics[width=0.7\linewidth]{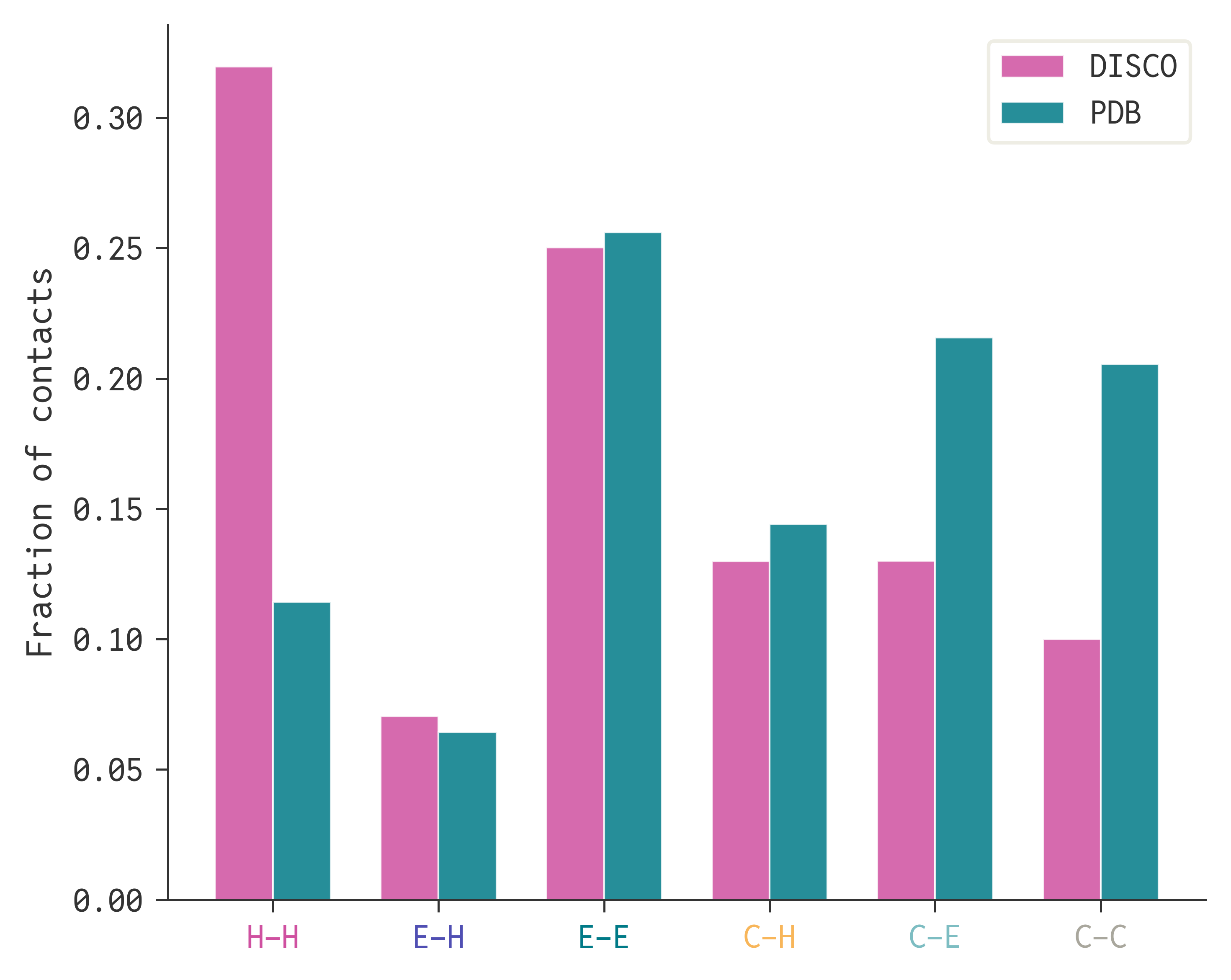}
    \caption{Distributions of long-range contact types in unconditionally generated samples. We include a comparison against 50,000 randomly sampled PDB structures from the training data.}
    \label{fig:contact_types}
\end{figure}

\begin{figure}[t]
    \centering
    \includegraphics[width=0.7\linewidth]{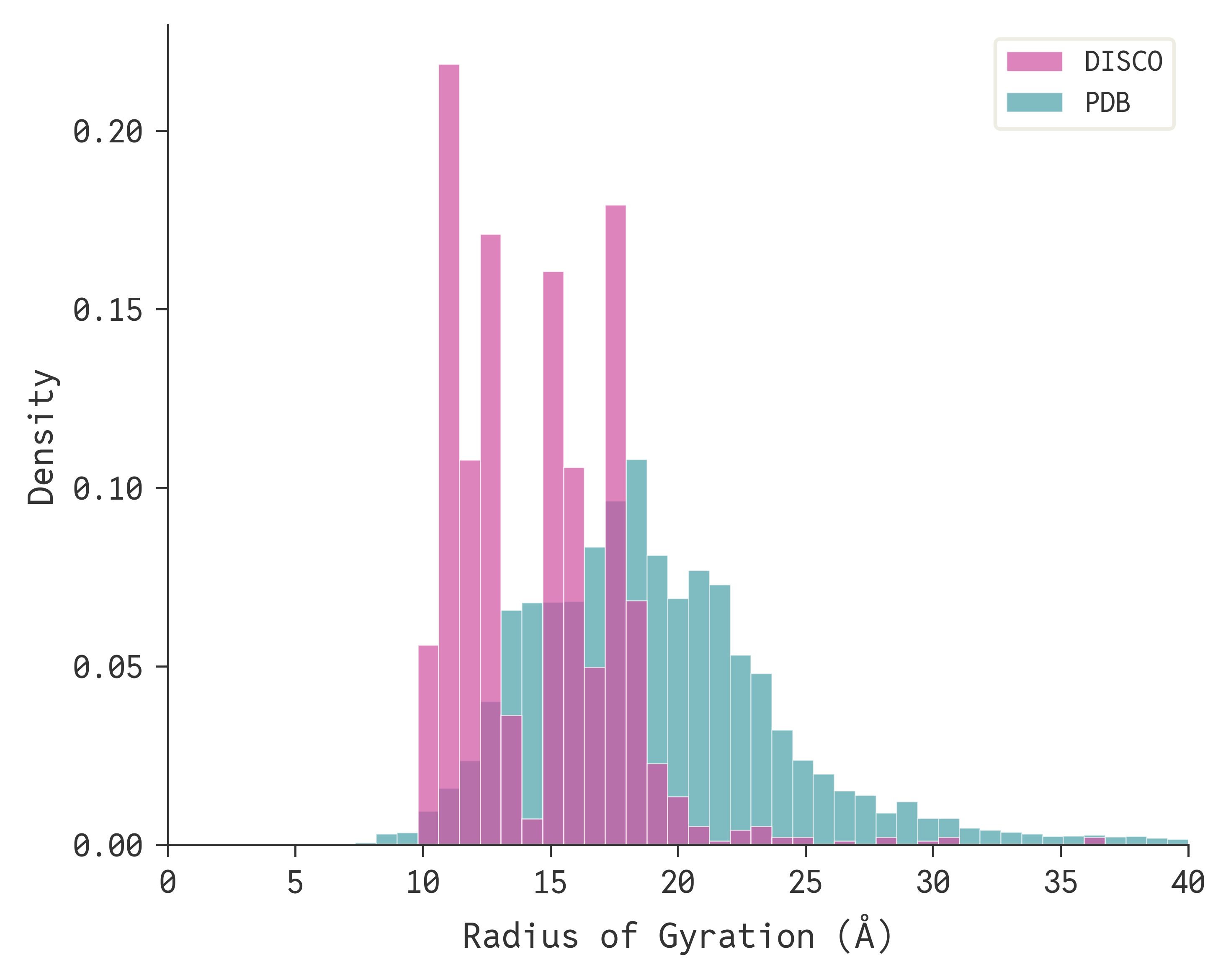}
    \caption{Distributions of radius of gyration in unconditionally generated samples. We include a comparison against 50,000 randomly sampled PDB structures from the training data.}
    \label{fig:rog}
\end{figure}

\clearpage
\subsubsection{Additional conditional generation results}
\begin{figure}[t]
    \centering
    \begin{subfigure}[b]{\textwidth}
        \centering
        \includegraphics[width=0.65\textwidth]{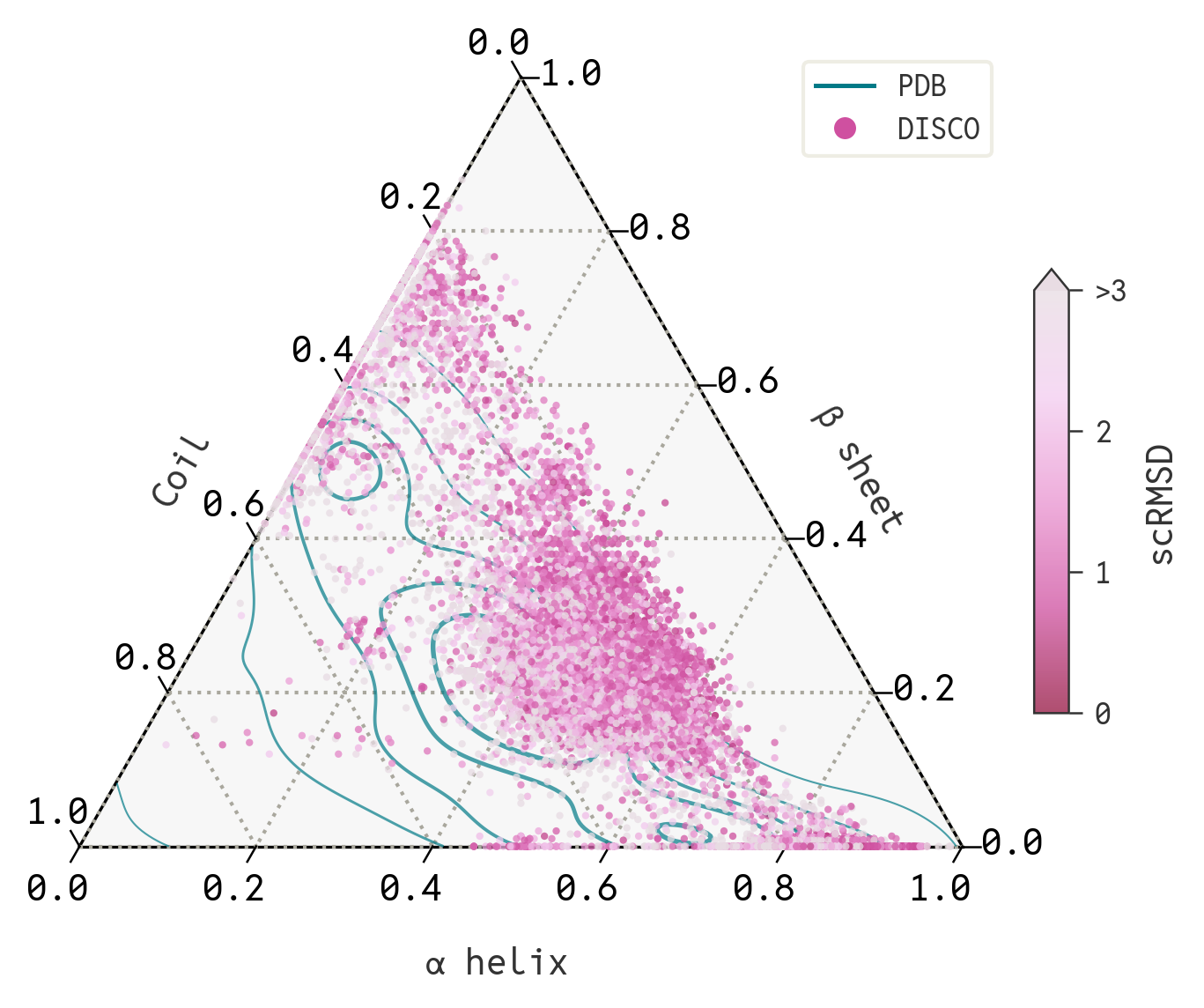}
        \label{fig:cond_ternary_all}
    \end{subfigure}
    \hfill
    \begin{subfigure}[b]{0.65\textwidth}
        \centering
        \includegraphics[width=\textwidth]{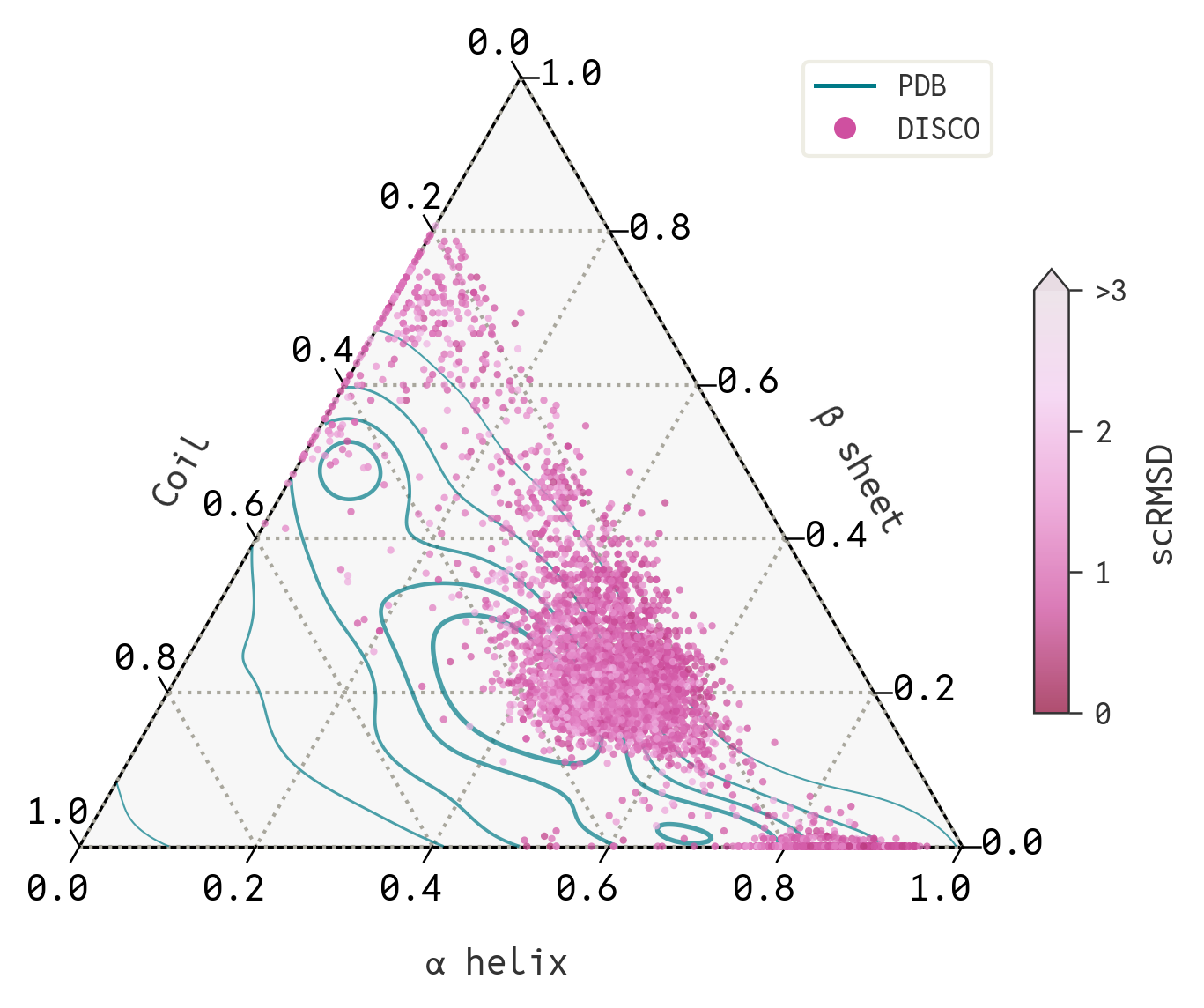}
        \label{fig:cond_ternary_codesignable}
    \end{subfigure}
    \caption{Secondary structure distribution for conditional generation using our sequence–structure model. Top shows results from all benchmark ligands without filtering, and bottom shows the effect of filtering by co-designability. We include a comparison against 50,000 randomly sampled PDB structures from the training data. Because current folding methods successfully refold only about 60\% of the PDB training set, co-designability is likely a biased filter.}
    \label{fig:cond_ss_distribution}
\end{figure}

We note that while co-designability is not as high as unconditional generation results, sample analyses of generated proteins generally showed no qualitative changes when filtered for co-designability. The only significant change can be seen in \cref{fig:cond_ss_distribution}. 

With the exception of motif novelty and diversity (and unless otherwise noted), all sample analysis are evaluated without co-designability filtering. Like unconditional generation analysis, general properties (e.g. conformer PoseBusters validity, steric clashes with ligands, logP of ligand vs pocket chemical features) reflect whether the model captures reasonable chemical fidelity; once this is satisfied \emph{and} the protein is co-designable, only then does a motif's novelty and diversity matter (e.g. one can easily sample random pockets that do not fold).  

\begin{figure}[t]
    \centering
    \includegraphics[width=0.7\linewidth]{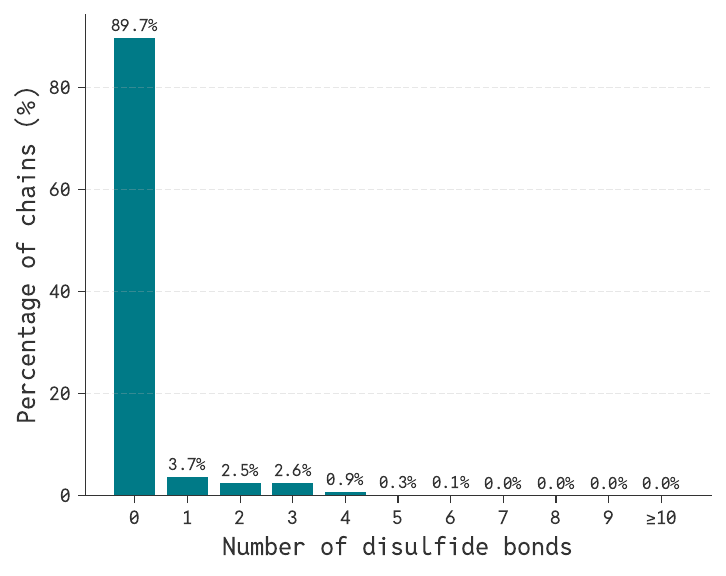}
    \caption{Distribution of the number of disulfide bonds present in each protein chain of the PDB training data, where the length of the chain is less than or equal to 100.}
    \label{fig:disulfide_distribution}
\end{figure}

\begin{figure}[t]
    \centering
    \includegraphics[width=\linewidth]{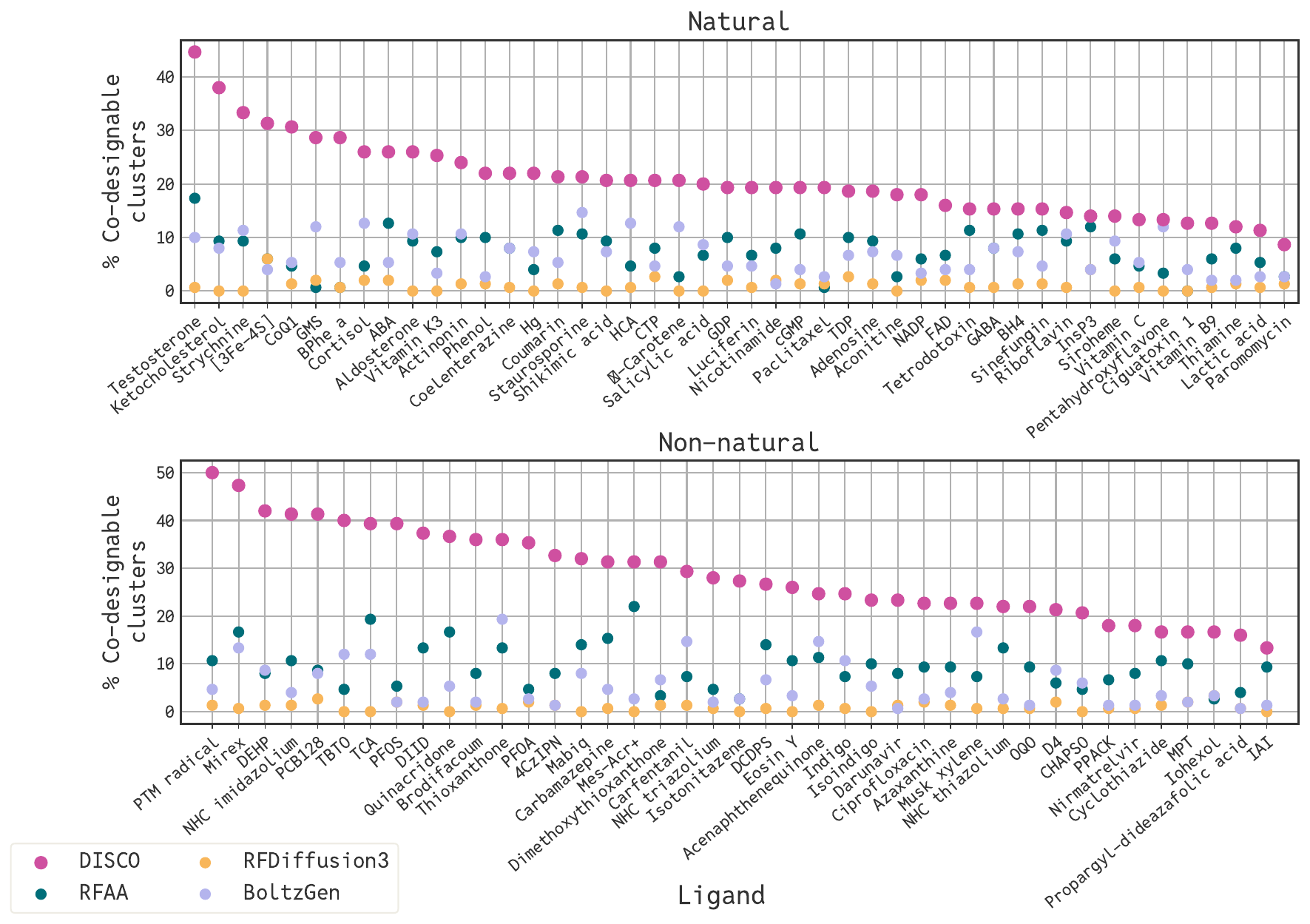}
    \caption{Benchmark performance across the remaining set of \benchmark, ranging from natural to non-natural ligands. We observe that for all ligands \nameshort generates the highest proportion of diverse, co-designable sequence-structure pairs.}
    \label{fig:p2_p3_ligands}
\end{figure}

\begin{figure}[h]
    \centering
    \begin{subfigure}[b]{0.65\textwidth}
        \centering
        \includegraphics[width=\textwidth]{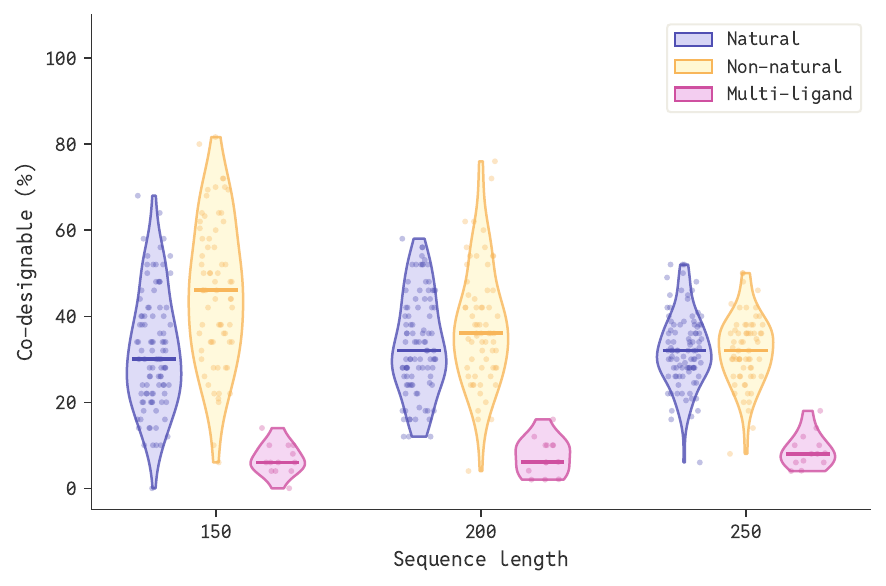}
    \end{subfigure}
    \hfill
    \begin{subfigure}[b]{0.65\textwidth}
        \centering
        \includegraphics[width=\textwidth]{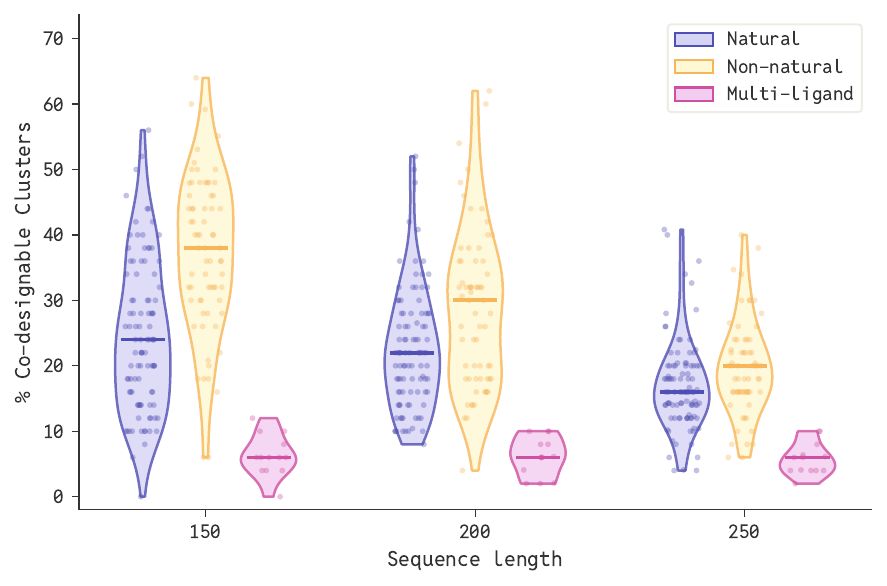}
    \end{subfigure}
    \caption{Distribution of average conditional generation co-designability and the proportion of co-designable clusters across protein lengths, where each scatter point corresponds to a ligand type.}
    \label{fig:conditional_metrics_by_length}
\end{figure}

\begin{figure}[t]
    \centering
    \begin{subfigure}[b]{0.48\textwidth}
        \centering
        \includegraphics[width=\textwidth]{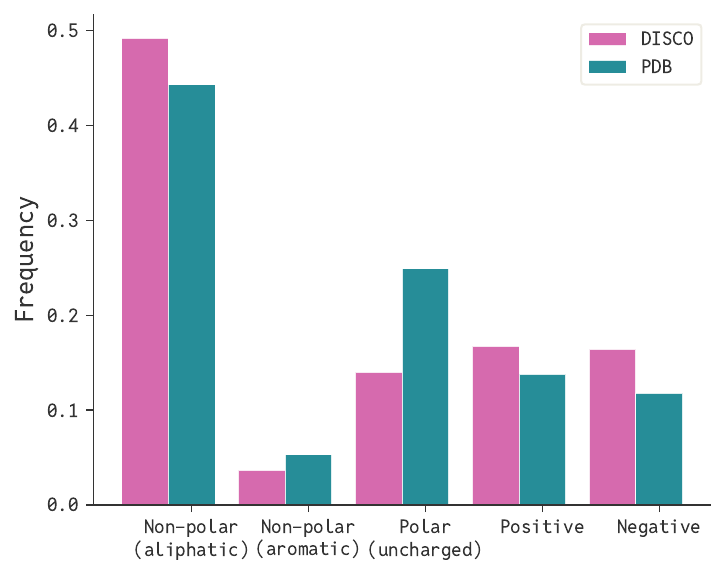}
    \end{subfigure}
    \hfill
    \begin{subfigure}[b]{0.48\textwidth}
        \centering
        \includegraphics[width=\textwidth]{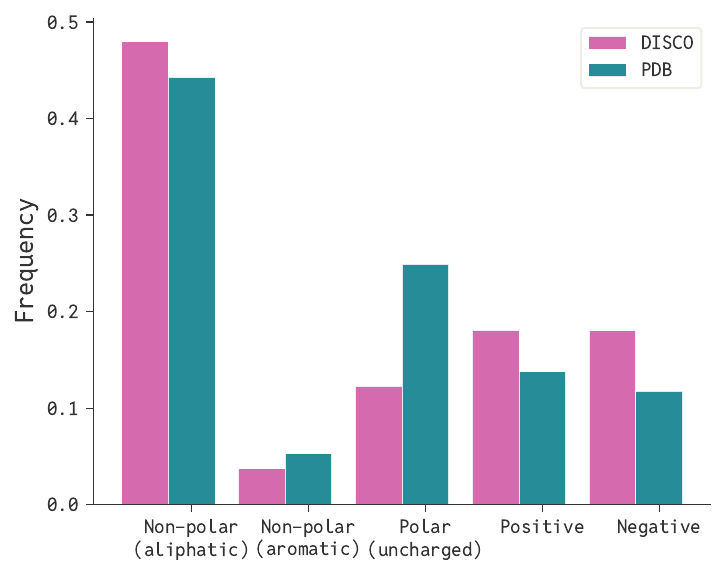}
    \end{subfigure}
    \begin{subfigure}[b]{\textwidth}
        \centering
        \includegraphics[width=\linewidth]{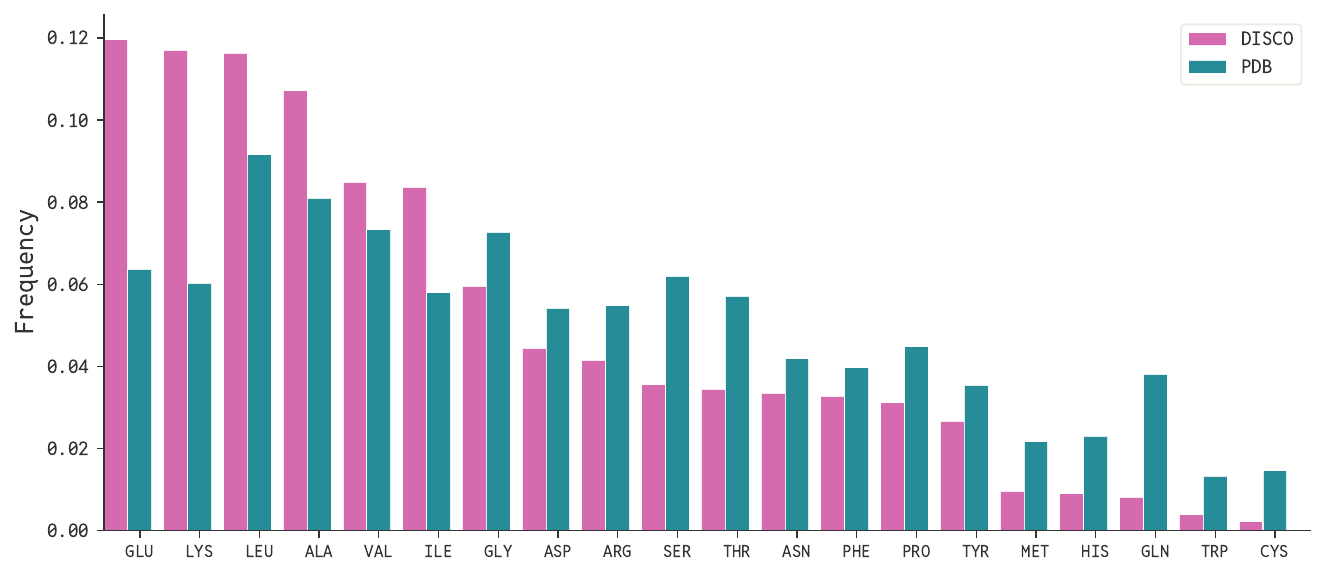}
    \end{subfigure}
    \caption{Distribution of amino acids for conditional generation using our sequence–structure model. Results are aggregated from all benchmark ligands. Shown are sequences designed by our model and sequences from 50,000 randomly sampled PDB structures from the training data. Top left shows all generated samples, top right shows only co-designable samples, and bottom shows all generated samples, but split into specific amino acids. Note that despite reduced co-designability, there is no significant change in the distribution of the types of amino acids.}
    \label{fig:cond_aa_distribution}
\end{figure}

% \todo{do we want to add LigandMPNN?}

\begin{figure}[t]
    \centering
    \begin{subfigure}[b]{\textwidth}
        \centering
        \includegraphics[width=0.65\textwidth]{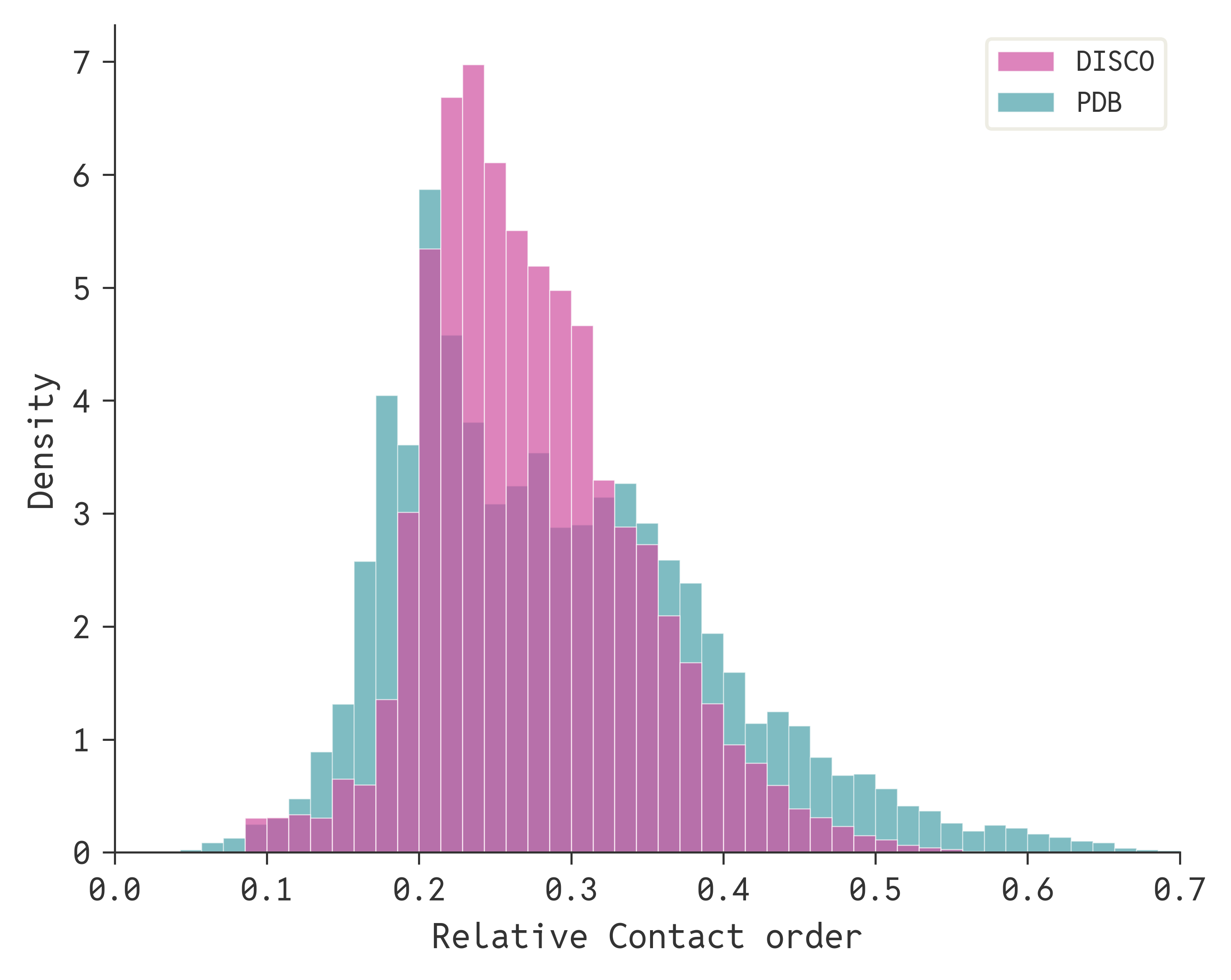}
        \label{fig:cond_contact_order_distribution_all}
    \end{subfigure}
    \hfill
    \begin{subfigure}[b]{0.65\textwidth}
        \centering
        \includegraphics[width=\textwidth]{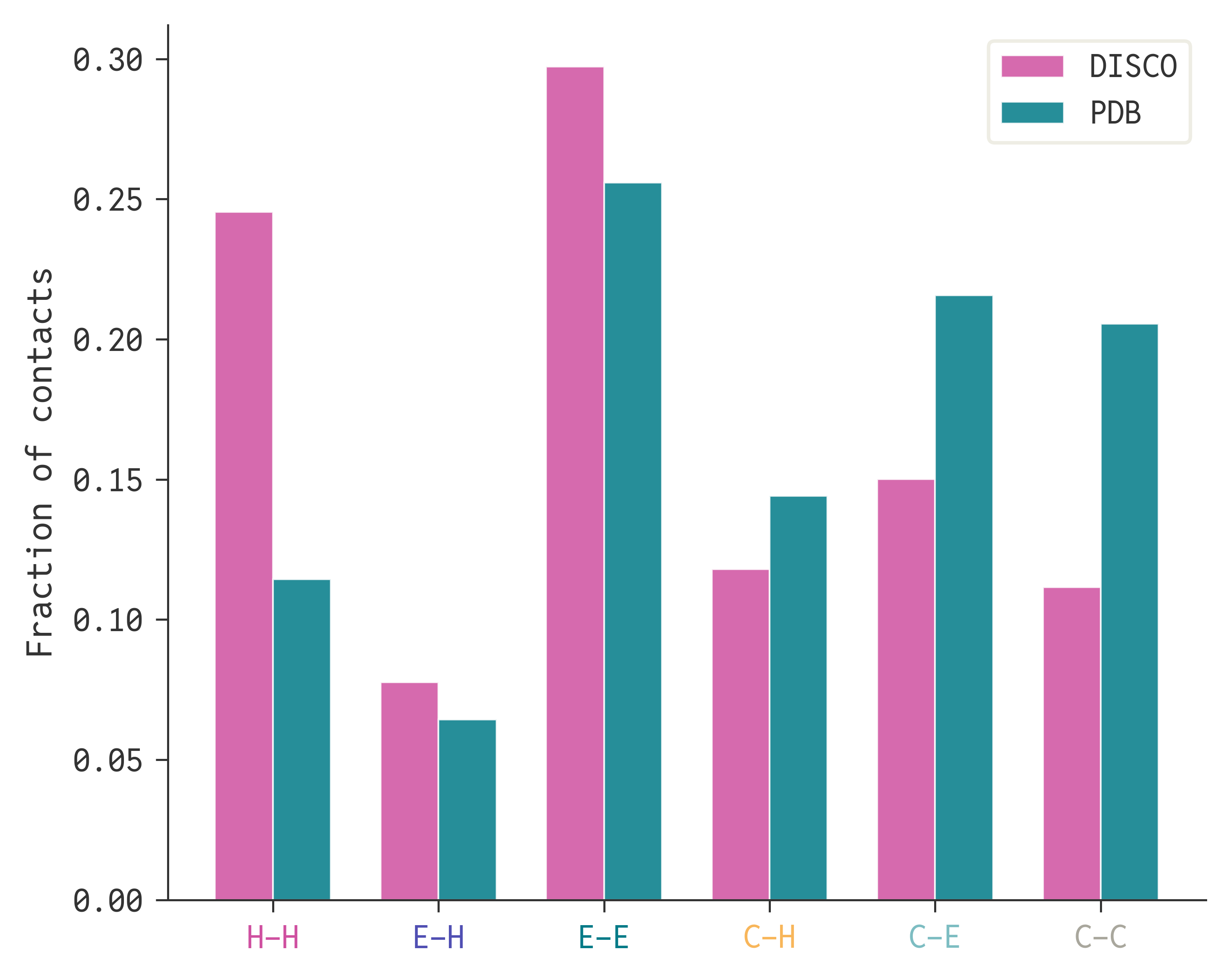}
        \label{fig:cond_contact_bar_distribution_all}
    \end{subfigure}
    \caption{Distributions of relative contact order for conditional generation using our sequence–structure model (top) and long-range contact types (bottom). Results are aggregated from all benchmark ligands without co-designability filtering. We include a comparison against 50,000 randomly sampled PDB structures from the training data. }
    \label{fig:cond_contact_distribution}
    % \centering
    % \includegraphics[width=\linewidth]{figures/conditional_contact_order_distributions_all.png}
    % \caption{Distributions of relative contact order for conditional generation using our sequence–structure model. Results are aggregated from all benchmark ligands without co-designability filtering. We include a comparison against 50,000 randomly sampled PDB structures from the training data. }
    % \label{fig:cond_contact_order_distribution}
\end{figure}

% \begin{figure}[t]
%     \centering
%     \includegraphics[width=\linewidth]{figures/conditional_contact_bars_all.png}
%     \caption{Distributions of long-range contact types for conditional generation using our sequence–structure model. Results are aggregated from all benchmark ligands without co-designability filtering. We include a comparison against 50,000 randomly sampled PDB structures from the training data.}
%     \label{fig:cond_contact_bar_distribution}
% \end{figure}

\begin{figure}[t]
    \centering
    \includegraphics[width=\linewidth]{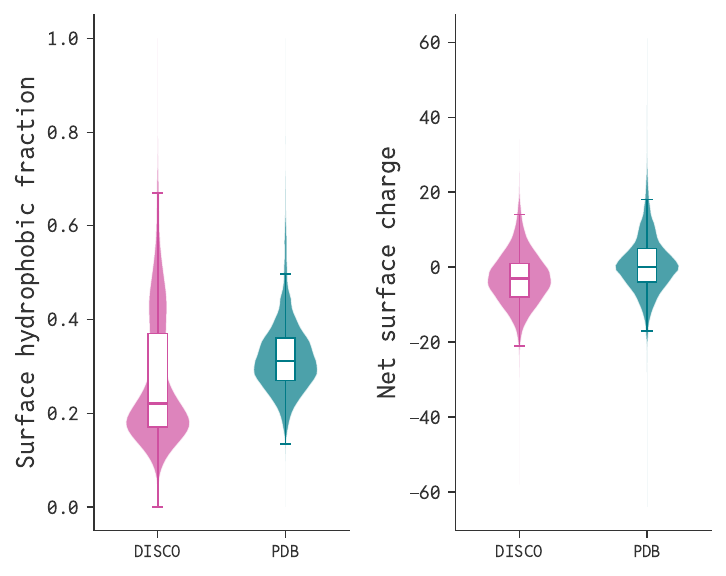}
    \caption{Distribution of surface hydrophobicity and net charge for conditional generation using our sequence–structure model. Results are aggregated from all benchmark ligands without co-designability filtering. We include a comparison against 50,000 randomly sampled PDB structures from the training data.}
    \label{fig:cond_surface_distribution}
\end{figure}

\begin{figure}[t]
    \centering
    \includegraphics[width=0.6\linewidth]{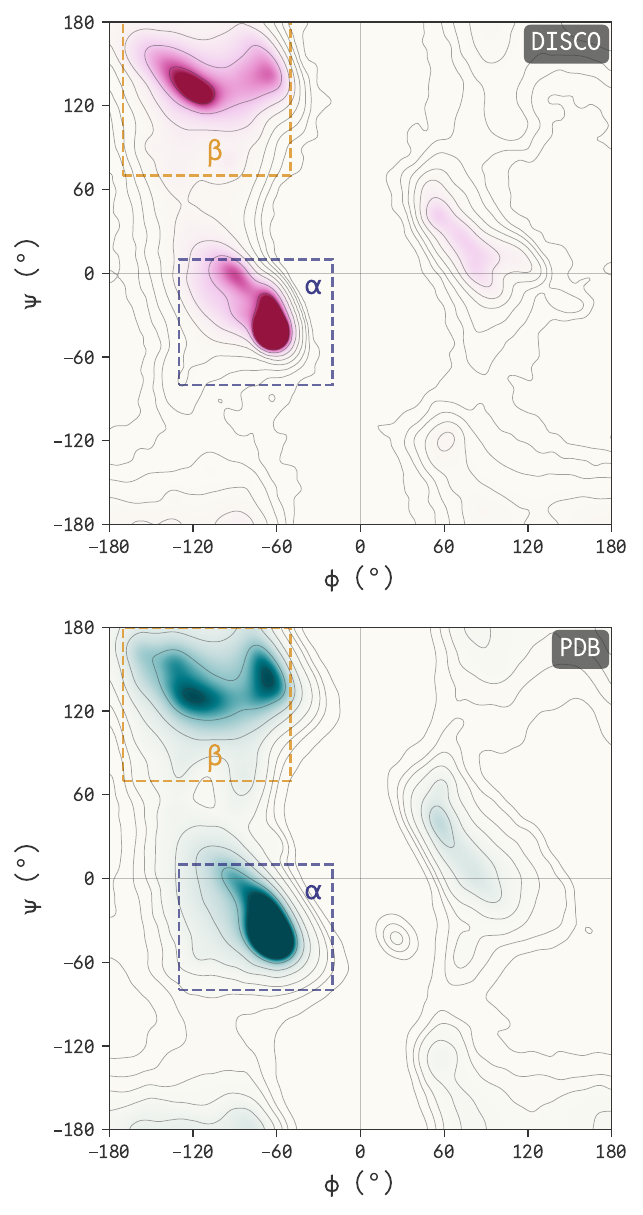}
    \caption{Ramachandran plot for conditional generation using our sequence–structure model. Results are aggregated from all benchmark ligands without co-designability filtering. We include a comparison against 50,000 randomly sampled PDB structures from the training data.}
    \label{fig:cond_rama_distribution}
\end{figure}

\begin{figure}[t]
    \centering
    \includegraphics[width=0.7\linewidth]{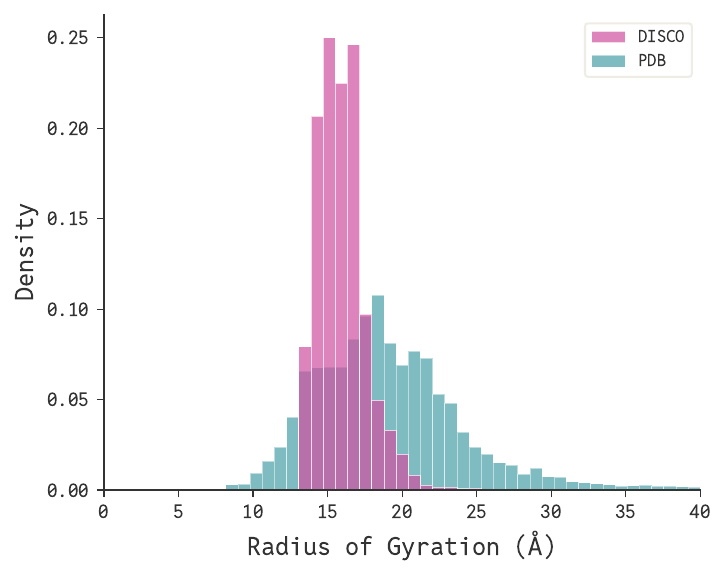}
    \caption{Distributions of radius of gyration for conditional generation using our sequence–structure model. Results are aggregated from all benchmark ligands without filtering. We include a comparison against 50,000 randomly sampled PDB structures from the training data.}
    \label{fig:cond_rog_distribution}
\end{figure}

\begin{figure}[htp]
    \centering
    \includegraphics[width=0.7\linewidth]{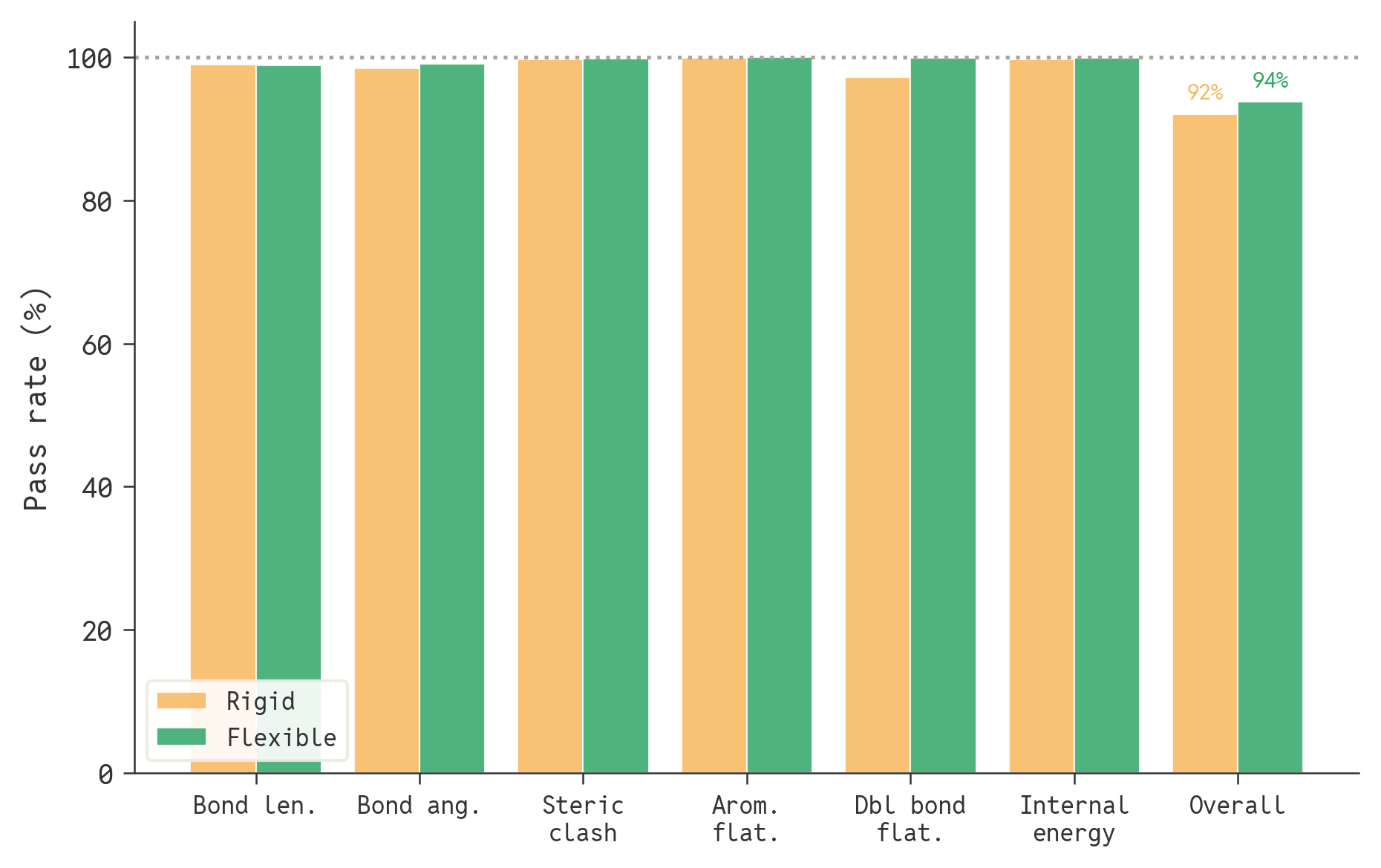}
    \caption{PoseBusters validity breakdown for conditional generation results without co-designability filtering.}
    \label{fig:posebusters_rate}
\end{figure}

\begin{figure}[t]
    \centering
    \includegraphics[width=0.7\linewidth]{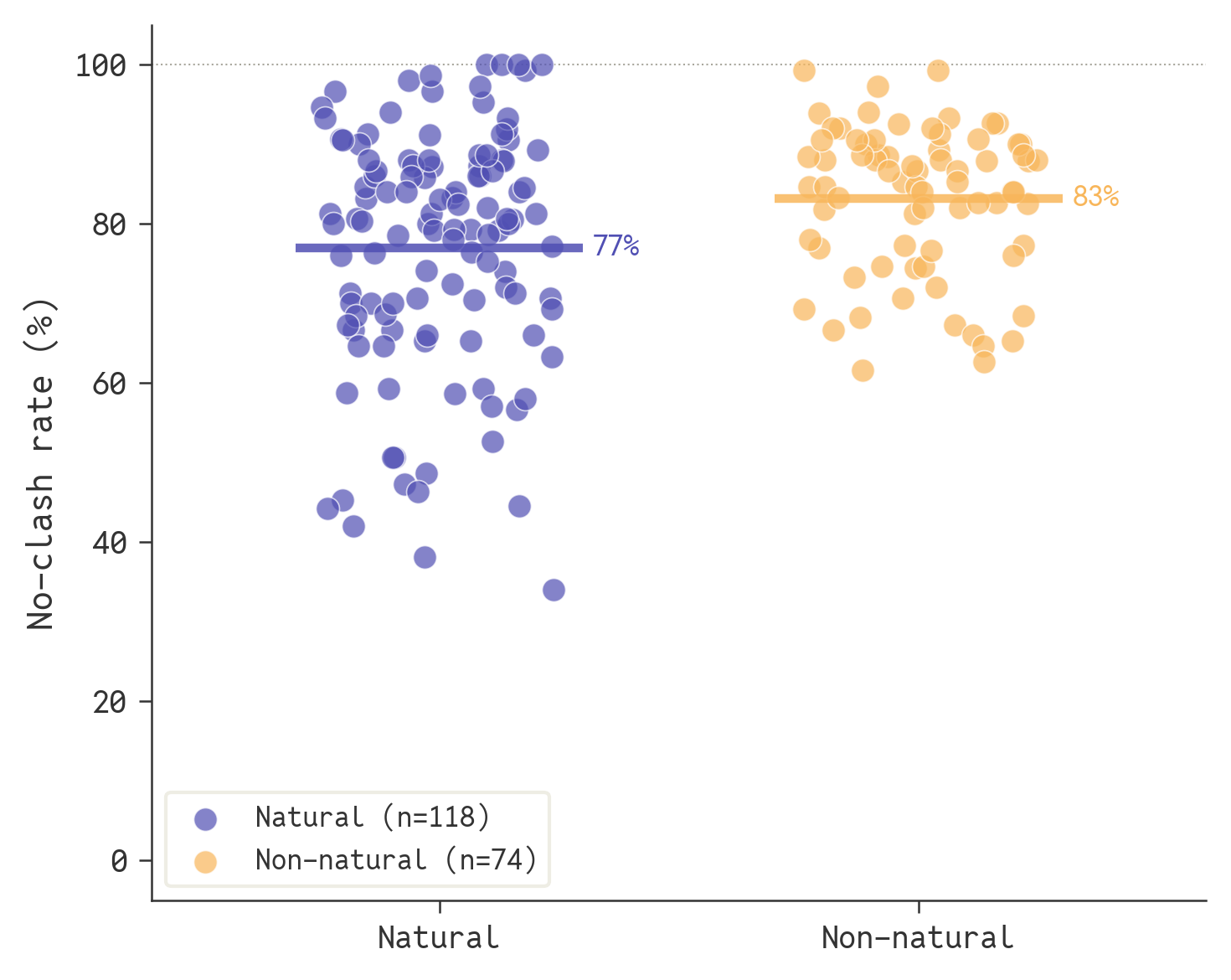}
    \caption{Average rate of any steric clash between the designed protein structure and the ligand(s), grouped by ligand type, without co-designability filtering.}
    \label{fig:steric_clashes}
\end{figure}

\begin{figure}[t]
    \centering
    \includegraphics[width=0.7\linewidth]{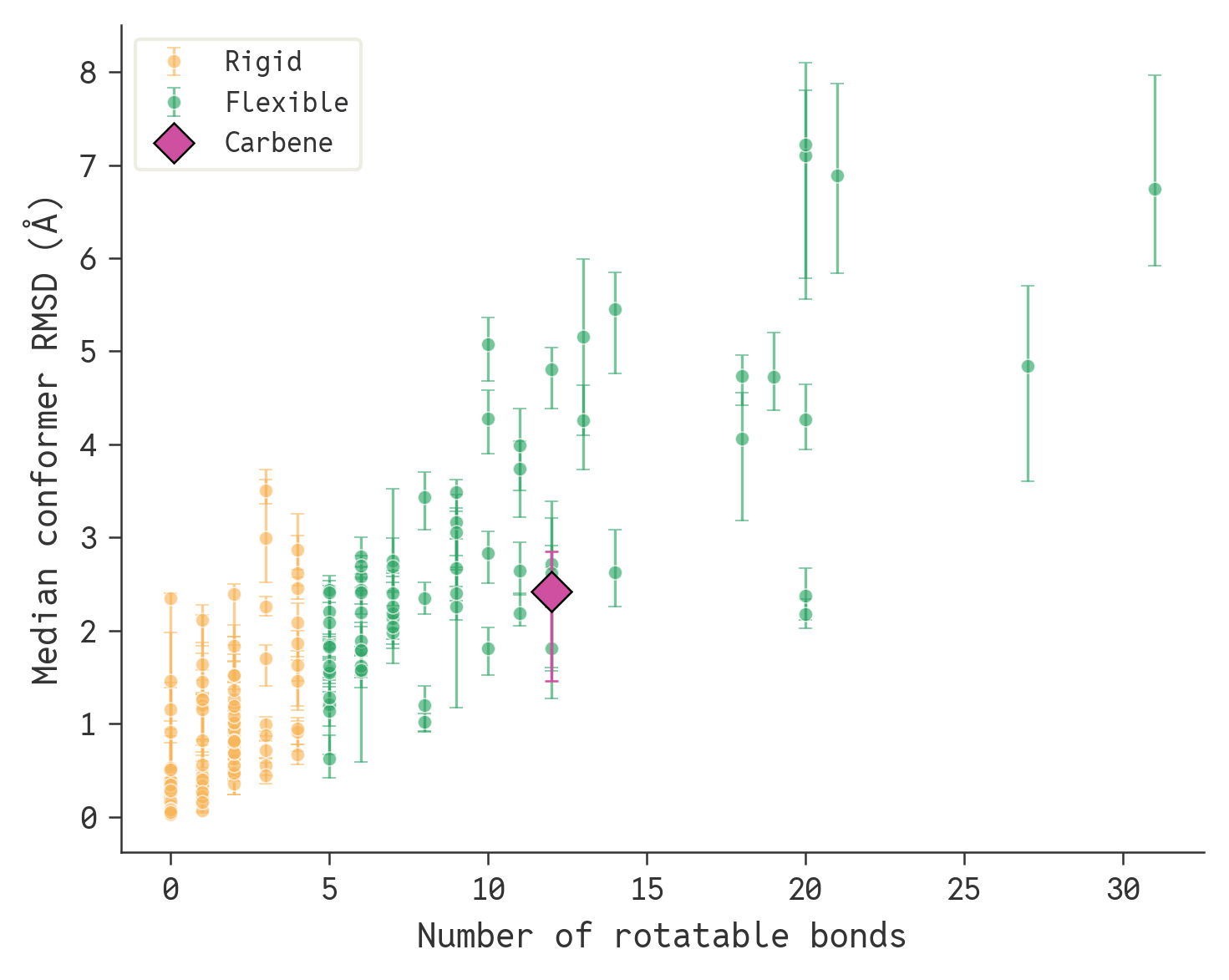}
    \caption{Number of rotatable bonds compared to the median of conformer RMSD (with standard deviation), split by ligand type, without co-designability filtering.}
    \label{fig:num_rot_vs_rmsd}
\end{figure}

\begin{figure}[h]
    \centering
    \begin{subfigure}[b]{\textwidth}
        \centering
        \includegraphics[width=0.7\textwidth]{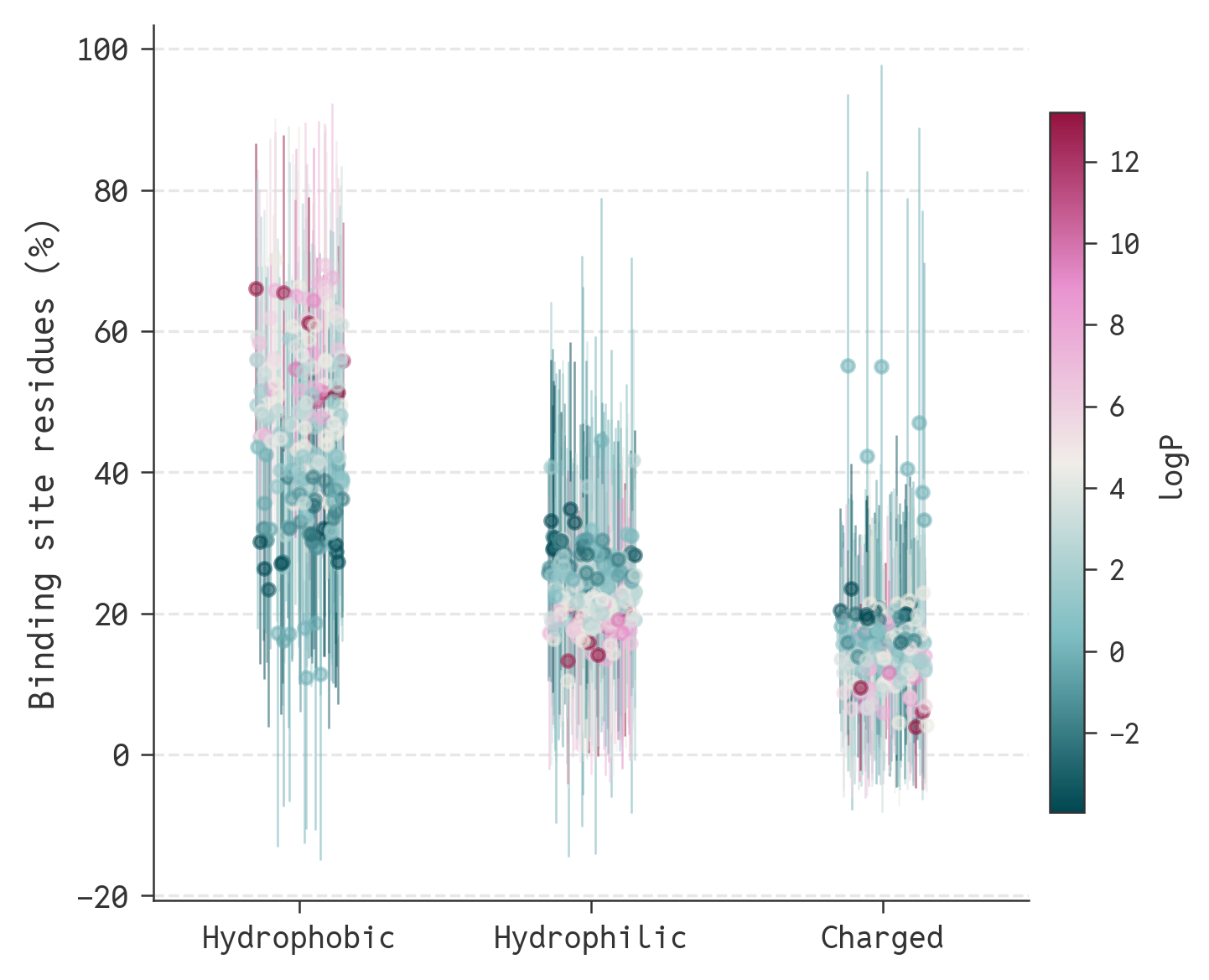}
    \end{subfigure}
    % \hfill
    % \begin{subfigure}[b]{0.7\textwidth}
    %     \centering
    %     \includegraphics[width=\textwidth]{figures/logp_vs_hydrophobicity_w_designability.png}
    % \end{subfigure}
    % \hfill
    % \begin{subfigure}[b]{0.7\textwidth}
    %     \centering
    %     \includegraphics[width=\textwidth]{figures/residue_composition_ternary_logp_sidechain.png}
    % \end{subfigure}
    \caption{The strip plot shows the average binding site residue composition, split by ligand type, without co-designability filtering. Note that the charged-residue outliers correspond to metals that require charged coordination.}% Bottom scatter plot shows ligand logP correlation with mean binding site hydrophobicity \emph{after} co-designability filtering, using Chai-1 predicted structures' side-chain ligand distance to determine the motif. } % , and the ternary plot with specific ligands labeled. % NOTE: I hid the ternary plot for now because that was determined using Chai-1 side-chain - ligand distance, since our comparison can only be to the backbone - ligand positions of BoltzGen/RFD3.
    \label{fig:binding_site_residue_composition}
\end{figure}

\begin{figure}[h]
    \centering
    \begin{subfigure}[b]{0.45\textwidth}
        \centering
        \includegraphics[width=\textwidth]{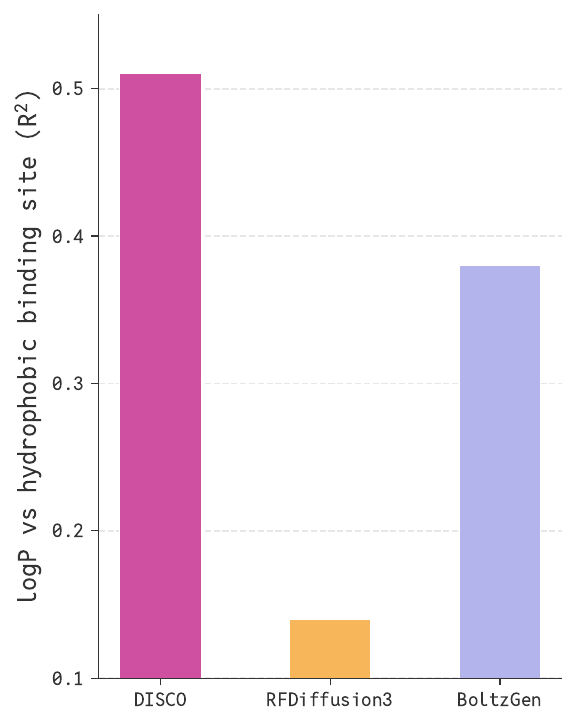}
    \end{subfigure}
    \hfill
    \begin{subfigure}[b]{0.45\textwidth}
        \centering
        \includegraphics[width=\textwidth]{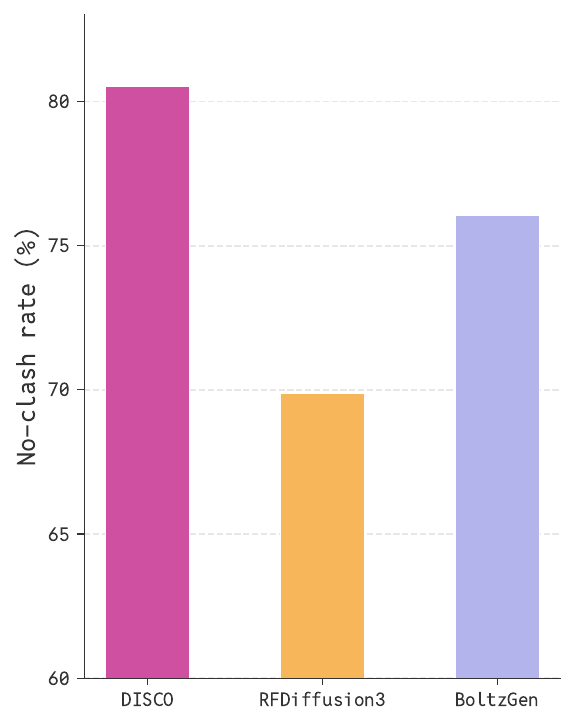}
    \end{subfigure}
    \caption{Comparison of chemical fidelity indicators of binding motif and ligand interaction, averaged over all ligand types that all three models share (i.e. no multi-ligands). Top shows the correlation of pocket hydrophobicity with ligand lipophilicity. Bottom shows whether the designed proteins have any backbone clashes with the ligand. Together with the co-designability cluster plot shown in the main text, these results establish that \nameshort yields realistic binding pockets that respond to ligand's chemical and physical properties.}
    \label{fig:binding_site_method_comparison}
\end{figure}

\begin{figure}[h]
    \centering
    \includegraphics[width=\textwidth]{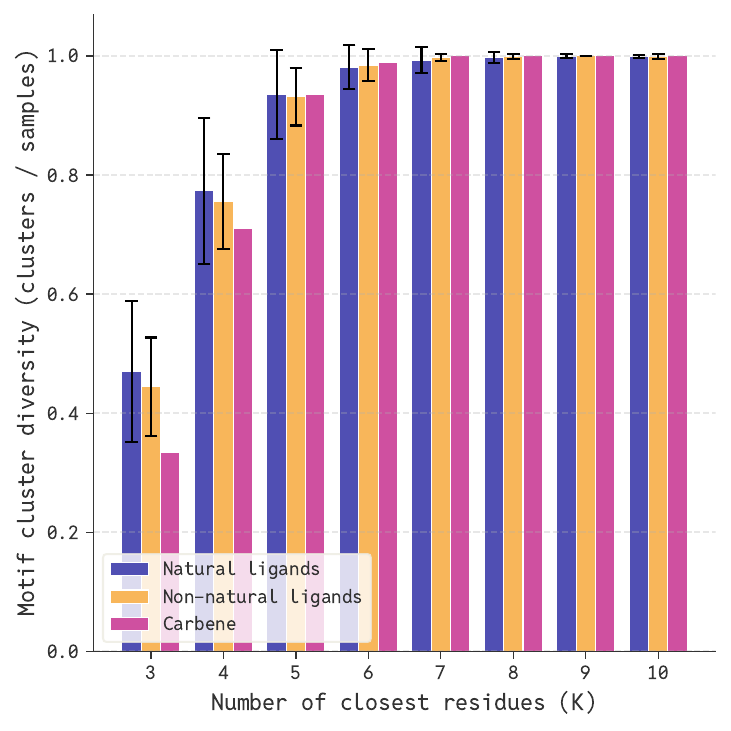}
    \caption{Distribution of motif cluster diversity (averaged per ligand type), across $K$ closest residues to the ligand.}
    \label{fig:binding_site_diversity_cluster}
\end{figure}

\begin{figure}[h]
    \centering
    \begin{subfigure}[b]{0.45\textwidth}
        \centering
        \includegraphics[width=\textwidth]{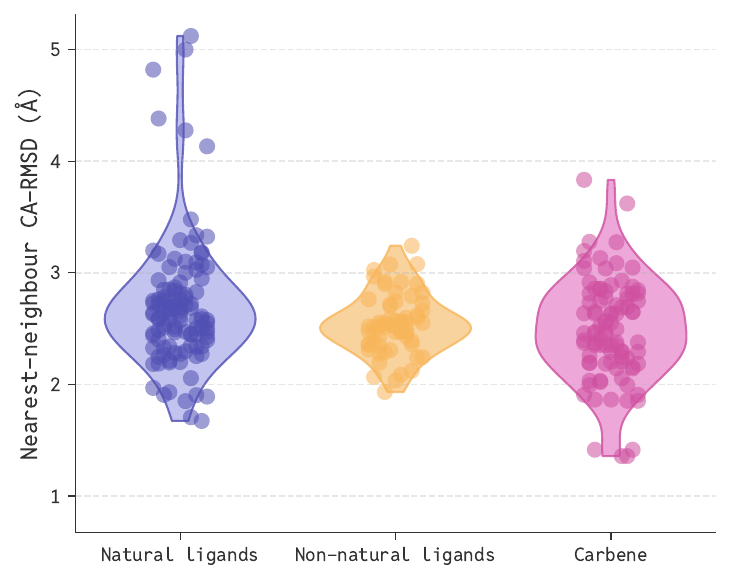}
        \label{fig:k5_ca_rmsd}
    \end{subfigure}
    \hfill
    \begin{subfigure}[b]{0.45\textwidth}
        \centering
        \includegraphics[width=\textwidth]{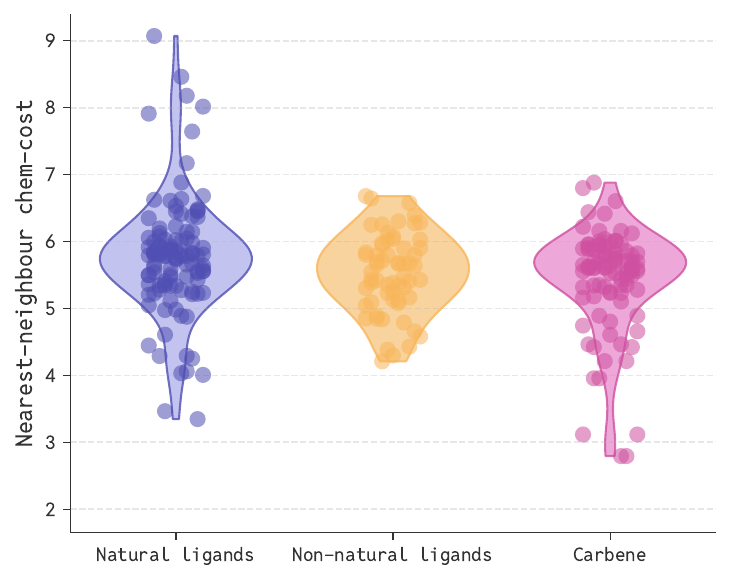}
        \label{fig:k5_chem_cost}
    \end{subfigure}    
    \hfill
    \begin{subfigure}[b]{0.45\textwidth}
        \centering
        \includegraphics[width=\textwidth]{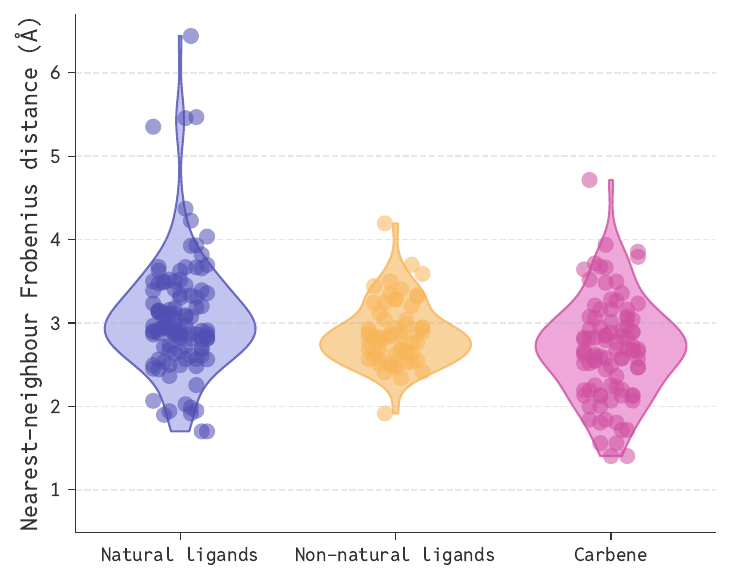}
        \label{fig:k5_frobenius}
    \end{subfigure}
    \hfill
    \begin{subfigure}[b]{0.45\textwidth}
        \centering
        \includegraphics[width=\textwidth]{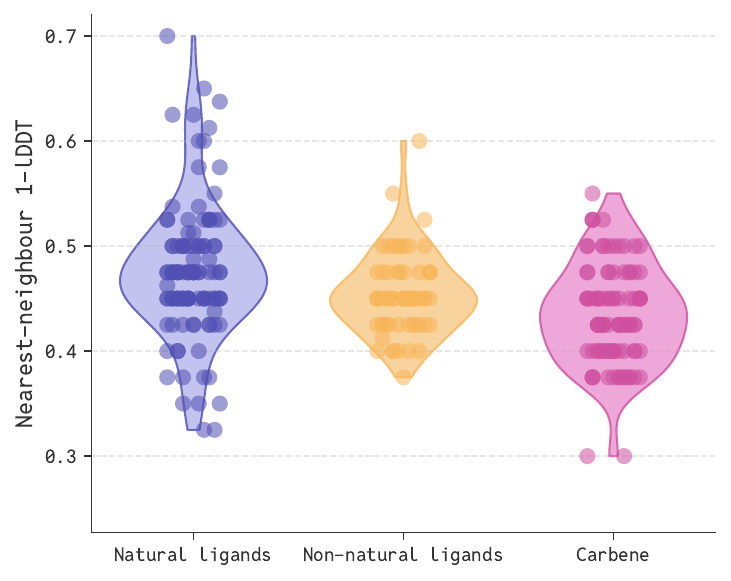}
        \label{fig:k5_lddt}
    \end{subfigure}
    \caption{Violin plot showing the full distribution of per-design nearest-neighbour diversity of the 5 closest residues to the ligand, each overlaid strip point shows a per-ligand median.}
    \label{fig:binding_site_diversity_k5}
\end{figure}

\begin{figure}[h]
    \centering
    \begin{subfigure}[b]{0.45\textwidth}
        \centering
        \includegraphics[width=\textwidth]{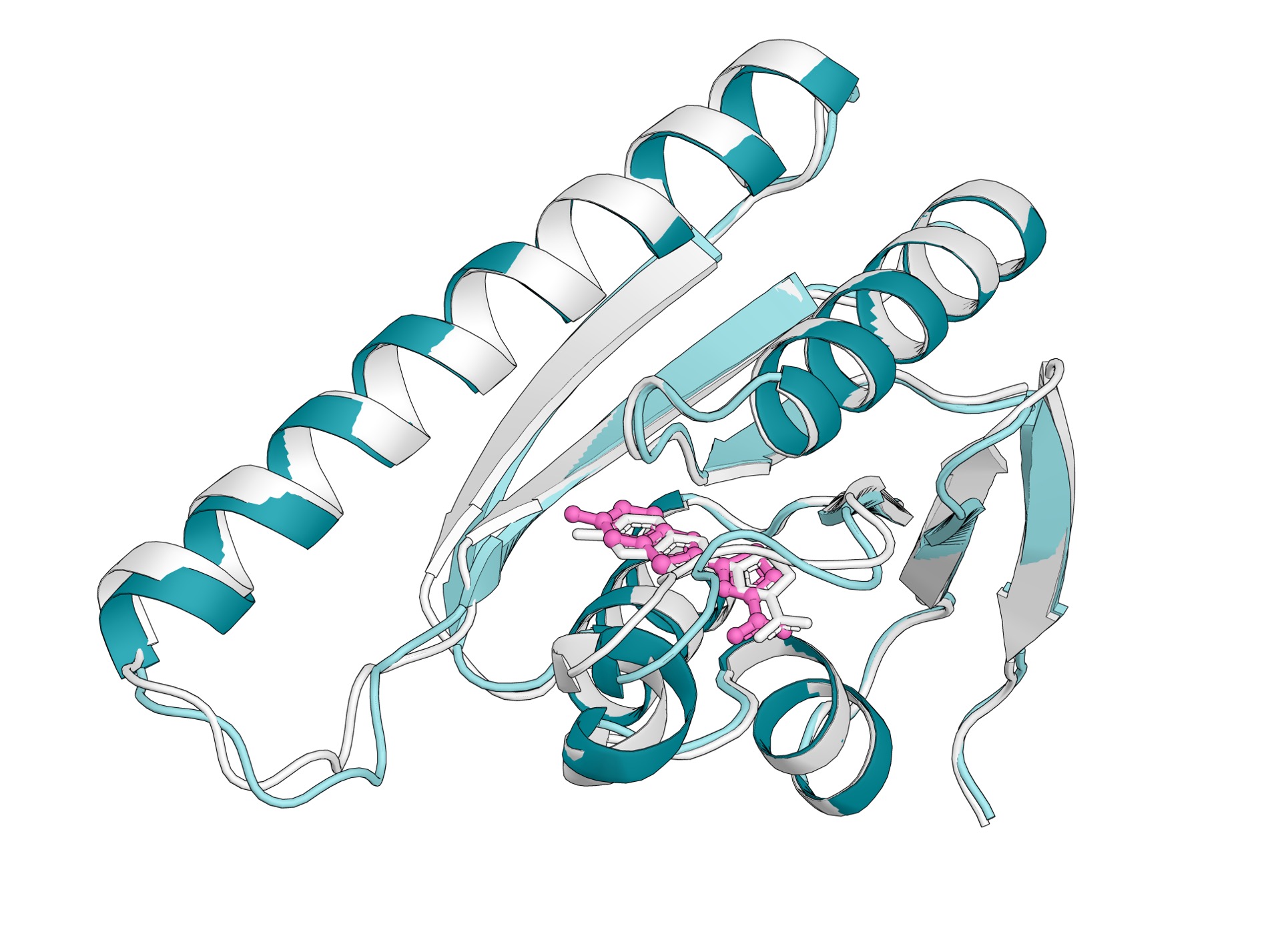}
    \end{subfigure}
    \hfill
    \begin{subfigure}[b]{0.45\textwidth}
        \centering
        \includegraphics[width=\textwidth]{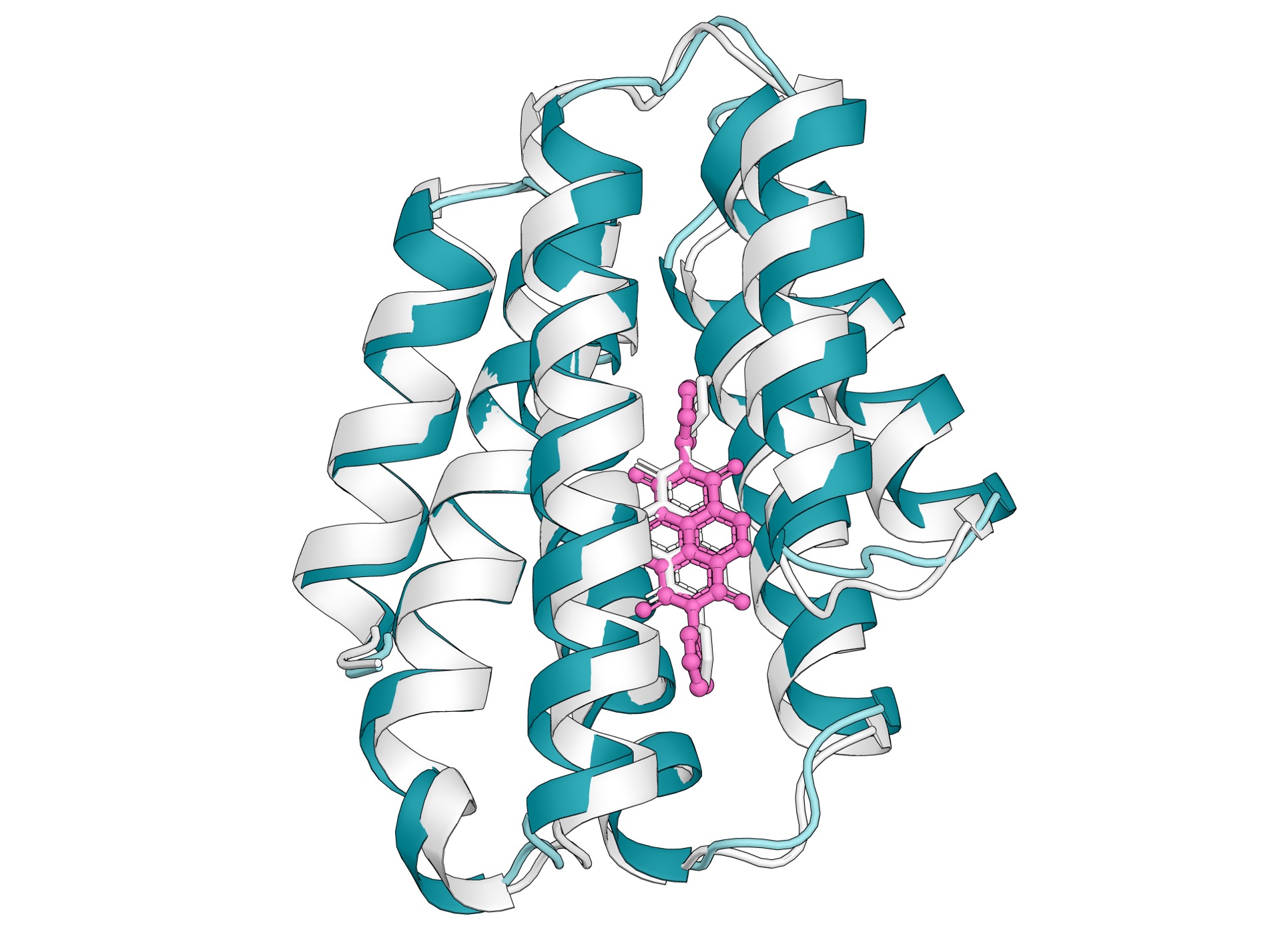}
    \end{subfigure}    
    \caption{Additional conditional design examples. Left shows conditioning on luciferin, a natural luminescent molecule, in a hybrid helix and sheet topology. Right shows a substituted naphthalene diimide, a non-natural electron acceptor, in a helix bundle.}
    \label{fig:cond_prot_examples}
\end{figure}

\begin{figure}[h]
    \centering
    \begin{subfigure}[b]{0.45\textwidth}
        \centering
        \includegraphics[width=\textwidth]{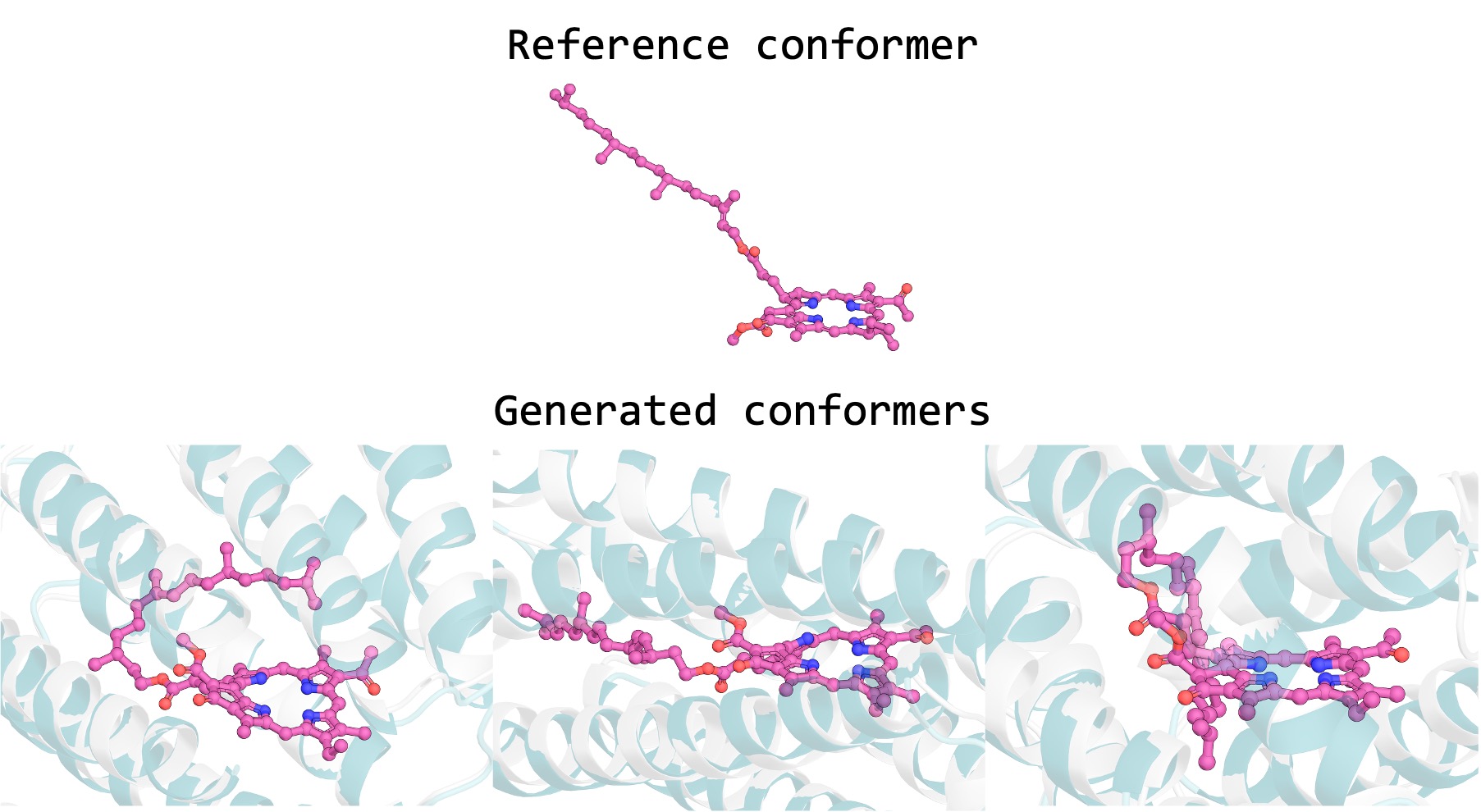}
    \end{subfigure}
    \hfill
    \begin{subfigure}[b]{0.45\textwidth}
        \centering
        \includegraphics[width=\textwidth]{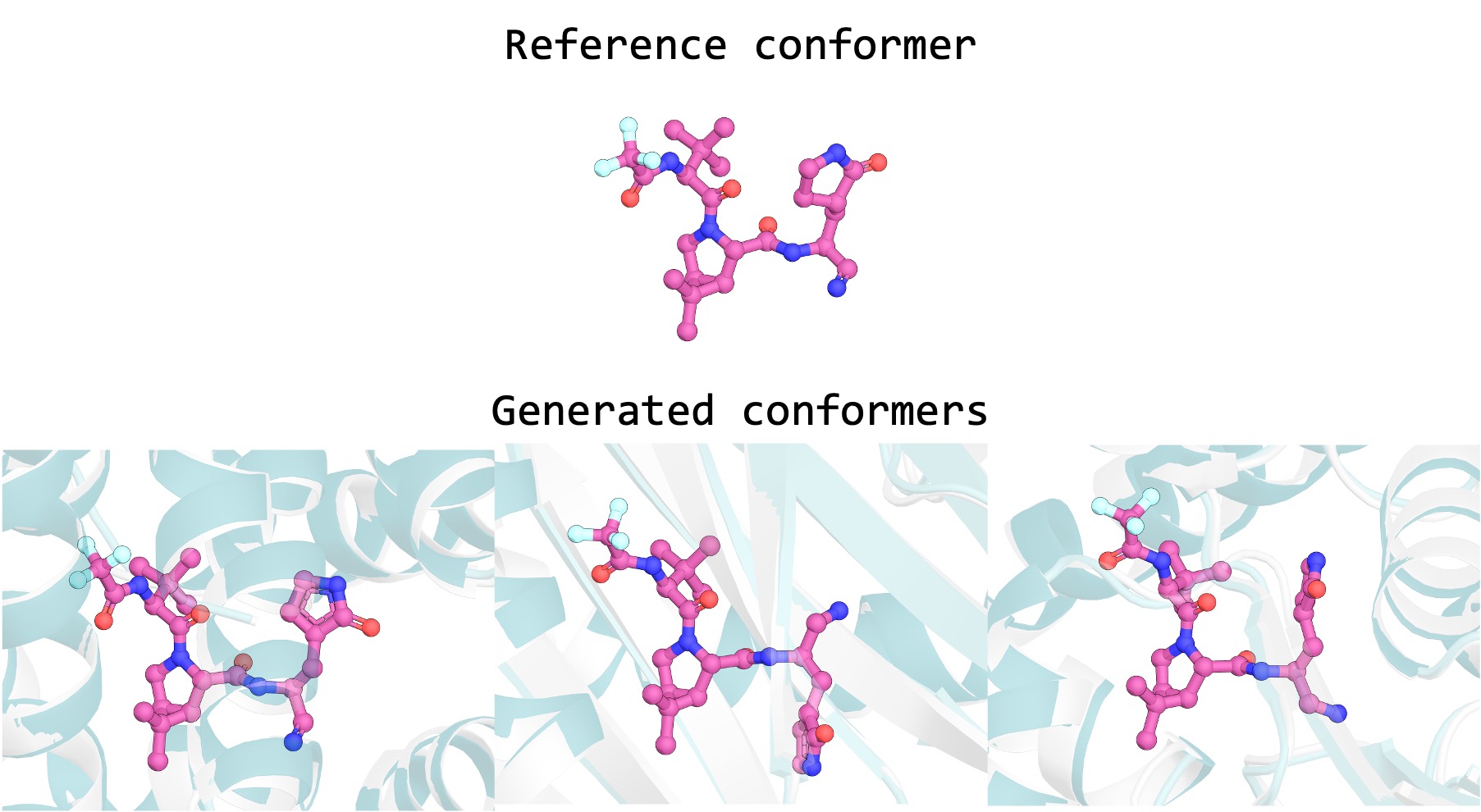}
    \end{subfigure}    
    \caption{Additional generated conformer examples, compared to the reference conformer. Left shows bacteriopheophytin A, and right shows Nirmatrelvir.}
    \label{fig:cond_conf_examples}
\end{figure}

\begin{figure}[h]
    \centering
    \begin{subfigure}[b]{0.45\textwidth}
        \centering
        \includegraphics[width=\textwidth]{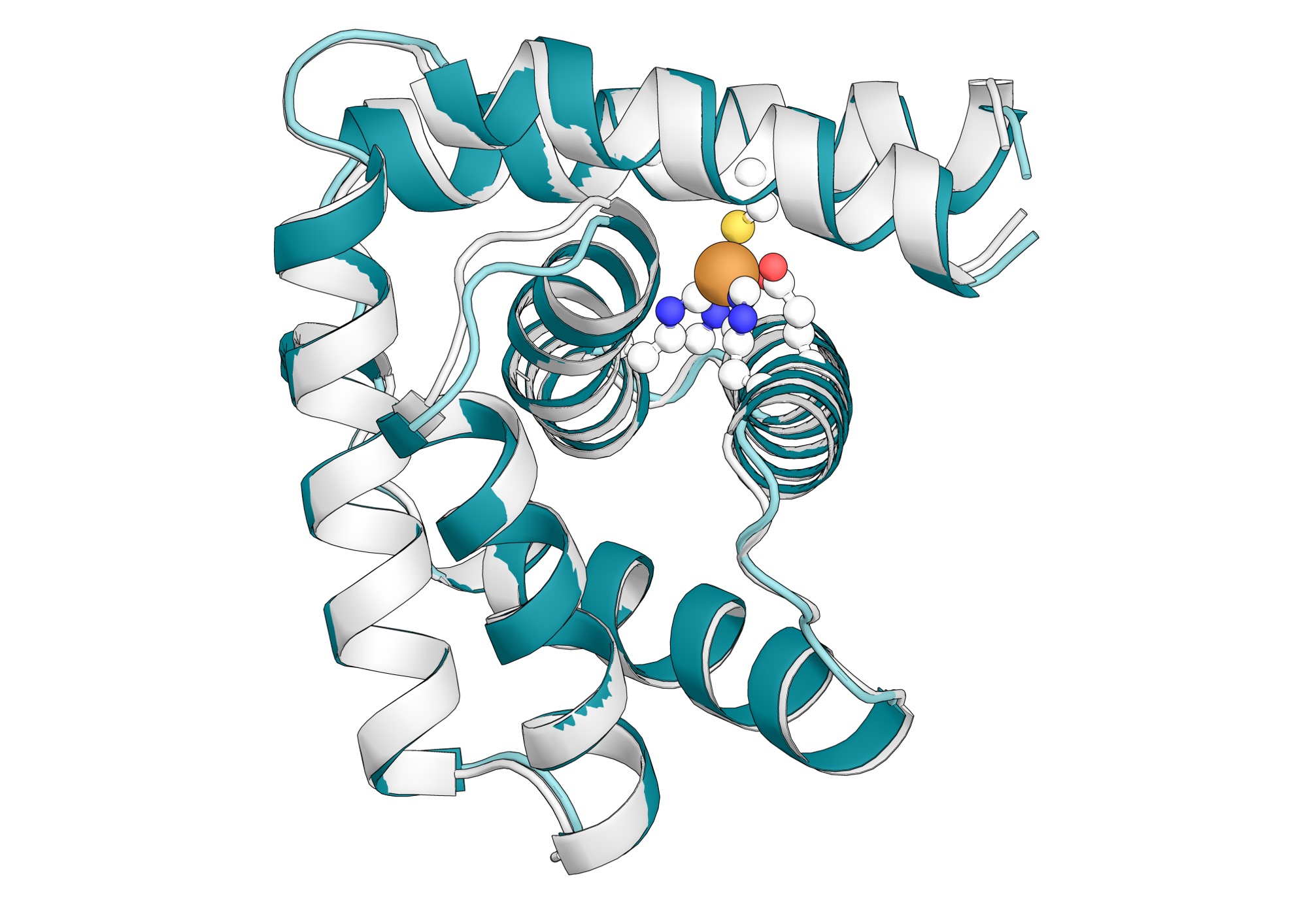}
    \end{subfigure}
    \hfill
    \begin{subfigure}[b]{0.45\textwidth}
        \centering
        \includegraphics[width=\textwidth]{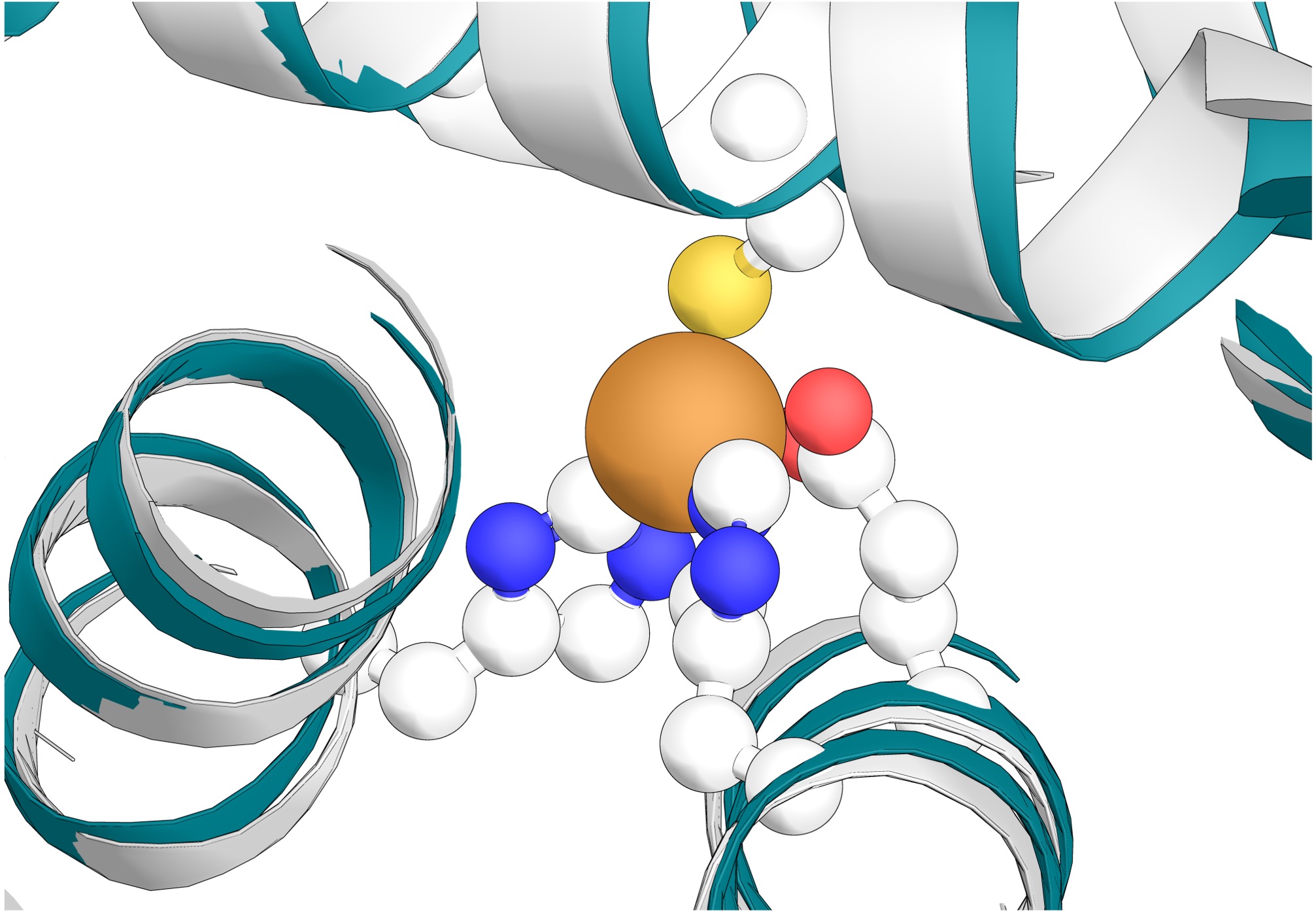}
    \end{subfigure}    
    \caption{An additional generated motif: the bonding geometries of a class I type II Cu center (2 His, 2 Cys, 1 Glu residues in tetrahedral coordination with Cu[2+]). This suggests the model can build informed residues to satisfy chemical constraints.}
    \label{fig:cond_chem_motif_examples}
\end{figure}

\begin{figure}[h]
    \centering
    \begin{subfigure}[b]{0.75\textwidth}
        \centering
        \includegraphics[width=\textwidth]{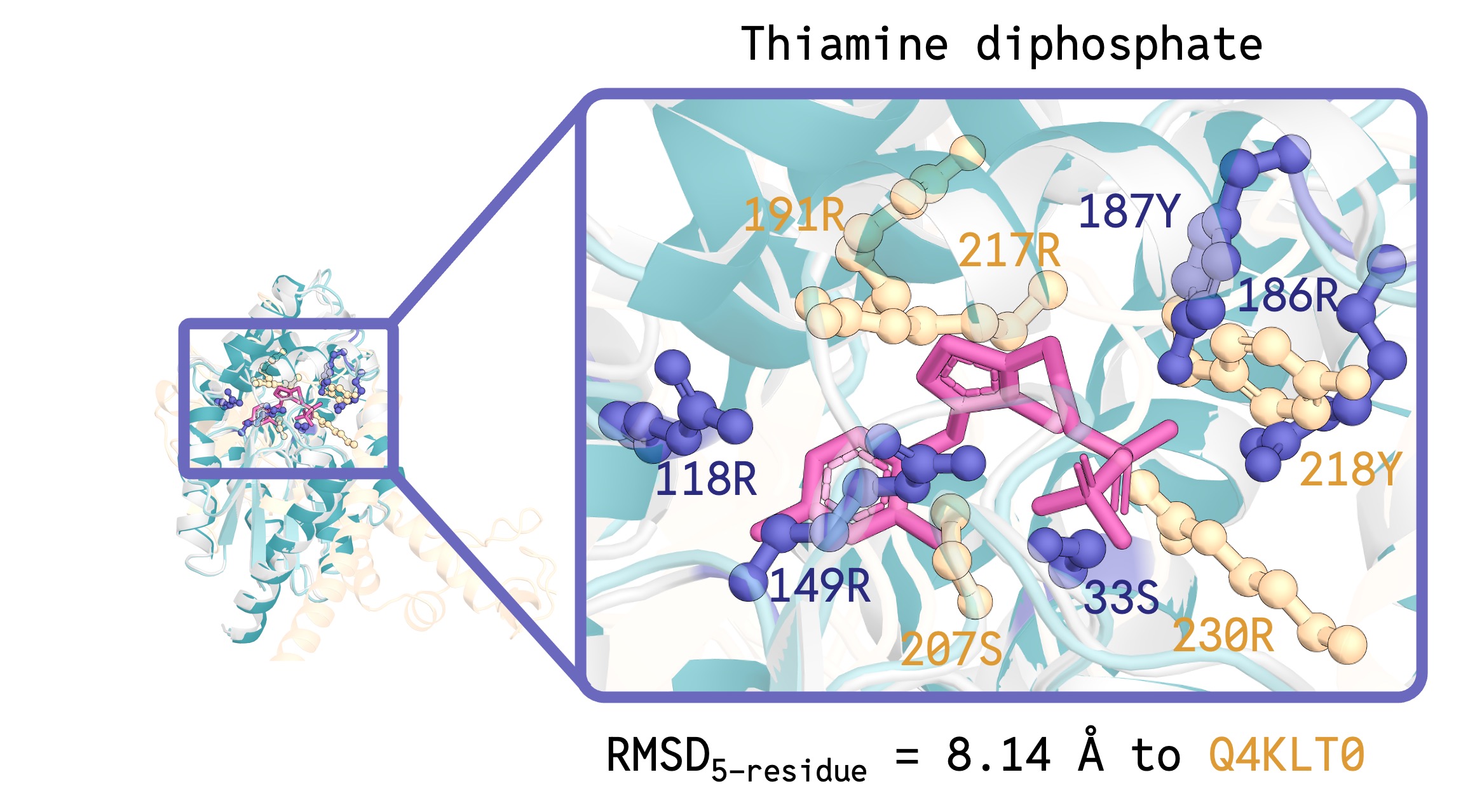}
    \end{subfigure}  
    \hfill
    \begin{subfigure}[b]{0.70\textwidth}
        \centering
        \includegraphics[width=\textwidth]{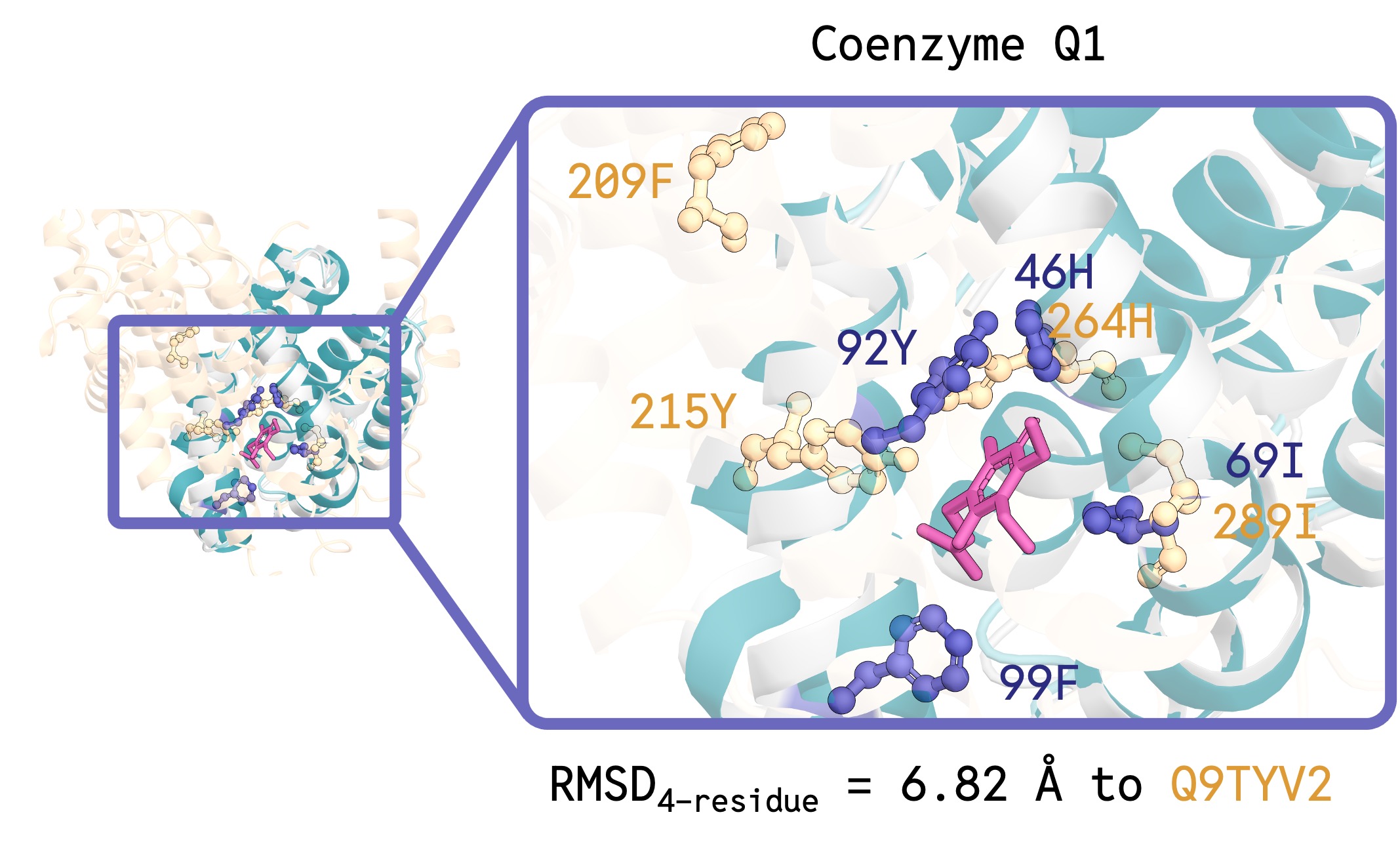}
    \end{subfigure}
    \caption{Additional generated motifs: two novel motifs (purple) aligned to the closest known motif (beige) in AlphaFoldDB. This indicates the model can build new residue motifs. Note that many generated motifs yield no match via Folddisco.}
    \label{fig:cond_novel_motif_examples}
\end{figure}
% Figure~\ref{fig:p2_p3_ligands} reports the percentage of co-designable clusters 
% produced by \nameshort , RFAA, RFDiffusion3, and BoltzGen across an extended benchmark 
% spanning both natural (top) and non-natural (bottom) ligands. \nameshort 
% achieves the highest proportion of diverse, co-designable sequence-structure pairs across all ligands in both categories, often by a substantial margin over competing methods. RFAA and BoltzGen show moderate but variable performance depending on the ligand, while RFDiffusion3 lags considerably behind across the board. Notably, this performance advantage holds not only for well-studied natural ligands such as \texttt{testosterone}, \texttt{coenzyme Q1}, and \texttt{riboflavin}, but also generalizes robustly to challenging non-natural substrates including \texttt{Ptm Radical}, \texttt{Mirex}, and \texttt{NHC Imidazolium} derivatives, demonstrating the broad applicability of \nameshort to ligand-binding protein design beyond the natural chemical repertoire.
\clearpage

\section{\textit{In vitro} methods and results}

\subsection{Materials and instrumentation}
 
All chemicals were obtained from commercial suppliers and used without further purification. Racemic product standards \textbf{1b}, \textbf{2b}, \textbf{4b} were prepared according to literature-known methods.\cite{kennemur2025enzymatic, allen2017n, doyle1981efficient}\medskip

\noindent M9-N buffer was used for enzymatic reactions and was prepared from a 5X stock solution of \ce{Na2HPO4}$\cdot$7\ce{H2O} (64~g), \ce{KH2PO4} (15~g), and \ce{NaCl} (2.5~g) dissolved in 1~L of double-distilled water (ddH\textsubscript{2}O).
The working solution (1X) was prepared by dilution with ddH\textsubscript{2}O, followed by addition of \ce{CaCl2} (1~mL, 0.1~M) and \ce{MgSO4} (2~mL, 1.0~M).
The pH was adjusted to 8.0.\medskip
 
\noindent The TAE buffer used in DNA gel electrophoresis was prepared from a 50X stock solution made by dissolving Tris free base (242~g), disodium EDTA (18.61~g), and glacial acetic acid (57.2~mL) in 1~L of ddH\textsubscript{2}O.
The 1X TAE buffer was then prepared by diluting 40~mL of the 50X stock solution with 1960~mL of ddH\textsubscript{2}O.\medskip

\noindent PCR amplifications were performed using the Phusion\textsuperscript{\textregistered} High-Fidelity PCR Kit (New England Biolabs) on an Eppendorf Mastercycler X50s.\medskip
 
\noindent Gibson Assembly reactions were carried out using an in-house solution containing isothermal master mix, T5 exonuclease, Phusion\textsuperscript{\textregistered} DNA polymerase, and Taq DNA ligase (New England Biolabs).\cite{gibson2009enzymatic}\medskip

\noindent Gene fragments (without adapters) were ordered from Twist Bioscience, where the vendor was chosen for speed, cost and quality reasons.\medskip % original words from Twist 

\noindent T7 Express competent Escherichia coli (BL21) (New England Biolabs) were used for all transformations.

\noindent All NMR spectra were obtained at the Caltech Liquid NMR Facility. NMR spectra were collected on a Bruker Prodigy 400 MHz instrument equipped with a cryoprobe operating at 400 MHz and 101 MHz for \textsuperscript{1}H and \textsuperscript{13}C, respectively. \textsuperscript{1}H spectra are referred to residual CDCl\textsubscript{3} solvent signals referenced at $\delta$ 7.26 ppm. Data for \textsuperscript{1}H NMR are reported as follows: chemical shift ($\delta$ ppm), integration, multiplicity (s = singlet, d = doublet, t = triplet, q = quartet, p = pentad, sext = sextet, hept = heptet, m = multiplet, br s = broad singlet), and coupling constant (Hz).\medskip

\noindent High-performance liquid-chromatography mass spectroscopy (HPLC--MS) for standard reaction screening analysis was performed using an Agilent 1200 series instrument equipped with an Agilent Poroshell C18 column (Agilent Poroshell 120 EC-C18, 4.6 x 50 mm, 2.7~\textmu m) with a Poroshell 120 guard column (4.6 x 5 mm, 2.7~\textmu m). Water and acetonitrile (MeCN) with 0.1\% (v/v) HPLC grade glacial acetic acid were used as eluents.\medskip 

\noindent Normal phase chiral HPLC--UV data was collected with an Agilent 1260 Infinity HPLC equipped with a Daivel Chiralpak IC column (4.6 x 250 mm, 5 µm) and a Daivel Chiralpak IB column (4.6 x 250 mm, 5 µm) run in series (for products \textbf{2b} and \textbf{3b}) or a Daicel Chiralcel OD-H (product \textbf{1b}). Hexanes and isopropanol were used as eluents. \medskip

\noindent Gas chromatography (GC) analyses were conducted on an Agilent Technologies 7820A GC system equipped with flame ionization detectors (FID). An Agilent J\&W HP-5 column was used for achiral analyses, while an Agilent J\&W CycloSil-B column was employed for chiral separations.
GC--MS analyses were carried out on an Agilent 7820A GC coupled to an Agilent 5977B MSD using an Agilent J\&W HP-5MS~UI column.\medskip

\noindent Chromatographic purification was accomplished by flash chromatography using a Biotage Isolera One instrument with Sfär High-Capacity Duo columns using AMD Silica Gel 60, 230–400 mesh. Ethyl acetate in hexanes was used as eluents. \medskip

\noindent Centrifugation was carried out in an Allegra 25R tabletop centrifuge equipped with a TS-5.1-500 swinging bucket rotor or a S5700 microplate swing bucket rotor.

% \subsection{Molecular Biology Reagents}
 
% PCR amplifications were performed using the Phusion\textsuperscript{\textregistered} High-Fidelity PCR Kit (New England Biolabs) on an Eppendorf Mastercycler X50s.
 
% Gibson Assembly reactions were carried out using an in-house solution containing isothermal master mix, T5 exonuclease, Phusion\textsuperscript{\textregistered} DNA polymerase, and Taq DNA ligase (New England Biolabs).
 
\subsection{Cloning and transformation in 96-well plates}
 
The pET22b(+) backbone (Novagen) was PCR-linearized, digested with \textit{Dpn}I, gel-purified and confirmed to produce minimal background transformation. All genes were inserted between the \textit{Nde}I and \textit{Xho}I restriction sites. Ordered gene fragments were delivered in lyophilized form (Twist Bioscience) and resuspended in PCR-grade water to a concentration of 50~ng/\textmu L. We note that for design sequences that did not begin with a methionine, an ATG codon was added as a start codon. A His-tag with the amino acid sequence LEHHHHHH was appended to the end of each sequence.
Assembly reactions were performed in 96-well format using Gibson Assembly. 
Each reaction contained 1~\textmu L insert DNA, 1~\textmu L vector (100~ng/\textmu L), and 5~\textmu L Gibson Assembly mix, followed by incubation at 50~\textdegree C for 1~h.
Transformation was performed by incubating 5~\textmu L BL21(DE3) competent cells with the assembly mixture on ice for 30~min, followed by heat shock at 42~\textdegree C for 10~s.
After recovery in LB medium (100~\textmu L), 10~\textmu L was used to inoculate 500~\textmu L LB medium supplemented with 100~\textmu g/mL carbenicillin (LB\textsubscript{carb}), and cultures were grown overnight at 37~\textdegree C.
Transformants were verified by in-house sequencing (LevSeq), showing ${\sim}95\%$ correct sequences.\cite{long2024levseq}
The remaining ${\sim}5\%$ likely resulted from sequencing, assembly, or transformation errors.
Variants were organized in 96-well plates, including: negative control: $\beta$-subunit of tryptophan synthase from \textit{Thermotoga maritima} (TmTrpB, wells A1, B2, C3, D4) and positive control: evolved protoglobin from \textit{Aeropyrum pernix} \textbf{5312}\cite{kennemur2025enzymatic} (wells E5 and F6).
Glycerol stocks were prepared for long-term storage.
 
\subsection{Protein expression and reaction screening in plates}
\label{sec:express}
 
Glycerol stocks were used to inoculate 500~\textmu L LB\textsubscript{carb} in 96-well plates, covered with a sterile, breathable film and grown at 37~\textdegree C with 220~rpm shaking.
From stationary-phase cultures, 50~\textmu L was transferred to 900~\textmu L TB\textsubscript{carb} and incubated for 2~h before induction with 50~\textmu L of isopropyl $\beta$-D-1-thiogalactopyranoside (IPTG)  (0.5~mM final) and 5-aminolevulinic acid (ALA) (1~mM final) in TB\textsubscript{carb}.
Cells were harvested by centrifugation (4,000$\times$\textit{g}, 10~min) and resuspended in 380~\textmu L M9-N buffer (pH~8.0) supplemented with 10~g/L glucose by orbital shaking at 900~rpm.
The plates were then transferred into a Coy anaerobic chamber ($\sim$ 0 -- 70 ppm O$_2$) to set up the reactions anaerobically.
Reactions were initiated by adding 20~\textmu L of a substrate solution in MeCN ({[substrate]}\textsubscript{final} = 10~mM, {[EDA]}\textsubscript{final} = 15~mM, final MeCN concentration: 5\%).
Plates were sealed with adhesive aluminum foil and incubated overnight at room temperature with shaking (700~rpm).\medskip

\noindent \textbf{Workup procedures (reaction dependent):}\smallskip
 
\noindent \textbf{For 1b, 3b and 4b:} The plates were removed from the Coy chamber, and HPLC-grade MeCN (1200~\textmu L) was added to each well.
The resulting suspensions were mixed by shaking on an orbital shaker at 500~rpm for 20~min at room temperature, and the plates were then centrifuged (5,000$\times$\textit{g}, 15~min, 4~\textdegree C) to remove protein and cell debris from the supernatant.
The supernatant (200~\textmu L/well) was transferred to a polypropylene 96-well microtiter plate (Agilent) and sealed with an Easy Pierce Heat Sealing Foil (Thermo Scientific\texttrademark) and subjected to reversed-phase HPLC--MS analysis using methods details in \autoref{tab:LCMS-1b}, \autoref{tab:LCMS-3b}, and \autoref{tab:LCMS-4b}. The product MS peak was integrated using the built-in analysis.\medskip

\begin{table}[h!]
    \centering
    \caption{LC-MS methods for analysis of compound \textbf{1b}. Solvent A: \ce{H2O} + 0.1\% AcOH, solvent B: MeCN + 0.1\% AcOH. Flow rate: 1.5~mL/min.}
    \label{tab:LCMS-1b}
    \begin{tabular}{c|c|c}
        Time (min) & A (\%) & B (\%) \\
        \hline
        0.0 & 50.0 & 50.0\\
        2.5 & 50.0 & 50.0
    \end{tabular}
\end{table}

\begin{table}[h!]
    \centering
    \caption{LC-MS methods for analysis of compound \textbf{3b}. Solvent A: \ce{H2O} + 0.1\% AcOH, solvent B: MeCN + 0.1\% AcOH. Flow rate: 1.5~mL/min.}
    \label{tab:LCMS-3b}
    \begin{tabular}{c|c|c}
        Time (min) & A (\%) & B (\%) \\
        \hline
        0.0 & 85.0 & 15.0\\
        0.4 & 85.0 & 15.0\\
        0.5 & 10.0 & 90.0\\
        2.1 & 5.0 & 95.0 \\
        2.5 & 5.0 & 95.0\\
        2.6 & 85.0 & 15.0 \\
        3.0 & 85.0 & 15.0 \\
    \end{tabular}
\end{table}

\begin{table}[h!]
    \centering
    \caption{LC-MS methods for analysis of compound \textbf{4b}. Solvent A: \ce{H2O} + 0.1\% AcOH, solvent B: MeCN + 0.1\% AcOH. Flow rate: 1~mL/min.}
    \label{tab:LCMS-4b}
    \begin{tabular}{c|c|c}
        Time (min) & A (\%) & B (\%) \\
        \hline
        0.0 & 50.0 & 50.0\\
        3.0 & 50.0 & 50.0
    \end{tabular}
\end{table}
 
\noindent \textbf{For 2b:} The plates were removed from the Coy chamber, and 400~\textmu L of a 1~mM solution of 1,2,3-trimethoxybenzene dissolved in 1:1 (v/v) EtOAc:cyclohexane were added to each well.
Plates were sealed using a 96-well sealing mat and then mixed by shaking on an orbital shaker (900~rpm) for 20~min.
Plates were then centrifuged (5,000$\times$\textit{g}, 10~min, room temperature), and the organic layer (200~\textmu L/well) was transferred to a 400~\textmu L glass insert within a 2.0~mL screw-cap vial for GC--MS analysis.
Analyses were performed on an Agilent J\&W HP-5MS UI column using the following oven program: 120~\textdegree C (1 min), ramped at 35~\textdegree C/min to 250~\textdegree C, and then ramped at 75~\textdegree C/min to 300~\textdegree C (hold 0.2 min) for a total run time of approximately 11.2 min. 
The internal standard and product peaks were integrated using built-in MassHunter Quantitative Analysis software (Agilent).

\subsection{Protein expression and whole-cell validation reactions}
 
Cultures expressing enzyme variants with promising activity were streaked from the corresponding glycerol stocks onto LB\textsubscript{carb} agar plates.
A single colony was used to inoculate 5~mL of LB\textsubscript{carb} and grown overnight at 37~\textdegree C with shaking at 250~rpm.
Subsequently, 0.5~mL of the starter culture were used to inoculate 50~mL of TB\textsubscript{carb} in a 125~mL sterile Erlenmeyer flask.
In parallel, plasmid DNA was isolated from the starter culture using a Monarch\textregistered  Plasmid Miniprep Kit (New England Biolabs), and the sequence of each variant was confirmed by Sanger sequencing (Quintara Biosciences, Culver City, CA).
Cultures were grown at 37~\textdegree C (220~rpm) to an OD\textsubscript{600} of 0.6--0.8, cooled on ice for 30~min, and induced with IPTG (0.5~mM final) and ALA (1.0~mM final).
Expression proceeded at 22~\textdegree C for 20--24~h.
Cells were harvested (4,000$\times$\textit{g}, 5~min), and pellets were resuspended in M9-N buffer (10~g/L glucose) and normalized to OD\textsubscript{600} = 30.
Aliquots (380~\textmu L) were transferred to microcentrifuge tubes and placed in a Coy anaerobic chamber.
Reactions were initiated by adding 20~\textmu L substrate solution ({[substrate]}\textsubscript{final} = 10~mM, {[EDA]}\textsubscript{final} = 15~mM, final MeCN concentration: 5\%).
Reactions were incubated overnight at room temperature (700~rpm), and each variant was tested in triplicate.

\subsubsection{Analysis by GC-FID (compounds 1b - 4b)}
Tubes were removed from the Coy chamber, and 400~\textmu L of a 1:1 (v/v) EtOAc:cyclohexane containing an internal standard (see section \nameref{sec:calib}) were added to each tube.
Tubes were vortexed and then centrifuged (20,000$\times$\textit{g}, 10~min, room temperature), and the organic layer (200~\textmu L) was transferred to a 400~\textmu L glass insert within a 2.0~mL screw-cap vial for GC--FID analysis. Analyses were performed on an Agilent J\&W HP-5 column using the following oven program: 100~\textdegree C (2 min), ramped at 50~\textdegree C/min to 300~\textdegree C (hold 2 min) for a total run time of 10 min. The internal standard and product peaks were integrated using built-in analysis software (Agilent) and quantified using calibration curves.

\subsubsection{Analysis by LC--MS (compound 4b for directed evolution)}
Tubes were removed from the Coy chamber, and 1.2 mL of MeCN was added to each tube. After mixing, the tubes were centrifuged (20,000xg, 15 minutes, 4 °C), 300~\textmu L of the supernatant and 50~\textmu L of a stock solution of papaverine hydrochloride (internal standard, 6.5 mg in 450 mL 2:1 MeCN:H$_2$O) were transferred into glass inserts within a 2.0~mL screw-cap vial for reversed-phase HPLC-MS analysis (method details are in table \ref{tab:LCMS-de}.). The internal standard (papaverine hydrochloride) and product peaks were integrated using built-in analysis software (Agilent) and quantified using calibration curves prepared and measured at the same time as sample analysis (see section \nameref{sec:calib2}).  

\begin{table}[h!]
    \centering
    \caption{LC-MS methods for analysis of compound \textbf{4b} (for directed evolution). Solvent A: \ce{H2O} + 0.1\% AcOH, solvent B: MeCN + 0.1\% AcOH. Flow rate: 1~mL/min.}
    \begin{tabular}{c|c|c}
        Time (min) & A (\%) & B (\%) \\
        \hline
        0.0 & 95.0 & 5.0\\
        1.0 & 35.0 & 65.0\\
        3.15 & 5.0 & 95.0\\
        3.35 & 5.0 & 98.0\\
        3.50 & 95.0 & 5.0\\
        4.00 & 95.0 & 5.0\\
    \end{tabular}
    \label{tab:LCMS-de}
\end{table}

\subsubsection{Analysis of enantioselectivity}
 \textbf{For 1b - 3b}\\
Tubes were then removed from the Coy chamber, and 400~\textmu L of a 1:1 (v/v) EtOAc:cyclohexane mixture were added to each tube.
Tubes were vortexed and then centrifuged (20,000$\times$\textit{g}, 10~min, room temperature), and the organic layer (200~\textmu L) was transferred to a 400~\textmu L glass insert within a 2.0~mL screw-cap vial for chiral HPLC-UV analysis. Analyses were performed using a Chiralpak IC column (4.6 x 250 mm, 5 µm) and a Chiralpak IB column (4.6 x 250 mm, 5 µm) run in series for products \textbf{2b} and \textbf{3b}. Conditions for \textbf{2b}: 30$\%$ isopropanol/hexanes, flow rate = 1.0 mL/min, runtime = 35 min, and UV-Vis detection at $\lambda$ = 235 nm. For \textbf{3b}: 6$\%$ isopropanol/hexanes, flow rate = 0.6 mL/min, runtime = 35 min, and UV-Vis detection at $\lambda$ = 235 nm. For \textbf{1b}, a Chiralcel OD-H column was used. Conditions: 6$\%$ isopropanol/hexanes, flow rate = 0.6 mL/min, runtime = 35 min, and UV-Vis detection at $\lambda$ = 235 nm.
The product peaks were integrated using built-in analysis software (Agilent).\medskip

\noindent \textbf{For 4b (for directed evolution})\\
Residual samples (1.3 mL) worked up for LC-MS analysis were concentrated to dryness under a steady stream of compressed air. A 1:1 (v:v) cyclohexane:EtOAc mixture (400~\textmu L) was then added to the tubes.  The tubes were vortexed and then centrifuged (20,000$\times$\textit{g}, 10~min, room temperature), and the organic layer (200~\textmu L) was transferred to a 400~\textmu L glass insert within a 2.0~mL screw-cap vial for  chiral GC-FID analysis. 
Analyses were performed on an Agilent CycloSil-B column using the following oven program: 50~\textdegree C (0 min), ramped at 10~\textdegree C/min to 170~\textdegree C (hold 19 min), then at 30~\textdegree C/min to 200~\textdegree C (hold 8 min), for a total run time of 40 min.

% \begin{figure}[h]
%     \centering
%     \includegraphics[width=\textwidth]{figures/bh_eda.png}
%     \caption{Reaction screening and validation results for B--H insertion with ethyl diazoacetate (\textbf{5a}) instead of ethyl diazopropanoate (\textbf{5b}). The enzymes are active but show different reactivity, indicating certain substrate selectivity. }
%     \label{fig:dr_styrene}
% \end{figure}\todo{remove or just put relative values}

\subsection{Hemochrome assay}
 
Protein concentration in the cell was determined using the hemochrome assay on the cell lysate.
Lysate was obtained by sonication using a Qsonica Q500 sonicator (9~min total, 1~s on, 2~s off, 35\% amplitude, on wet ice).
The cell debris was removed by centrifugation (5,000$\times$\textit{g}, 10~min, 4~\textdegree C).
To a cuvette, 500~\textmu L of the lysate and 500~\textmu L of Solution~I [0.2~M NaOH, 40\% (v/v) pyridine, 0.5~mM \ce{K3Fe(CN)6}] were added.
The UV--Vis spectrum (500--600~nm) of the oxidized state Fe(III) was recorded immediately.
Sodium dithionite (10~\textmu L of 0.5~M solution in 0.5~M NaOH) was added and mixed thoroughly by pipetting, and the UV--Vis spectrum of the reduced state Fe(II) was recorded immediately.
The protein concentration was calculated using the extinction coefficient and dilution factor:
\begin{equation}
  \varepsilon_{557_\text{reduced} - 540_\text{oxidized}} = 23.98 \text{~mM}^{-1}\text{cm}^{-1}
\end{equation}

\clearpage
\subsection{Selectivity results}

\begin{figure}[ht!]
    \centering
    \includegraphics[width=0.5\textwidth]{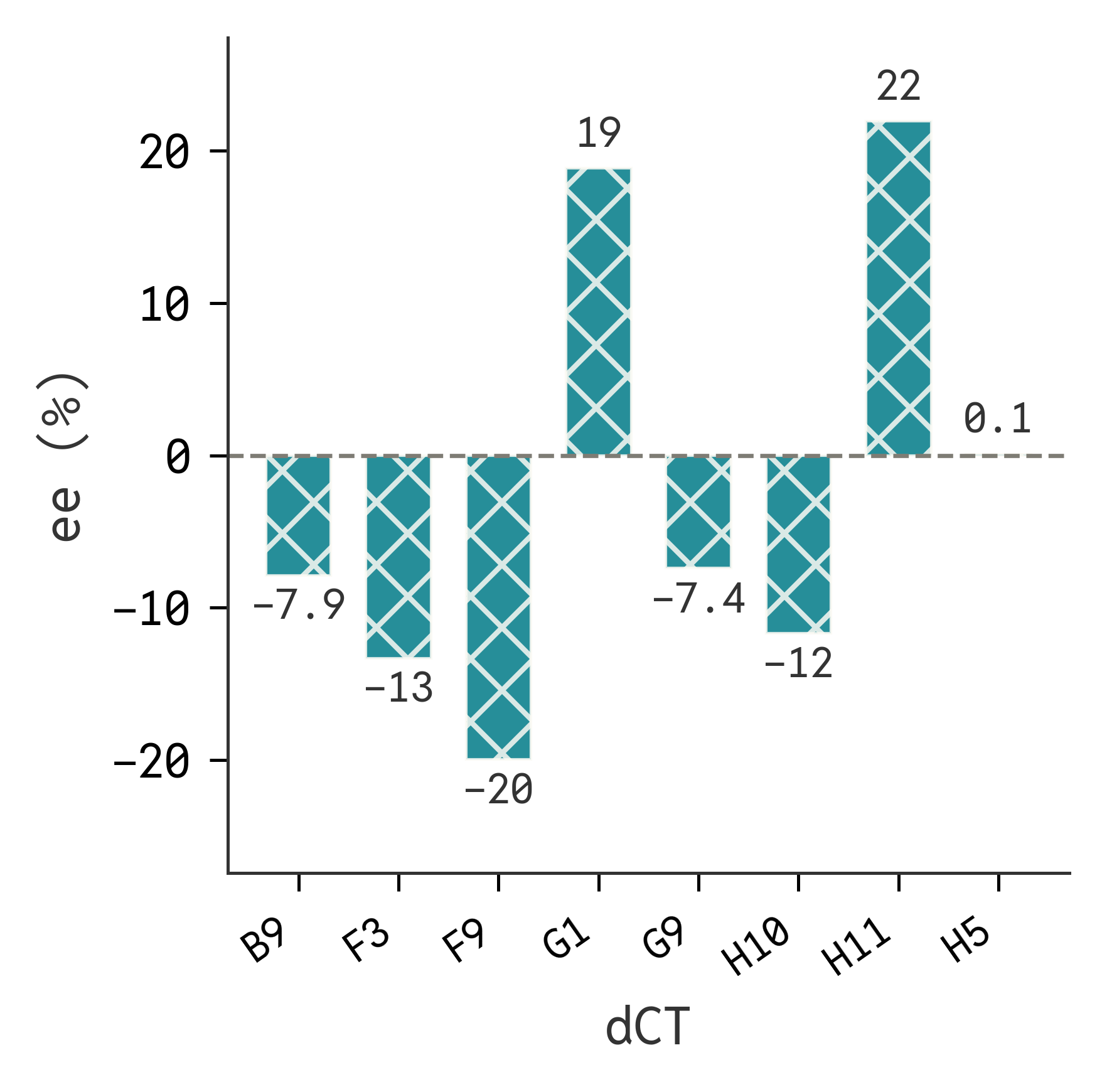}
    \caption{Enantiomeric excess (\%) of methoxystyrene cyclopropanation product (trans, \textbf{1b}). The corresponding enantiomeric ratio of dCT-H11 is 61:39.}
    \label{fig:ee_styrene}
\end{figure}

\begin{figure}[ht!]
    \centering
    \includegraphics[width=0.5\textwidth]{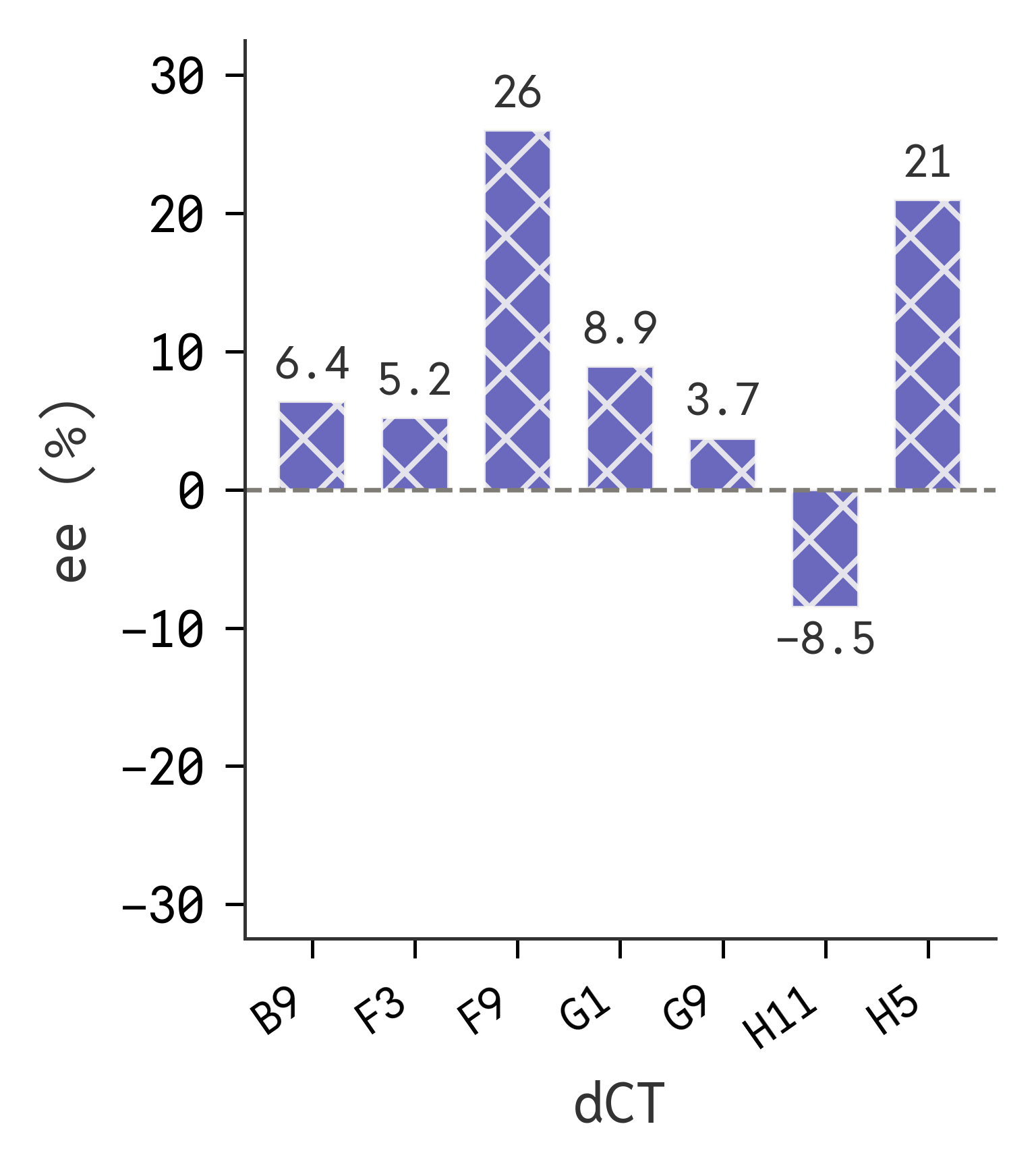}
    \caption{Enantiomeric excess (\%) of the product of B--H insertion \textbf{2b}. The corresponding enantiomeric ratio of dCT-F9 is 63:37.}
    \label{fig:ee_bh}
\end{figure}

\begin{figure}[ht!]
    \centering
    \includegraphics[width=0.5\textwidth]{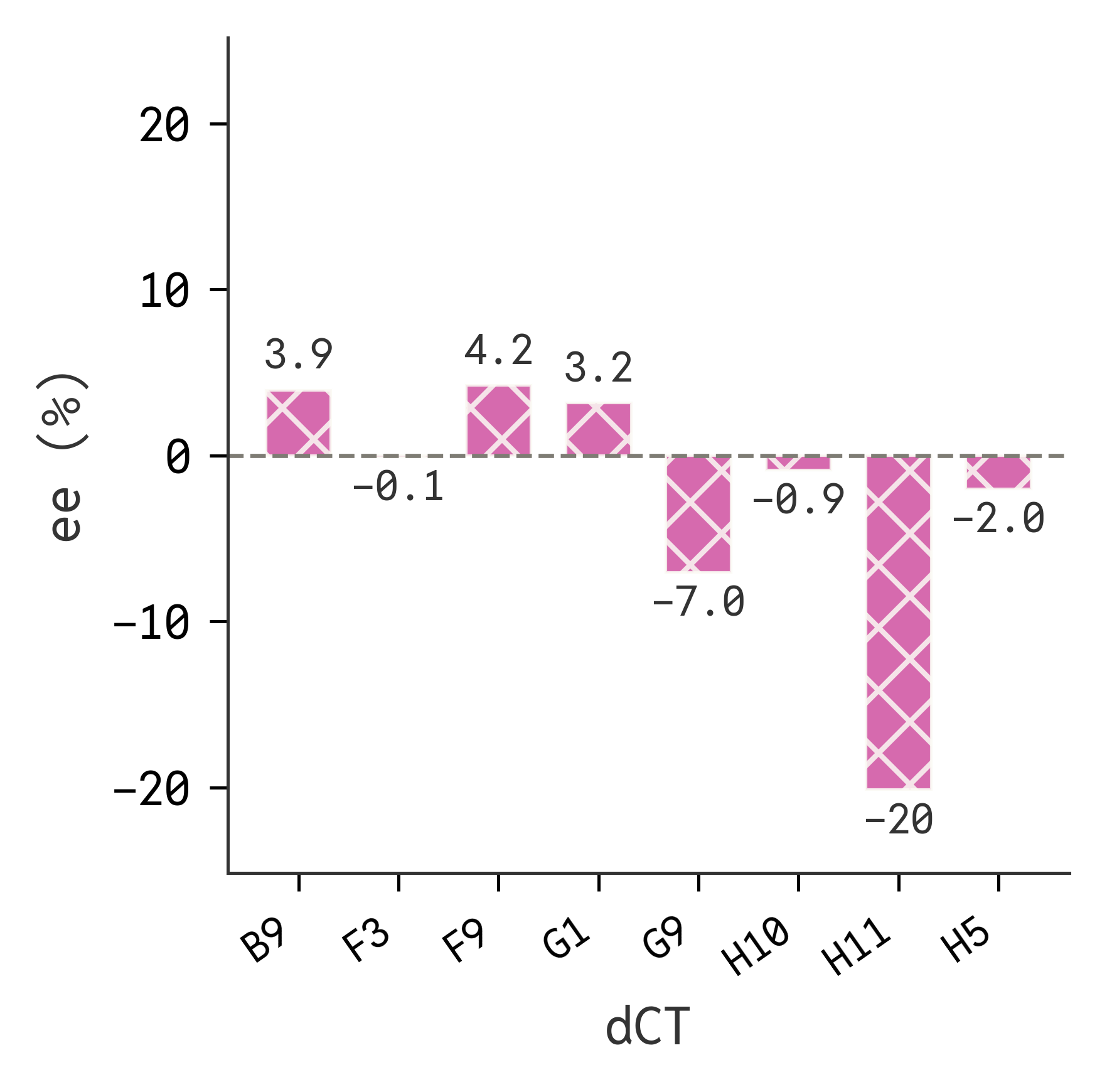}
    \caption{Enantiomeric excess (\%) of the product of C--H insertion \textbf{3b}. The corresponding enantiomeric ratio of dCT-H11 is 40:60.}
    \label{fig:ee_ch}
\end{figure}

\begin{figure}[ht!]
    \centering
    \includegraphics[width=0.4\textwidth]{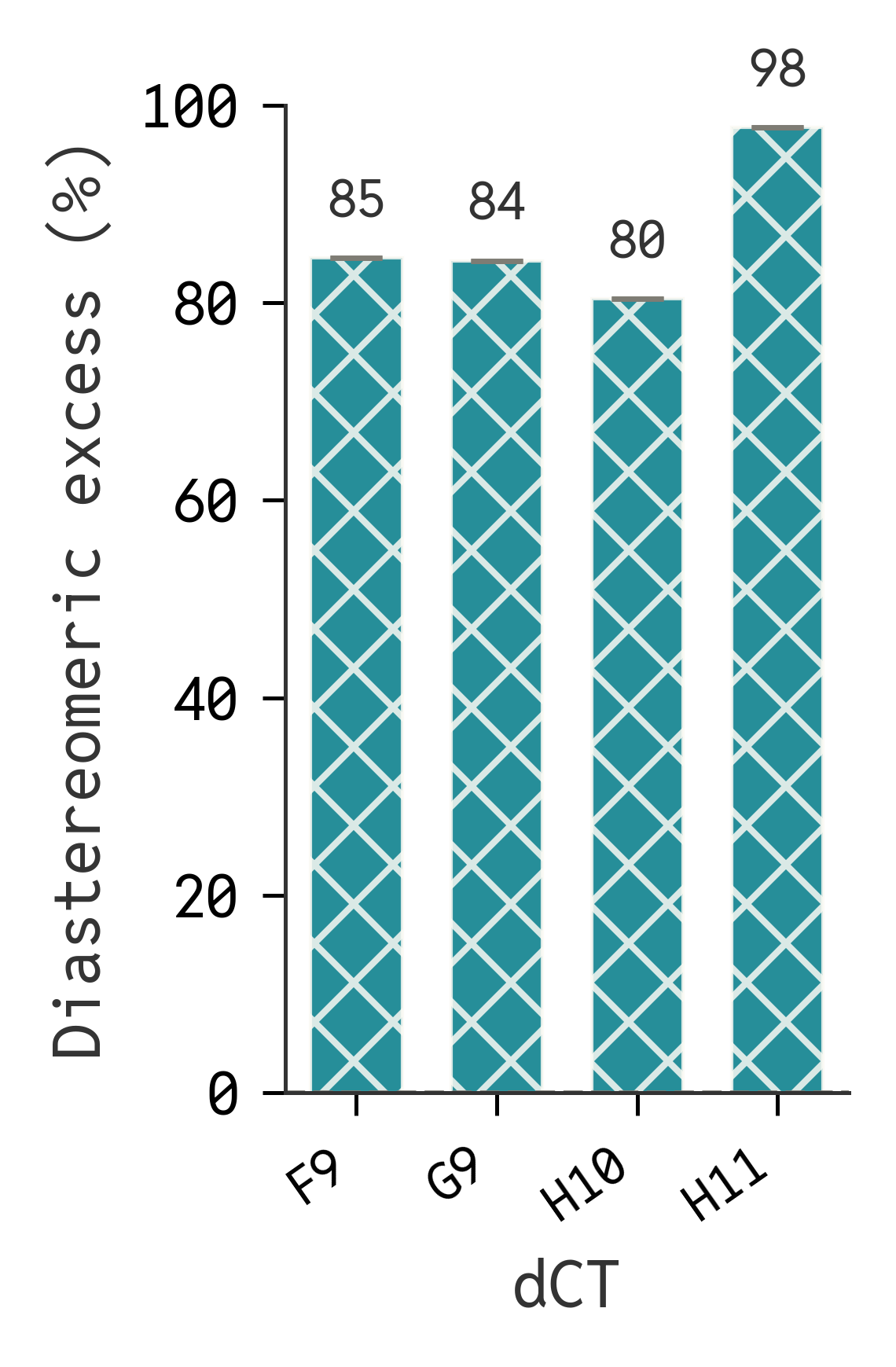}
    \caption{Diastereomeric excess (\%) of methoxystyrene cyclopropanation product (trans vs cis, \textbf{1b}) in selected validation results. The corresponding diastereomeric ratio of dCT-H11 is 99:1 (trans:cis).}
    \label{fig:dr_styrene}
\end{figure}

\clearpage
\subsection{Directed evolution}
\begin{figure}[h!]
    \centering
    % \begin{subfigure}[b]{0.60\textwidth}
    %     \centering
    %     \includegraphics[width=\textwidth]{figures/bars_Azetidine_r1_yield.png}
    % \end{subfigure}    
    % \hfill
    \begin{subfigure}[b]{0.5\textwidth} % 0.35
        \centering
        \includegraphics[width=\textwidth]{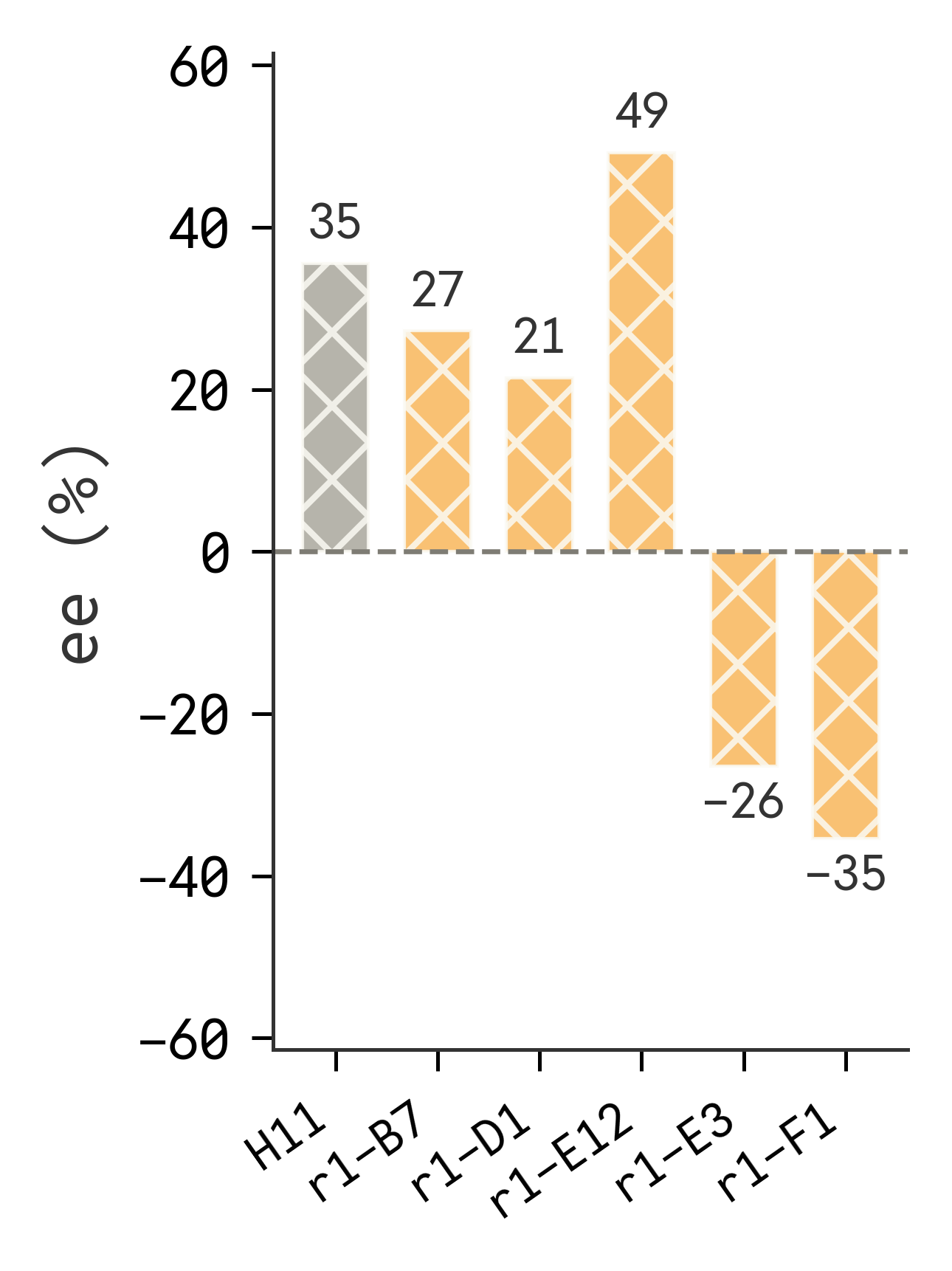}
    \end{subfigure}
    \caption{Selected validation results of the evolution of the spirocyclopropanation reaction \textbf{4b}. The parent (dCT-H11) is shown in gray. The corresponding enantiomeric ratios of dCT-H11, dCT-r1-E12, and dCT-r1-F1 are 68:32, 75:25 and 25:75, respectively.} %Left shows yield, and right shows enantiometric excess. \todo{leave validation yield out?}}
    \label{fig:ee_azetidine}
\end{figure}

% \todo{daniel - add citation}
\noindent Random mutations were introduced using error-prone PCR (epPCR). PCRs were run using Taq polymerase (New England Biolabs) in the presence of 300, 400, and 500 µM MnCl$_2$ with the primers shown in table \ref{tab:primers} following a reported procedure.\cite{boville2018engineered} Primers 005 and 006 were used to amplify the DNA region encoding the full-length protein of interest. Primers 007 and 008 were used to amplify the backbone (pET22b(+) vector) fragment containing a gene encoding for ampicillin resistance using Phusion\texttrademark  \, polymerase and the template plasmid DNA. PCR products were isolated, and fragments were then assembled into a circular plasmid using the Gibson assembly protocol, incubating at 50~\textdegree C for 1 hour. Then, 2.5~\textmu L of the Gibson product were used to transform 10~\textmu L of competent \textit{E. coli }cells. Transformed cells were supplemented with 150 \textmu L SOC medium and immediately plated on LB\textsubscript{carb} agar plates. Plates were incubated at 30~\textdegree C for 16–18 hours. Single colonies were then used to inoculate 500~\textmu L LB\textsubscript{carb} 96-well plates; eight wells on each plate were inoculated with single colonies of the parent enzyme (parent controls), and two wells were inoculated with a sterile toothpick (sterile controls).  Expression, reaction setup and work-up in 96-well plates  was done as described in section \ref{sec:express}. A total of eight plates were screened by reversed-phase HPLC--MS, and sequences were verified by in-house sequencing (LevSeq).\cite{long2024levseq} The highest-performing variants were validated in triplicate as described in section \ref{sec:express}, and the enantioselectivity was measured (see figure \ref{fig:ee_azetidine}).

\begin{table}[h!]
    \centering
\caption{Primers used in error-prone PCR.}
\label{tab:primers}
    \begin{tabular}{ccc}
         \textbf{Primers}& \textbf{Sequence (5' $\rightarrow$ 3')} & \\ \hline
         \textbf{005}& GAA ATA ATT TTG TTT AAC TTT AAG AAG GAG ATA TAC ATA TG & \\
         \textbf{006}& GCC GGA TCT CAG TGG TGG TGG TGG TGG TGC TCG AG & \\
         \textbf{007}& CAT ATG TAT ATC TCC TCC TTA AAG TTA AAC AAA ATT ATT TC & \\
         \textbf{008}& CTC GAG CAC CAC CAC CAC CAC CAC TGA GAT CCG GC & \\
    \end{tabular}

\end{table}

\begin{table}[h!]
    \centering
\caption{Mutations of highest-performing variants from error-prone PCR.}
\label{tab:mutations}
    \begin{tabular}{ccc}
         \textbf{Variant}& \textbf{Mutations} & \\ \hline
         \textbf{dCT-r1-B7}& A2P L68R E73D F82L N90I L103P S167F & \\
         \textbf{dCT-r1-D1}& E39D D110E T121A F173S E191V H199L & \\
         \textbf{dCT-r1-E3}& A66V W98R E144 V156A & \\
         \textbf{dCT-r1-E12}& V26A K27N M43L R77C F120Y G150D I152T L161P R180C L195H & \\
         \textbf{dCT-r1-F1}& V8E K27E W98R L114P L165F & \\
    \end{tabular}

\end{table}

\clearpage
\subsection{DNA sequences of the best designs}
\seqsplit{
\noindent >dCT-H11\\
ATGGCGCTGCCAGATATCGAAGTGTACCCGGATCTGGAACTGAGCGAGCGTCAGCGTGAACGCTTAGTCAAAATAGTGAAAGCGGCCATTGAAGAAATCTATGAACGCGGTTACGAAGCCACGTCGATGGAAGATATCGCCGAGCGCGCCGGCATTTCACGTGAGACTGTTCGCTACTATTTTCCTTCTAAAGAGGCTCTGCTGAAAGCAGCTGCAGAATTTTTAGCGCGCCGTCATGAACGTTTCTTCGAAGAATGGAGTAACATTAATAGCCCCGAAGAGCTTGAAGATTGGTTGCAGAAACTGCTTGATAGCTTGGAACGGGTCGATCTGGAGAAACTGGTGAACCTGGCGGACTTCACCGAAGCAGAGCCGGAGCTCAAAGAGAAAATCAAAGAACTGCTGTATCATCGGCTGAAAAAACTCTTGGAAGAAGGGAAGAAAAAGGGCCAAATTAAAGAGGACGTGGATGTTGAAGAACTGACCAAGGAATTACTGTCCATGATTGATCAGCTGTTTAGTACGTATATCAAAACACGTGACCGCGCGGAATTTGAAAAGCTCAAAAAGGAGATTTTGGACCTTTTACGTAAACACGTAGTTCTCGAGCACCACCACCACCACCACTGA}
\seqsplit{
\\\\>dCT-H10\\
ATGGGGGTGGCCGCGGAATTAGAACGTGCAGCGCGGGCTTTACGTGAAGCTCTTCGCGCAGCTGTTCGTGAGGCTGGTTTACGGCTGCCTGGTGAGGAAGCCGAAGCGTTCGAACGCCTGCTGGCCTTACATTGCGAGGTGGCTGACGGTCTGTTGCGCGTTCTGGCGGCAGCGTTAACTCGTGATCCTGCAGAGTGTGCTGCTGCTGCGCGTGCCGTAGGTGAATTGGCTCTGCGCCTGTTGGAAGCAAGTGCTGCGCTTGTTGTACCTCCAGGACGTGCAGCGCTTGCGCGTGCGTTTGTTGAACGGTTGCGTGGCTTTGTCGAAGGCTTCGTGGCGTTAGCGGAGCGTGTTGGCGGAGCGTCTGCAGATGCTTTAGAGGAAGCAGCTCGTCGTCTGGCGGCCGCTGCTTTGGCAGCAGATCGTGTAGTTCTGGCCTTCTTTGAAGAAGCGCTTGCTGAAGCCGAGCGTGCGGGCGATGAACGTTACGCGCGCTTATGCCGTGAACGTCTGGCAGTGTATCCTCGCGTGAGCCGTGCGGCATTAGACGCGTTTGCGGCATTAGCAGTTCGTCTGATTCGCGCCGCTGCTCGTGCGCTGGCCCTCGAGCACCACCACCACCACCACTGA}
\seqsplit{
\\\\>dCT-F9\\
ATGTCTTATGAAGAGCTGCTCGAGGAAATTGCACGCGAGATCCTGCGCAGTGCGCGCGCTGTGTATGAGGGATCAATCGAAGAACATCGTGAACATGTTCGCCGTGCACGCGAATTGCTTGAGCGTCTGGGCAATGTCCGTGTAGTTGTCGAAGTGCGTGATCCGGATGGCGTTCCCCTGTTAGTAGTCCGTTTTGAAGTGGAAAACGGCGAATGGCGTATCGACGCGATATACAACCAAGTGTCGCGCTTAACCGATCAGGAGTTCTACGAGTTCGTGCGGGACAATCCGGGTATTGACCCACGGGAACTCCTGCCTCACCTTGCCGAATTTGTTCGTCGCCTGAAAGAATCCGGTAGCGTTGAGGTGCGGGTACGCATTGAATTGACGGATGAAGAAGTGGTGTTAGAACTGGTCGAACGTCTGGAGGGGCGTCCGGCCGAAGAACTGTATCTCGAGCACCACCACCACCACCACTGA}
\seqsplit{
\\\\>dCT-G9\\
ATGGAATACGATTATAAAGCTCGCAGCCTTAAAAAAGCCGAGTTACTGACTAACGCTCTGAAAGGTGAAAACGCTTCGCCAGAAGTTAAAGCGTTTGTAGATGCGCTTAAAAAGAAAGTGGGAGATGATATTGTGCCGGCGGTCGAAGCACTGAAGAATGTTAGCGAAGACCTCAAGAAAGAAATTGTAGAGCTGCTTCGTAAAGCATTGGATAGTGCCAAAACCAAGGACGAACTGATCGCCGCGTTAACCAAACTGCATGGCTTAGTCTTCGACTCCCTGGGGCTGCCGAATGCCGATGCAATAATCAAACAGGTGACCGATGCCCTGAAGCAGAAGTCTCTGGAGGAGCTCAAAGAAGTGTTGAAAACAGTCGTTGATGGTCACACGGGCAAAACGGACGAAAAACAAGCAATCGAAAAAGCAATGGCGTTGGCCGCGGCGATTTTTTCACTCGAGCACCACCACCACCACCACTGA}
\seqsplit{
\\\\>dCT-G1\\
ATGACTTTTGAAGAACTTAAAAAGAAAACGAAAGAGGAGGTGCGTGAAATTGTTGAAAGTTGCGATCCTAAAGAACTTGTCGAGCGTCTCCAGGAACTGGTGCGTGAAACACATAACTCGGTGCTTCCGCCGACCATAGTAGAATTTCAAAGCAATCTGTTTGAACGGCTGCTGGCTGCGAAAACGGTCGAAGAGAAAGCGCTGTTGCTGGAGAAGTTGGTTTTTATTGGTATCTATGTCAAGATCGTTCGCAAATACGATGAGGAGCTGTTCAAAAAGGTACAGGCCATTTTACGGCGTTTCTTGGAGGAAATCGACACCTTAACCGAAGAAAAAATTAAAGAATTCCTGGATCGCATTTATGAACTGGTGGAACCACTGAAACGCGCCTGGCTCGAAGAAGGCGAGAAAAACGCACGCAAAGTGCTGGAAGCAATCAAGAAATTATGTCTCGAGCACCACCACCACCACCACTGAGA}

\clearpage
\subsection{Calibration curves for GC-FID analysis}
\label{sec:calib}

\textbf{For 1b, 3b and 4b:}

\noindent The analyte of interest and internal standard (1,2-diphenylethane) were dissolved in a 1:1 (v:v) solution of cyclohexane and EtOAc to create two separate 100~mM solutions. Seven different standard-to-internal standard (S/IS) ratios were prepared: 1:1, 0.5:1, 0.25:1, 0.125:1, and 0.0625:1 by mixing specific volumes of the 100~mM solutions in a GC vial. The total volume for each sample was adjusted to 500~\textmu L by adding cyclohexane:EtOAc (1:1, v/v). An aliquot of the prepared solutions was transferred to a 400~\textmu L glass insert within a 2.0 mL screw-cap vial and subjected to GC-FID analysis with an achiral stationary phase.\medskip 

\noindent \textbf{For 2b:} 

\noindent The analyte of interest was prepared under conditions analogous to those used for enzymatic reactions. A 200~mM stock solution in MeCN was prepared, from which successive stocks of 100, 20, 10, 2 and 1~mM were generated. From each stock, 20~\textmu L were added to 380~\textmu L of reaction buffer (M9-N, pH 8.0, 10 g/L glucose), followed by 400~\textmu L of a 1:1 (v/v) cyclohexane:EtOAc solution containing the internal standard (1,2,3-trimethoxystyrene, 1 mM). Tubes were vortexed and centrifuged (20,000$\times$g, 10 min, room temperature), and 200~\textmu L of the organic layer were transferred to a 400~\textmu L glass insert in a 2.0~mL screw-cap vial for GC--FID analysis.

\noindent The IS and S peaks were integrated using built-in analysis software (Agilent). Calibration curves were generated by plotting the known concentrations of the product, based on prepared S/IS ratios, as a function of the ratio of its FID signal (peak area) relative to that of the internal standard. The slope of the calibration curve was used to calculate the product yield in enzymatic reactions.

\clearpage
\subsubsection{Calibration curves for \textbf{1b}}
\begin{figure}[ht!]
    \centering
    \includegraphics[width=0.7\linewidth,trim={0 0 0 0.85cm},clip]{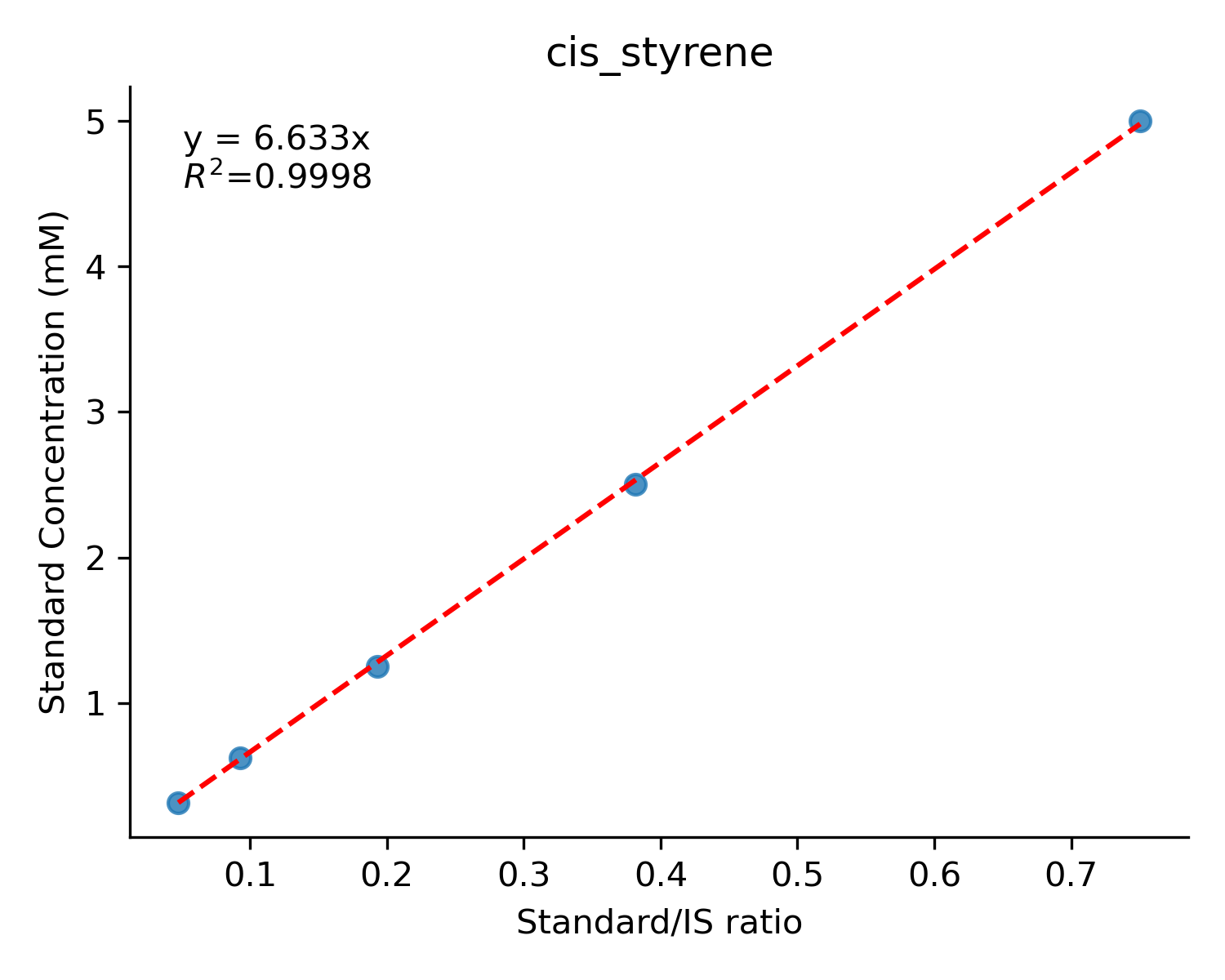}
    \caption{Calibration curve of the cis isomer of \textbf{1b} using 1,2-diphenylethane as IS.}
    % \label{fig:placeholder}
\end{figure}

\begin{figure}[ht!]
    \centering
    \includegraphics[width=0.7\linewidth,trim={0 0 0 0.85cm},clip]{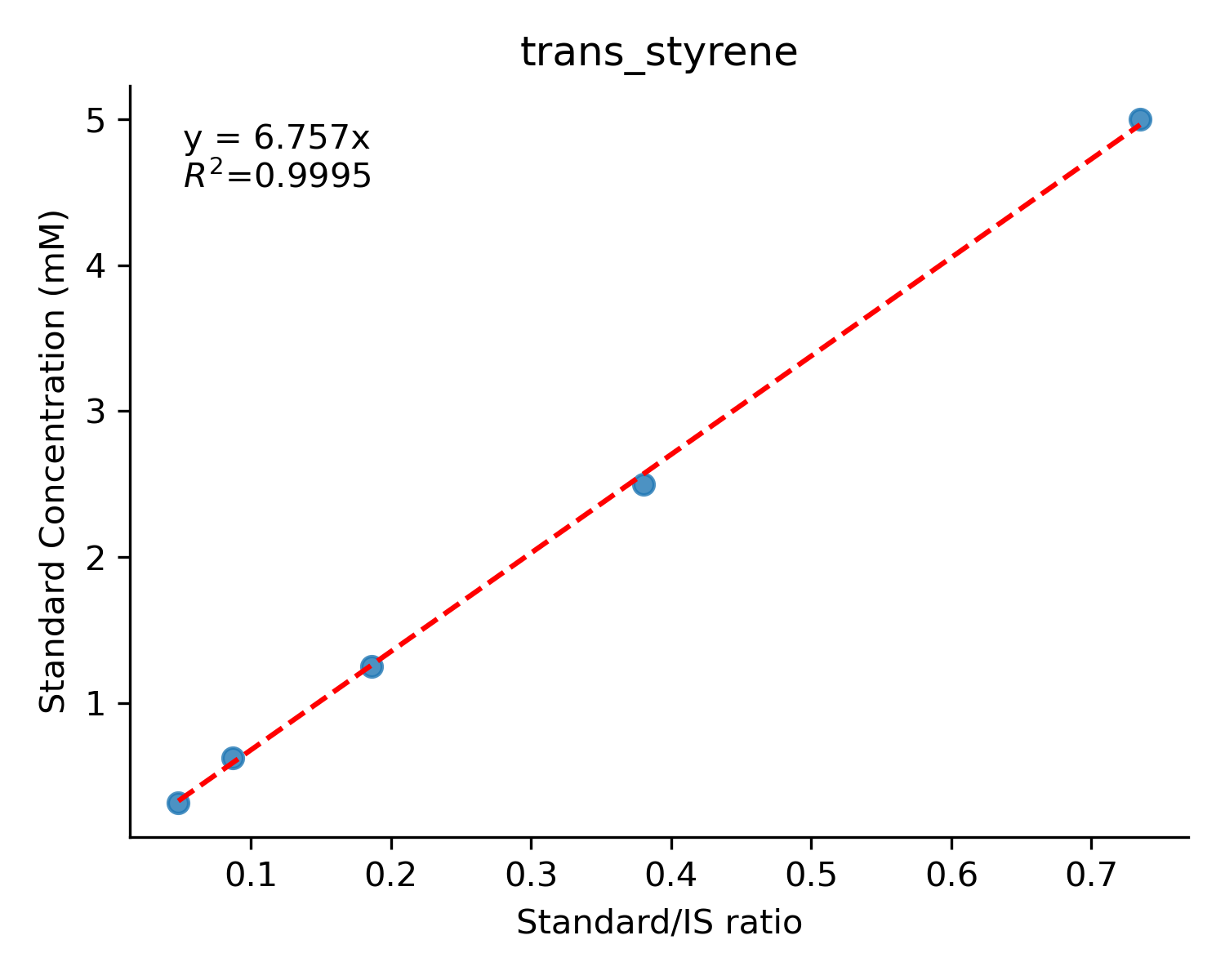}
    \caption{Calibration curve of the trans isomer of \textbf{1b} using 1,2-diphenylethane as IS.}
    % \label{fig:placeholder}
\end{figure}

\clearpage
\subsubsection{Calibration curves for \textbf{2b}}
\begin{figure}[ht!]
    \centering
    \includegraphics[width=0.7\linewidth,trim={0 0 0 0.85cm},clip]{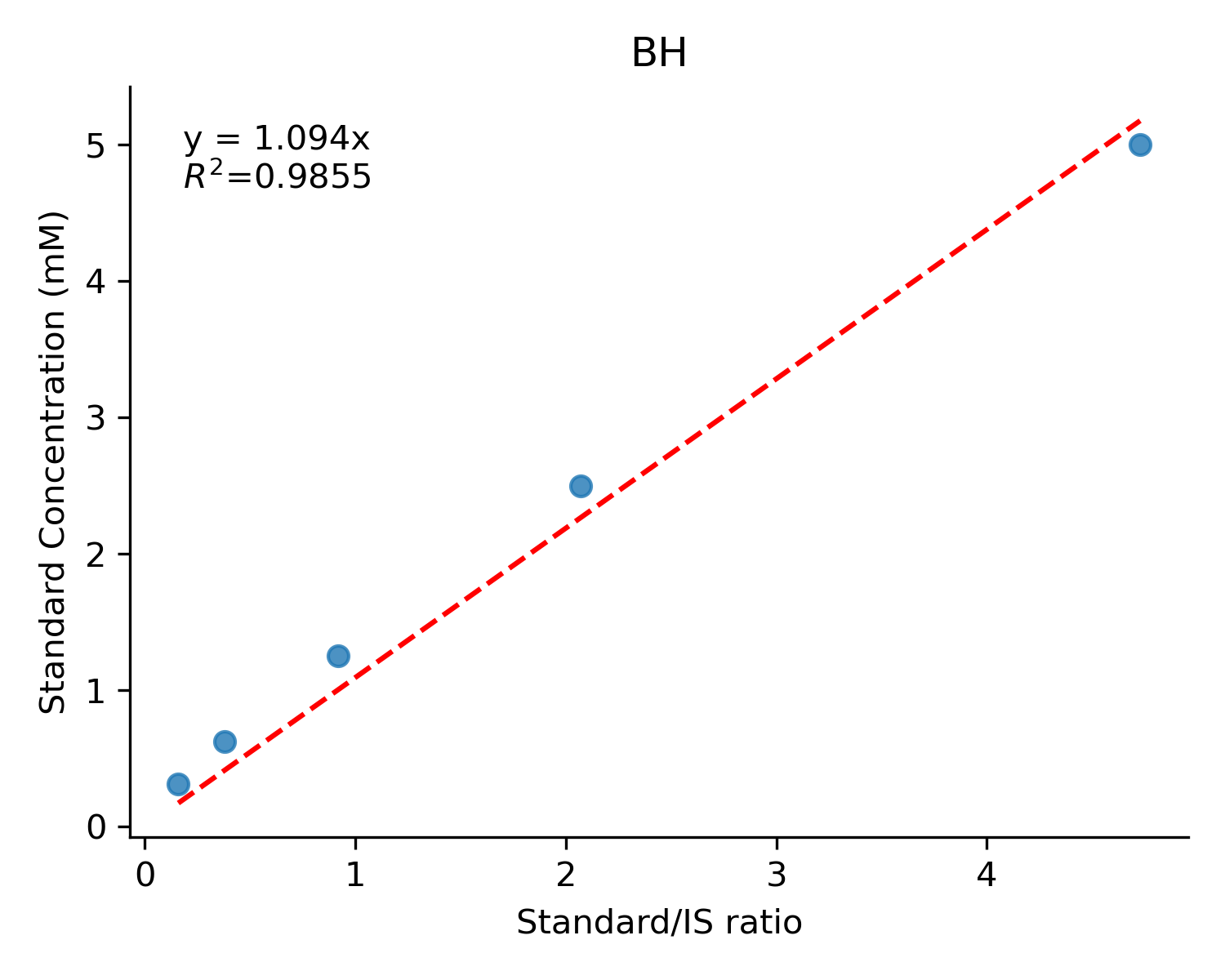}
    \caption{Calibration curve of \textbf{2b} using 1,2,3-trimethoxystyrene as IS.}
    % \label{fig:placeholder}
\end{figure}

% \clearpage
\subsubsection{Calibration curves for \textbf{3b}}
\begin{figure}[ht!]
    \centering
    \includegraphics[width=0.7\linewidth,trim={0 0 0 0.85cm},clip]{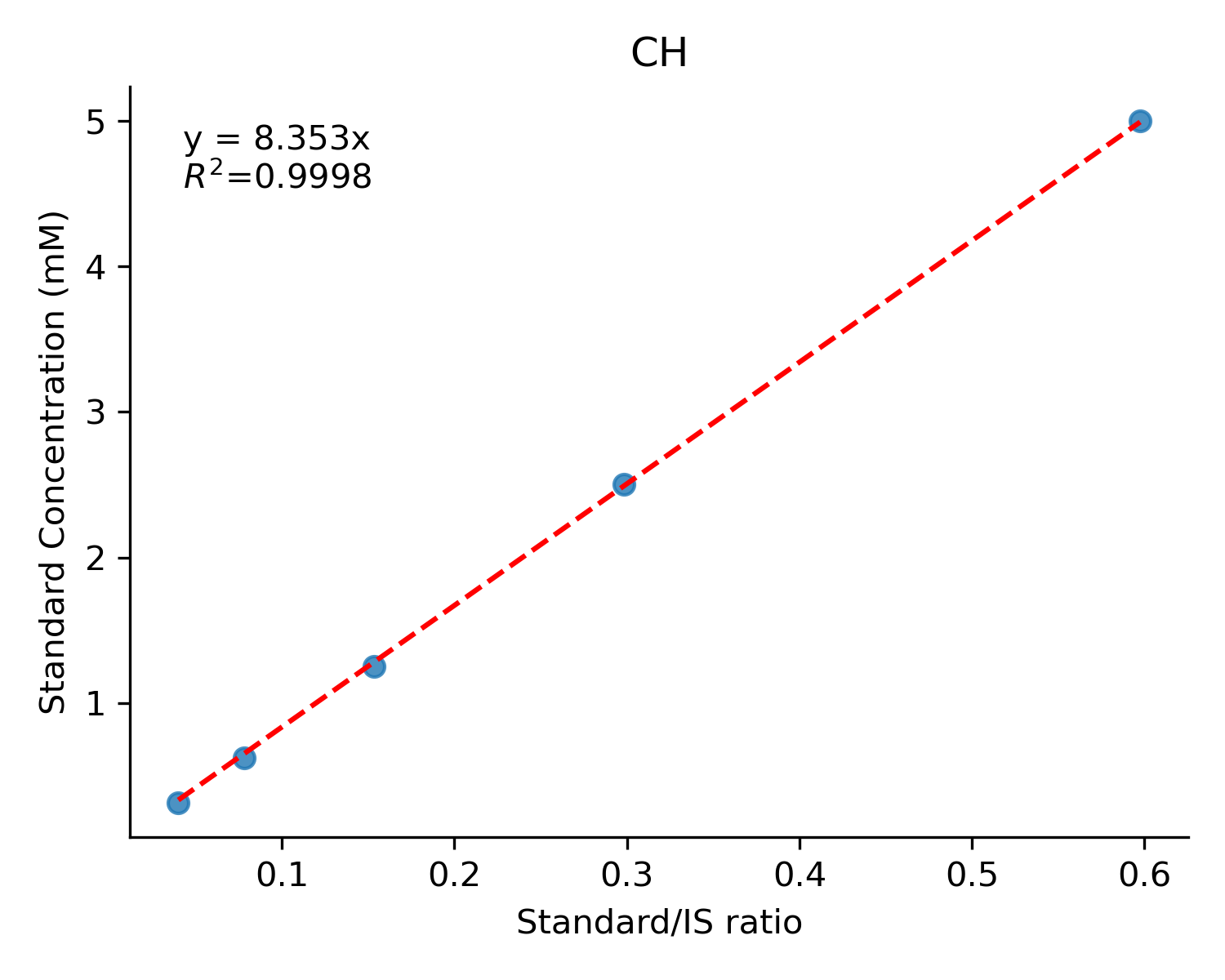}
    \caption{Calibration curve of \textbf{3b} using 1,2-diphenylethane as IS.}
    % \label{fig:placeholder}
\end{figure}

\clearpage
\subsubsection{Calibration curves for \textbf{4b}}
\begin{figure}[ht!]
    \centering
    \includegraphics[width=0.7\linewidth,trim={0 0 0 0.85cm},clip]{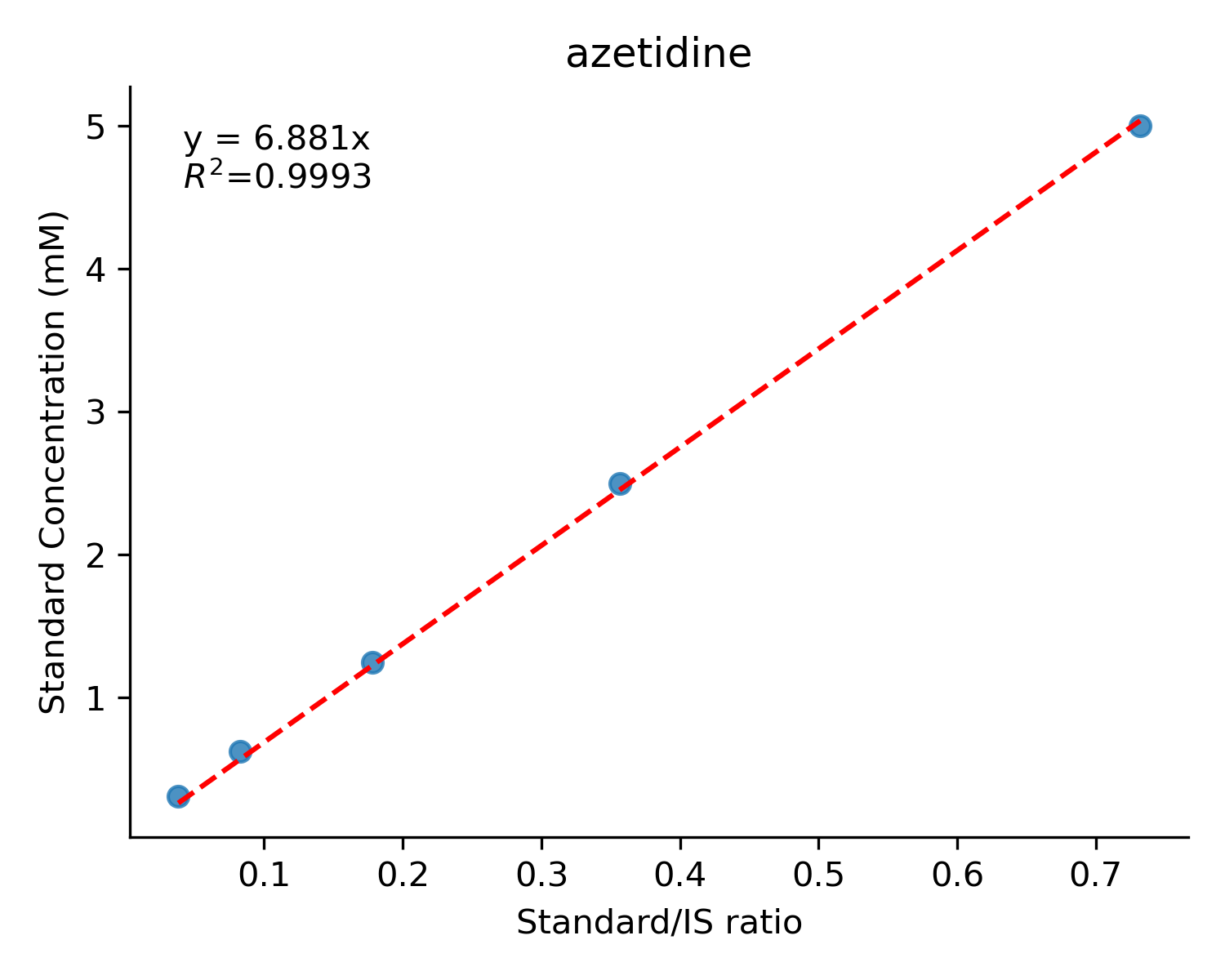}
    \caption{Calibration curve of Compound \textbf{4b} using 1,2-diphenylethane as IS.}
    % \label{fig:placeholder}
\end{figure}

\clearpage
\subsection{Calibration curves for LC-MS analysis}
\label{sec:calib2}
A stock solution of \textit{tert}-butyl-3-methyleneazetidine-1-carboxylate (\textbf{4a}) was prepared in acetonitrile (1~mM final concentration) and used to create successive stocks of 10, 30, 50, 100, 200, 300~\textmu M concentration. An aliquot of these stocks (300~\textmu L) and 50~\textmu L of a stock solution of papaverine hydrochloride (internal standard, 6.5 mg in 2:1 MeCN:H$_2$O) were transferred to 400~\textmu L glass inserts in 2.0 mL screw-cap vials for LC--MS analysis.

\begin{figure}[ht!]
    \centering
    \includegraphics[width=0.7\linewidth]{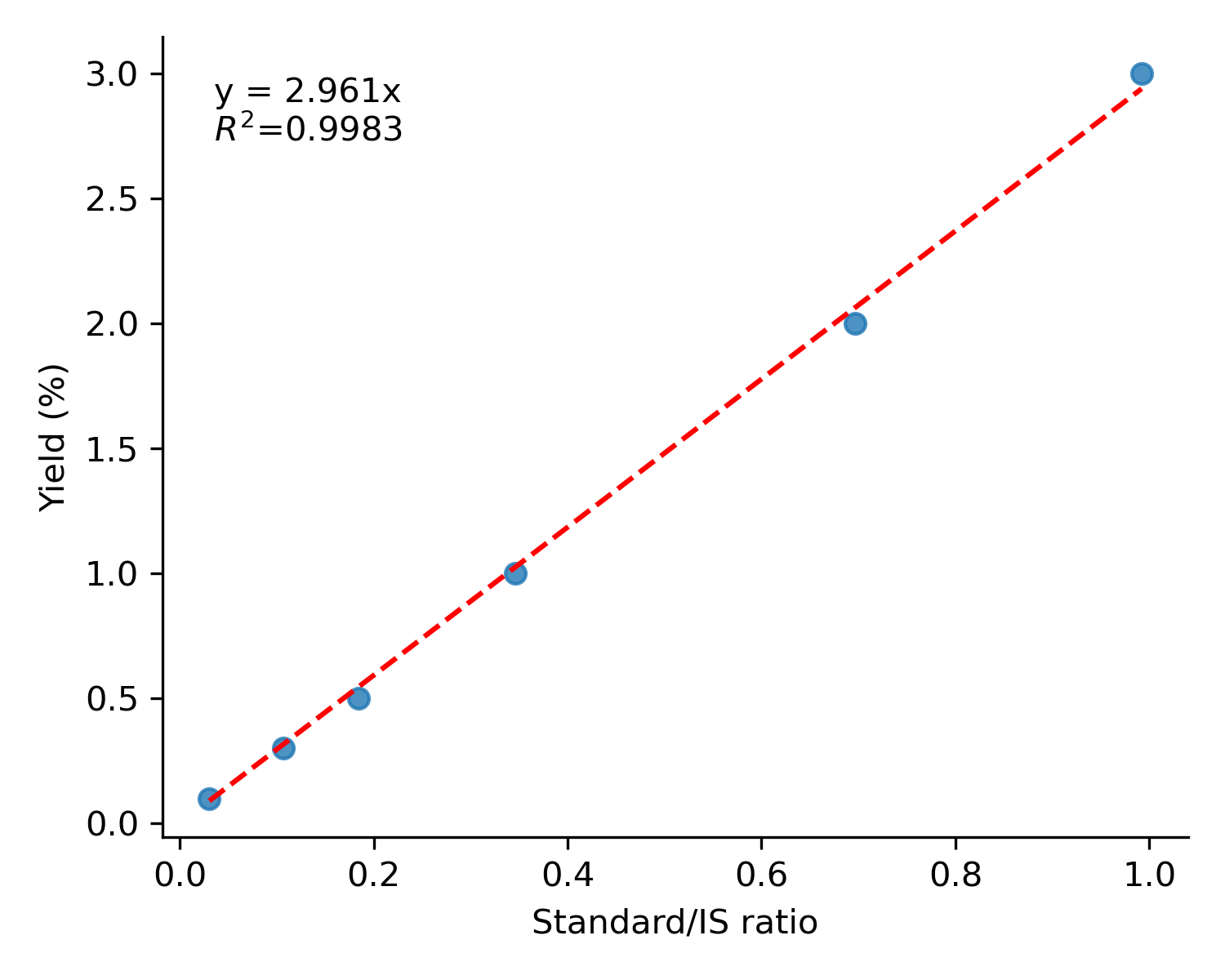}
    \caption{Calibration curve of \textbf{4b} on LC-MS for directed evolution assay.}
    % \label{fig:placeholder}
\end{figure}

%\clearpage 
\subsection{Synthesis of product standard 3b}
\begin{figure}[h!]
    \centering
    \includegraphics[width=1\linewidth]{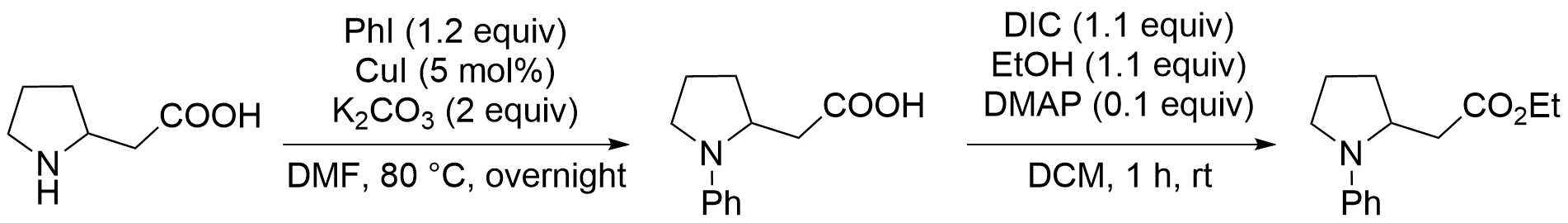}
    \caption{Synthesis of \textbf{3b}.}
    \label{fig:3b_synthesis}
\end{figure}

\noindent \textbf{Step 1:} A sealed tube flushed with nitrogen was charged with  pyrrolidine (5 mmol), potassium carbonate (2 equiv, 10 mmol), copper (I) iodide (0.25 mmol, 5 mol$\%$), iodobenzene (1.2 equiv, 6 mmol) and DMF (7.5 ml). The mixture was heated at 90~\textdegree C for 48 hours, then cooled to room temperature. Water was added, and the pH value was adjusted to <3 with concentrated HCl. The aqueous phase was extracted four times with ethyl acetate. The combined organic layers were washed with brine, dried over magnesium sulfate, filtered and concentrated under reduced pressure. Purification by flash column chromatography (0 to 100$\%$ EtOAc/hexane gradient) afforded the product. \smallskip

\noindent \textbf{Step 2:} A round-bottom flask was charged with carboxylic acid (1.0 equiv), ethanol (1.1 equiv), and DMAP (0.1 equiv). Dichloromethane was added (0.4 M), and the mixture was stirred vigorously. DIC (1.1 equiv) was then added dropwise via syringe, and the reaction mixture was stirred until consumption of the acid was complete, as determined by TLC. The mixture was filtered through a fritted funnel and rinsed with CH\textsubscript{2}Cl\textsubscript{2}/Et\textsubscript{2}O. The solvent was removed under reduced pressure, and the crude product was purified by flash column chromatography to afford the corresponding ester as a yellow oil. \smallskip

\noindent \textbf{\textsuperscript{1}H NMR} (400 MHz, CDCl\textsubscript{3}) $\delta$ 7.26 – 7.22 (m, 2H), 6.71 – 6.65 (m, 1H), 6.65 – 6.55 (m, 2H), 4.20 – 4.15 (m, 3H), 3.46 – 3.39 (m, 1H), 3.22 – 3.13 (m, 1H), 2.79 (dd, \textit{J} = 15.0, 2.9 Hz, 1H), 2.26 – 2.17 (m, 1H), 2.17 – 1.94 (m, 4H), 1.28 (t, \textit{J} = 7.1 Hz, 3H). \textbf{\textsuperscript{13}C} NMR (101 MHz, CDCl\textsubscript{3}) $\delta$ 172.1, 146.5, 129.5, 129.4, 121.6, 116.0, 111.9, 55.4, 47.9, 37.8, 31.0, 23.1, 14.3.  \smallskip

\noindent Spectroscopic data are in agreement with the literature.\cite{zhang2019enzymatic}

\begin{figure}
    \centering
    \includegraphics[width=1\linewidth]{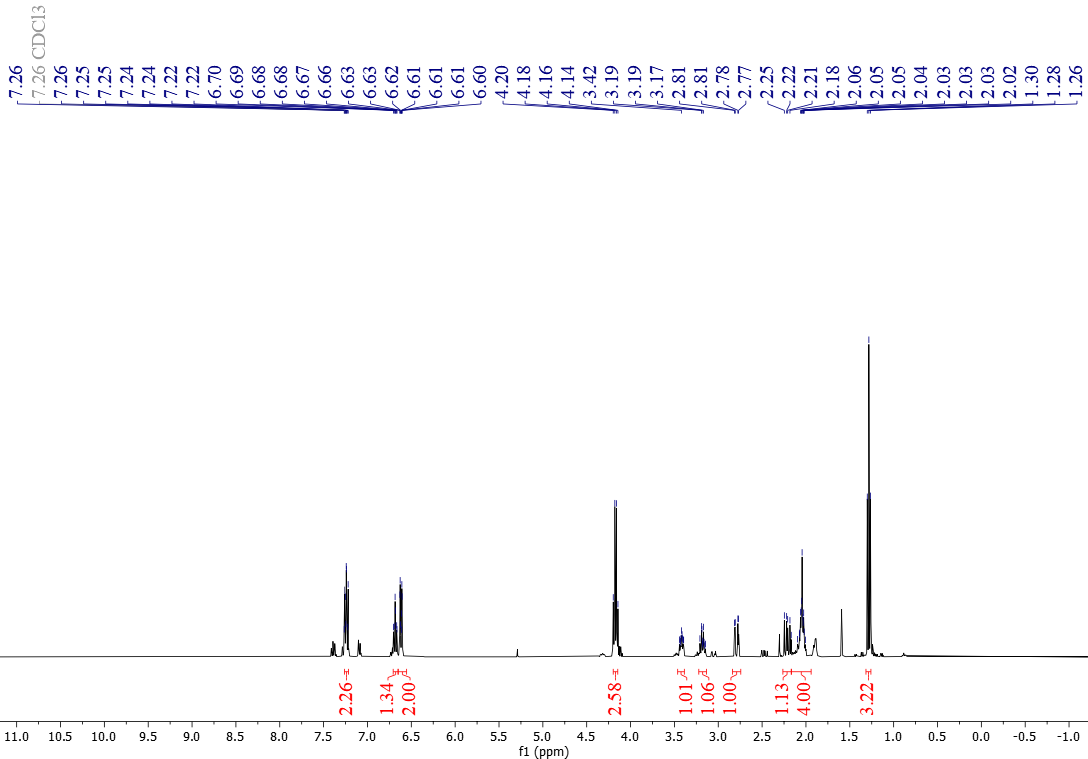}
    \caption{\textsuperscript{1}H NMR (400 MHz, CDCl3) of \textbf{3b}.}
    \label{fig:1H_NMR_3b}
\end{figure}

\begin{figure}
    \centering
    \includegraphics[width=1\linewidth]{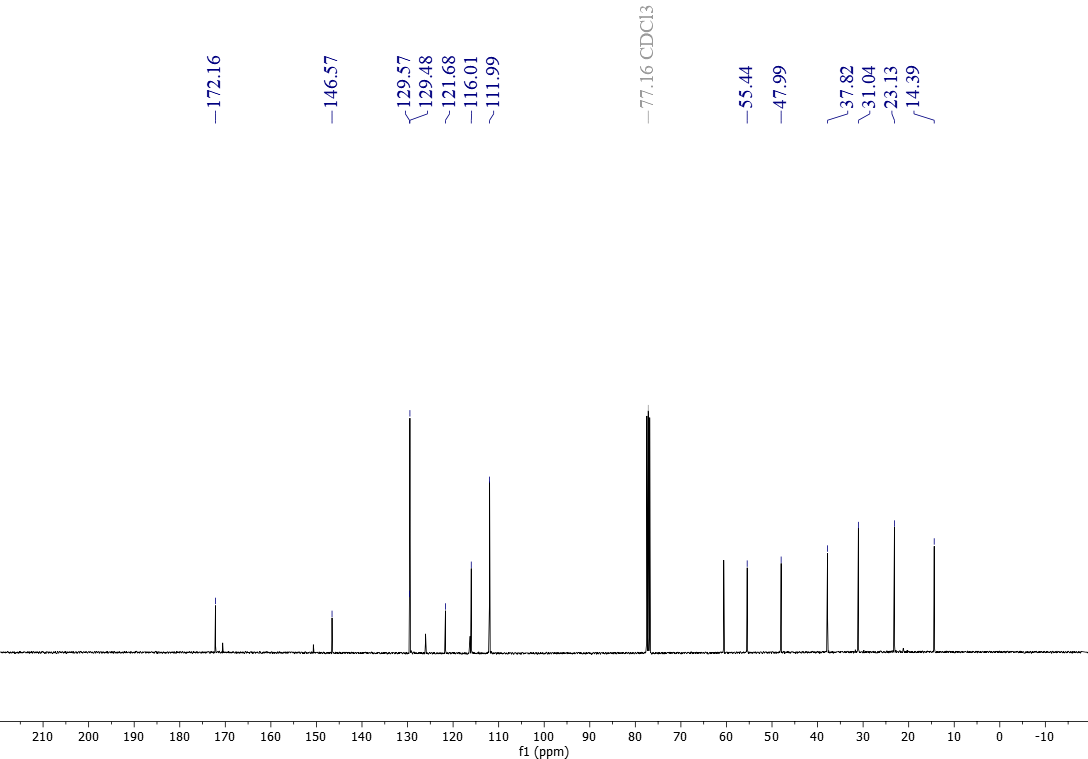}
    \caption{\textsuperscript{13}C NMR (101 MHz, CDCl3) of \textbf{3b}.}
    \label{fig:13C_NMR_3b}
\end{figure}

\clearpage 
\subsection{HPLC--UV chromatograms for enantioselectivity determination}
\label{sec:raw_ee}

\begin{figure}[h!]
    \centering
    \includegraphics[width=1\linewidth]{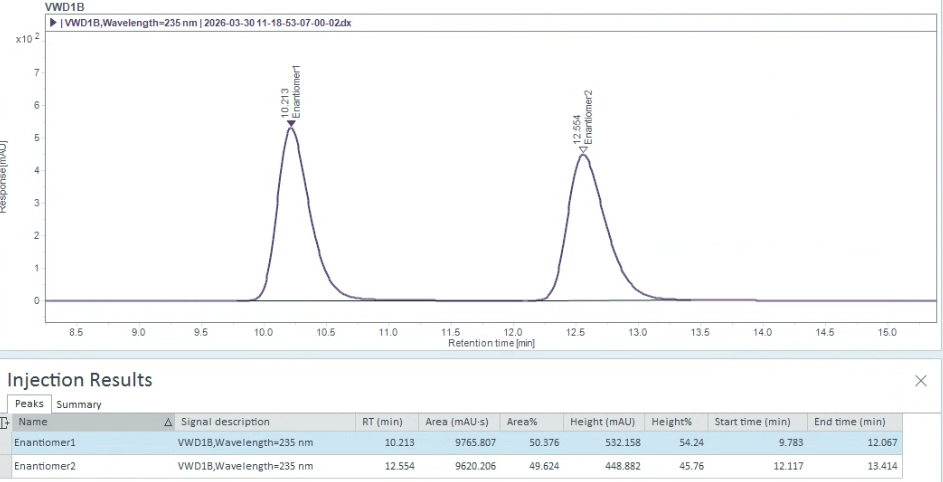}
    \caption{HPLC-UV chromatogram of an authentic standard of compound \textbf{1b}.}
    \label{fig:CP_Ins_Pdt_Std}
\end{figure}

\begin{figure}[h!]
    \centering
    \includegraphics[width=1\linewidth]{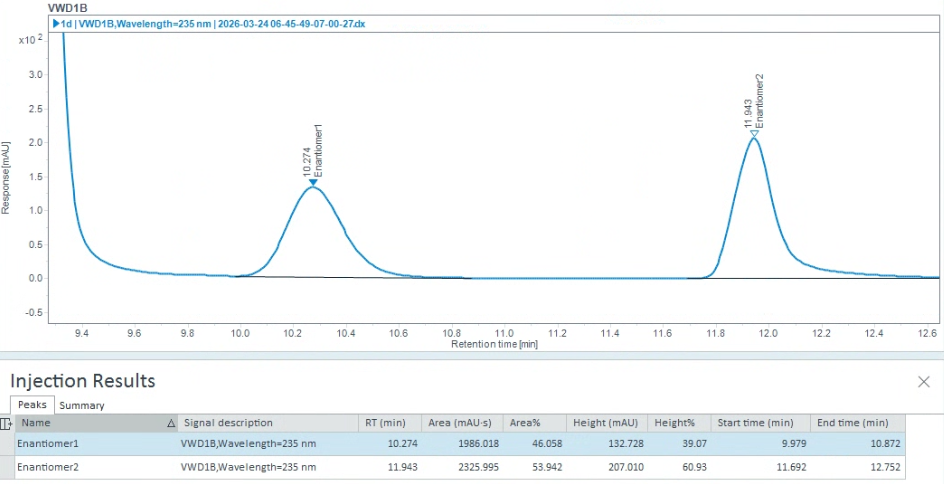}
    \caption{HPLC-UV chromatogram of compound \textbf{1b} (reaction with dCT-B9).}
    \label{fig:CP_Ins_1_B9}
\end{figure}

\begin{figure}
    \centering
    \includegraphics[width=1\linewidth]{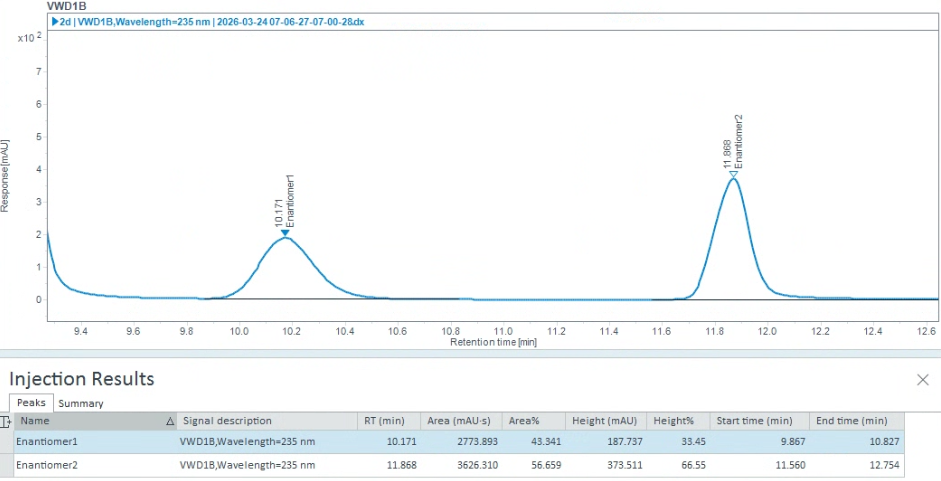}
    \caption{HPLC-UV chromatogram of compound \textbf{1b} (reaction with dCT-F3).}
    \label{fig:CP_Ins_2_F3}
\end{figure}

\begin{figure}
    \centering
    \includegraphics[width=1\linewidth]{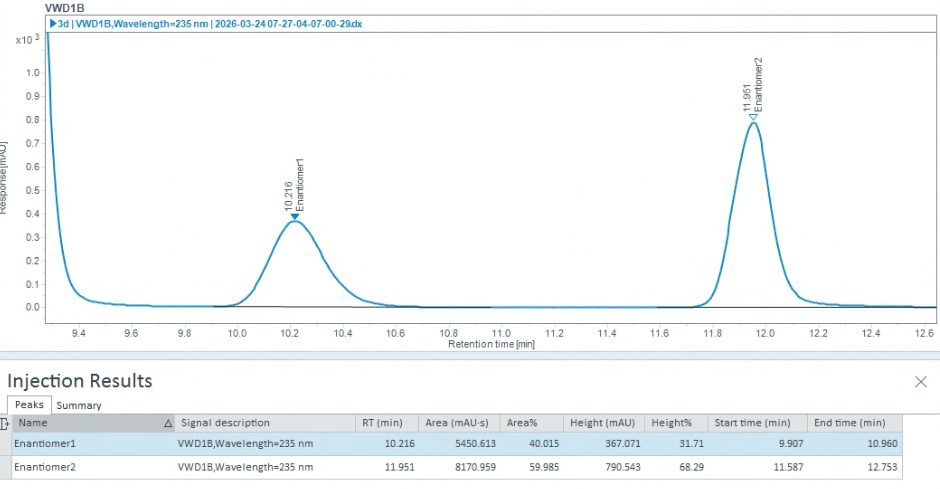}
    \caption{HPLC-UV chromatogram of compound \textbf{1b} (reaction with dCT-F9).}
    \label{fig:CP_Ins_3_F9}
\end{figure}

\begin{figure}
    \centering
    \includegraphics[width=1\linewidth]{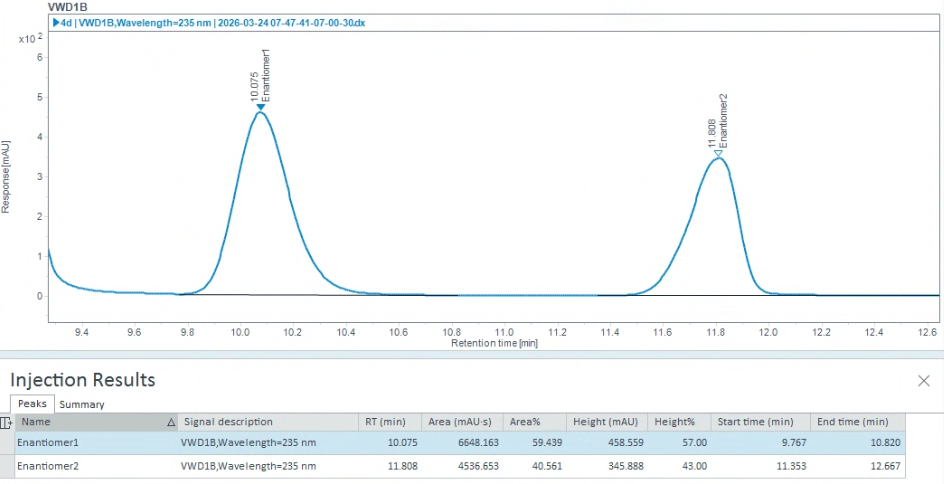}
    \caption{HPLC-UV chromatogram of compound \textbf{1b} (reaction with dCT-G1).}
    \label{fig:CP_Ins_4_G1}
\end{figure}

\begin{figure}
    \centering
    \includegraphics[width=1\linewidth]{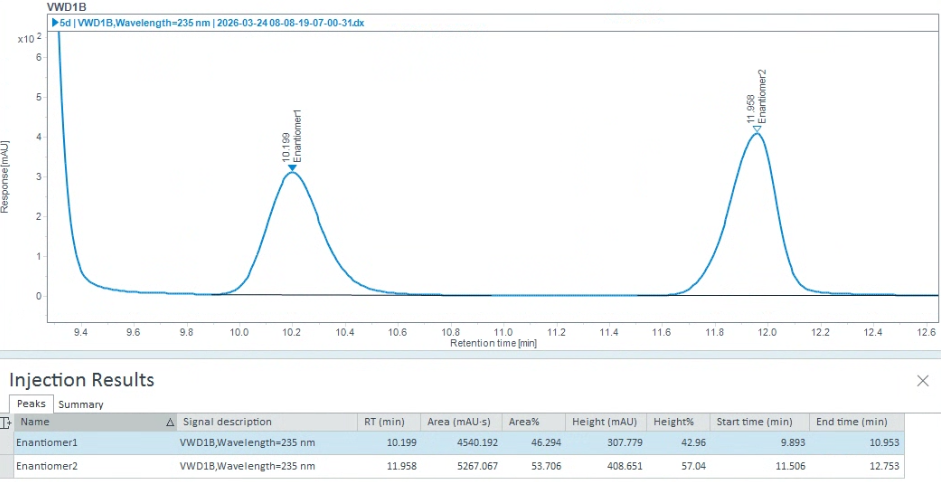}
    \caption{HPLC-UV chromatogram of compound \textbf{1b} (reaction with dCT-G9).}
    \label{fig:CP_Ins_5_G9}
\end{figure}

\begin{figure}
    \centering
    \includegraphics[width=1\linewidth]{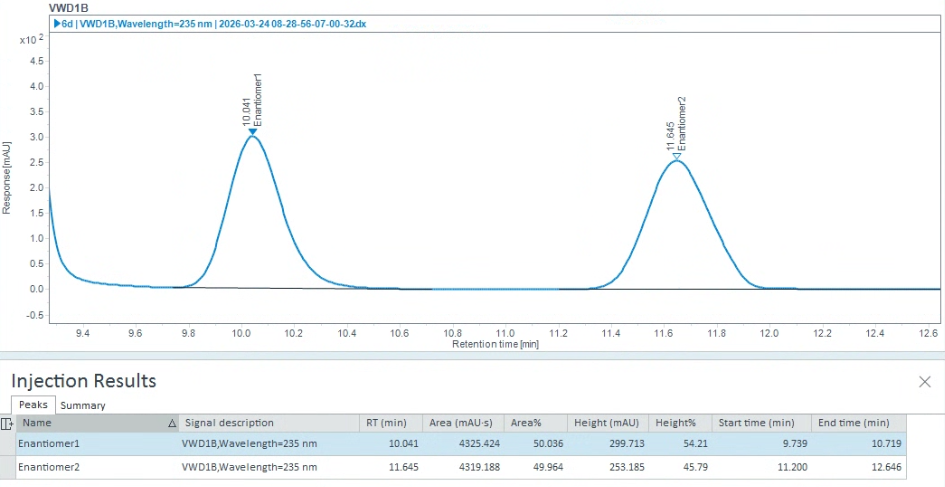}
    \caption{HPLC-UV chromatogram of compound \textbf{1b} (reaction with dCT-H5).}
    \label{fig:CP_Ins_6_H5}
\end{figure}

\begin{figure}
    \centering
    \includegraphics[width=1\linewidth]{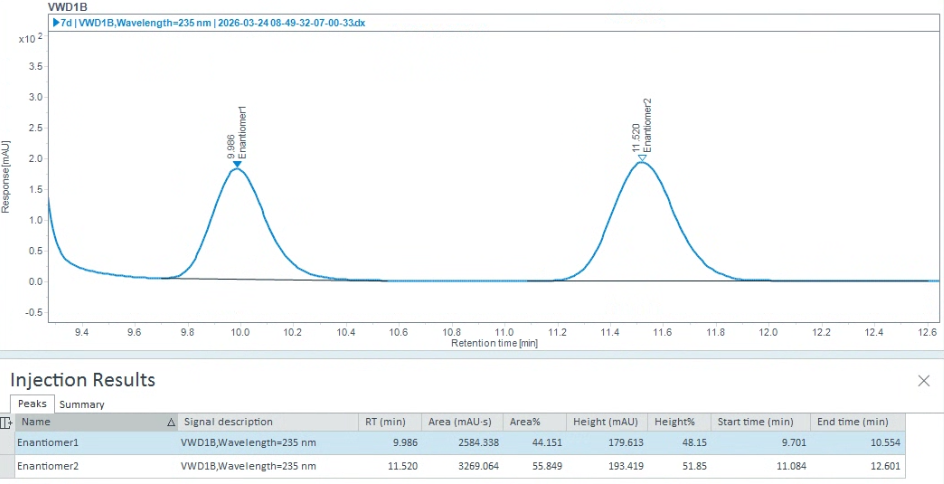}
    \caption{HPLC-UV chromatogram of compound \textbf{1b} (reaction with dCT-H10).}
    \label{fig:CP_Ins_7_H10}
\end{figure}

\begin{figure}
    \centering
    \includegraphics[width=1\linewidth]{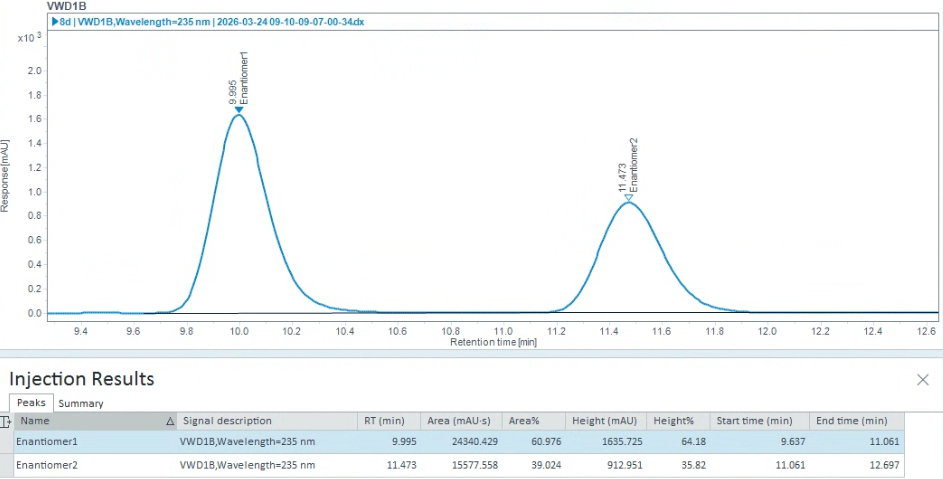}
    \caption{HPLC-UV chromatogram of compound \textbf{1b} (reaction with dCT-H11).}
    \label{fig:CP_Ins_8_H11}
\end{figure}

\begin{figure}
    \centering
    \includegraphics[width=1\linewidth]{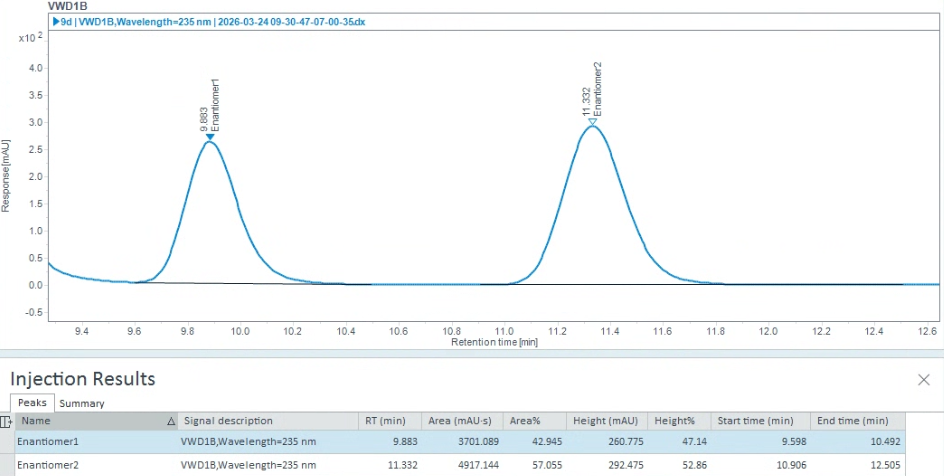}
    \caption{HPLC-UV chromatogram of compound \textbf{1b} (reaction with TmTrpB control).}
    \label{fig:CP_Ins_9_TrpB}
\end{figure}

\begin{figure}
    \centering
    \includegraphics[width=1\linewidth]{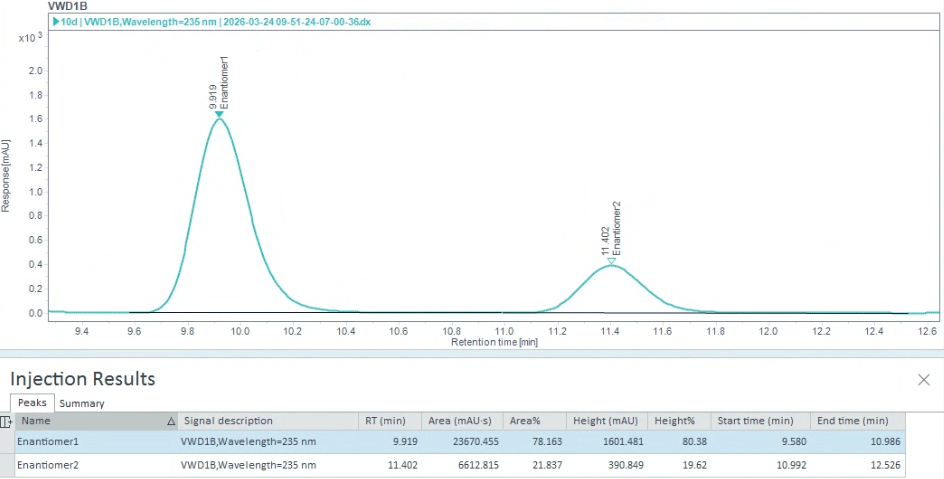}
    \caption{HPLC-UV chromatogram of compound \textbf{1b} (reaction with ApePgb 5312 control).}
    \label{fig:CP_Ins_10_ctrl}
\end{figure}

\begin{figure}[h!]
    \centering
    \includegraphics[width=1\linewidth]{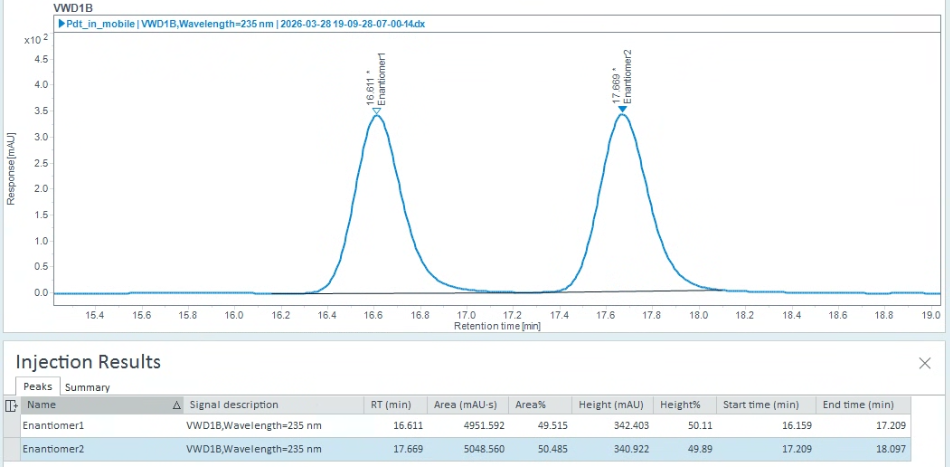}
    \caption{HPLC-UV chromatogram of an authentic standard of compound \textbf{3b}.}
    \label{fig:ee_3b}
\end{figure}

\begin{figure}[htbp!]
    \centering
    \includegraphics[width=1\linewidth]{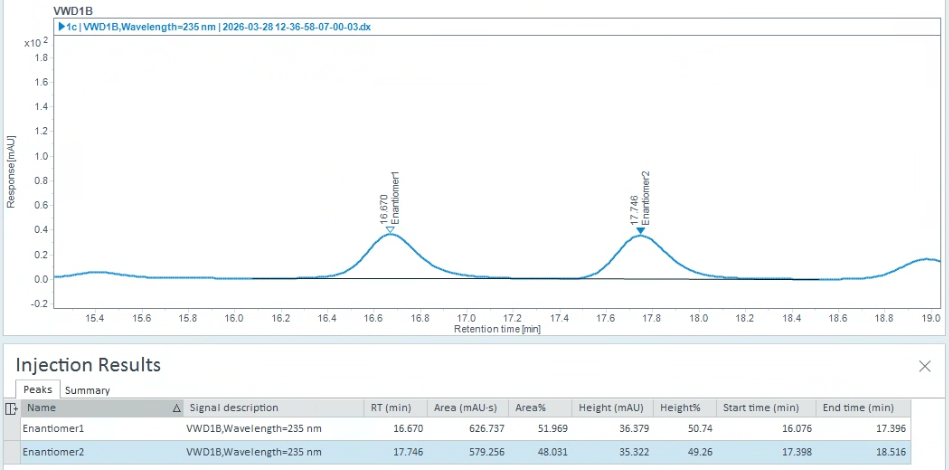}
    \caption{HPLC-UV chromatogram of compound \textbf{3b} (reaction with dCT-B9).}
    \label{fig:ee_3b_dCT-B9}
\end{figure}

\begin{figure}[ht!]
    \centering
    \includegraphics[width=1\linewidth]{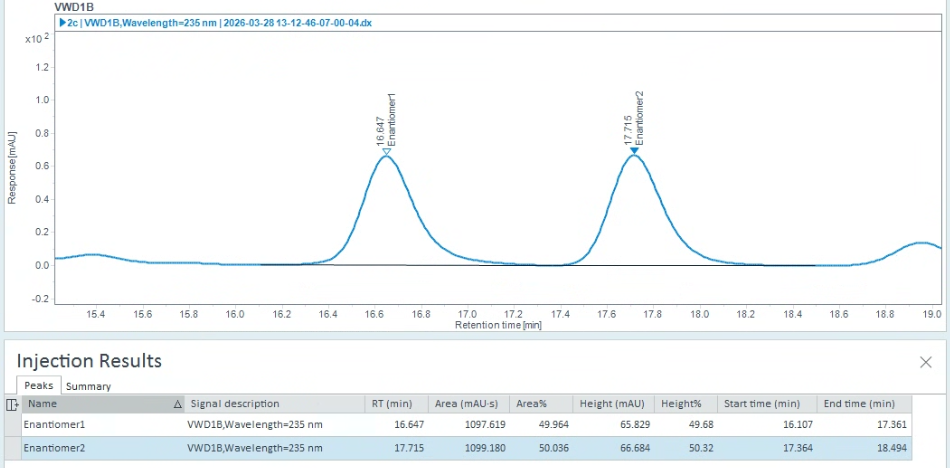}
    \caption{HPLC-UV chromatogram of compound \textbf{3b} (reaction with dCT-F3).}
    \label{fig:ee_3b_dCT-F3}
\end{figure}

\begin{figure}[ht!]
    \centering
    \includegraphics[width=1\linewidth]{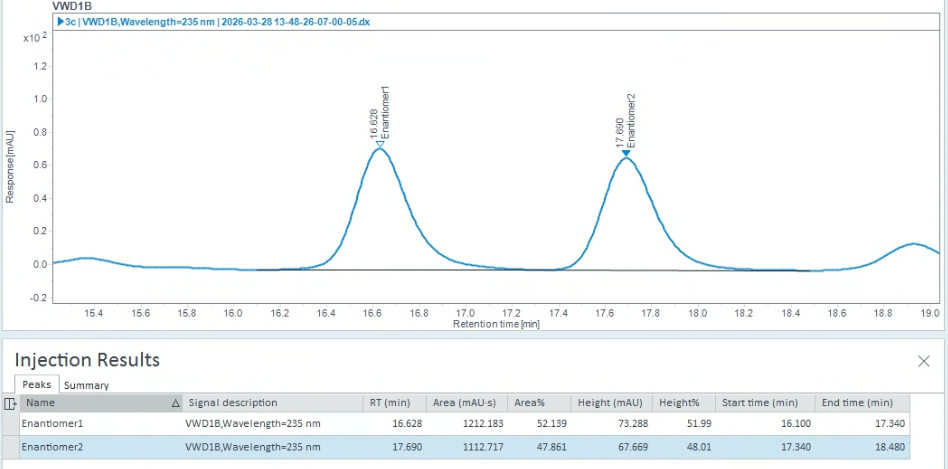}
    \caption{HPLC-UV chromatogram of compound \textbf{3b} (reaction with dCT-F9).}
    \label{fig:ee_3b_dCT-F9}
\end{figure}

\begin{figure}[ht!]
    \centering
    \includegraphics[width=1\linewidth]{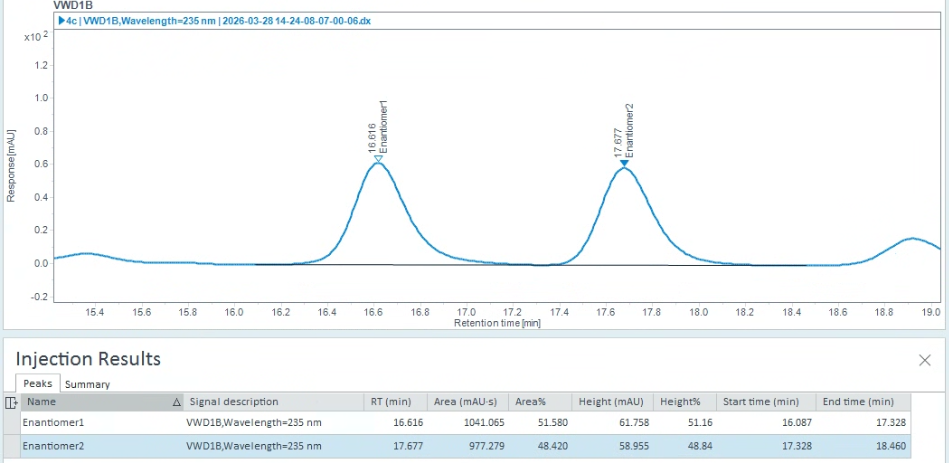}
    \caption{HPLC-UV chromatogram of compound \textbf{3b} (reaction with dCT-G1).}
    \label{fig:ee_3b_dCT-G1}
\end{figure}

\begin{figure}[ht!]
    \centering
    \includegraphics[width=1\linewidth]{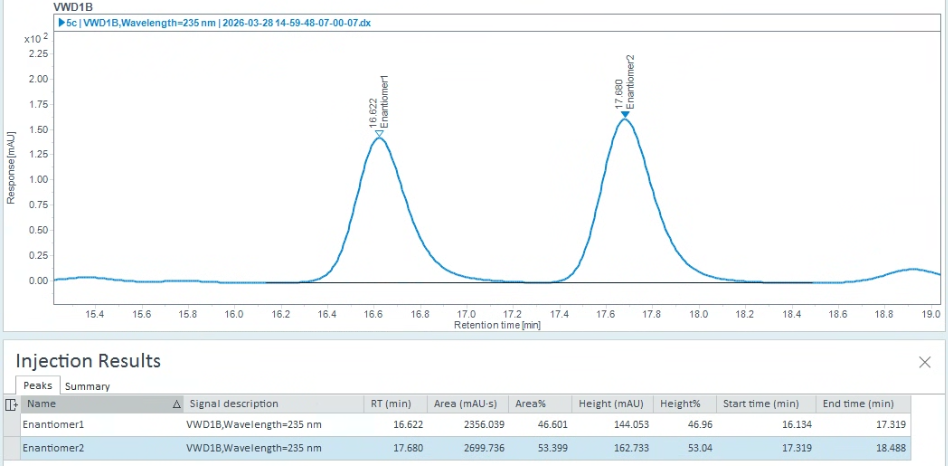}
    \caption{HPLC-UV chromatogram of compound \textbf{3b} (reaction with dCT-G9).}
    \label{fig:ee_3b_dCT-G9}
\end{figure}

\begin{figure}[ht!]
    \centering
    \includegraphics[width=1\linewidth]{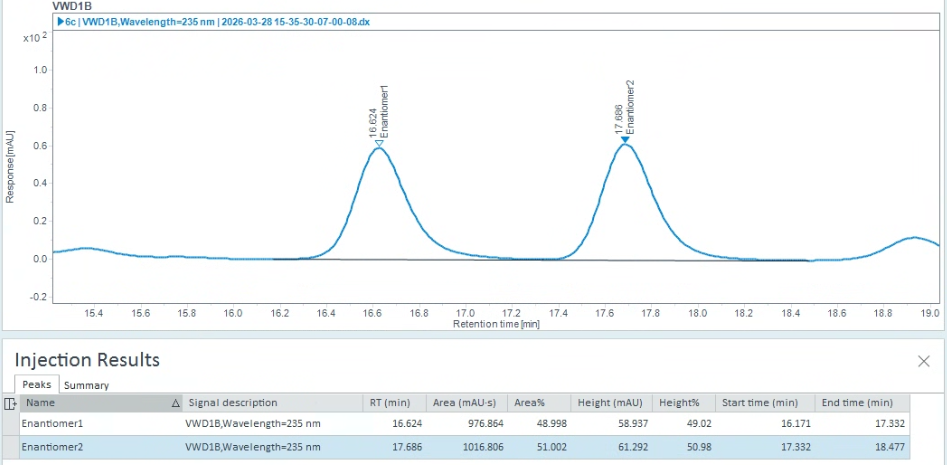}
    \caption{HPLC-UV chromatogram of compound \textbf{3b} (reaction with dCT-H5).}
    \label{fig:ee_3b_dCT-H5}
\end{figure}

\begin{figure}[ht!]
    \centering
    \includegraphics[width=1\linewidth]{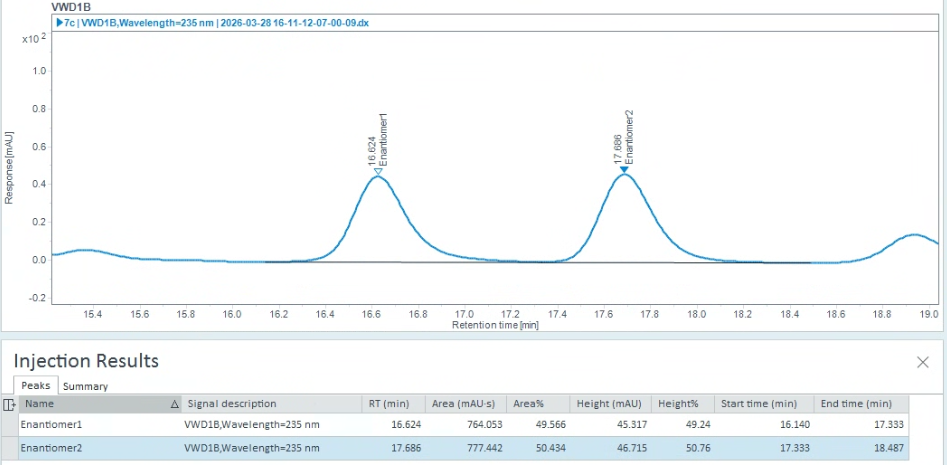}
    \caption{HPLC-UV chromatogram of compound \textbf{3b} (reaction with dCT-H10).}
    \label{fig:ee_3b_dCT-H10}
\end{figure}

\begin{figure}[ht!]
    \centering
    \includegraphics[width=1\linewidth]{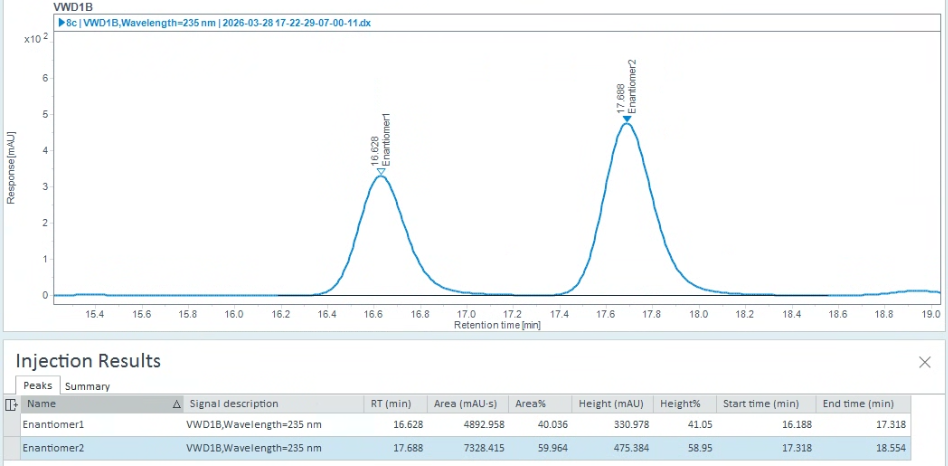}
    \caption{HPLC-UV chromatogram of compound \textbf{3b} (reaction with dCT-H11).}
    \label{fig:ee_3b_dCT-H11}
\end{figure}

\begin{figure}[ht!]
    \centering
    \includegraphics[width=1\linewidth]{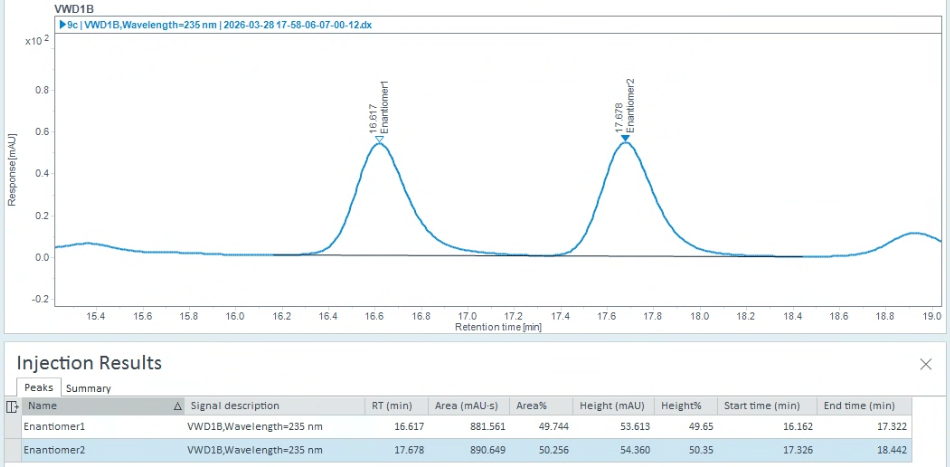}
    \caption{HPLC-UV chromatogram of compound \textbf{3b} (reaction with TmTrpB control).}
    \label{fig:ee_3b_TmTrpB}
\end{figure}

\begin{figure}[ht!]
    \centering
    \includegraphics[width=1\linewidth]{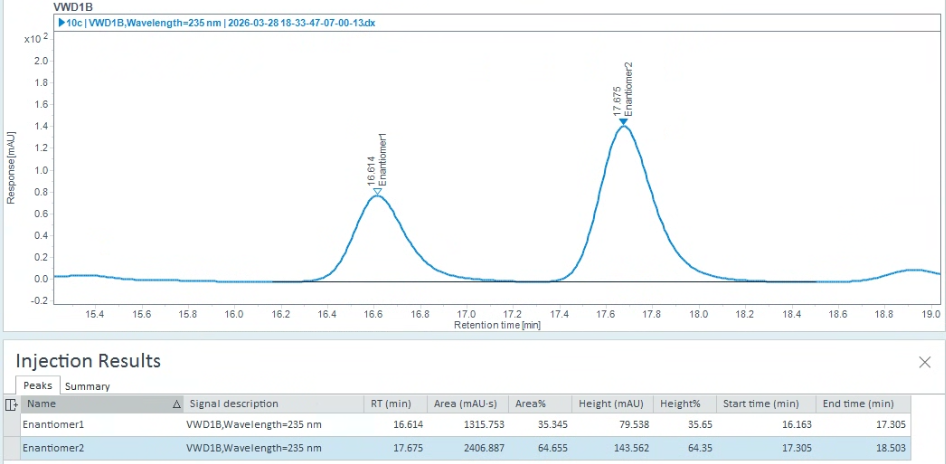}
    \caption{HPLC-UV chromatogram of compound \textbf{3b} (reaction with ApePgb 5312 control).}
    \label{fig:ee_3b_pos_ctrl}
\end{figure}

\begin{figure}[ht!]
    \centering
    \includegraphics[width=1\linewidth]{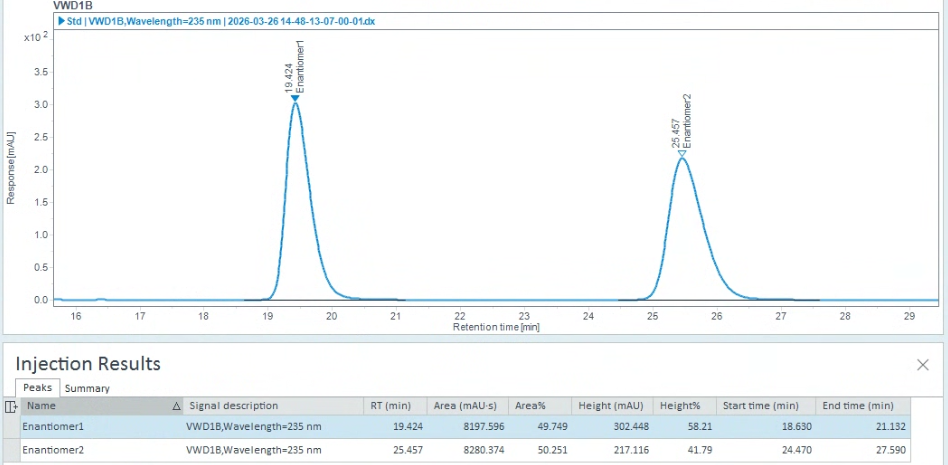}
    \caption{HPLC-UV chromatogram of an authentic product standard of compound \textbf{2b}.}
    \label{fig:ee_2b_std}
\end{figure}

\begin{figure}[ht!]
    \centering
    \includegraphics[width=1\linewidth]{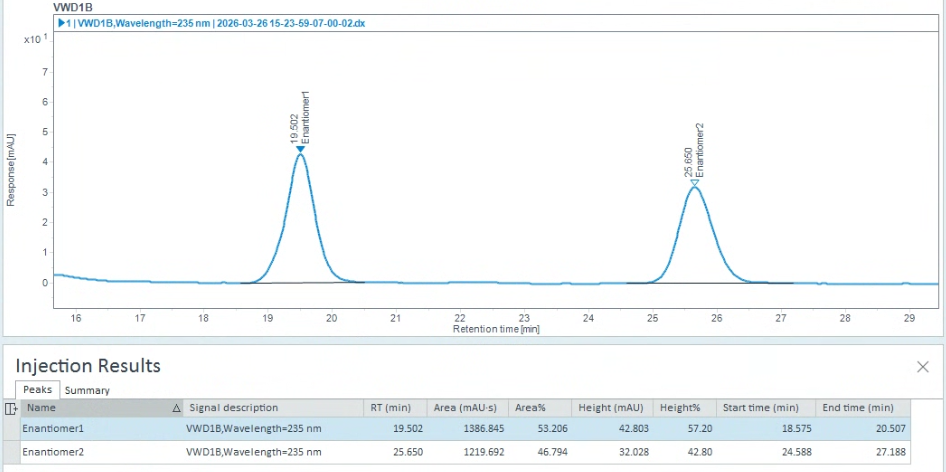}
    \caption{HPLC-UV chromatogram of compound \textbf{2b} (reaction with dCT-B9).}
    \label{fig:ee_2b_B9}
\end{figure}

\begin{figure}[ht!]
    \centering
    \includegraphics[width=1\linewidth]{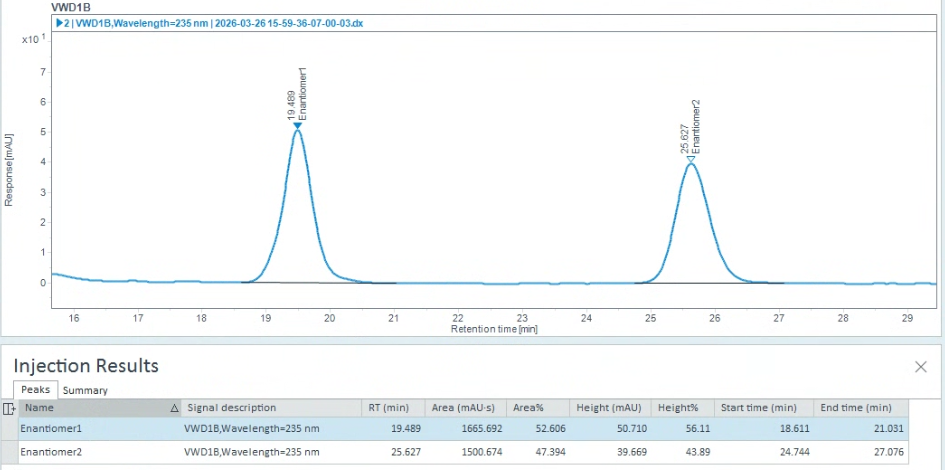}
    \caption{HPLC-UV chromatogram of compound \textbf{2b} (reaction with dCT-F3).}
    \label{fig:ee_2b_F3}
\end{figure}

\begin{figure}[ht!]
    \centering
    \includegraphics[width=1\linewidth]{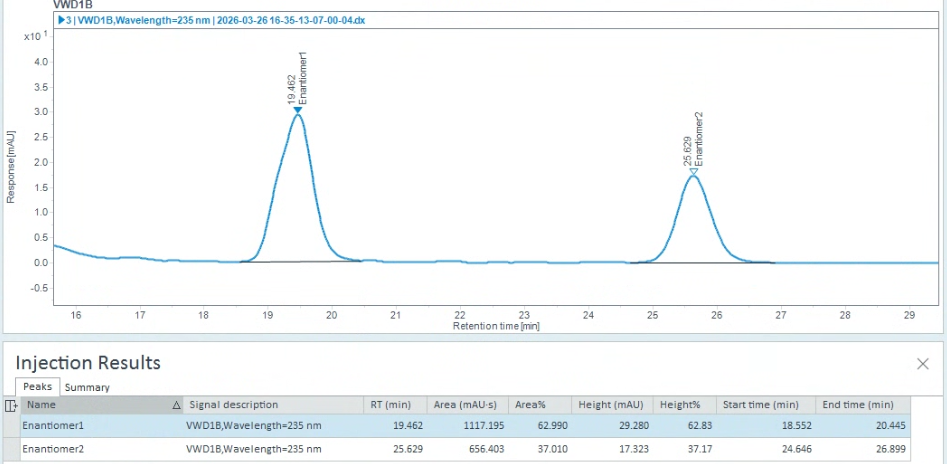}
    \caption{HPLC-UV chromatogram of compound \textbf{2b} (reaction with dCT-F9).}
    \label{fig:ee_2b_F9}
\end{figure}

\begin{figure}[ht!]
    \centering
    \includegraphics[width=1\linewidth]{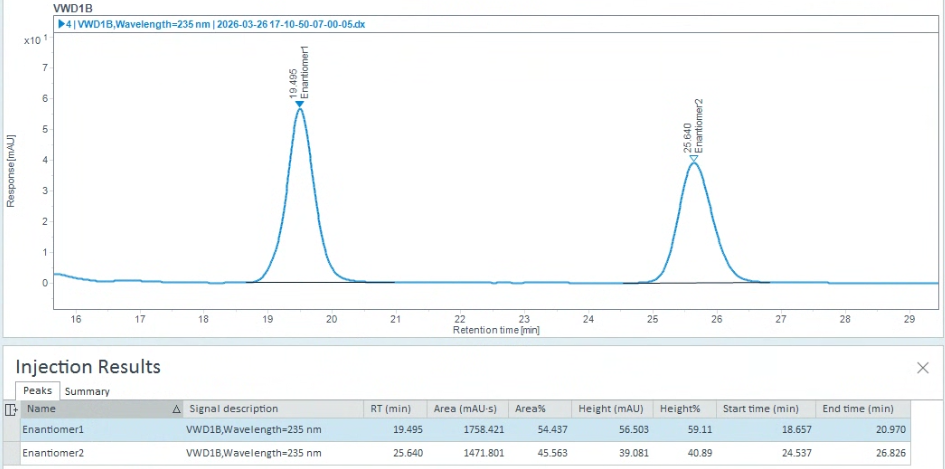}
    \caption{HPLC-UV chromatogram of compound \textbf{2b} (reaction with dCT-G1).}
    \label{fig:ee_2b_G1}
\end{figure}

\begin{figure}[ht!]
    \centering
    \includegraphics[width=1\linewidth]{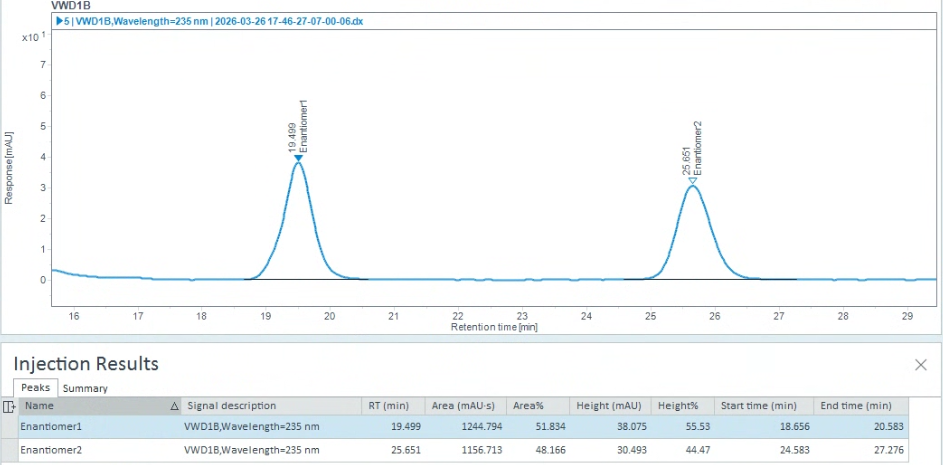}
    \caption{HPLC-UV chromatogram of compound \textbf{2b} (reaction with dCT-G9).}
    \label{fig:ee_2b_G9}
\end{figure}

\begin{figure}[ht!]
    \centering
    \includegraphics[width=1\linewidth]{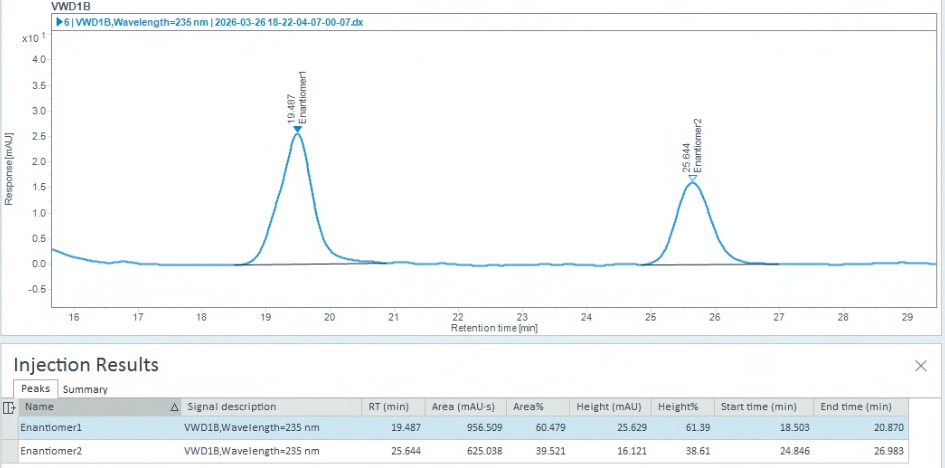}
    \caption{HPLC-UV chromatogram of compound \textbf{2b} (reaction with dCT-H5).}
    \label{fig:ee_2b_H5}
\end{figure}

% \begin{figure}[ht!]
%     \centering
%     \includegraphics[width=1\linewidth]{figures/BH_Ins_7_H10.png}
%     \caption{HPLC-UV chromatogram of compound \textbf{2b} (reaction with dCT-H10).}
%     \label{fig:placeholder}
% \end{figure}

\begin{figure}[ht!]
    \centering
    \includegraphics[width=1\linewidth]{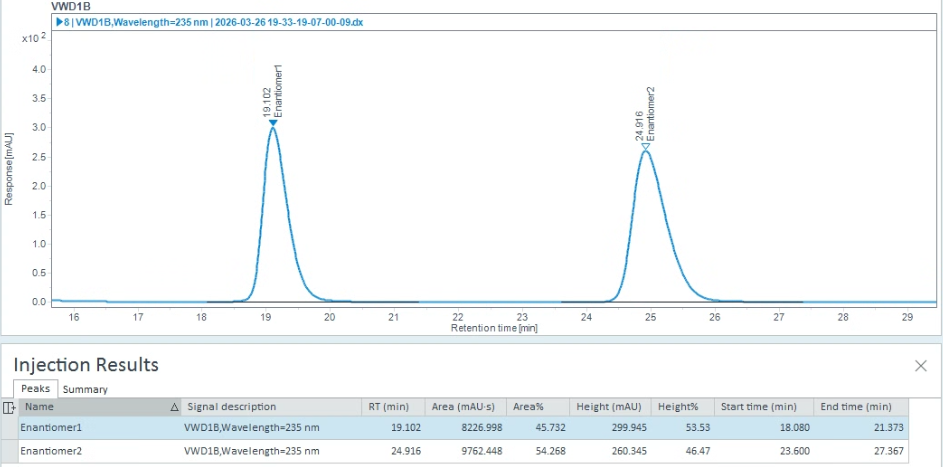}
    \caption{HPLC-UV chromatogram of compound \textbf{2b} (reaction with dCT-H11).}
    \label{fig:BH_Ins_8_H11}
\end{figure}

\begin{figure}[ht!]
    \centering
    \includegraphics[width=1\linewidth]{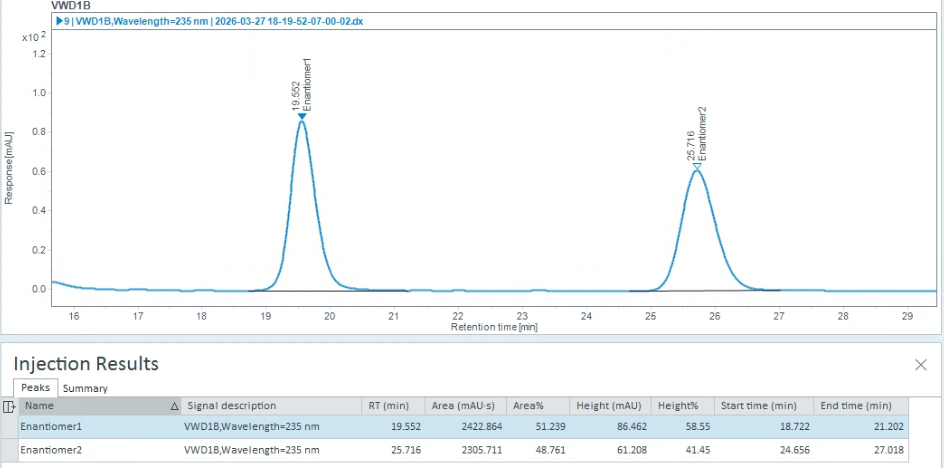}
    \caption{HPLC-UV chromatogram of compound \textbf{2b} (reaction with TmTrpB control).}
    \label{fig:BH_Ins_9_TrpB}
\end{figure}

\begin{figure}[ht!]
    \centering
    \includegraphics[width=1\linewidth]{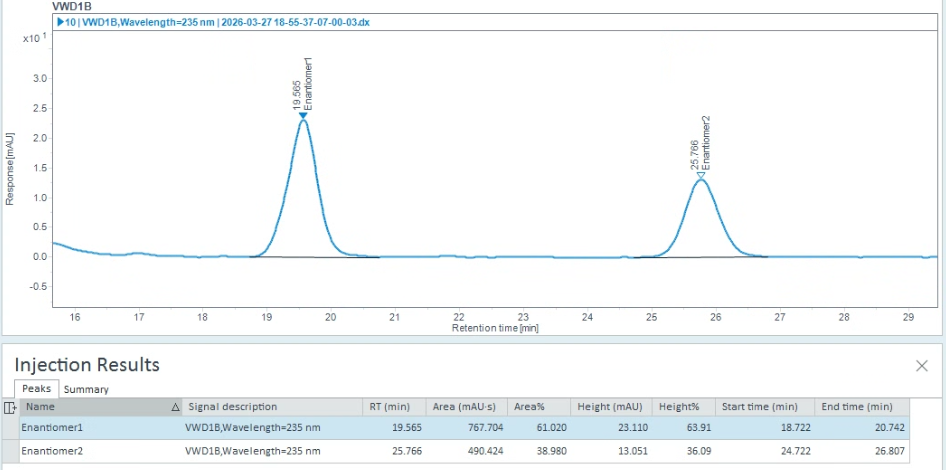}
    \caption{HPLC-UV chromatogram of compound \textbf{2b} (reaction with ApePgb 5312 control).}
    \label{fig:BH_Ins_10_ctrl}
\end{figure}

\clearpage 
\subsection{Chiral GC--FID traces for enantioselectivity determination}

\begin{figure}[ht!]
    \centering
    \includegraphics[width=0.95\linewidth]{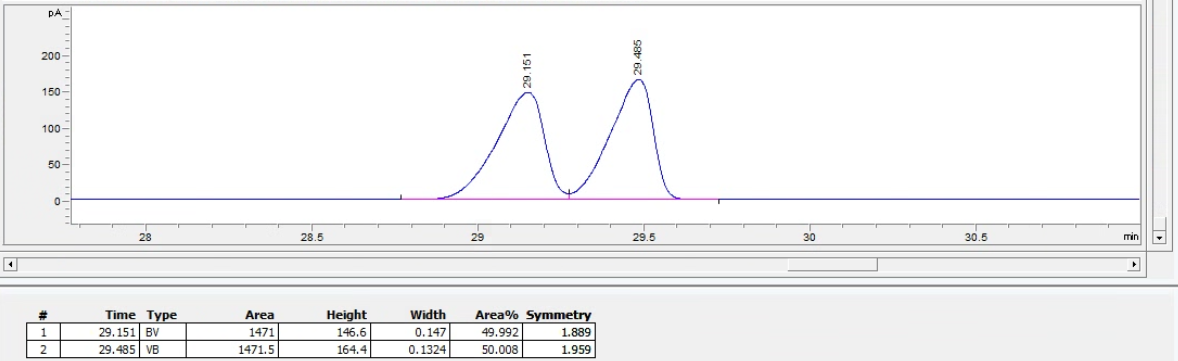}
    \caption{Chiral GC-FID trace for racemic standard of \textbf{4b}.}
    \label{fig:Azet_Evo_Std_Racemate_2}
    \vspace{-1em}
\end{figure}

\begin{figure}[ht!]
    \centering
    \includegraphics[width=0.95\linewidth]{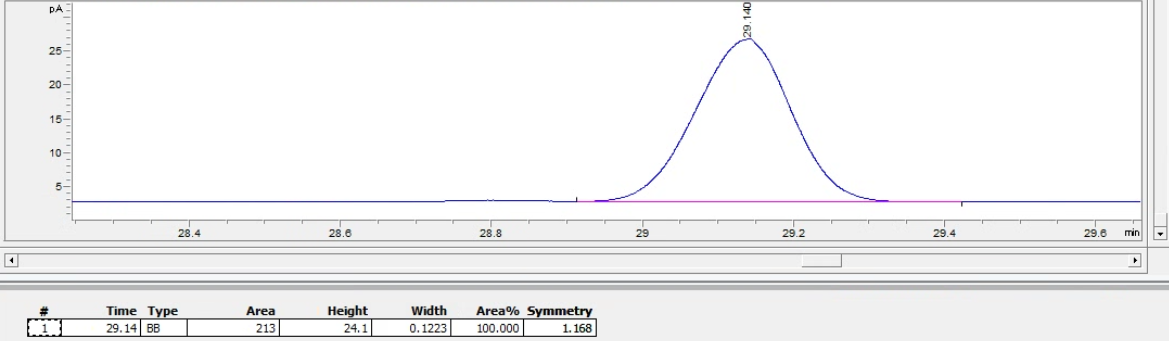}
    \caption{Chiral GC-FID trace for an enantiopure standard of (R)-\textbf{4b}.}
    \label{fig:Azet_Evo_Std_R_2}
    \vspace{-1em}
\end{figure}

\begin{figure}[ht!]
    \centering
    \includegraphics[width=0.95\linewidth]{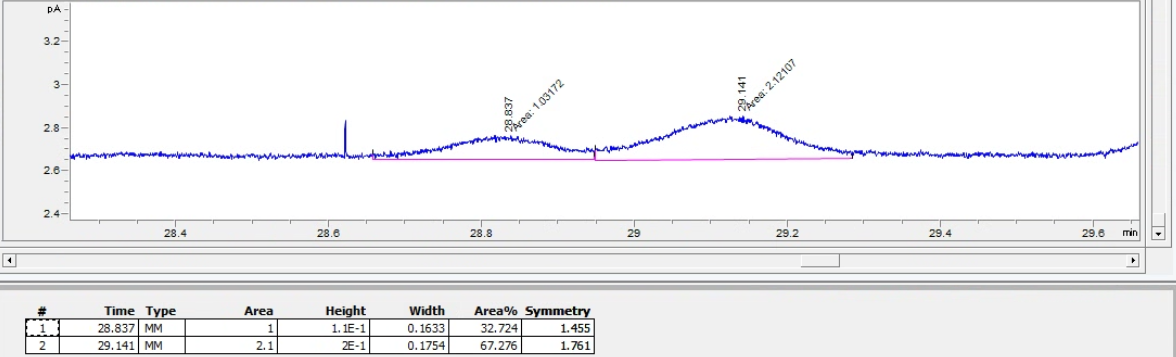}
    \caption{Chiral GC-FID trace for \textbf{4b} (reaction with dCT-H11).}
    \label{fig:Azet_Evo_H11_parent_2}
    \vspace{-1em}
\end{figure}

\begin{figure}[ht!]
    \centering
    \includegraphics[width=0.95\linewidth]{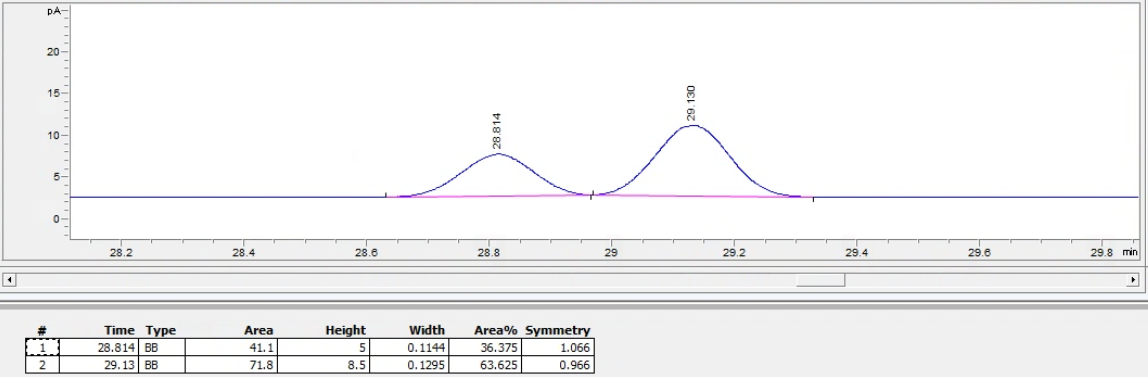}
    \caption{Chiral GC-FID trace for \textbf{4b} (reaction with evolved variant dCT-r1-B7).}
    \label{fig:Azet_Evo_B7_2}
\end{figure}

\begin{figure}[ht!]
    \centering
    \includegraphics[width=0.95\linewidth]{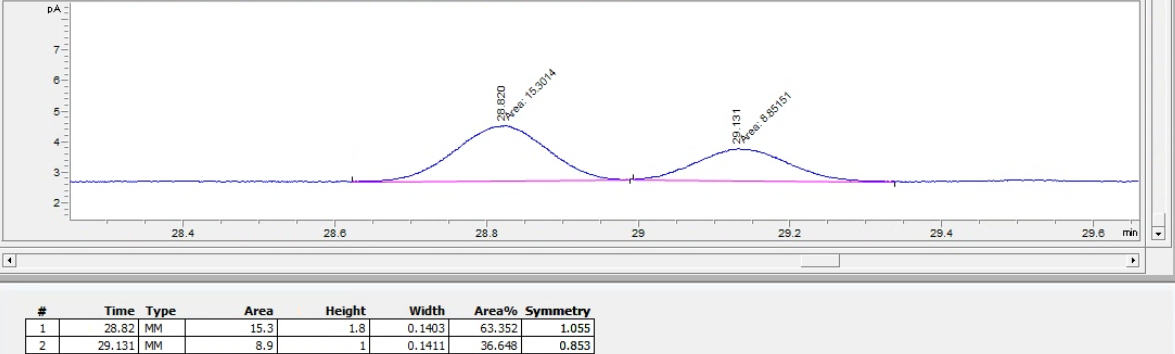}
    \caption{Chiral GC-FID trace for \textbf{4b} (reaction with evolved variant dCT-r1-E3).}
    \label{fig:Azet_Evo_E3_2}
    \vspace{-1em}
\end{figure}

\begin{figure}[ht!]
    \centering
    \includegraphics[width=0.95\linewidth]{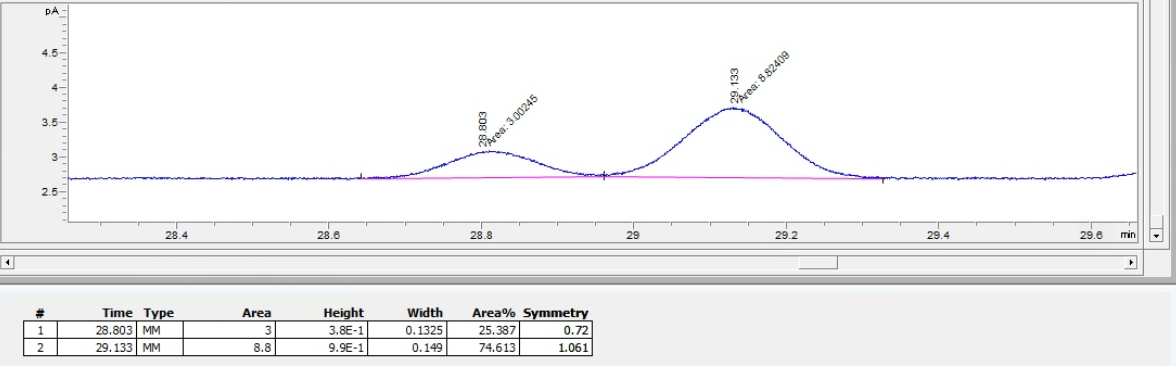}
    \caption{Chiral GC-FID trace for \textbf{4b} (reaction with evolved variant dCT-r1-E12).}
    \label{fig:Azet_Evo_E12_2}
    \vspace{-1em}
\end{figure}

\begin{figure}[ht!]
    \centering
    \includegraphics[width=0.95\linewidth]{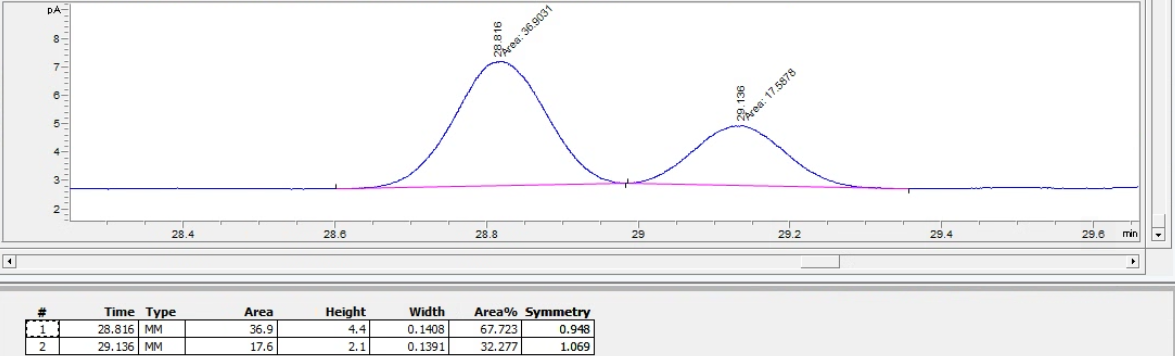}
    \caption{Chiral GC-FID trace for \textbf{4b} (reaction with evolved variant dCT-r1-F1).}
    \label{fig:Azet_Evo_F1_2}
    \vspace{-1em}
\end{figure}

\begin{figure}[ht!]
    \centering
    \includegraphics[width=0.95\linewidth]{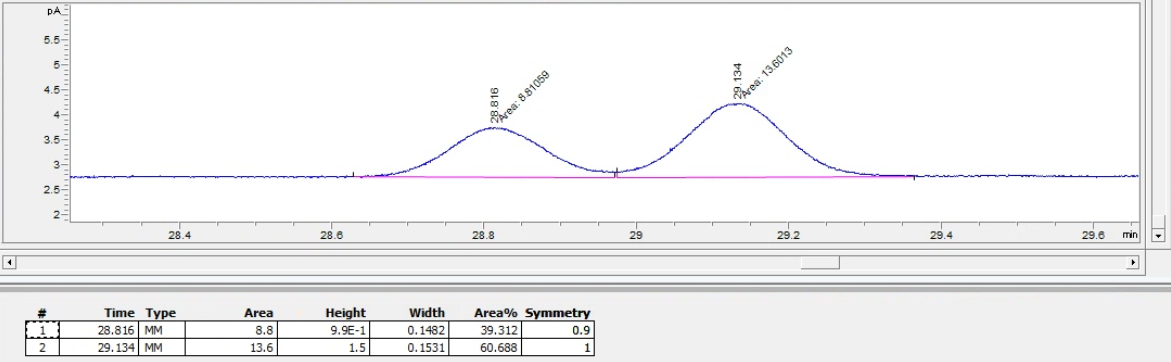}
    \caption{Chiral GC-FID trace for \textbf{4b} (reaction with evolved variant dCT-r1-D1).}
    \label{fig:Azet_Evo_D1_2}
    \vspace{-1em}
\end{figure}

\end{document}